\documentclass{article}

\usepackage{arxiv}

\usepackage[utf8]{inputenc} 
\usepackage[T1]{fontenc}    
\usepackage{url}            
\usepackage{booktabs}       
\usepackage{tabularx}
\usepackage{amsfonts}       
\usepackage{amsmath}
\usepackage{amssymb}
\usepackage{amsthm}
\usepackage{nicefrac}       
\usepackage{microtype}      
\usepackage{graphicx}
\usepackage{placeins}
\usepackage{float}
\usepackage{xcolor}
\usepackage{enumitem}
\usepackage[authoryear]{natbib}
\usepackage{hyperref}       
\hypersetup{
	colorlinks=true,
	citecolor=blue,
	linkcolor=blue,
	urlcolor=blue
}
\graphicspath{ {./images/} }

\newcommand{\R}{\mathbb{R}}
\newcommand{\E}{\mathbb{E}}
\newcommand{\cV}{\mathcal{V}}
\newcommand{\Proj}{\Pi}
\newcommand{\bm}[1]{\boldsymbol{#1}}

\newcommand{\Prob}{\mathbb{P}}
\newcommand{\Beta}{\mathcal{B}}
\newcommand{\sinth}[2]{\sin_\Theta(#1,#2)}

\newtheorem{theorem}{Theorem}[section]
\newtheorem{condition}[theorem]{Condition}
\newtheorem{definition}[theorem]{Definition}
\newtheorem{proposition}[theorem]{Proposition}
\newtheorem{corollary}[theorem]{Corollary}
\newtheorem{lemma}[theorem]{Lemma}
\newtheorem{remark}[theorem]{Remark}

\newtheoremstyle{example}
  {\topsep}       
  {\topsep}       
  {\normalfont}   
  {}              
  {\bfseries}     
  {.}             
  {.5em}          
  {}

\theoremstyle{example}
\newtheorem{example}{Example}

\title{The Geometry of Multitask Learning: Sharp Rates for Shared Subspace Estimation}

\author{
	Tony Cai \\
    Department of Statistics and Data Science\\
	University of Pennsylvania\\
	\texttt{tcai@wharton.upenn.edu} \\
	\And 
	Abhinav Chakraborty\\
	Department of Statistics\\
	Columbia University\\
	\texttt{ac4662@columbia.edu} \\
	\And
	Samory Kpotufe\\
	Department of Statistics\\
	Columbia University\\
	\texttt{samory@columbia.edu} 
}

\begin{document}
	\maketitle
	\begin{abstract}
		
		Learning a shared subspace in multitask learning is an important statistical problem, yet exact rates remain poorly understood in many of the formal settings considered in the literature. To address this gap, which arises from limitations of existing upper-bound techniques, we propose a general multitask lemma. The lemma provides sufficient conditions for fast subspace estimation in generic multitask settings, in the form of geometric conditions on the task regression functions and their estimators. We subsequently show that the multitask lemma yields sharp minimax rates---previously unknown---for linear regression and classification in the regimes considered, and rates matching up to logarithmic factors for additive-noise regression.
	\end{abstract}
	
	\section{Introduction}
	\label{sec:introduction}

Multitask representation learning exploits common structure across a collection of related prediction problems.  Rather than learning each task separately in the ambient covariate space, one first estimates a shared low-dimensional representation and then carries out task-specific prediction within that representation.  This principle, often referred to as \emph{pretraining} or \emph{feature learning}, is now the prevalent approach in modern machine learning: e.g., various downstream tasks such as classification, image segmentation, depth perception across evolving categories of objects all rely on pretrained image features  \citep{radford2021learning,oquab2024dinov2,bolya2025perception,zhuang2025argus}; this is the same with modern text analysis and completion (question answering) with LLMs which all rely on pretrained features of natural language  \citep{mccann2018natural,raffel2020exploring,sanh2022multitask,chung2024scaling}.  Multitask representation learning also arises in scientific applications in which the shared representation is itself the inferential target.  Examples include identifying common functional patterns across neuroimaging studies \citep{marquand2014bayesian,mensch2021extracting,wang2026fmri}, recovering biological pathways shared across treatment-response problems \citep{yuan2016multitask,wu2025anticancer}, and estimating latent risk factors across different financial assets \citep{lettau2020latent,wan2024mining}.

These modern applications motivate understanding \emph{how accurately} a shared representation can be learned from data across tasks. We study this question in settings where each task's regression function $f_t(x), t \in [T]$, depends on the covariates only through their projection onto a common low-dimensional subspace $\mathcal V$. In other words, learning the shared representation amounts to estimating $\mathcal V$ by somehow aggregating data from the collection of tasks. The accuracy of this estimation depends not only on the number of tasks and sample sizes per task, but also on how strongly different directions in $\mathcal V$ are represented across the task collection. A weakly represented direction can become a statistical bottleneck, even when other directions are well represented. The central question of this paper is therefore
\begin{quote}
    \emph{How does the geometry of the task collection determine the statistical rate for estimating the shared representation subspace $\mathcal V$?}
\end{quote}

	Despite extensive work establishing prediction and representation-learning guarantees in multitask settings \citep{baxter2000model,pontil2013excess,maurer2016benefit,tripuraneni2021provable,thekumparampil2021statistically}, such a fundamental question is still not well understood, even in basic parametric models, leaving open the optimality or lack-thereof of existing estimation procedures. For instance, in multitask linear regression, where \(f_t(x)=\beta_t^\top x\) and \(\mathcal V=\operatorname{span}(\beta_1,\ldots,\beta_T)\), exact minimax rates are known only in restrictive isotropic regimes where the nonzero eigenvalues of the task covariance \(\Sigma_{\bm\beta}=\frac{1}{T}\sum_{t=1}^T\beta_t\beta_t^\top\) are of comparable order \citep{niu2024learning,tian2025learning}. For multitask nonlinear models with $f_t(x) = g(\beta_t^\top x)$, including classification and additive-noise regression, existing upper bounds are likewise not sharp, and matching lower bounds are not available \citep{niu2024learning}. 
    
    This paper establishes for the first time how the smallest nonzero eigenvalue of the task covariance governs the rate of shared-subspace estimation in several well-studied parametric settings considered in the literature. We obtain sharp minimax rates for linear regression and classification in the regimes considered, and matching rates up to logarithmic factors for additive-noise regression. These results follow from new techniques for deriving tight upper bounds, paired with new lower bounds in nonlinear settings. Furthermore, the analysis yields new insights into the optimality or lack thereof of traditional estimators, including basic algorithmic modifications that lead to optimality in the linear (Section \ref{ssec:linear-reg-overview}) and nonlinear (Appendix~\ref{sec:nonlinear-general-design}) case.
	\subsection{Problem Formulation}
	\label{subsec:problem-formulation}
	
	We now formalize the shared-subspace model and notation previewed above.
	We consider $T$ prediction tasks.  For each task $t\in[T]$, we observe $n$ independent pairs
	\[
	\{(X_{t,i},Y_{t,i}):i\in[n]\},
	\]
	and the regression function has the shared single-index form
	\begin{equation}
		f_t(x)=\E(Y_t\mid X_t=x)=g(\beta_t^{\top}x),
		\qquad \beta_t\in\R^d,
		\label{eq:intro-model}
	\end{equation}
	where the link $g$ is common across tasks. 
	For ease of presentation, we use a common link in the main text; Appendix~\ref{sec:task-dependent-nonlinear-links} considers the more general case of task-dependent links \(g_t\).
	 We write
	\[
	\mathcal V:=\operatorname{span}(\beta_1,\ldots,\beta_T)\subset\R^d
	\]
	for the $k$-dimensional subspace spanned by the task parameters.
	The identity link gives linear regression, while suitable nonlinear links cover binary classification and non-linear  regression.  Throughout the principal results, the rank $k$ is treated as known and the different task datasets are independent.
	
	As previewed above, the geometry of the task collection is encoded by the task covariance
	\begin{equation}
		\Sigma_{\bm\beta}
		:=\frac{1}{T}\sum_{t=1}^T\beta_t\beta_t^{\top},
		\label{eq:intro-task-covariance}
	\end{equation}
	whose range is $\mathcal V$.  Let
	$
	\lambda_1\ge\cdots\ge\lambda_k>0$
	denote its nonzero eigenvalues.  The smallest nonzero eigenvalue $\lambda_k$ measures the strength of the weakest represented direction.  A small $\lambda_k$ means that some direction in $\mathcal V$ is only weakly expressed across the task collection, making the full subspace difficult to identify even when other directions are strongly represented.
	
	For a subspace $\mathcal U\subseteq\mathbb R^d$, let $\Proj_{\mathcal U}$ denote the orthogonal projector onto $\mathcal U$.  We measure the error in estimating $\mathcal V$ using the operator sine-theta loss
	\[
	\sinth{\widehat{\mathcal V}}{\mathcal V}
	:=
	\left\|
	\Proj_{\widehat{\mathcal V}}-\Proj_{\mathcal V}
	\right\|_{\mathrm{op}}.
	\]
   
    The $\sin_{\Theta}$ error naturally upper-bounds the downstream prediction risk.
    In particular, Proposition~\ref{prop:downstream-risk} shows that the excess squared prediction loss on a downstream task—after projection onto $\widehat{\mathcal V}$—can be upper-bounded by the sum of  $\sin_\Theta^2(\widehat{\mathcal V},\mathcal V)$ and estimation error. For instance, for least-squares linear regression with \(m\) samples from the downstream task, under the assumptions of Remark~\ref{rem:downstream-least-squares}, such a downstream upper bound takes the form
    \[
    O\!\left(\frac{k}{m}+\sin_\Theta^2(\widehat{\mathcal V},\mathcal V)\right).
    \]
It follows that, to improve performance in downstream regression, e.g., achieve an excess risk of $O(k/m)$ down from 
the usual high-dimensional rates of $O({d}/{m})$, it suffices to estimate $\mathcal V$ at a rate $\sin_\Theta^2(\widehat{\mathcal V},\mathcal V) = O(k/m)$. However, the optimal rate for estimating $\cal V$ remains unclear despite much work on the subject. 
    
	Indeed, even in the basic linear-regression setting, where
	$f_t(x)=\beta_t^\top x$, existing upper and lower bounds on $\sin_\Theta(\widehat{\mathcal V},\mathcal V)$ leave a potentially
	large gap in terms of the dependence of the spectrum of $\Sigma_{\bm \beta}$ (i.e., the geometry of task parameters $\beta_t$), especially in regimes where $\lambda_1 \gg \lambda_k$, i.e., when not all directions in $\cal V$ are well represented in the data. In particular, previous works either assume $\lambda_1\asymp\lambda_k$ \citep{tian2025learning}, i.e., that somehow all directions in the unknown subspace $\cal V$ are equally represented, or show upper-bounds of the form 
	\begin{equation}\label{eq:intro-upper-bound}
	\left[
	\sqrt{\frac{\lambda_1d}{\lambda_k^2nT}}
	+
	\sqrt{\frac{d}{\lambda_k^2n^2T}}
	\right]\wedge 1,
	\end{equation}
     while the best known lower-bounds of \cite{niu2024learning} suggest the faster rates (removing a factor $\sqrt{\lambda_1/\lambda_k}$)
	\begin{equation}
		\mathcal R_{\star}(d,T,n,\lambda_k)
		:=
		\left[
		\sqrt{\frac{d}{\lambda_knT}}
		+
		\sqrt{\frac{d}{\lambda_k^2n^2T}}
		\right]\wedge1.
		\label{eq:intro-benchmark}
	\end{equation}
	Optimal rates (and therefore procedures) are even less understood in more general settings \eqref{eq:intro-model} with nonlinear link functions $g$. Namely, while various works such as \citep{niu2024learning} establish upper-bounds of the form \eqref{eq:intro-upper-bound}, no lower-bounds are known in general settings. 

    The following simple example highlights the importance of this gap, by relating $\lambda_1, \lambda_k$ more concretely to (the a priori unknown) geometry of $\beta_t$'s.  
	
	\begin{example}
		\label{example:linear-reg}
		 Assume $n$ samples per task, and consider the following simple linear regression example: let $\cV=\mathrm{span}\{e_1,e_2\} \subset \R^d$, and suppose that $(T-1)$ tasks have unknown $\beta_t$ equal to $e_1$ while only one task has $\beta_t = e_2$. Intuitively the direction $e_2$ is thus harder to identify, and hence, so is $\cal V$. Such difficulty in identifying the rare direction $e_2$ is encoded in the top two non-zero eigenvalues of $\Sigma_{\bm \beta}$, namely, $\lambda_1= 1- \frac{1}{T}$ and $\lambda_2 = \frac{1}{T}$. We can now get a qualitative sense of the gap between the above upper and lower-bounds: 
		the upper-bound in~\eqref{eq:intro-upper-bound} specializes here to $\sqrt{{\lambda_1 d}/{\lambda_2^2 nT}} \approx \sqrt{{dT}/{n}}$, \emph{which becomes vacuous as $T$ grows toward $n$};  
		instead, if an upper bound matching the lower-bound ~\eqref{eq:intro-benchmark} were achievable, it would yield, for this example much faster rates of the form $\sqrt{{d}/{\lambda_2 nT} } = \sqrt{{d}/{n} }$, when $T = O(n)$, or of the form $\sqrt{{dT}/{n^2}}$ when $T = \Omega(n)$, i.e., still considerably faster than $\sqrt{{dT}/{n}}$. 
        
	\end{example}

    From the example, we see that the gap is consequential on how much information might be pulled out of the $T$ tasks, i.e., we could get significantly further speedup in downstream prediction tasks if it turns out that the lower bound ~\eqref{eq:intro-benchmark} is in fact achievable in any of the settings considered.

	\subsection{Our Contribution}
	\label{subsec:our-contribution}

    Our main aim in this work is to provide a general theoretical framework for deriving sharp rates for shared representation learning across multiple tasks, that properly account for the geometry of the tasks, and provide insight on the design of procedures that achieve these rates. Within this framework, we characterise the minimax rate of estimation for $\mathcal{V}$ in several canonical parametric settings with regression functions of the form $f_t(x) := g(\beta_t^\top  x), t \in [T]$. Our main contributions are summarized below.
    

 	\paragraph{$\bullet$ Multitask Lemma.} We start with a generic estimation result that applies to general
 multitask settings with unknown \(\mathcal V\), as defined in
 Section~\ref{subsec:problem-formulation}, and is agnostic to the choice of task
 statistics \(\widehat\beta_t\), \(t\in[T]\). Such task statistics \(\widehat\beta_t\) may directly estimate $\beta_t$ or may instead estimate suitable surrogates $\widetilde\beta_t:=\E[\widehat\beta_t]$ of $\beta_t$'s, provided $\cal V$ can be recovered from $\widetilde{\cal V} \doteq \text{span}\{\widetilde \beta_t\}_{t \in [T]} $. The importance of this result, as we will see, is in providing useful guidance in designing problem-specific task statistics $\widehat \beta_t$, so as to yield tighter rates of estimation than previously known when instantiated in various settings of interest, e.g., in linear and generalized linear regression and classification settings as alluded to above. For simplicity, we refer to this generic result as a \emph{multitask lemma}. 

 Formally, the multitask lemma establishes tight bounds on the (sin $\Theta$) distance between {
 $\widehat {\cal V} \doteq \widehat{\mathcal{V}}(\{\widehat \beta_t\}_{t \in [T]})$
 }
 and $\widetilde{\cal V}$ in terms of conditions on how task-wise {errors} 
 \(\varepsilon_t:=\widehat\beta_t-\widetilde\beta_t\) (which are assumed to have sub-gaussian tails) interact with $\widetilde{\cal V}$ and its complement, while also taking into account the \emph{signal strength} as captured by the geometry of $\widetilde\beta_t$ (encoded in the spectrum of 
 $\Sigma_{\bm{\tilde\beta}}
 :=
\frac1T\sum_{t=1}^T\tilde\beta_t\tilde\beta_t^\top$). Namely, the lemma reveals the following quantity of interest: let $
 \Sigma_{\bm\varepsilon}
 :=
 \frac{1}{T}\sum_{t=1}^T
 \E[\varepsilon_t\varepsilon_t^\top]
 $ denotes the average of error covariances, and define 
 \begin{equation}\label{eq:mixing}
 	\delta_1
 	:=
 	{\widetilde\lambda_k}^{-1}
 		\left\|
 		\Proj_{\widetilde{\mathcal V}}
 		\Sigma_{\bm\varepsilon}
 		\Proj_{\widetilde{\mathcal V}^{\perp}}
 		\right\|_{\mathrm{op}
 	},
 \end{equation}
where ${\widetilde\lambda_k}$ is the smallest non-zero eigenvalue of $\Sigma_{\bm{\tilde\beta}}$. Roughly, the multitask lemma implies that the benchmark $\mathcal R_{\star}(d,T,n,\lambda_k)$ of \eqref{eq:intro-benchmark} is achievable whenever we can design taskwise statistics $\widehat\beta_t$ such that $\delta_1$ is small, while also having $\widetilde\lambda_k \asymp \lambda_k$.  

The result relies on a matrix perturbation bound of \citet{cai2018rate} 
which provides refinements over classical Davis-Kahan results on the (sin $\Theta$) distance between singular subspaces of a low-rank matrix \(A\) and those of its perturbation \(\widehat A:=A+E\). While the perturbation bound extends easily to stochastic settings where $E$ is random noise with i.i.d. entries, the main difficulty in the multi-task setting is that the errors \(\varepsilon_t:=\widehat\beta_t-\widetilde\beta_t\), which induce the perturbation \(E\) in \(\widehat A=\widetilde A+E\), where
$
\widetilde A
:=
[\widetilde\beta_1,\ldots,\widetilde\beta_T]^\top
$
is the population surrogate matrix, cannot be expected to have i.i.d. entries, hence the need to derive more careful conditions in terms of their relationship to the spaces being estimated. 
The result is presented in Section \ref{ssec:overview-mtl} with further discussion. 


		\paragraph{$\bullet$ First Optimal Rates in Linear Regression via Whitening.}
		Our first application of the multitask lemma concerns the linear multitask model where $f_t(x) = \beta_t^\top x$.
Namely, we show that, for well-conditioned design matrices, an appropriate \emph{whitened} version of traditional OLS estimators achieves the known minimax lower-bounds of \citep{niu2024learning}, closing the existing gap with available upper-bounds which were loose in terms of the geometry of $\beta_t$'s (i.e. the spectrum of $\Sigma_{\bm \beta}$; recall Example \ref{example:linear-reg}).

		Surprisingly, traditional OLS estimator \(\widehat\beta_t^{(\mathrm{OLS})}\) do not appear themselves to achieve the optimal rates of $\mathcal R_\star(d,T,n,\lambda_k)$, 
        at least in light of past rates, and also the rates resulting from 
		instantiating the multitask lemma with $\widehat \beta_t = \widehat\beta_t^{(\mathrm{OLS})}$. To see why, recall that fast estimation depends on the $\delta_1$ term introduced above, which controls the correlation between $\cal V$ and ${\cal V}^\perp$ components of task-wise error terms $\varepsilon_t, t \in [T]$. As it turns out, using OLS estimators, the task-wise errors have average covariance $\Sigma_{\bm\varepsilon}=(\sigma_\xi^2/n)\widehat S$, for \(\widehat S:=\frac1T\sum_{t=1}^T(\frac1n \sum_{i=1}^n X_{t,i} X_{t,i}^\top)^{-1}\). In other words, the geometry of $\varepsilon_t$'s depend  directly on the $X$-marginal, and therefore can be arbitrarily misaligned with $\cal V, \ {\cal V}^\perp$ (which depend on the conditional $Y \mid X$ through $\beta_t$'s).
		In fact, Remark~\ref{rem:ols-nonvanishing-delta} gives an example where the OLS errors yield $\delta_1=\Theta(1)$, and the resulting bias prevents spectral aggregation of the unwhitened OLS estimates from consistently recovering $\mathcal V$ as $T$ grows. 
		 Our solution is to try to induce $\Sigma_{\bm \varepsilon}  \propto I_d$, by proper whitening, which would lead to $\delta_1=0$. To this end we consider the task-wise statistics
		\(\widehat\beta_t:=\widehat S^{-1/2}
	\widehat\beta_t^{(\mathrm{OLS})}\) which result in $ \widetilde{\cal V} = \widehat S^{-1/2} \cal V $, and $\widetilde \lambda_k \approx \lambda_k$, allowing for the recovery of $\cal V$ at optimal rate. See Section~\ref{ssec:linear-reg-overview} for technical details.

    \paragraph{ $\bullet$ Sharp and near-sharp rates for nonlinear models.}
    
    Our second application of the multitask lemma concerns nonlinear multitask models with regression functions \(f_t(x)=g(\beta_t^\top x)\) and isotropic Gaussian covariates, a standard setting for such models \citep{niu2024learning,collins2024provable}. Concretely, we consider two distinct settings: \emph{binary classification}, where \(Y_t\mid X_t=x\sim\operatorname{Ber}(g(\beta_t^\top x))\), and \emph{additive-noise regression}, where \(Y_t=g(\beta_t^\top X_t)+\xi_t\), with \(\xi_t\) mean-zero, sub-Gaussian, and independent of \(X_t\).
    Using the first-moment estimator described below, we derive upper bounds that improve on previously known guarantees \citep{niu2024learning}. We complement these guarantees with the first minimax lower bounds for both problems. Together, these results yield an exact minimax characterization for binary classification and matching upper and lower bounds up to logarithmic factors for additive-noise regression.
	
	To instantiate the multitask lemma, we use a classical first-moment estimator from the literature on single-index models \citep{li1989regression,naik2000partial}. Specifically, for each task \(t\), we consider the estimator
	$
	\widehat\beta_t=\frac1n\sum_{i=1}^nY_{t,i}X_{t,i}.	$
	Under mild regularity conditions on $g$, Stein’s identity gives \(\widetilde\beta_t=\E[\widehat\beta_t]\propto\beta_t\), so \(\widetilde{\mathcal V}=\mathcal V\), while Gaussian rotational symmetry of $X_t$ ensures \(\delta_1=0\) even though \(\Sigma_{\bm\varepsilon}\) need not be isotropic. For binary classification, bounded responses also provide the required sub-Gaussian error control, completing the conditions needed to apply the multitask lemma. In additive-noise regression, however, the products \(Y_{t,i}X_{t,i}\) need not be sub-Gaussian, even when \(Y_{t,i}\) is sub-Gaussian. We overcome this difficulty by applying the multitask lemma to a response-truncated estimator \(\widehat\beta_t^{(L)}\), obtained by replacing responses with magnitude exceeding \(L\) by zero in \(\widehat\beta_t\). Using a general alignment result (Proposition~\ref{prop:response-transformation-alignment}), we show that this truncation preserves the exact alignment i.e. \(\widetilde\beta^{(L)}_t=\E[\widehat\beta^{(L)}_t]\propto\beta_t\) so \(\widetilde{\cal V} = \mathcal{V}\). We further verify that \(\delta_1=0\), allowing us to apply the multitask lemma to the truncated estimator. We choose \(L\) so that the truncated and original first-moment estimators yield the same estimated subspace with high probability, transferring the bound at the cost of additional logarithmic factors. The favorable error geometry above relies on isotropy of the design covariance. In the non-isotropic, unknown-covariance case, whitening with sample splitting preserves the same two-term classification upper-bound rate, up to constants, for a common, well-conditioned Gaussian covariance (Appendix~\ref{sec:nonlinear-general-design}).
	
	While the preceding construction yields the upper bounds, establishing the corresponding lower bounds requires a different set of ideas. The existing lower-bound argument (known only for linear regression) relies on a Gaussian prior for the task coefficients and Gaussian observation noise, which preserve Gaussian structure (owing to the linearity) and permit explicit divergence calculations \citep{niu2024learning}. Nonlinear links destroy this structure, preventing a direct extension of the argument. We instead use discrete priors and develop new divergence bounds for the resulting mixture distributions, obtaining the first minimax lower bounds for both binary classification and additive-noise regression; see Section~\ref{ssec:gsi-overview} for details.
	
	Taken together, these results provide a unified theory of shared-subspace
	estimation with sharp and near-sharp rates across linear and nonlinear
	multitask models. In all three settings, the rate has the same two-term
	structure as in \(\mathcal R_\star(d,T,n,\lambda_k)\),
	with a phase transition at \(n\lambda_k\asymp1\). Above this threshold the
	first term dominates, whereas below it the second term governs the estimation
	error. Our numerical experiments illustrate this transition and the
	rare-direction geometry of Example~\ref{example:linear-reg}. The following
	remark gives a simple explanation for the two regimes.
	
	\begin{remark}[Gaussian matrix-denoising interpretation]
		\label{rem:existing-phase-transition}
	
		The same two-term rate and phase transition can already be seen in the
		most basic Gaussian mean-vector model. Consider  observing $\{Z_t\}_{t\in[T]}$ according to the model
		\[
		Z_t=\beta_t+\sigma\varepsilon_t,
		\qquad
		\varepsilon_t\stackrel{\mathrm{i.i.d.}}{\sim}N(0,I_d),
		\qquad t\in[T],
		\]
		where \(\operatorname{span}(\beta_1,\ldots,\beta_T)=\mathcal V\). To express
		this model in the matrix notation introduced above, let
		$
		\widehat A:=[Z_1,\ldots,Z_T]^\top,$
	$
		\widetilde A:=\E[\widehat A]
		=[\beta_1,\ldots,\beta_T]^\top.
$
		Then
	$
		\widehat A=\widetilde A+E,
		E:=\sigma[\varepsilon_1,\ldots,\varepsilon_T]^\top.
		$
		The signal matrix \(\widetilde A\) has rank \(k\), with right singular
		subspace \(\operatorname{span}(\widetilde A^\top)=\mathcal V\), while the
		entries of \(E\) are i.i.d.\ \(N(0,\sigma^2)\). Thus, estimating
		\(\mathcal V\) from \(\widehat A\) is precisely an i.i.d.\ Gaussian
		matrix-denoising problem, whose minimax rate is known to be
		\[
		\left[
		\sigma\sqrt{\frac{d}{T\lambda_k}}
		+
		\sigma^2\sqrt{\frac{d}{T\lambda_k^2}}
		\right]\wedge1;
		\]
		see \citet{cai2018rate} and Appendix~\ref{app:gaussian-means}.  The first term dominates when
		\(\lambda_k\gg\sigma^2\), the second dominates when
		\(\lambda_k\ll\sigma^2\), and the transition occurs at
		\(\lambda_k\asymp\sigma^2\).

		In the multitask models studied here, estimating a task parameter from \(n\)
		observations typically incurs squared \(\ell_2\) error of order \(d/n\),
		corresponding to an effective coordinatewise variance
		\(\sigma^2\asymp1/n\). Substituting this scale into the Gaussian-denoising rate
		gives exactly \(\mathcal R_\star(d,T,n,\lambda_k)\), and the phase transition
		\(\lambda_k\asymp\sigma^2\) becomes \(n\lambda_k\asymp1\).

		In this Gaussian model, the errors are i.i.d.\ In multitask regression,
		however, the task-wise errors are generally heterogeneous and dependent. Our multitask lemma shows
		that the same two-term rate and phase transition persist in this more general multitask
		setting.

	\end{remark}

The preceding remark isolates, in a canonical Gaussian model, the core interaction between the weakest direction in the task geometry, quantified by $\lambda_k$, and task-wise estimation errors that produces the two-term rate and phase transition. To study this geometric effect sharply in multitask learning, we work in a structured setting: the representation is shared exactly, the rank $k$ is known, every task has the same sample size, and the task datasets are independent.  The nonlinear applications use isotropic Gaussian designs. The main-text results are presented for a common link, while Appendix~\ref{sec:task-dependent-nonlinear-links} extends the upper bounds to task-dependent links under uniform conditions.  These assumptions are reasonable for a sharp-rate analysis and provide a tractable setting for isolating the geometric
		effect of interest.  Approximately shared or nonlinear representations, unequal or dependent tasks, unknown rank, and non-Gaussian designs may introduce additional statistical terms or require different task-statistics and subspace estimation arguments.

Table~\ref{tab:intro-summary} summarizes the model-specific conclusions.  Link-, noise-, and design-dependent constants are suppressed.
	
	\begin{table}[H]
		\centering
		\small
		\caption{Principal model-specific results. The benchmark $\mathcal R_{\star}$ is defined in \eqref{eq:intro-benchmark}. Displayed restrictions concern the minimax comparisons; the upper bounds hold under the assumptions of the corresponding upper-bound theorems.}
		\label{tab:intro-summary}
		\begin{tabularx}{\textwidth}{>{\raggedright\arraybackslash}p{0.19\textwidth}>{\raggedright\arraybackslash}p{0.2\textwidth}>{\raggedright\arraybackslash}p{0.14\textwidth}>{\raggedright\arraybackslash}X}
			\toprule
			Model & Taskwise construction & Rate & Optimality status\\
			\midrule
			Linear regression with well-conditioned designs &
			Whitened least-squares estimator &
			$\mathcal R_{\star}$ &
			Matching upper and lower bounds for $n\gtrsim d$, $k\gtrsim\log T$, and $T\le e^{cn}$; full conditions in Corollary~\ref{cor:overview-linear-minimax}\\[1mm]
			Binary classification with Gaussian design &
			First-moment estimator &
			$\mathcal R_{\star}$ &
			Matching upper and lower bounds for $n\gtrsim k$ and $T\lesssim d\,e^{cn}$; full conditions in Corollary~\ref{cor:overview-cls-minimax}\\[1mm]
			Additive-noise regression with Gaussian design &
			First-moment estimator &
			$\mathcal R_{\star}$ up to explicit logarithmic factors &
			Upper and lower bounds matching up to logarithmic factors for $n\gtrsim k$ and $T\lesssim d\,e^{c\sqrt n}$; full conditions in Corollary~\ref{cor:overview-add-minimax}\\
			\bottomrule
		\end{tabularx}
	\end{table}
	\FloatBarrier


	\subsection{Other Related Work}
	
	\paragraph{Statistical learning theory for multitask learning.}
	\citet{baxter2000model}, \citet{pontil2013excess}, and \citet{maurer2016benefit} give foundational statistical guarantees for multitask and representation learning, explaining when shared structure can improve learning across tasks. These works motivate our shared-subspace viewpoint, but they focus on inductive bias or prediction risk rather than exact minimax rates for estimating the subspace itself.
	
	\paragraph{Shared linear representations in multitask learning.}
	\citet{tripuraneni2021provable} and
	\citet{thekumparampil2021statistically} study algorithms for linear
	shared-representation learning, while \citet{du2021fewshot},
	\citet{chua2021how}, and \citet{collins2021exploiting} analyze related
	transfer, fine-tuning, and federated formulations. These works provide
	important upper bounds and downstream guarantees, but do not characterize
	the subspace-estimation rate under general anisotropic task geometry.
    \citet{tripuraneni2021provable} study linear shared-representation learning under isotropic Gaussian design. They show for a certain estimator \(\widehat{\mathcal V}\),
\(\sin\Theta(\widehat{\mathcal V},\mathcal V)\lesssim_{\log}
\sqrt{d/(Tn\lambda_k^2)}\), while their lower bound is
\(\max\{(Tn\lambda_k)^{-1/2},\sqrt{dk/(Tn)}\}\).
These bounds do not match in general, leaving the exact subspace-estimation
rate unresolved. \citet{niu2024learning} improve on this line of work by
developing sharper spectral methods for shared linear and generalized-linear
representations. As discussed in \ref{subsec:problem-formulation}, their upper bound is sharp when the
	nonzero task-covariance eigenvalues are of comparable order, but in the
	general anisotropic regime it retains an additional dependence on the
	largest eigenvalue. In their formulation, moreover,
	the target is the covariance-transformed subspace
	\(\mathcal V_m=\operatorname{span}(\Sigma_{X,1}\beta_1,\ldots,
	\Sigma_{X,T}\beta_T)\) (here $\Sigma_{X,t} = \E [X_tX_t^\top]$ is the covariance of the design $X_t$), which need not coincide with
	\(\mathcal V=\operatorname{span}(\beta_1,\ldots,\beta_T)\). When the design
	is isotropic (i.e $\Sigma_{X,t} = I_d, t\in[T]$), the two targets coincide, and our linear result removes the
	additional largest-eigenvalue dependence. For the nonlinear models, we
	obtain an exact minimax characterization for classification and matching
	upper and lower bounds up to logarithmic factors for additive-noise
	regression. Remark~\ref{rem:niu-linear-comparison} gives a detailed
	comparison in the linear setting.
	
	
	\paragraph{Heterogeneity, robustness, and misspecification.}
	Several works extend the shared linear-representation framework of \citet{tripuraneni2021provable} to more general data-generating settings. \citet{duchi2022subspace} accommodate heterogeneous, user-dependent noise and, in their linear regression formulation, allow the response noise to depend on the realized covariates. \citet{zhang2024sample} consider task-specific, non-isotropic covariate distributions as well as dependence among observations within each task. These results substantially broaden the range of admissible noise and covariate structures, but their guarantees do not capture the sharp dependence on task geometry in~\eqref{eq:intro-benchmark}, achieved by the results summarized in Table~\ref{tab:intro-summary}. Separately, \citet{tian2025learning} allow task representations to be similar rather than exactly shared; however, their clean rates require the task coefficients to be well spread and therefore do not cover the rare-direction anisotropic examples emphasized in the introduction.
	\paragraph{Matrix perturbation tools.}
	Our deterministic reduction builds on the refined singular-subspace
	perturbation inequality of \citet{cai2018rate}. The same work also provides
	stochastic guarantees for low-rank matrix denoising with i.i.d. mean-zero
	sub-Gaussian noise entries. Multitask estimators instead produce heterogeneous
	taskwise errors with potentially dependent coordinates, so those stochastic
	guarantees do not apply directly; our multitask lemma provides the additional
	control required in this setting.

	
	\subsection{Organization and Notation}
	\paragraph{Organization.}
	We first formalize the shared-subspace estimation problem in
	Section~\ref{sec:setup}, including the task covariance that captures the
	geometry of the task parameters and the \(\sin\Theta\) loss.  The
	main results are previewed in Section~\ref{sec:overview}: the general
	multitask lemma for recovering a surrogate subspace from taskwise summaries,
	the model-specific identification of that surrogate with the target subspace
	\(\mathcal V\), and the resulting rates. Section~\ref{sec:general} develops
	and proves the multitask lemma. Section~\ref{sec:instantiations} then applies
	the framework to linear regression, using whitened least-squares taskwise
	summaries, and to generalized linear models with Gaussian design, using
	first-moment summaries; the latter includes classification and additive-noise
	regression. These applications establish the stated upper and minimax lower
	bounds.
	Section~\ref{sec:discussion} concludes with the implications and limitations
	of our results. Appendix~\ref{app:gaussian-means} isolates the
	common two-term rate and its phase transition in a Gaussian mean-vector
	benchmark. Proofs are deferred to the remaining appendices.
	
	\paragraph{Notation.}
	For a positive integer \(m\), write \([m]:=\{1,\ldots,m\}\). For a matrix
	\(M\), let \(\|M\|_{\mathrm{op}}\) and \(\|M\|_{\mathrm{F}}\) denote its
	operator and Frobenius norms, respectively. If \(M\) is rectangular, its
	singular values are denoted by
	\(\sigma_1(M)\ge\sigma_2(M)\ge\cdots\); in particular, \(\sigma_k(M)\) is the
	\(k\)th largest singular value. If \(M\) is symmetric, we write
	\(\lambda_{\max}(M)\) and \(\lambda_{\min}(M)\) for its largest and smallest
	eigenvalues, and use \(\lambda_j(M)\) for the \(j\)th largest eigenvalue when
	the ordering is needed. For a subspace \(\mathcal U\), \(\Proj_{\mathcal U}\)
	denotes the orthogonal projector onto \(\mathcal U\). For a matrix \(Y\),
	\(\Proj_Y\) denotes the orthogonal projector onto the column space of \(Y\).
	We write \(\operatorname{span}(v_1,\ldots,v_m)\) for the linear span of vectors
	and \(\operatorname{span}(V)\) for the column span of a matrix \(V\).
	The notation \(a\lesssim b\) means \(a\le Cb\) for a universal or
	model-dependent constant \(C\) whose value may change from line to line; we use
	\(\gtrsim\) and \(\asymp\) analogously. We write \(\|\cdot\|_{\psi_2}\) for
	the sub-Gaussian Orlicz norm, and \(\E\) and \(\Prob\) for expectation and
	probability.
	
	\section{Preliminaries}
	\label{sec:setup}
	\paragraph{Setup.}
	{Recall from Section~\ref{subsec:problem-formulation} that, for each task
		\(t\in[T]\), we observe \(n\) i.i.d.\ pairs
		\(\{(X_{t,i},Y_{t,i}):i\in[n]\}\) drawn from a task-specific distribution
		satisfying the shared single-index model~\eqref{eq:intro-model}, with the
		datasets independent across tasks. Writing
		\(\bm\beta:=\{\beta_t:t\in[T]\}\) for the collection of task parameters,
		recall that the target subspace \(\mathcal V\) is the span of \(\bm\beta\)
		and that \(\Sigma_{\bm\beta}\) is its task covariance, as defined
		in~\eqref{eq:intro-task-covariance}. Also recall the operator sine-theta loss
		\(\sinth{\widehat{\mathcal V}}{\mathcal V}\) used to measure
		subspace-estimation error.}
	
	\paragraph{Task Statistics and Surrogate Quantities.}
	
	{As previewed in Section~\ref{subsec:our-contribution}, for each
		\(t\in[T]\), let \(\widehat\beta_t\in\R^d\) be a task statistic constructed
		from the data of task \(t\). These task statistics need not be unbiased. We use
		the collection \(\{\widehat\beta_t:t\in[T]\}\) to estimate the shared subspace
		\(\mathcal V\) and now formally define their population surrogates.}
	
	\begin{definition}[Surrogate subspace and surrogate task covariance]
		\label{def:surrogate-quantities}
		For each task $t\in[T]$, write
		$\tilde\beta_t:=\E[\widehat\beta_t]$ for the population mean of
		$\widehat\beta_t$, or \emph{surrogate task vector}, and let
		$\bm{\tilde\beta}:=\{\tilde\beta_t:t\in[T]\}$. The associated surrogate
		subspace is
		\[
		\widetilde{\mathcal V}:=\operatorname{span}(\bm{\tilde\beta}),
		\]
		which we assume has dimension $k$. Its
		surrogate task covariance is
		\[
		\Sigma_{\bm{\tilde\beta}}
		:=
		\frac1T\sum_{t=1}^T\tilde\beta_t\tilde\beta_t^\top,
		\]
		with nonzero eigenvalues
		$\widetilde\lambda_1\ge\cdots\ge\widetilde\lambda_k>0$. Let
		$\widetilde V\in\R^{d\times k}$ be an orthonormal basis for
		$\widetilde{\mathcal V}$, and let
		$\widetilde V_\perp\in\R^{d\times(d-k)}$ be an orthonormal basis for its
		orthogonal complement.
	\end{definition}
	
	\paragraph{Downstream Error.}
	
	\begingroup
	We next formalize the downstream-error interpretation of the operator
	sine-theta loss discussed in Section~\ref{subsec:problem-formulation}.
	For a future task \(T+1\), let
	\(\widehat V\in\R^{d\times k}\) be an orthonormal basis for an estimated
	subspace \(\widehat{\mathcal V}\), and let
	\(\widehat f:\R^k\to\R\) be a measurable predictor based on the learned
	features \(\widehat V^\top X_{T+1}\). Conditionally on
	\((\widehat{\mathcal V},\widehat f)\), define the downstream excess risk
	\[
	\mathcal E(\widehat f;\widehat{\mathcal V})
	:=
	\E\!\left[
	\left(Y_{T+1}-\widehat f(\widehat V^\top X_{T+1})\right)^2
	\right]
	-
	\E\!\left[
	\left(Y_{T+1}-g(\beta_{T+1}^\top X_{T+1})\right)^2
	\right],
	\]
	and let
	\[
	f^\star_{\widehat{\mathcal V}}(z)
	:=
	\E[Y_{T+1}\mid \widehat V^\top X_{T+1}=z]
	\]
	denote the optimal predictor based on the learned features.
	\endgroup
	
	\begin{proposition}[Downstream error decomposition]\label{prop:downstream-risk}
		{Let the future task satisfy the shared single-index
			model~\eqref{eq:intro-model}, with \(\beta_{T+1}\in\mathcal V\) and an
			\(L\)-Lipschitz link \(g\). Assume that
			\(\E[X_{T+1}]=0\),
			\(\operatorname{Cov}(X_{T+1})=\Sigma_X\), and
			\(\E[Y_{T+1}^2]<\infty\). Assume also that
			\((\widehat{\mathcal V},\widehat f)\) is independent of the test pair
			\((X_{T+1},Y_{T+1})\). Then}
		\[
		\mathcal E(\widehat f;\widehat{\mathcal V})
		\le
		\E\!\left[
		\left(
		\widehat f(\widehat V^\top X_{T+1})
		-
		f^\star_{\widehat{\mathcal V}}(\widehat V^\top X_{T+1})
		\right)^2
		\right]
		+
		L^2\|\Sigma_X\|_{\mathrm{op}}\|\beta_{T+1}\|_2^2\,
		\sinth{\widehat{\mathcal V}}{\mathcal V}^2.
		\]
	\end{proposition}
	{Thus, the downstream excess risk decomposes into prediction error within
		the learned representation and representation error controlled by
		\(\sinth{\widehat{\mathcal V}}{\mathcal V}^2\). The following remark
        makes this bound concrete for downstream least-squares regression.}

	\begin{remark}[Downstream least-squares prediction]
        \label{rem:downstream-least-squares}
        Consider a downstream task \(Y=\beta^\top X+\xi\), independent of the
        data used to estimate \(\widehat{\mathcal V}\), where
        \(X\sim N(0,I_d)\), \(\beta\in\mathcal V\), and \(\xi\) is independent
        of \(X\), has mean zero, and has variance \(\sigma^2\).
        Fit least squares using the \(k\) learned features and
        \(m\ge2k+2\) fresh i.i.d.\ samples. Conditional on
        \(\widehat{\mathcal V}\), its expected excess squared prediction loss
        is bounded by
        \[
        C\left[
        \sigma^2\frac{k}{m}
        +\|\beta\|_2^2\sinth{\widehat{\mathcal V}}{\mathcal V}^2
        \right],
        \]
        where the expectation is over the downstream training and test data.
        For bounded noise variance and parameter norm, this gives the rate
        stated in the introduction. The \(k/m\) term comes from the
        least-squares calculation; Proposition~\ref{prop:downstream-risk}
        controls the representation contribution.
    \end{remark}

    \section{Overview of Main Results}
	\label{sec:overview}
	
	This section presents the main results starting from a general subspace-recovery guarantee to model-specific minimax rates.
	We first state
	{the multitask lemma, which gives a high-probability bound for recovering the
		surrogate subspace from a stacked matrix of task statistics.} We then
	instantiate the lemma for linear regression and generalized linear models with
	Gaussian design, identify the surrogate and target subspaces directly in each
	model, and state the corresponding minimax bounds for classification and
	additive-noise regression.

	\subsection{General Multitask Lemma}\label{ssec:overview-mtl}
	{The multitask lemma gives sufficient conditions under which generic task
		statistics yield fast recovery of their shared surrogate subspace.}
	
	\paragraph{Setting:} Recall from Section~\ref{sec:setup} the surrogate task vectors
	$\tilde\beta_t=\E[\hat\beta_t]$, the surrogate subspace
	$\widetilde{\mathcal V}$, and the surrogate covariance
	$\Sigma_{\bm{\tilde\beta}}$. Its smallest nonzero eigenvalue
	\(\widetilde\lambda_k\) measures the strength of the weakest direction represented
	by the population surrogates.
	{Recall from Section~\ref{subsec:our-contribution} the population
		surrogate matrix \(\widetilde A\), its perturbed counterpart \(\widehat A\),
		and the task-wise errors
		\(\varepsilon_t:=\widehat\beta_t-\widetilde\beta_t\). Stacking these errors
		into the perturbation matrix \(E\), we have}
	{
	\begin{equation}\label{eq:mtl-matrix-decomposition}
	\widehat A=\widetilde A+E,
	\qquad
	E:=
	[\varepsilon_1, \hdots,
	\varepsilon_T]^\top.
	\end{equation}
	}
	{We now formally define the averaged error covariance previewed in
		Section~\ref{subsec:our-contribution}.}
	\begin{definition}[Averaged error covariance]
		\label{def:averaged-error-covariance}
		Write $\bm\varepsilon:=\{\varepsilon_t:t\in[T]\}$. The averaged error
		covariance is
		\[
		\Sigma_{\bm\varepsilon}
		:=
		\frac1T\sum_{t=1}^T \E[\varepsilon_t\varepsilon_t^\top].
		\]
	\end{definition}
	\paragraph{Estimator:} Let $\widehat V_k^{(s)}$ be an orthonormal basis for the top-$k$ right singular
	subspace of $\widehat A$, and define the surrogate-subspace estimator
	\begin{equation}\label{eq:surrogate-subspace-estimator}
	\widehat{\mathcal V}^{(s)}:=\operatorname{span}(\widehat V_k^{(s)}).
	\end{equation}
	The following conditions control the stochastic fluctuations of the errors and the
	population eigengap.

	\begin{condition}[Multitask lemma conditions]\label{cond:surrogate-recovery}
		Assume the following.
		\begin{enumerate}
			\item {The error vectors \(\varepsilon_t\) are independent across
			\(t\in[T]\), and there exists \(\sigma>0\) such that
			\[
			\|\langle \varepsilon_t,a\rangle\|_{\psi_2}
			\le C\sigma\|a\|_2
			\qquad
			\text{for all \(t\in[T]\) and \(a\in\R^d\)}.
			\]}
			\item Define
			\[
			\rho:=\lambda_k\!\left(
			\Proj_{\widetilde{\mathcal V}}
			\bigl[\Sigma_{\bm{\tilde\beta}}+\Sigma_{\bm\varepsilon}\bigr]
			\Proj_{\widetilde{\mathcal V}}
			\right),
			\qquad
			\nu:=\lambda_{\max}\!\left(
			\Proj_{\widetilde{\mathcal V}^\perp}
			\Sigma_{\bm\varepsilon}
			\Proj_{\widetilde{\mathcal V}^\perp}
			\right).
			\]
			Assume the population gap condition
			$\rho-\nu\ge \gamma_0\widetilde\lambda_k$ for a fixed constant
			$\gamma_0\in(0,1]$.
		\end{enumerate}
	\end{condition}
	
	{Recall from~\eqref{eq:mixing} the normalized error-covariance mixing
		parameter \(\delta_1\). Because the errors are centered, the numerator in its
		definition can equivalently be written as
		\[
		\frac1T\sum_{t=1}^T
		\operatorname{Cov}\!\left(
		\Proj_{\widetilde{\mathcal V}}\varepsilon_t,\,
		\Proj_{\widetilde{\mathcal V}^{\perp}}\varepsilon_t
		\right).
		\]
	}
	
	The dimensionless quantity \(\delta_1\) captures the population bias
		induced by error-covariance mixing across the surrogate subspace and its
		orthogonal complement. It vanishes when the averaged error covariance is
		isotropic, or more generally block diagonal with respect to
		\(\widetilde{\mathcal V}\oplus\widetilde{\mathcal V}^{\perp}\). Across the
		examples considered in this paper, this cross-covariance is either zero (or
		small) because of the model structure, or is made zero (or small) by
		preprocessing.
		In particular, Gaussian rotational symmetry gives \(\delta_1=0\)
		for classification and additive-noise regression under isotropic Gaussian
		design. Whitening also gives \(\delta_1=0\) in linear regression
		(Section~\ref{ssec:linear-reg-overview}) and in classification with a
		common non-isotropic Gaussian covariance
		(Appendix~\ref{sec:nonlinear-general-design}).
		
	
	\begin{theorem}[Multitask lemma]\label{thm:overview-mtl}\label{thm:mtl-main}
		Under Condition~\ref{cond:surrogate-recovery}, for every \(\tau\ge1\), with probability
		at least $1-e^{-\tau}$,
		\[
		\sinth{\widehat{\mathcal V}^{(s)}}{\widetilde{\mathcal V}}
		\le
		C\left(
		\left[
		\sigma\sqrt{\frac{(d+\tau)\rho}{T\widetilde\lambda_k^2}}
		+
		\sigma^2\sqrt{\frac{d+\tau}{T\widetilde\lambda_k^2}}
		+\sigma^2\cdot
		\frac{d+\tau}{T\widetilde\lambda_k}
		+
		{\delta_1}
		\right]\wedge 1
		\right).
		\]
		Here and below, \(C\) may depend on the fixed sub-Gaussian norm constant and
		on \(\gamma_0\).
	\end{theorem}
	
	{The theorem highlights two central principles for designing task
		statistics: the error covariance should have weak cross-subspace interaction,
		as measured by \(\delta_1\), and the task statistics should preserve the
		surrogate signal strength, with
		\(\widetilde\lambda_k\asymp\lambda_k\). Together with the remaining
		model-specific conditions, these properties yield the
		benchmark rate in~\eqref{eq:intro-benchmark}.}
	
	\begin{remark}[From the surrogate to the target subspace]
		\label{rem:surrogate-to-target}
		{The multitask lemma estimates the surrogate subspace
			\(\widetilde{\mathcal V}\), which need not equal the target subspace
			\(\mathcal V\). Each application must therefore identify how to recover
			\(\mathcal V\) from \(\widetilde{\mathcal V}\). In linear regression, this is
			achieved by the inverse whitening map defined in
			Section~\ref{ssec:linear-reg-overview}; in the generalized linear models,
			the surrogate task vectors remain aligned with the target task vectors, so
			\(\widetilde{\mathcal V}=\mathcal V\).}
	\end{remark}
	
	{We next instantiate the multitask lemma for linear regression and
		generalized linear models, verifying the conditions of the multitask lemma and
		relating the resulting surrogate subspace to the target subspace in each
		setting.}
	
	\subsection{Instantiations of the Multitask Lemma}\label{ssec:overview-instantiations}
	{We now instantiate the multitask lemma in three settings. For linear
		regression, we construct whitened least-squares task statistics. For
		classification and additive-noise regression with Gaussian covariates, we use
		first-moment task statistics, together with truncation in the additive-noise
		setting. In each case, we verify the conditions of the multitask lemma and
		relate the surrogate subspace to the target subspace.}
	
	
	\subsubsection{Linear Regression}
	\label{ssec:linear-reg-overview}
	\paragraph{Setting.}
	{As previewed in Section~\ref{subsec:our-contribution}, our first
		application is linear regression. We now specify the observation model:}
	\[
	Y_{t,i}=X_{t,i}^\top\beta_t+\xi_{t,i},
	\qquad
	\beta_t\in\mathcal V.
	\]
	{The noises \(\xi_{t,i}\) are independent of the designs and independent
		across tasks and samples, have mean zero and common variance
		\(\sigma_\xi^2\), and satisfy
		\(\|\xi_{t,i}\|_{\psi_2}\le K_\xi\sigma_\xi\) for a fixed numerical constant
		\(K_\xi\).}
	We assume $n\ge d$ and that, for fixed constants
	$0<\kappa_{\min}<\kappa_{\max}$, the empirical covariance of each task is
	uniformly well conditioned:
	\begin{equation}\label{eq:design-condition}
		\kappa_{\min} I_d
		\preceq \frac1n\sum_{i=1}^n X_{t,i}X_{t,i}^\top
		\preceq \kappa_{\max} I_d
		\qquad\text{for all }t\in[T].
	\end{equation}
	
	\paragraph{Estimator.}
	{As previewed in Section~\ref{subsec:our-contribution}, we whiten the
		ordinary least-squares task statistics. For each task \(t\), define}
	\begin{equation}\label{eq:ols-task-statistic}
	\widehat\beta_t^{(\mathrm{OLS})}
	:=
	\left(\sum_{i=1}^n X_{t,i}X_{t,i}^\top\right)^{-1}
	\sum_{i=1}^n X_{t,i}Y_{t,i}.
	\end{equation}
	{We formally define the average inverse empirical covariance by}
	\begin{equation}\label{eq:linear-whitening-matrix}
	\widehat S
	:=
	\frac1T\sum_{t=1}^T
	\left(\frac1n\sum_{i=1}^n X_{t,i}X_{t,i}^\top\right)^{-1},
	\end{equation}
	{and use the whitened task statistic}
	\begin{equation}\label{eq:linear-whitened-task-statistic}
	\widehat\beta_t
	:=
	\widehat S^{-1/2}\widehat\beta_t^{(\mathrm{OLS})}.
	\end{equation}
	
	\begin{remark}[Why whitening is needed]
		\label{rem:linear-whitening-motivation}
		{As discussed in Section~\ref{subsec:our-contribution}, the raw OLS
			task statistics can have an averaged error covariance that mixes
			\(\mathcal V\) with \(\mathcal V^\perp\), producing a nonnegligible
			\(\delta_1\). Whitening makes the averaged error covariance isotropic and
			hence sets \(\delta_1=0\). The detailed calculation, including an example
			in which \(\delta_1\) remains bounded away from zero without whitening, is
			given in Section~\ref{sec:linear}; see also
			Remark~\ref{rem:ols-nonvanishing-delta}.}
	\end{remark}
	
	{Applying the spectral construction of
		Section~\ref{ssec:overview-mtl} to these whitened task statistics yields the
		surrogate-subspace estimator
		\(\widehat{\mathcal V}^{(s)}
		=\operatorname{span}(\widehat V_k^{(s)})\).
		We then undo the whitening transformation and define the final target-subspace
		estimator by}
		\begin{equation}\label{eq:linear-target-subspace-estimator}
		\widehat{\mathcal V}
		:=
		\operatorname{span}\!\left(
		\widehat S^{1/2}\widehat V_k^{(s)}
		\right).
		\end{equation}
	
	\paragraph{Verification of the multitask lemma conditions.}
	{Conditionally on the designs, the population surrogate is
		\[
		\widetilde\beta_t
		:=
		\E[\widehat\beta_t\mid X_1,\ldots,X_T]
		=
		\widehat S^{-1/2}\beta_t.
		\]
		Consequently,
		\begin{equation}\label{eq:linear-surrogate-transformation}
		\widetilde{\mathcal V}=\widehat S^{-1/2}\mathcal V,
		\qquad
		\mathcal V=\widehat S^{1/2}\widetilde{\mathcal V}.
		\end{equation}
		Thus the final estimator simply undoes the population transformation induced
		by whitening.  The design condition~\eqref{eq:design-condition} makes
		\(\widehat S^{1/2}\) well conditioned, so applying it changes the
		sine-theta error by at most a fixed factor.  It also gives
		\[
		\widetilde\lambda_k
		=
		\lambda_k\!\left(
		\widehat S^{-1/2}\Sigma_{\bm\beta}\widehat S^{-1/2}
		\right)
		\asymp
		\lambda_k(\Sigma_{\bm\beta}).
		\]
		Finally, the averaged covariance of the whitened errors is
		\((\sigma_\xi^2/n)I_d\), so \(\delta_1=0\).}
	
	\begin{theorem}[Linear regression upper bound]\label{thm:overview-linear}\label{thm:linear}
		Condition on the design matrices $X_1,\ldots,X_T$, and assume
		\eqref{eq:design-condition}. Then, for every \(\tau\ge1\), with
		probability at least $1-e^{-\tau}$,
		\[
		\sinth{\widehat{\mathcal V}}{\mathcal V}
		\le
		C\left(
		\left[
		\sigma_\xi\sqrt{\frac{d+\tau}{Tn\lambda_k}}
		+
		\sigma_\xi^2\sqrt{\frac{d+\tau}{Tn^2\lambda_k^2}}
		\right]\wedge 1
		\right),
		\]
		where $C$ depends only on the design-conditioning constants
		\(\kappa_{\min},\kappa_{\max}\).
	\end{theorem}
	
	For the corresponding minimax statement, define
	\[
	\mathcal P(\lambda_k)
	:=
	\left\{
	\begin{array}{l}
		(\mathcal V,\beta_1,\dots,\beta_T):
		\mathcal V\subseteq\R^d,\ \dim(\mathcal V)=k,\\[0.3em]
		\beta_t\in\mathcal V,\ \|\beta_t\|_2\le 1\ \text{for all }t\in[T],\,
		\lambda_k\!\left(\Sigma_{\bm \beta}\right)
		\ge \lambda_k
	\end{array}
	\right\}.
	\]
	{Here \(\lambda_k\) is the prescribed signal-strength threshold defining
		the class, whereas \(\lambda_k(\Sigma_{\bm\beta})\) denotes the realized
		smallest nonzero eigenvalue of $\Sigma_{\bm\beta}$.}

	\begin{corollary}[Linear regression minimax rate]\label{cor:overview-linear-minimax}
		Let \(\mathfrak X(\kappa_{\min},\kappa_{\max})\) be the class of fixed designs satisfying
		\eqref{eq:design-condition}, and let \(\mathfrak E_{\rm sg}\) be the class of
		errors that are independent across tasks and samples, have mean zero and a
		common variance \(\sigma_\xi^2\le1\), and satisfy
		\(\|\xi_{t,i}\|_{\psi_2}\le K_\xi\sigma_\xi\) for a fixed numerical constant
		\(K_\xi\).
		Assume that \(\mathcal P(\lambda_k)\) is nonempty.
		Assume $k\ge c_0\log T$, $d\ge(1+\rho_1)k$, $T\ge(1+\rho_2)k$, and
		\(n\ge C_1d\). Assume also that \(T\le e^{c_1 n}\), for a sufficiently
		small constant \(c_1>0\). Then
		\[
		\inf_{\widehat{\mathcal V}}\ 
		\sup_{\substack{
				X_{1:T}\in\mathfrak X(\kappa_{\min},\kappa_{\max}),\,
				\xi\in\mathfrak E_{\rm sg}\\
				(\mathcal V,\beta_1,\dots,\beta_T)\in \mathcal P(\lambda_k)}}
		\E\bigl[\sinth{\widehat{\mathcal V}}{\mathcal V}\bigr]
		\asymp
		\left[\sqrt{\frac{d}{Tn\lambda_k}} + \sqrt{\frac{d}{Tn^2\lambda_k^2}}\right]\wedge 1,
		\]
		up to constants depending only on the fixed model constants. 
	\end{corollary}
	
	The upper bound is attained by the whitened spectral estimator above uniformly
	over deterministic designs in
	\(\mathfrak X(\kappa_{\min},\kappa_{\max})\) and errors in
	\(\mathfrak E_{\rm sg}\). The lower bound follows by restricting this larger
	class to Gaussian errors and to a Gaussian-design instance satisfying
	\eqref{eq:design-condition}, using Theorem~5.1 of \citet{niu2024learning}.
	This closes the gap left by prior upper bounds, which additionally scaled with $\lambda_1/\lambda_k$ and could blow up when $\lambda_k = o(\lambda_1)$. Notably, the minimax risk above depends on the task covariance spectrum only through $\lambda_k$.
	
	\begin{remark}[Comparison with \cite{niu2024learning}]
		\label{rem:niu-linear-comparison}
		{\citet{niu2024learning} estimate the span of the response--covariate
			cross-moments
			\[
			m_t:=\E[Y_{t,i}X_{t,i}]
			=\Sigma_{X,t}\beta_t,
			\qquad
			\Sigma_{X,t}:=\E[X_{t,i}X_{t,i}^\top].
			\]
			When the covariances \(\Sigma_{X,t}\) vary across tasks, the resulting
			subspace \(\operatorname{span}(m_1,\ldots,m_T)\) need not equal the target
			subspace \(\mathcal V=\operatorname{span}(\beta_1,\ldots,\beta_T)\).
			This distinction matters for downstream prediction. Since a downstream task
			parameter \(\beta\) lies in \(\mathcal V\), its regression function depends
			on \(x\) only through \(\Proj_{\mathcal V}x\). By contrast,
			\(\operatorname{span}(m_1,\ldots,m_T)\) need not contain the task parameters,
			so projection onto this subspace may discard directions that are relevant for
			downstream prediction.
			Their first-moment estimator does not require well-conditioned empirical
			design matrices because it targets the different population quantity
			\(\operatorname{span}(m_1,\ldots,m_T)\). Our estimator instead reconstructs
			the task parameters through least squares and therefore requires the
			full-rank design condition~\eqref{eq:design-condition}. This allows it to
			target \(\mathcal V\) directly and attain the optimal rate, closing the gap
			left by previous upper bounds that were loose in their dependence on the
			geometry of the task parameters.}
	\end{remark}
	
	\subsubsection{Generalized Linear Models with Gaussian Design}
	\label{ssec:gsi-overview}
	
	\paragraph{Setting.}
	{As previewed in Section~\ref{subsec:our-contribution}, our second
		application concerns nonlinear models with isotropic Gaussian covariates. We
		assume}
	\begin{equation}\label{eq:gsi-model}
	X_{t,i}\sim N(0,I_d),
	\qquad
	\E[Y_{t,i}\mid X_{t,i}]
	=
	g(X_{t,i}^\top\beta_t),
	\qquad
	Y_{t,i}\mid X_{t,i}
	\stackrel d=
	Y_{t,i}\mid X_{t,i}^\top\beta_t,
	\end{equation}
	{where \(g:\R\to\R\) is shared across tasks,
		\(\beta_t\in\mathcal V\), and \(\|\beta_t\|_2\le1\) for every \(t\in[T]\).
		This setting includes binary classification and additive-noise regression.}

    Appendix~\ref{sec:task-dependent-nonlinear-links} extends the classification
    and additive-noise upper bounds to task-dependent links \(g_t\), under
    uniform alignment and regularity conditions. The same first-moment estimator
    applies without knowledge of the links and retains the respective rates.
	
	\paragraph{Estimator.}
	{As previewed in Section~\ref{subsec:our-contribution}, we formally
		define the first-moment task statistic by}
	\begin{equation}\label{eq:gsi-first-moment-estimator}
		\widehat\beta_t
		:=
		\frac1n\sum_{i=1}^n Y_{t,i}X_{t,i}.
	\end{equation}
	{This task statistic does not require knowledge of \(g\); the link
		assumptions enter only the analysis and the constants in the guarantees.
		Applying the spectral construction of Section~\ref{ssec:overview-mtl} gives
		the surrogate-subspace estimator \(\widehat{\mathcal V}^{(s)}\). Once the
		alignment established below shows that
		\(\widetilde{\mathcal V}=\mathcal V\), this is also the target-subspace
		estimator, and we write}
	\begin{equation}\label{eq:gsi-subspace-estimator}
		\widehat{\mathcal V}
		:=
		\widehat{\mathcal V}^{(s)}.
	\end{equation}
	\paragraph{Population alignment.}
	{Let \(Z\sim N(0,1)\) and define
		$
		a_t:=\E[g'(\|\beta_t\|_2Z)].
		$
		Stein's identity gives
		\begin{equation}\label{eq:gsi-population-alignment}
		\widetilde\beta_t
		=
		\E[\widehat\beta_t]
		=
		a_t\beta_t.
		\end{equation}
		Thus each surrogate task vector is aligned with its corresponding target task
		vector. Define the uniform alignment constant}
	\begin{equation}\label{eq:ag-def}
		a_g
		:=
		\inf_{\beta\in B_2(1)} \bigl|\E[g'(\|\beta\|_2 Z)]\bigr|.
	\end{equation}
	{Throughout this section, we assume \(a_g>0\); see
		Remark~\ref{rem:alignment-examples} for the logistic and Cauchit links. Since
		\(\|\beta_t\|_2\le1\), we have \(|a_t|\ge a_g\), and consequently
		\[
		\widetilde{\mathcal V}=\mathcal V,
		\qquad
		\widetilde\lambda_k
		\ge
		a_g^2\lambda_k(\Sigma_{\bm\beta}).
		\]
	}
	
	\paragraph{Classification.}
	{Take \(Y_{t,i}\in\{0,1\}\) with
		\[
		\Prob(Y_{t,i}=1\mid X_{t,i})
		=
		g(X_{t,i}^\top\beta_t).
		\]
		The bounded Bernoulli responses yield uniformly sub-Gaussian errors for the
		first-moment task statistics, while Gaussian rotational symmetry gives
		\(\delta_1=0\). Together with the alignment and signal-strength properties
		established above, this verifies the conditions of the multitask lemma without
		truncation and yields the following recovery guarantee.}
	
	\par\vspace{\baselineskip}

	\begin{theorem}[Classification upper bound]\label{thm:overview-cls}\label{thm:gsi}
		Let \(\widehat{\mathcal V}\) be the first-moment spectral estimator
			defined in~\eqref{eq:gsi-first-moment-estimator}--\eqref{eq:gsi-subspace-estimator}.
			Assume \(a_g>0\) (as defined in~\eqref{eq:ag-def}), \(g\in C^2(\R)\) with bounded first and second derivatives, and
			\(n\ge n_{\mathrm{cls}}\), where \(n_{\mathrm{cls}}\ge2\) is a fixed constant depending only on
			\(g\). Then, for every \(\tau\ge1\), with probability
		at least $1-e^{-\tau}$,
		\[
		\sinth{\widehat{\mathcal V}}{\mathcal V}
		\le
		Ca_g^{-2}\left(
		\left[
		\sqrt{\frac{d+\tau}{Tn\lambda_k}}
		+
		\sqrt{\frac{d+\tau}{Tn^2\lambda_k^2}}
		\right]\wedge1
		\right),
		\]
		where $C$ depends only on fixed link constants.
	\end{theorem}
	
	Throughout the classification and additive-noise minimax results, we assume
	that \(\mathcal P(\lambda_k)\) is nonempty. In particular, the
	standing convention \(\lambda_k>0\) and Lemma~\ref{lem:shared-lb-frame} imply
	\(T\ge k\) and \(0<\lambda_k\le 1/k\).
	{To state the classification minimax risk, we impose the following
		condition on the link function.}
	\begin{definition}[Admissible link functions]\label{def:admissible-link}
		A link function $g:\R\to(0,1)$ is called \emph{admissible} if it is twice
		continuously differentiable and satisfies
		\[
		\sup_{t\in\R}\left|\frac{g'(t)}{g(t)(1-g(t))}\right|\le L_1,
		\qquad
		\sup_{t\in\R}\left|\frac{g''(t)}{g(t)(1-g(t))}\right|\le L_2.
		\]
	\end{definition}
	Examples include the logistic link $g(u)=1/(1+e^{-u})$ and the Cauchit link
	$g(u)=\tfrac12+\tfrac1\pi\arctan(u)$.
	\begin{corollary}[Classification minimax rate]\label{cor:overview-cls-minimax}
		{Let \(g\) be admissible and assume the conditions of
			Theorem~\ref{thm:gsi}. Assume \(1\le k\le d/2\). For the matching lower
			bound, assume in addition that}
		$n\ge C_0k$ and $Te^{-c_0n}\le c_0d$, for constants
		$c_0,C_0>0$ depending only on $L_1,L_2$. Then
		\[
		\inf_{\widehat{\mathcal V}}\ 
		\sup_{(\mathcal V,\beta_1,\dots,\beta_T)\in\mathcal P(\lambda_k)}
		\E\bigl[\sinth{\widehat{\mathcal V}}{\mathcal V}\bigr]
		\asymp
		\left[
		\sqrt{\frac{d}{Tn\lambda_k}}
		+
		\sqrt{\frac{d}{Tn^2\lambda_k^2}}
		\right]\wedge1,
		\]
		up to constants depending only on $a_g,L_1,L_2$. The upper bound is attained
		by the first-moment spectral estimator above and only uses the assumptions of
		Theorem~\ref{thm:overview-cls}; the additional conditions are used for the
		Fano-type local-packing lower bound of Theorem~\ref{thm:gsi-lower}.
	\end{corollary}
	
	This is, to our knowledge, the first sharp minimax rate for shared-subspace
	recovery in this classification setting, with the same functional form as the
	linear regression risk of Corollary~\ref{cor:overview-linear-minimax}.
	
	\paragraph{Additive-noise regression.}
	{As previewed in Section~\ref{subsec:our-contribution}, we now specialize
		the common nonlinear setting to additive-noise regression:}
	\[
	Y_{t,i}=g(X_{t,i}^\top\beta_t)+\xi_{t,i},
	\qquad
	\xi_{t,i}\perp X_{t,i},
	\qquad
	\E[\xi_{t,i}]=0,
	\qquad
	\|\xi_{t,i}\|_{\psi_2}\le K_\xi.
	\]
	{We assume \(g\in C^2(\R)\) with}
	\begin{equation}\label{eq:bounded-derivatives}
		\|g'\|_\infty\le L_1,
		\qquad
		\|g''\|_\infty\le L_2.
	\end{equation}
	{As briefly described in Section~\ref{subsec:our-contribution}, unlike
		in classification, the products \(Y_{t,i}X_{t,i}\) need not be sub-Gaussian,
		so the multitask lemma cannot be applied directly to the first-moment task
		statistics. We instead apply it to a response-truncated version.
		Proposition~\ref{prop:response-transformation-alignment} shows that response
		truncation preserves alignment of the surrogate vectors with the task
		parameters, while Gaussian rotational symmetry gives \(\delta_1=0\). For the chosen truncation level,
		their covariance has rank \(k\) and smallest nonzero eigenvalue at least
		\(a_g^2\lambda_k/2\), so their span remains \(\mathcal V\). Finally, the
		truncated and original first-moment spectral estimators agree with high
		probability, transferring the resulting guarantee to the original estimator
		at the cost of logarithmic factors.}
	
	\begin{theorem}[Additive-noise upper bound]\label{thm:overview-add-upper}\label{thm:gsi-subg}
		Let \(\widehat{\mathcal V}\) be the first-moment spectral estimator
			defined in~\eqref{eq:gsi-first-moment-estimator}--\eqref{eq:gsi-subspace-estimator}.
			Assume
			\eqref{eq:bounded-derivatives},  $a_g>0$ (as defined in~\eqref{eq:ag-def}), and \(n\ge n_{\mathrm{add}}\), where
			\(n_{\mathrm{add}}\) is the finite model-dependent threshold defined
			in~\eqref{eq:n-add-definition}. Set
		$\Gamma_{n,T,\lambda_k}:=\log\!\left(\frac{e nT}{\lambda_k}\right).
		$
		Then, for every \(\tau\ge1\), with probability at least \(1-e^{-\tau}\),
		\[
		\sinth{\widehat{\mathcal V}}{\mathcal V}
		\le
		C\left(
		\left[
		\frac{K_Y\sqrt{\Gamma_{n,T,\lambda_k}+\tau}}{a_g}
		\sqrt{\frac{d+\tau}{Tn\lambda_k}}
		+
		\frac{K_Y^2(\Gamma_{n,T,\lambda_k}+\tau)}{a_g^2}
		\sqrt{\frac{d+\tau}{Tn^2\lambda_k^2}}
		\right]\wedge1
		\right),
		\]
		where \(K_Y:=C_0\bigl(|g(0)|+L_1+K_\xi\bigr)\) for a universal constant
		\(C_0>0\), and \(C\) depends only on fixed model constants.
	\end{theorem}
	
	Combining this high-probability upper bound with the Gaussian additive-response
	lower bound yields the following minimax comparison, up to the logarithmic
	factor induced by truncation.
	
	\begin{corollary}[Additive-noise minimax bounds]\label{cor:overview-add-minimax}
		Fix a Gaussian additive-noise model with
		\(\xi_{t,i}\sim N(0,\sigma^2)\). Assume the conditions of
		Theorem~\ref{thm:overview-add-upper} hold, with \(n\ge C_0k\),
		\(Te^{-c\sqrt n}\le c_0d\), and \(1\le k\le d/2\), for constants
		\(c,c_0,C_0>0\) depending only on the fixed model constants. Then
		{
			\[
			\mathfrak R_{\mathrm{add}}(\lambda_k)
			:=
			\inf_{\widehat{\mathcal V}}\ 
			\sup_{(\mathcal V,\beta_1,\dots,\beta_T)\in
				\mathcal P(\lambda_k)}
			\E\bigl[\sinth{\widehat{\mathcal V}}{\mathcal V}\bigr]
			\]
			satisfies
			\[
			c\left(
			\left[
			\sqrt{\frac{d}{Tn\lambda_k}}
			+
			\sqrt{\frac{d}{Tn^2\lambda_k^2}}
			\right]\wedge1
			\right)
			\le
			\mathfrak R_{\mathrm{add}}(\lambda_k)
			\le
			C\left(
			\left[
			\sqrt{\Gamma_{n,T,\lambda_k}}
			\sqrt{\frac{d}{Tn\lambda_k}}
			+
			\Gamma_{n,T,\lambda_k}
			\sqrt{\frac{d}{Tn^2\lambda_k^2}}
			\right]\wedge1
			\right),
			\]
			where
			\(\Gamma_{n,T,\lambda_k}=\log(e nT/\lambda_k)\).
			The constants may depend on
			\(a_g,L_1,L_2,K_\xi,\sigma\), but not on \(n,T,d,\lambda_k\). 
		}
	\end{corollary}
	
    The additional conditions \(n\ge C_0k\), \(Te^{-c\sqrt n}\le c_0d\),
    and \(k\le d/2\) are used for the lower bound; the upper bound requires
    only the assumptions of Theorem~\ref{thm:overview-add-upper}.

	{Thus the additive-noise result is near-minimax up to the displayed logarithmic
		factors. The upper bound is attained by the first-moment spectral estimator,
		while the lower bound is Theorem~\ref{thm:gsi-additive-lower}.}
	
	{Taken together, all three settings share the same two-term rate,
		$\sqrt{d/(Tn\lambda_k)}+\sqrt{d/(Tn^2\lambda_k^2)}$, with a transition at
		$n\lambda_k=1$. The resulting bounds depend on the task geometry
		through $\lambda_k$ alone, closing the gap left by previous upper bounds
		that are loose in the spectrum of $\Sigma_{\bm\beta}$.}
	
	\section{Analysis of the Multitask Lemma}\label{sec:general}
	
	Having stated the multitask lemma as a statistical recovery result in
	Section~\ref{ssec:overview-mtl}, we now explain its proof. Matrix perturbation
	provides the deterministic reduction, while the main statistical work is to
	develop concentration bounds for the heterogeneous taskwise error matrix that
	arises in multitask learning.
	
	\paragraph{Deterministic reduction.}
	Recall the task statistics \(\widehat\beta_t\) and their population surrogate
	quantities from Definition~\ref{def:surrogate-quantities}, the stacked matrix
	decomposition \(\widehat A=\widetilde A+E\)
	from~\eqref{eq:mtl-matrix-decomposition}, the averaged error covariance
	\(\Sigma_{\bm\varepsilon}\) from
	Definition~\ref{def:averaged-error-covariance}, and the surrogate-subspace
	estimator \(\widehat{\mathcal V}^{(s)}\)
	from~\eqref{eq:surrogate-subspace-estimator}. The purpose of this section
	is to show how the heterogeneous taskwise errors affect the surrogate signal
	and, in turn, to establish recovery by controlling the resulting signal and
	error blocks. Let \(\widetilde V\) be an orthonormal basis of
	\(\widetilde{\mathcal V}\), and let \(\widetilde V_\perp\) be an orthonormal
	basis of \(\widetilde{\mathcal V}^{\perp}\). The two block matrices entering
	the perturbation argument are
	\[
	Y:=\widehat A\widetilde V,
	\qquad
	Y_\perp:=\widehat A\widetilde V_\perp.
	\]
	Whenever
	\(\sigma_k^2(Y)>\sigma_{k+1}^2(\widehat A)\), Proposition~1 of
	\citet{cai2018rate} gives the refined perturbation bound
	\begin{equation}\label{eq:mtl-deterministic-reduction}
		\sinth{\widehat{\mathcal V}^{(s)}}{\widetilde{\mathcal V}}
		\lesssim
		\frac{\sigma_k(Y)\,\|\Proj_YY_\perp\|_{\mathrm{op}}}
		{\sigma_k^2(Y)-\sigma_{k+1}^2(\widehat A)}.
	\end{equation}
	The deterministic inequality~\eqref{eq:mtl-deterministic-reduction} is the
	ingredient borrowed from \citet{cai2018rate}. Our contribution begins with the
	probabilistic step needed to apply it in the multitask setting: deriving
	concentration bounds for independent but non-identically distributed taskwise
	error rows whose coordinates may be dependent. Specifically, we lower bound
	the spectral gap in the denominator and upper bound the cross-block term
	\(\|\Proj_YY_\perp\|_{\mathrm{op}}\) in the numerator.
	
	\begin{remark}[Population eigengap]
		Recall the quantities \(\rho\) and \(\nu\) from the multitask lemma conditions
		in Condition~\ref{cond:surrogate-recovery}. In terms of \(Y\) and
		\(Y_\perp\), they admit the equivalent block representations
		\[
		\rho=\lambda_{\min}\!\left(\frac1T\E[Y^\top Y]\right),
		\qquad
		\nu = \lambda_{\max}\!\left(\frac1T\E[Y_\perp^\top Y_\perp]\right).
		\]
		Consequently, the normalized population counterpart of the empirical
		denominator \(\sigma_k^2(Y)-\sigma_{k+1}^2(\widehat A)\) is
		\(\rho-\nu\). The condition
		\(\rho-\nu\gtrsim\widetilde\lambda_k\) is the population separation that ensures,
		after concentration, a positive empirical denominator of order
		\(T\widetilde\lambda_k\).
	\end{remark}
	
	\paragraph{Concentration of matrix functionals.}
	To motivate the concentration argument, we first contrast the multitask setting
	with the entrywise-independent matrix-denoising setting.
	\begin{remark}[Comparison with i.i.d.\ matrix denoising]
		In the matrix-denoising verification of \citet{cai2018rate}, the perturbation
		matrix has entrywise i.i.d.\ sub-Gaussian entries. After reducing an operator
		norm to fixed directions, a typical scalar quantity is
		\[
		u^\top E^\top E u
		=
		\|E u\|_2^2
		=
		\operatorname{vec}(E)^\top
		(uu^\top\otimes I_T)
		\operatorname{vec}(E).
		\]
		The entrywise i.i.d.\ assumption gives independent coordinates of
		\(\operatorname{vec}(E)\), which is what allows Hanson--Wright type
		control of this quadratic form. Under the multitask lemma conditions, this
		coordinate independence need not hold: each row is a taskwise error
		\(\varepsilon_t\), whose coordinates may be dependent and whose distribution
		may vary with \(t\). We therefore use independence across tasks and
		Bernstein-type bounds for the resulting row-wise sums.
	\end{remark}
	
	Accordingly, the proof reduces to three high-probability estimates
	for \(Y^\top Y\), \(Y_\perp^\top Y_\perp\), and \(Y^\top Y_\perp\), adapted to
	the row-wise, potentially anisotropic structure of the taskwise errors.
	The first keeps the empirical signal block close to its population counterpart.
	
	\begin{lemma}[Signal-block concentration]
		Under the multitask lemma conditions in
		Condition~\ref{cond:surrogate-recovery}, for every \(\tau\ge1\), with probability at least
		\(1-e^{-\tau}\),
		\[
		\sigma_k^2(Y)
		\ge
		T\rho
		-
		C\left[
		\sigma\sqrt{T(d+\tau)\rho}
		+
		\sigma^2\sqrt{T(d+\tau)}
		+
		\sigma^2(d+\tau)
		\right].
		\]
	\end{lemma}
	
	The second estimate controls the empirical error floor in the orthogonal block.
	
	\begin{lemma}[Orthogonal-block concentration]
		Under the multitask lemma conditions in
		Condition~\ref{cond:surrogate-recovery}, for every \(\tau\ge1\), with probability at least
		\(1-e^{-\tau}\),
		\[
		\sigma_{k+1}^2(\widehat A)
		\le
		T\nu+C\sigma^2\bigl(\sqrt{T(d+\tau)}+(d+\tau)\bigr).
		\]
	\end{lemma}
	
	Combining the previous two lemmas with the population gap condition gives the
	denominator lower bound
	\[
	\sigma_k^2(Y)-\sigma_{k+1}^2(\widehat A)\gtrsim T\widetilde\lambda_k
	\]
	on the event where the fluctuation level is smaller than a constant multiple of
	\(\widetilde\lambda_k\); outside this regime the final bound is already trivial
	after truncation by one.
	
	It remains to control the numerator, which measures how much the estimated
	signal block \(Y\) leaks into the orthogonal block \(Y_\perp\).
	The required bound combines three contributions: one linear
		signal--error term, one centered quadratic fluctuation---whose square-root
		and linear deviation regimes produce the two \(\sigma^2\) terms---and the
		population cross-covariance in~\eqref{eq:mixing}, yielding the
		following estimate.
	
	\begin{lemma}[Cross-block concentration]
		Under the multitask lemma conditions in
		Condition~\ref{cond:surrogate-recovery}, for every \(\tau\ge1\) such that
		\(\delta_{\tau}\le 1/2\), with probability at least
		\(1-e^{-\tau}\),
		\[
		\|\Proj_YY_\perp\|_{\mathrm{op}}
		\le
		C\left[
		\sigma\sqrt{d+\tau}
		+
		\frac{\sigma^2\sqrt{d+\tau}}{\sqrt{\rho}}
		+
		\frac{\sigma^2(d+\tau)}{\sqrt{T\rho}}
		+
		\frac{\sqrt{T}\,\delta_1\widetilde\lambda_k}{\sqrt{\rho}}
		\right].
		\]
	\end{lemma}
	Here \(\delta_\tau\) denotes the high-probability relative deviation of the
	empirical signal block \(T^{-1}Y^\top Y\) from its population counterpart; its
	precise definition and bound are given in Lemma~\ref{lem:mtl-signal}
	and~\eqref{eq:mtl-delta} in the appendix.
	
	The final term is the population contribution.
		By~\eqref{eq:mixing}, the averaged covariance cross-block satisfies
		\[
		\bigl\|
		\widetilde V^\top\Sigma_{\bm\varepsilon}\widetilde V_\perp
		\bigr\|_{\mathrm{op}}
		=
		\delta_1\widetilde\lambda_k.
		\]
		Accordingly, the corresponding unnormalized population block
		\(\E[Y^\top Y_\perp]\) has operator norm
		\(T\delta_1\widetilde\lambda_k\). Combining this with the inverse signal
		scale \((T\rho)^{-1/2}\) produces the displayed contribution
		\(\sqrt{T}\,\delta_1\widetilde\lambda_k/\sqrt{\rho}\); substitution into the
		deterministic reduction then yields the dimensionless population-bias term
		\(\delta_1\) in the multitask lemma.
	
	\paragraph{Resulting bound.}
	Substituting the three estimates into the deterministic
	reduction~\eqref{eq:mtl-deterministic-reduction} yields the terms in
	Theorem~\ref{thm:mtl-main} according to their probabilistic source. The linear
	signal--error interaction produces
	\(\sigma\sqrt{(d+\tau)\rho/(T\widetilde\lambda_k^2)}\), the centered quadratic
	fluctuation produces the two \(\sigma^2\) contributions
	\(\sigma^2\sqrt{(d+\tau)/(T\widetilde\lambda_k^2)}\) and
	\(\sigma^2(d+\tau)/(T\widetilde\lambda_k)\), and the population cross-block
	produces the additive term \(\delta_1\). This proves the multitask lemma in the
	nontrivial concentration regime; the truncation by one covers the complementary
	regime in which the fluctuations are too large relative to the population
	eigengap.
	
	\section{Instantiations of the Multitask Lemma}\label{sec:instantiations}
	We now apply the general framework of Section~\ref{sec:general} to both linear
	and nonlinear settings. Section~\ref{sec:linear} treats linear regression, where
	the whitened least-squares estimator is used as the task statistic and its
	averaged error covariance is exactly isotropic. Section~\ref{sec:gsi} treats
	generalized linear models with Gaussian design, where the first-moment estimator
	is used as the task statistic; Stein's identity connects its population
	surrogate to the true task parameter.
	The resulting upper and lower bounds establish the exact minimax rates for
	linear regression and classification, and match up to logarithmic factors for
	additive-noise regression.
	\subsection{Linear Regression}\label{sec:linear}
	
	Recall the linear model from Section~\ref{ssec:overview-instantiations}, and
	condition throughout on the designs satisfying~\eqref{eq:design-condition}.
	{Recall the OLS task statistic from~\eqref{eq:ols-task-statistic}, the
		whitening matrix \(\widehat S\) from~\eqref{eq:linear-whitening-matrix}, and
		the whitened task statistic from~\eqref{eq:linear-whitened-task-statistic}.
		To see concretely why whitening is needed, define
		\(\eta_t:=\widehat\beta_t^{(\mathrm{OLS})}-\beta_t\). The raw OLS statistics
		have the correct population means and directional sub-Gaussian scale, but their
		averaged error covariance is
		\[
		\frac1T\sum_{t=1}^T\E[\eta_t\eta_t^\top\mid X_{1:T}]
		=
		\frac{\sigma_\xi^2}{n}\widehat S,
		\]
		which need not be block diagonal with respect to
		\(\mathcal V\oplus\mathcal V^\perp\). Consequently, heterogeneous designs can
		produce \(\delta_1\ne0\), as the following example demonstrates.

		\begin{remark}[Why unwhitened OLS aggregation can be inconsistent]
            \label{rem:ols-nonvanishing-delta}
            Consider \(d=2\), \(\mathcal V=\operatorname{span}(e_1)\), and
            \(\beta_t=(\sigma_\xi/\sqrt n)e_1\) for every task. Suppose that
            \[
            \left(\frac1n\sum_{i=1}^nX_{t,i}X_{t,i}^\top\right)^{-1}
            =
            \begin{pmatrix}1&c\\c&1\end{pmatrix},
            \qquad c\in(0,1),
            \]
            for every \(t\). Fix \(c\) and \(n\). These designs are uniformly
            well conditioned, but the OLS error covariance has a nonzero
            off-diagonal entry \(c\sigma_\xi^2/n\), coupling \(\mathcal V\)
            and \(\mathcal V^\perp\). Since
            \(\lambda_1(\Sigma_{\bm\beta})=\sigma_\xi^2/n\), the normalized
            mixing term is \(\delta_1=c\).

            To see how this affects subspace estimation, consider the matrix
            used by unwhitened spectral aggregation:
            \[
            \frac1T\sum_{t=1}^T
            \widehat\beta_t^{(\mathrm{OLS})}
            \widehat\beta_t^{(\mathrm{OLS})\top}
            \ \xrightarrow[T\to\infty]{\mathbb P}
            \underbrace{\frac{\sigma_\xi^2}{n}
            \begin{pmatrix}1&0\\0&0\end{pmatrix}}_{\text{signal}}
            +
            \underbrace{\frac{\sigma_\xi^2}{n}
            \begin{pmatrix}1&c\\c&1\end{pmatrix}}_{\text{OLS error covariance}}.
            \]
            The limiting matrix has its leading eigenvector at angle
            \(\theta=\frac12\arctan(2c)>0\) from \(e_1\). Consequently,
            the subspace error converges in probability to \(\sin\theta>0\),
            even as the number of tasks grows.

        \end{remark}
		
		Conditionally on the designs, the whitened errors are independent across
		tasks and satisfy
		\(\|\langle\varepsilon_t,a\rangle\|_{\psi_2}
		\le C\sigma_\xi\|a\|_2/\sqrt n\). Moreover, their averaged error covariance is
		\[
		\Sigma_{\bm\varepsilon}
		=
		\frac{\sigma_\xi^2}{n}I_d.
		\]
		Consequently, \(\delta_1=0\), while
		\(\rho=\widetilde\lambda_k+\sigma_\xi^2/n\) and
		\(\nu=\sigma_\xi^2/n\), so
		\(\rho-\nu=\widetilde\lambda_k\). Thus both multitask lemma conditions hold
        with \(\sigma\asymp\sigma_\xi/\sqrt n\). Combining the lemma with the
        signal-strength comparison and inverse-whitening argument in
        Section~\ref{ssec:linear-reg-overview} yields the upper bound in
        Theorem~\ref{thm:linear}.}
    
	We now show this rate is minimax optimal, i.e., it matches the minimax risk
	lower bound. This lower bound is obtained by specializing Theorem 5.1 of
	\cite{niu2024learning}.
	{We use the parameter class
		\(\mathcal P(\lambda_k)\) defined in
		Section~\ref{ssec:overview-instantiations}.}
	
	\begin{theorem}[Linear regression minimax lower bound]\label{thm:linear-lb}
		Consider the isotropic Gaussian linear multitask model
		\[
		Y_{t,i}=X_{t,i}^\top \beta_t+\xi_{t,i},
		\qquad
		X_{t,i}\stackrel{\mathrm{i.i.d.}}{\sim}N(0,I_d),
		\qquad
		\xi_{t,i}\stackrel{\mathrm{i.i.d.}}{\sim}N(0,1),
		\]
		with \(n\) samples per task and \(T\) tasks. Assume that
		\(\mathcal P(\lambda_k)\) is nonempty, that
		\(n\ge d\), \(k\ge c_0\log T\), \(d\ge (1+\rho_1)k\), and
		\(T\ge (1+\rho_2)k\) for some constants \(c_0,\rho_1,\rho_2>0\). Then
		there exists a constant \(c>0\), depending only on
		\(c_0,\rho_1,\rho_2\), such that
		\[
		\inf_{\widehat{\mathcal V}}
		\sup_{(\mathcal V,\beta_1,\dots,\beta_T)\in \mathcal P(\lambda_k)}
		\E\big[\sinth{\widehat{\mathcal V}}{\mathcal V}\big]
		\ge
		c\left(
		\left[
		\sqrt{\frac{d}{Tn\lambda_k}}
		+
		\sqrt{\frac{d}{Tn^2\lambda_k^2}}
		\right]\wedge 1
		\right).
		\]
	\end{theorem}
	\noindent Since isotropic Gaussian design satisfies~\eqref{eq:design-condition}
	with high probability for $n\gtrsim d$, Theorems~\ref{thm:linear}
	and~\ref{thm:linear-lb} yield the expectation minimax characterization stated
	in Corollary~\ref{cor:overview-linear-minimax}. The proof of that corollary
	integrates the conditional upper-tail bound uniformly over the fixed-design and
	sub-Gaussian-noise classes, and combines it with the Gaussian-design lower
	bound above.
	
	\subsection{Nonlinear Models with Isotropic Gaussian Covariates}\label{sec:gsi}
	
    We now verify the multitask lemma conditions for the nonlinear models with
    isotropic Gaussian covariates introduced in Section~\ref{ssec:gsi-overview}.
    Recall the first-moment task statistic and target-subspace estimator
    from~\eqref{eq:gsi-first-moment-estimator}
    and~\eqref{eq:gsi-subspace-estimator}. The population alignment established
    in~\eqref{eq:gsi-population-alignment}, together with \(a_g>0\), gives
    \(\widetilde{\mathcal V}=\mathcal V\). It remains to control the taskwise
    errors and verify the population gap for classification and additive-noise
    regression.

	\subsubsection{Classification}
	Recall the general model in~\eqref{eq:gsi-model}. We specialize
	its conditional response law to
	\[
	Y_t\mid X_t\sim\operatorname{Ber}\!\left(g(X_t^\top\beta_t)\right)
	\]
	for every \(t\in[T]\). In this isotropic Gaussian setting, no whitening
	is required.
	\paragraph{Verification of the multitask conditions.}
	{By~\eqref{eq:gsi-population-alignment},
		\[
		\Sigma_{\bm{\tilde\beta}}
		=
		\frac1T\sum_{t=1}^T a_t^2\beta_t\beta_t^\top
		\succeq
		a_g^2\Sigma_{\bm\beta}.
		\]
		Thus \(\widetilde\lambda_k\ge a_g^2\lambda_k\).
		Bounded Bernoulli responses verify Condition~\ref{cond:surrogate-recovery}(1) with
		\(\sigma^2\asymp n^{-1}\). Gaussian
		rotational symmetry gives
		\[
		\operatorname{Cov}(\hat\beta_t)
		=
		\frac{\E[Y_{t,1}]}{n}I_d
		+
		\frac{\E[g''(\|\beta_t\|_2Z)]-a_t^2}{n}\,\beta_t\beta_t^\top.
		\]
		Thus the averaged error covariance has the form
		\begin{equation}\label{eq:gsi-covariance-block-structure}
			\Sigma_{\bm\varepsilon}
			=
			\omega^2 I_d+VR_\star V^\top
		\end{equation}
		for some \(\omega^2\ge0\) and symmetric \(R_\star\), where \(V\) is an
		orthonormal basis for \(\mathcal V\). Hence
		\[
		\Proj_{\mathcal V}\Sigma_{\bm\varepsilon}
		\Proj_{\mathcal V^\perp}=0,
		\]
		so \(\delta_1=0\) and the multitask bound has no error-covariance mixing
		term. Finally,
		the calculation in Appendix~\ref{sec:proof-of-classification-upper-bound}
		gives \(\rho-\nu\ge\frac12(1-n^{-1})\widetilde\lambda_k\) for
		\(n\ge n_{\mathrm{cls}}\). Since \(n_{\mathrm{cls}}\ge2\), this verifies
		Condition~\ref{cond:surrogate-recovery}(2) with the fixed constant \(\gamma_0=1/4\).}
	
	{The preceding verification establishes the classification upper bound
		stated in Theorem~\ref{thm:gsi}.}
	
	\begin{remark}[Symmetric links]
		{For links satisfying
			\(g(-z)=1-g(z)\), the function \(g''\) is odd. Hence
			\(\E[g''(rZ)]=0\) for every \(r\). The calculation in
			Appendix~\ref{sec:proof-of-classification-upper-bound} therefore shows
			that \(n_{\mathrm{cls}}\) can be chosen as \(2\). Consequently, the classification
			upper bound in Theorem~\ref{thm:gsi}
			applies for every \(n\ge2\).}
	\end{remark}
	
	\begin{remark}[{Examples satisfying alignment}]
		\label{rem:alignment-examples}
		{Consider the logistic link
			\(g(u)=(1+e^{-u})^{-1}\) and the Cauchit link
			\(g(u)=\frac12+\frac1\pi\arctan(u)\), the standard logistic and Cauchy
			distribution functions, respectively. Their derivatives are strictly positive:
			\[
			g'(u)=\frac{e^{-u}}{(1+e^{-u})^2}
			\qquad\text{and}\qquad
			g'(u)=\frac{1}{\pi(1+u^2)},
			\]
			respectively. Hence, for either link,
			\[
			\E[g'(rZ)]\ge \Prob(|Z|\le1)\inf_{|u|\le1}g'(u)>0,
			\qquad 0\le r\le1,
			\]
			so $a_g>0$ uniformly over \(\|\beta\|_2\le1\).}
	\end{remark}
	
	\paragraph{Minimax Lower Bound}
	{The preceding analysis establishes the classification upper bound under
	the stated alignment and regularity conditions. We now show that both terms in
	that rate are unavoidable over
	the parameter class \(\mathcal P(\lambda_k)\). The lower bound uses the
	admissibility condition in Definition~\ref{def:admissible-link}, which is
	tailored to comparing Bernoulli likelihoods. Under the joint assumptions of the
	upper and lower bounds, the result therefore completes the exact minimax
	characterization for classification.}

	\begin{theorem}[Classification lower bound under Gaussian design]\label{thm:gsi-lower}
		Let \(g\) be admissible in the sense of
		Definition~\ref{def:admissible-link}.
		{Assume that \(\mathcal P(\lambda_k)\) is nonempty.}
		Assume \(1\le k\le d/2\).
		There exist constants \(c,c_0,C_0>0\), depending only on \(L_1\) and \(L_2\),
		such that whenever
		\(
		n\ge C_0 k
		\)
		and
		\(
		T e^{-c_0 n}\le c_0 d,
		\)
		the following holds:
		\[
		\inf_{\widehat{\mathcal V}}
		\sup_{(\mathcal V,\beta_1,\dots,\beta_T)\in
			\mathcal P(\lambda_k)}
		\E\!\left[\sinth{\widehat{\mathcal V}}{\mathcal V}\right]
		\ge
		c\left(
		\left[
		\sqrt{\frac{d}{Tn\lambda_k}}
		+
		\sqrt{\frac{d}{Tn^2\lambda_k^2}}
		\right]\wedge 1
		\right).
		\]
	\end{theorem}

	{\textbf{Proof strategy and novelty.}
	The two terms in Theorem~\ref{thm:gsi-lower} arise from different lower-bound
	mechanisms. The first follows by adapting the deterministic packing argument of
	\citet{niu2024learning}; in our classification model, the main change is the
	Kullback--Leibler calculation for Bernoulli observations. The weak-signal term,
	which dominates when \(n\lambda_k\lesssim1\), requires a different argument.
	In linear regression, \citet{niu2024learning} use a Gaussian prior on the task
	coefficients: after averaging over those coefficients, the resulting observation
	law remains Gaussian and can be analyzed explicitly. The nonlinear Bernoulli
	response destroys this Gaussian structure. Our new ingredient is a hidden-sign
	mixture construction, together with divergence bounds that retain the
	cancellation created by averaging over the latent task coefficients.}

	{\textbf{A common packing inside the classification model.}
	Both lower-bound constructions use the same finite collection of candidate
	subspaces. For \(\ell\in[M]\), let
	\[
	V^{(\ell)}=[e_1,\ldots,e_{k-1},u_\ell],
	\qquad
	\mathcal V_\ell:=\operatorname{span}(V^{(\ell)}),
	\]
	where \(V^{(\ell)}\) is an orthonormal basis, \(e_1,\ldots,e_{k-1}\) are fixed
	coordinate vectors, and only the last unit
	vector \(u_\ell\) varies. By the local packing construction in
	Lemma~\ref{lem:gsi-lb-packing}, the vectors can be chosen so that
	\(\log M\gtrsim d-k\) and, for a prescribed separation \(\varepsilon\),
	\[
	\sinth{\mathcal V_\ell}{\mathcal V_m}^2
	\asymp\varepsilon,
	\qquad \ell\ne m.
	\]
	Thus the subspaces are separated at sine-theta distance of order
	\(\sqrt\varepsilon\), while the number of alternatives is exponential in
	\(d-k\). Fano's inequality reduces the problem to choosing \(\varepsilon\) so
	that the pairwise Kullback--Leibler divergences are of order at most \(d-k\).
	Each candidate is an instance of the classification model already studied in
	this section: for a coefficient \(\beta\), let \(\mathbb P_\beta\) denote the
	one-observation law
	\[
	X\sim N(0,I_d),
	\qquad
	Y\mid X\sim\operatorname{Ber}\!\bigl(g(X^\top\beta)\bigr).
	\]
	After choosing task coefficients
	\(\beta_t^{(\ell)}\in\mathcal V_\ell\), the full-data law under candidate
	\(\ell\) is
	\(\bigotimes_{t=1}^T(\mathbb P_{\beta_t^{(\ell)}})^{\otimes n}\). Thus the
	packing changes the shared subspace in the original model; it does not introduce
	a different observation model.}

	{\textbf{The first, deterministic term.}
	For the first term in Theorem~\ref{thm:gsi-lower},
	Lemma~\ref{lem:shared-lb-frame} guarantees deterministic task coefficients in
	each \(\mathcal V_\ell\) such that
	\[
	\frac1T\sum_{t=1}^T
	\beta_t^{(\ell)}(\beta_t^{(\ell)})^\top
	=
	\lambda_k\Proj_{\mathcal V_\ell},
	\qquad
	\|\beta_t^{(\ell)}\|_2\le1.
	\]
	Thus every candidate belongs to \(\mathcal P(\lambda_k)\). Let
	\(P_\ell^{\rm det}\) be its full-data law. With
	\(h(s):=\log\!\bigl(g(s)/(1-g(s))\bigr)\),
	Definition~\ref{def:admissible-link} gives
	\(\|h'\|_\infty\le L_1\), and the standard Bernoulli
	Kullback--Leibler bound yields
	\[
	\mathrm{KL}(P_\ell^{\rm det}\|P_m^{\rm det})
	\lesssim
	Tn\lambda_k
	\sinth{\mathcal V_\ell}{\mathcal V_m}^2.
	\]
	Taking \(\varepsilon\asymp d/(Tn\lambda_k)\), up to truncation by a constant,
	yields the first lower-bound term \(\sqrt{d/(Tn\lambda_k)}\).}

	{\textbf{The second term: a hidden-sign construction.}
	To obtain the second term in Theorem~\ref{thm:gsi-lower}, work in the
	weak-signal regime
	\(n\lambda_k\lesssim1\), and independently for each
	task draw
	\[
	\omega_t=(\xi_{t,1},\ldots,\xi_{t,k-1},\zeta_t)\in\{\pm1\}^k,
	\]
	with all coordinates independent and uniform on \(\{\pm1\}\). Under candidate
	subspace \(\ell\), take
	\[
	\beta_t^{(\ell)}
	=
	\sqrt{2\lambda_k}\,V^{(\ell)}\omega_t
	\in\mathcal V_\ell.
	\]
	Under the weak-signal condition and the scaling \(n\gtrsim k\) in
	Theorem~\ref{thm:gsi-lower},
	\(\|\beta_t^{(\ell)}\|_2^2=2k\lambda_k\le1\). Moreover,
	on the event
	\[
	\lambda_{\min}\!\left(\frac1T\sum_{t=1}^T
	\omega_t\omega_t^\top\right)\ge\frac12,
	\]
	the conditional parameter tuple
	\((\mathcal V_\ell,\beta_1^{(\ell)},\ldots,\beta_T^{(\ell)})\) belongs to
	\(\mathcal P(\lambda_k)\); when
	\(T\gtrsim k\), this event has probability bounded away from zero. The
	lower-bound reduction works with the joint law of the signs and data and
	restricts the risk comparison to this event, so only valid parameter
	tuples enter the minimax supremum. Let
	\[
	P_\ell^{\rm mix}
	:=
	\E_\omega\!\left[
	\bigotimes_{t=1}^T
	(\mathbb P_{\beta_t^{(\ell)}})^{\otimes n}
	\right]
	\]
	be the full-data law obtained after averaging over the unobserved signs
	\(\omega_1,\ldots,\omega_T\). Averaging is essential: it hides
	the individual task directions and creates the cancellation that is absent from
	a comparison conditional on all the signs. The goal is to show that, for
	\(\ell\ne m\),
	\[
	\mathrm{KL}(P_\ell^{\rm mix}\|P_m^{\rm mix})
	\lesssim
	Tn^2\lambda_k^2
	\sinth{\mathcal V_\ell}{\mathcal V_m}^2,
	\]
	up to an exponentially small residual. This is the pairwise divergence scale
	that yields the second term through Fano's inequality.}

	{\textbf{Reduction to a one-task mixture comparison.}
	Bounding this KL divergence directly is difficult because each
	\(P_\ell^{\rm mix}\) averages a nonlinear Bernoulli likelihood over all of the
	latent task signs. We therefore reveal
	\(\xi_t=(\xi_{t,1},\ldots,\xi_{t,k-1})\) but keep \(\zeta_t\), the sign
	multiplying the varying direction, unobserved. Revealing \(\xi_t\) can only make
	the hypotheses easier to distinguish, so the divergence in this augmented
	experiment upper-bounds the target KL divergence. Conditional on
	\(\xi_t=\xi\), the comparison between candidates
	\(\ell\) and \(m\) reduces to a rank-one problem around the common baseline
	\[
	\beta_0(\xi):=\sqrt{2\lambda_k}\sum_{j=1}^{k-1}\xi_j e_j,
	\]
	with the two varying directions \(u_\ell\) and \(u_m\). Define the corresponding
	symmetric mixtures
	\[
	Q_{u_\ell,\xi}
	=
	\frac12\mathbb P_{\beta_0(\xi)+\sqrt{2\lambda_k}u_\ell}^{\otimes n}
	+
	\frac12\mathbb P_{\beta_0(\xi)-\sqrt{2\lambda_k}u_\ell}^{\otimes n},
	\qquad
	Q_{u_m,\xi}
	=
	\frac12\mathbb P_{\beta_0(\xi)+\sqrt{2\lambda_k}u_m}^{\otimes n}
	+
	\frac12\mathbb P_{\beta_0(\xi)-\sqrt{2\lambda_k}u_m}^{\otimes n}.
	\]
	Thus, for each task, \(Q_{u_\ell,\xi}\) and \(Q_{u_m,\xi}\) are precisely the
	conditional one-task laws given \(\xi_t=\xi\), after averaging over the hidden
	sign \(\zeta_t\).
	The revealed-sign reduction, independence across tasks, and
	\(\mathrm{KL}\le\chi^2\) give the explicit comparison
	\[
	\mathrm{KL}(P_\ell^{\rm mix}\|P_m^{\rm mix})
	\le
	T\,\E_\xi\!\left[
	\mathrm{KL}(Q_{u_\ell,\xi}\|Q_{u_m,\xi})
	\right]
	\le
	T\,\E_\xi\!\left[
	\chi^2(Q_{u_\ell,\xi}\|Q_{u_m,\xi})
	\right].
	\]
	Thus it remains to control a one-task \(\chi^2\)-divergence.}

	{\textbf{Controlling the one-task \(\chi^2\)-divergence.}
	Set \(R_\xi:=\mathbb P_{\beta_0(\xi)}^{\otimes n}\), and write
	\(q_{\ell,\xi}:=dQ_{u_\ell,\xi}/dR_\xi\) and
	\(q_{m,\xi}:=dQ_{u_m,\xi}/dR_\xi\). Splitting the definition of \(\chi^2\)
	according to whether the denominator \(q_{m,\xi}\) is at least \(1/2\) gives
	\[
	\begin{aligned}
	\chi^2(Q_{u_\ell,\xi}\|Q_{u_m,\xi})
	\le{}&
	2\|q_{\ell,\xi}-q_{m,\xi}\|_{L^2(R_\xi)}^2\\
	&+\left(
	\E_{R_\xi}[(q_{\ell,\xi}-q_{m,\xi})^4]
	\E_{R_\xi}[q_{m,\xi}^{-2}\mathbf 1_{\{q_{m,\xi}<1/2\}}]
	\right)^{1/2}.
	\end{aligned}
	\]
	This inequality explains the role of an \(L^2\) bound relative to the common
	reference law \(R_\xi\); ``centered-reference'' in
	Proposition~\ref{prop:gsi-lb-centered} refers to this common-baseline
	comparison, not to mean-centering. Since
	\(u_\ell,u_m\) are orthogonal to \(\beta_0(\xi)\), for sufficiently small
	\(\lambda_k\) with \(n\lambda_k\lesssim1\),
	Proposition~\ref{prop:gsi-lb-centered} gives
	\[
	\left\|
	q_{\ell,\xi}-q_{m,\xi}
	\right\|_{L^2(R_\xi)}^2
	\lesssim
	n^2\lambda_k^2\sinth{\mathcal V_\ell}{\mathcal V_m}^2.
	\]
	The averaging over the latent sign cancels the first-order dependence on the varying
	direction, leaving a bound quadratic in \(\lambda_k\). This cancellation
	is the source of the \(n^2\lambda_k^2\), rather than \(n\lambda_k\), divergence
	scale. The fourth-moment and lower-tail estimates in
	Appendix~\ref{subsec:classification-lb} make the second term in the preceding
	\(\chi^2\) inequality exponentially small. Consequently, uniformly in \(\xi\),
	\[
	\chi^2(Q_{u_\ell,\xi}\|Q_{u_m,\xi})
	\lesssim
	n^2\lambda_k^2\sinth{\mathcal V_\ell}{\mathcal V_m}^2
	+e^{-c/\lambda_k}.
	\]}

	{\textbf{From the one-task bound to the lower-bound rate.}
	Averaging over \(\xi\) and applying the preceding KL comparison yields
	\[
	\mathrm{KL}(P_\ell^{\rm mix}\|P_m^{\rm mix})
	\lesssim
	Tn^2\lambda_k^2
	\sinth{\mathcal V_\ell}{\mathcal V_m}^2
	+
	T e^{-c/\lambda_k}.
	\]
	In the weak-signal regime, \(1/\lambda_k\gtrsim n\), so the condition in
	Theorem~\ref{thm:gsi-lower},
	\(Te^{-c_0n}\lesssim d\) makes the residual exponential term negligible
	compared with the packing size. Taking
	\(\varepsilon\asymp d/(Tn^2\lambda_k^2)\) in Fano's inequality yields the
	second lower-bound term \(\sqrt{d/(Tn^2\lambda_k^2)}\). When
	\(n\lambda_k\gtrsim1\), this term is no larger than the first. Thus the
	deterministic adaptation supplies the first term, while the new hidden-sign
	mixture analysis supplies the genuinely nonlinear weak-signal term.}
	
	\begin{remark}[{Assumptions for the two lower-bound mechanisms}]
		{As far as link regularity is concerned, the first lower-bound term only uses the bound on
		\(g'(t)/[g(t)(1-g(t))]\). The second term is a weak-signal phenomenon and
		uses, in addition, the bound on \(g''(t)/[g(t)(1-g(t))]\). Two different
		probability events enter the randomized argument. The Rademacher covariance
		good event ensures that the sampled task coefficients satisfy the covariance
		constraint in \(\mathcal P(\lambda_k)\). Separately, the lower-tail event
		\(\{q_{m,\xi}<1/2\}\) appears when converting the common-reference \(L^2\)
		bound into a \(\chi^2\) bound and produces the exponentially small residual.
		The precise roles of the dimensional and sample-size conditions in
		Theorem~\ref{thm:gsi-lower} are recorded in
		Appendix~\ref{subsec:classification-lb}.}
	\end{remark}
	
	{Combining Theorems~\ref{thm:gsi} and~\ref{thm:gsi-lower} under their
	joint assumptions shows that the classification rate is exactly minimax, as
	stated in Corollary~\ref{cor:overview-cls-minimax}. In particular, the
	weak-signal term is unavoidable rather than an artifact of the first-moment
	estimator. We next turn to additive-noise regression, where the corresponding
	upper and lower bounds match up to logarithmic factors.}
	
	\subsubsection{Additive-Noise Regression}
	
	Recall the general model in~\eqref{eq:gsi-model}. We specialize
	its conditional response law to the additive-noise model
	\[
	Y_t=g(X_t^\top\beta_t)+\xi_t,
	\qquad
	\xi_t\perp X_t,
	\qquad
	\E[\xi_t]=0,
	\qquad
	\|\xi_t\|_{\psi_2}\le K_\xi
	\]
	for every \(t\in[T]\). Thus the common Gaussian design and shared-subspace
	structure remain unchanged; only the response distribution is specialized.
	As in the overview, assume that \(g\in C^2(\R)\)
	satisfies~\eqref{eq:bounded-derivatives}.
	Let \(K_Y:=C\bigl(|g(0)|+L_1+K_\xi\bigr)\), as in the overview.
	
	{Recall that the estimator in Theorem~\ref{thm:gsi-subg} is the
		first-moment spectral estimator defined
		in~\eqref{eq:gsi-first-moment-estimator}--\eqref{eq:gsi-subspace-estimator}.}
	
	\paragraph{Verification of the multitask conditions.}
	Unlike the classification case, Condition~\ref{cond:surrogate-recovery}(1) (sub-Gaussian error) is not immediate:
	although \(Y\) is sub-Gaussian, the products \(YX\) defining
	\(\hat\beta_t\) need not be sub-Gaussian. We therefore apply the multitask
	lemma to an auxiliary truncated first-moment estimator. For a truncation level
	\(L>0\), define
	\[
	\phi_L(y):=y\mathbf 1\{|y|\le L\},
	\qquad
	\hat\beta_t^{(L)}
	:=
	\frac1n\sum_{i=1}^n \phi_L(Y_{t,i})X_{t,i}.
	\]
	{Truncation is only a proof device; the estimator of interest remains
        the original first-moment spectral estimator \(\widehat{\mathcal V}\).}
	Because \(\phi_L(Y_{t,i})\) is bounded by \(L\), the centered truncated
	first moment \(\hat\beta_t^{(L)}-\E[\hat\beta_t^{(L)}]\) is sub-Gaussian at
	scale \(\sigma_L\asymp L/\sqrt n\), verifying Condition~\ref{cond:surrogate-recovery}(1).
	Let \(\widehat V_k^{(L)}\) be an orthonormal basis matrix for the top-\(k\)
	right singular subspace of the matrix with rows
	\((\hat\beta_t^{(L)})^\top\). For a suitable choice of \(L\),
	the truncated and untruncated first moments agree simultaneously with high
	probability, and therefore \(\widehat V_k^{(L)}=\widehat V_k\) on that event.
	It therefore suffices to analyze \(\widehat V_k^{(L)}\) and transfer the
	resulting bound to \(\widehat V_k\) on this agreement event.
	
	\begingroup
	We next verify that truncation preserves the population signal direction. The
	following proposition isolates the Gaussian rotational-symmetry argument used
	for this purpose.

	\begin{proposition}[Exact alignment under response transformations]
		\label{prop:response-transformation-alignment}
		Let \(X\sim N(0,I_d)\), let \(\beta\in\R^d\), and suppose that
		\(Y\mid X\stackrel{d}{=}Y\mid X^\top\beta\), the standard distributional
		single-index assumption that the conditional law of \(Y\) depends on \(X\)
		only through \(X^\top\beta\). For any measurable function \(\phi\) satisfying
		\(\E[|\phi(Y)|\|X\|_2]<\infty\), there exists a scalar
		\(a_\phi(\beta)\) such that
		\[
		\E[\phi(Y)X]=a_\phi(\beta)\beta.
		\]
		If \(\beta\neq0\), then
		\(a_\phi(\beta)=
		\E[\phi(Y)X^\top\beta]/\|\beta\|_2^2\); for \(\beta=0\), one has
		\(\E[\phi(Y)X]=0\).
	\end{proposition}

	\begin{proof}
		Suppose first that \(\beta\neq0\). Let
		\(u=\beta/\|\beta\|_2\) and decompose
		\[
		X=\alpha u+W,
		\qquad
		\alpha=u^\top X,
		\qquad
		W=(I_d-uu^\top)X.
		\]
		By Gaussian rotational symmetry, \(W\) is centered and independent of
		\(\alpha\). Since \(W\) is independent of \(\alpha\), and the conditional law
		of \(Y\) given \(X\) depends only on
		\(X^\top\beta=\|\beta\|_2\alpha\), \(W\) is independent of
		\((\alpha,Y)\). Hence \(\E[\phi(Y)W]=0\). Consequently,
		\[
		\E[\phi(Y)X]
		=
		\E[\phi(Y)\alpha]u
		=
		\frac{\E[\phi(Y)X^\top\beta]}{\|\beta\|_2^2}\,\beta.
		\]
		When \(\beta=0\), the single-index assumption makes \(Y\) independent of
		\(X\), and the result follows from \(\E[X]=0\).
	\end{proof}

	Applying Proposition~\ref{prop:response-transformation-alignment} with
	\(\phi=\phi_L\) gives, for \(\beta_t\neq0\),
	\[
	\tilde\beta_t^{(L)}
	:=
	\E[\hat\beta_t^{(L)}]
	=
	a_{t,L}\beta_t,
	\qquad
	a_{t,L}
	:=
	\frac{\E[\phi_L(Y_t)X_t^\top\beta_t]}{\|\beta_t\|_2^2}.
	\]
	For \(\beta_t=0\), \(\tilde\beta_t^{(L)}=0\). Thus truncation preserves each
	population direction exactly. Alignment alone, however, does not guarantee
	that the transformed task vectors retain enough magnitude to span
	\(\mathcal V\). To control their magnitude, we compare \(a_{t,L}\) with the
	corresponding untruncated coefficient. Stein's identity gives
	\endgroup
	\[
	\tilde\beta_t:=\E[\hat\beta_t]
	=a_t\beta_t,
	\qquad
	a_t=\E[g'(\|\beta_t\|_2Z)].
	\]
	The uniform sub-Gaussian tail bound for \(Y_t\) gives
	\[
	|a_{t,L}-a_t|\,\|\beta_t\|_2
	=
	\|\tilde\beta_t^{(L)}-\tilde\beta_t\|_2
	\le
	b_L,
	\qquad
	b_L:=CK_Ye^{-cL^2/K_Y^2}.
	\]
	Thus truncation preserves the alignment coefficients at the scale relevant
	for the population signal. More precisely,
	\[
	\left\|
	\frac1T\sum_{t=1}^T
	\tilde\beta_t^{(L)}(\tilde\beta_t^{(L)})^\top
	-
	\frac1T\sum_{t=1}^T
	\tilde\beta_t\tilde\beta_t^\top
	\right\|_{\mathrm{op}}
	\le
	2L_1b_L+b_L^2.
	\]
	For a suitable choice of \(L\), the right-hand side is at most
	\(\widetilde\lambda_k/2\), so by Weyl's inequality
	\(\widetilde\lambda_k^{(L)}\ge\widetilde\lambda_k/2
	\ge a_g^2\lambda_k/2\). Moreover, the same rotational decomposition gives
	the covariance block structure in~\eqref{eq:gsi-covariance-block-structure},
	so the truncated surrogate has \(\delta_1=0\).
	{Rotational symmetry also
		implies that every
		\(\tilde\beta_t^{(L)}\) is a scalar multiple of \(\beta_t\), so
		\(\widetilde{\mathcal V}^{(L)}\subseteq\mathcal V\).  The lower bound
		\(\widetilde\lambda_k^{(L)}\ge a_g^2\lambda_k/2>0\) shows that the surrogate
		subspace has dimension \(k\).  Therefore
		\(\widetilde{\mathcal V}^{(L)}=\mathcal V\), and no transformation of the
		estimated subspace is required.}
	The remaining one-dimensional covariance
	calculation verifies Condition~\ref{cond:surrogate-recovery}(2) for
	\(n\ge n_{\mathrm{add}}\), where \(n_{\mathrm{add}}\) is a finite constant
	depending only on \(g\), defined explicitly in
	\eqref{eq:n-add-definition} in Appendix~\ref{sec:proof-of-additive-upper-bound}.
	Applying the multitask lemma to the truncated estimator yields a bound for
	\(\widehat V_k^{(L)}\), which the preceding agreement event passes to
	\(\widehat V_k\), establishing Theorem~\ref{thm:gsi-subg}.
    The logarithmic factor is the cost of choosing \(L\) large
	enough both to make \(b_L\) negligible and to ensure this agreement. Details are in
	Appendix~\ref{sec:proof-of-additive-upper-bound}.


	\paragraph{Minimax Lower Bound}
	
	For Gaussian additive responses, we reuse the discrete-prior geometric packing
	of candidate shared subspaces and task coefficients underlying the classification
	lower bound, Theorem~\ref{thm:gsi-lower}; see
	Lemma~\ref{lem:gsi-lb-packing}. The parameter geometry is unchanged, but the
	response channel is now additive Gaussian rather than Bernoulli, so the
	classification divergence bound no longer applies. We derive the corresponding
	mixture-divergence bound for the additive Gaussian channel, yielding the
	following lower bound over \(\mathcal P(\lambda_k)\).
	
	\begin{theorem}[Gaussian additive-response lower bound]\label{thm:gsi-additive-lower}
		Assume that \(g\in C^2(\R)\) satisfies~\eqref{eq:bounded-derivatives}.
		Fix \(\sigma>0\), and consider the isotropic Gaussian additive multitask model
		\[
		Y_{t,i}=g(X_{t,i}^\top\beta_t)+\xi_{t,i},
		\qquad
		X_{t,i}\stackrel{\mathrm{i.i.d.}}{\sim}N(0,I_d),
		\qquad
		\xi_{t,i}\stackrel{\mathrm{i.i.d.}}{\sim}N(0,\sigma^2),
		\]
		with \(\xi_{t,i}\perp X_{t,i}\). Then there exist constants \(c,c_0,C_0>0\),
		depending only on \(L_1\), \(L_2\), and \(\sigma\), such that
		whenever \(1\le k\le d/2\),
		\[
		n\ge C_0k,
		\qquad
		T e^{-c\sqrt n}\le c_0 d,
		\]
		we have
		\[
		\inf_{\widehat{\mathcal V}}
		\sup_{(\mathcal V,\beta_1,\dots,\beta_T)\in
			\mathcal P(\lambda_k)}
		\E\!\left[\sinth{\widehat{\mathcal V}}{\mathcal V}\right]
		\ge
		c\left(
		\left[
		\sqrt{\frac{d}{Tn\lambda_k}}
		+
		\sqrt{\frac{d}{Tn^2\lambda_k^2}}
		\right]\wedge 1
		\right).
		\]
	\end{theorem}
	
Together, Theorems~\ref{thm:gsi-subg} and~\ref{thm:gsi-additive-lower} yield
the additive-noise minimax comparison stated in
Corollary~\ref{cor:overview-add-minimax}. Taken together, the results of this
section establish exact minimax rates for linear regression and
classification, while the additive-noise upper and lower bounds match up to
logarithmic factors.
	
	{
		\section{Numerical illustrations}\label{sec:numerics}

We first examine the two main rate phenomena: the improved dependence on
anisotropic task geometry and the transition between the two terms in the
rate. Figure~\ref{fig:numerical-main} illustrates these phenomena in the
Gaussian linear model. We also examine the task-spectrum phenomenon in
logistic classification and illustrate the benefit of whitening under
non-isotropic linear designs.

\paragraph{Setup and rate comparison.}
Throughout, the target is
$\mathcal V=\operatorname{span}(e_1,e_2)$, and performance is measured by
the $\sin\Theta$ loss $\sinth{\widehat{\mathcal V}}{\mathcal V}$.
For the experiments in Figure~\ref{fig:numerical-main}, we use
\[
X_{t,i}\sim N(0,I_d),\qquad
Y_{t,i}=X_{t,i}^{\top}\beta_t+\xi_{t,i},\qquad
\xi_{t,i}\sim N(0,1),
\]
independently across tasks and observations. We use the whitened OLS
spectral estimator from Section~\ref{sec:instantiations}, including the
map back by $\widehat S^{1/2}$. The earlier upper bound represented by
\citet{niu2024learning} and our bound have respective dependences
\[
O_{\log}\!\left(
\sqrt{\frac{d\lambda_1}{Tn\lambda_k^2}}
+\sqrt{\frac{d}{Tn^2\lambda_k^2}}
\right),
\qquad
\left(
\sqrt{\frac{d}{Tn\lambda_k}}
+\sqrt{\frac{d}{Tn^2\lambda_k^2}}
\right)\wedge1.
\]
Our bound removes the additional dependence on $\lambda_1$ and exhibits
a transition at $n\lambda_k\asymp1$. With the other parameters fixed,
the loss scales as $\lambda_k^{-1/2}$ above this threshold and as
$\lambda_k^{-1}$ below it. Each point in
Figure~\ref{fig:numerical-main} averages 50 independent Monte Carlo
repetitions; error bars are pointwise 95\% Monte Carlo intervals for
the mean loss.

\paragraph{Rare-task design.}
We take $d=20$ and $n=400$, and vary $T/n$ from $1/8$ to $4$.
As in the introduction, $T-1$ task vectors lie along $e_1$ and one lies
along $e_2$, with independent randomized signs. Thus
$(\lambda_1,\lambda_2)=(1-1/T,1/T)$. The earlier bound has dependence
$\sqrt{dT/n}+\sqrt{dT/n^2}$, whereas our bound has dependence
$\sqrt{d/n}+\sqrt{dT/n^2}$. Figure~\ref{fig:numerical-main}(left)
shows that the earlier dependence grows too quickly, while our two-term
expression follows the observed behavior more closely. Both rate curves
are normalized to the empirical loss at the first grid point; they compare
parameter dependence rather than unknown constants or the performance of
competing estimators.

\paragraph{Phase transition.}
We fix $d=5$, $n=1000$, $T=50000$, and $\lambda_1=0.5$, and vary
$n\lambda_2$ over nine logarithmically spaced values from $10^{-2}$ to
$10^2$. We use equal numbers of the four sign patterns with coordinate
magnitudes $\sqrt{\lambda_1}$ and $\sqrt{\lambda_2}$, so that the task
covariance has the prescribed eigenvalues exactly. The large value of $T$
makes both regimes visible without the loss saturating at one over most
of the grid. Separate empirical log--log fits below and above
$n\lambda_2=1$ have slopes $-0.91$ and $-0.54$, respectively, close to
the predicted slopes $-1$ and $-1/2$
(Figure~\ref{fig:numerical-main}, right). These fits are descriptive
rather than numerical theorem bounds.

\begin{figure}[!t]
  \centering
  \includegraphics[width=.48\textwidth]{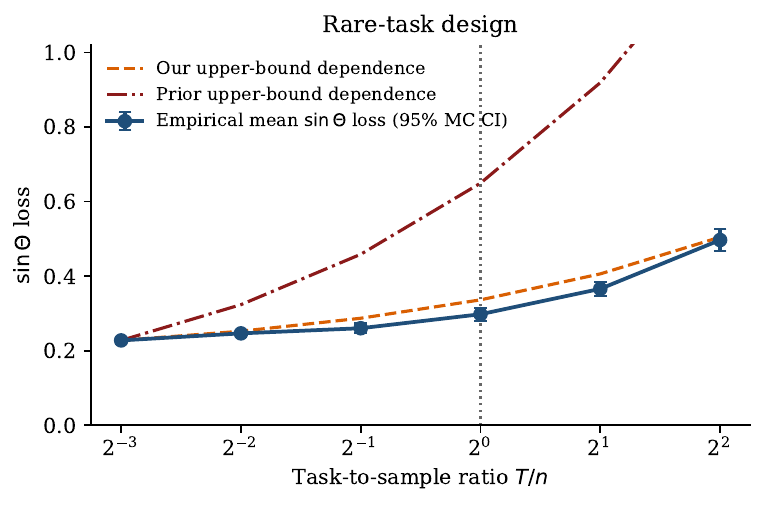}
  \includegraphics[width=.48\textwidth]{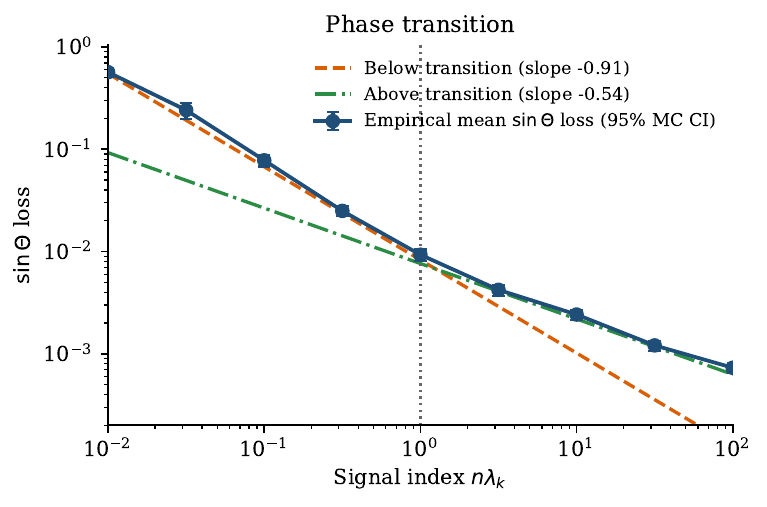}
  \caption{The two main rate phenomena in the Gaussian linear model.
  Left: rare-task design with $d=20$, $n=400$, and $T/n$ varying from
  $1/8$ to $4$. Both rate curves are calibrated at the first grid point
  to compare parameter dependence. Right: phase transition with $d=5$,
  $n=1000$, $T=50000$, and $n\lambda_2$ varying from $10^{-2}$ to $10^2$.
  Dashed and dash-dotted lines are empirical log--log fits below and above
  $n\lambda_2=1$, respectively.}
  \label{fig:numerical-main}
\end{figure}

\paragraph{Logistic classification.}
We also examine the dependence on the task spectrum in a nonlinear model
(Figure~\ref{fig:numerical-logistic-whitening}, panel~(a)). We use isotropic Gaussian
covariates and Bernoulli responses with logistic link
$g(u)=(1+e^{-u})^{-1}$, and apply the first-moment spectral estimator
in~\eqref{eq:gsi-first-moment-estimator}--\eqref{eq:gsi-subspace-estimator}.
For each fixed $\lambda_2\in\{0.01,0.03\}$, fractions $\lambda_1$ and
$\lambda_2$ of the tasks have coefficients $\pm e_1$ and $\pm e_2$,
respectively, with balanced signs; the remaining coefficients are zero.
This gives the exact task spectrum $(\lambda_1,\lambda_2)$ and keeps
all nonzero task norms, hence their logistic alignment, fixed.
After an initial decrease, the loss changes little as $\lambda_1$
increases, while a larger $\lambda_2$ substantially improves recovery,
consistent with the sharper spectrum dependence in
Theorem~\ref{thm:overview-cls}.

\paragraph{Benefit of whitening.}
Figure~\ref{fig:numerical-logistic-whitening}(b) compares raw and whitened OLS under fixed non-isotropic designs
with unit-variance Gaussian noise. Half the tasks have empirical design
covariance $C_{0.6}$ and half have $C_{0.8}$, where $C_\rho$ has unit
diagonal and only the off-diagonal pairs $(1,3)$ and $(2,4)$, with entries
$\rho$ and $0.8\rho$, respectively. Each group uses balanced sign patterns
with task spectrum $(0.3,0.1)$. These designs are uniformly well conditioned
but make the averaged OLS error covariance mix the target subspace with
its orthogonal complement. As $T$ increases, raw aggregation approaches
its population loss of $0.2004$, whereas the whitened loss decreases to
$0.0054$. This illustrates the removal of the persistent bias described
in Remark~\ref{rem:linear-whitening-motivation}.

\begin{figure}[!t]
  \centering
  \includegraphics[width=\textwidth]{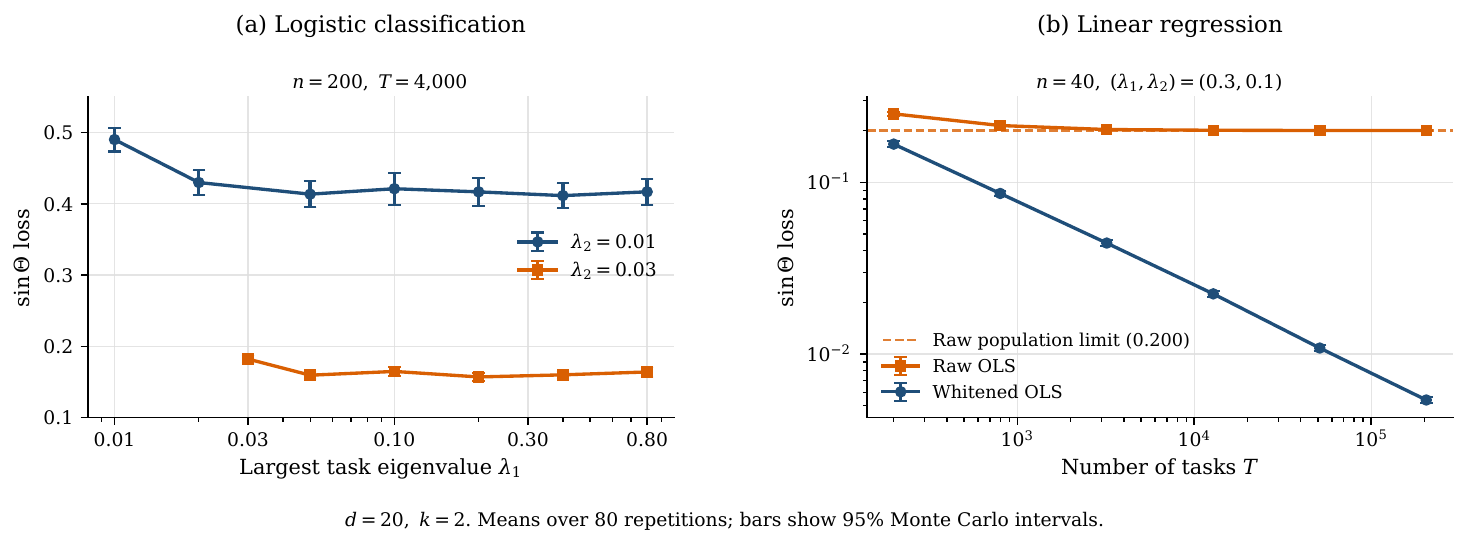}
  \caption{Task-spectrum dependence in logistic classification and the
  benefit of whitening in linear regression. (a) Logistic first-moment
  recovery with $d=20$, $n=200$, and $T=4000$, varying $\lambda_1$ at
  two fixed values of $\lambda_2$. (b) Raw versus whitened OLS with
  $d=20$, $n=40$, and $(\lambda_1,\lambda_2)=(0.3,0.1)$, as $T$ varies
  from $200$ to $204800$ under non-isotropic fixed designs. The dashed
  line is the raw population loss. Each point averages 80 independent
  Monte Carlo repetitions; error bars are pointwise 95\% Monte Carlo
  intervals for the mean $\sin\Theta$ loss.}
  \label{fig:numerical-logistic-whitening}
\end{figure}

	}

	\section{Discussion}\label{sec:discussion}
	{We obtained tight statistical rates for shared-subspace estimation in
		multitask settings, where multiple prediction problems share a common
		low-dimensional representation. Our general theoretical framework separates
		subspace estimation from the construction of taskwise estimators and
		accommodates the heterogeneous errors that arise across tasks. In broad
		parameter regimes, we obtain exactly matching upper and lower bounds, up to
		constants, for linear regression and classification. For additive-noise
		regression, the upper and lower bounds match up to logarithmic factors.}
	
Taken together, our results show that the statistical difficulty is determined
not only by the number of tasks, but also by the geometry of their parameters.
In the models considered here, the weakest represented direction, quantified by
\(\lambda_k\), governs the rate, whereas the largest task-covariance eigenvalue
does not. More precisely, the minimax rate contains the two terms
$
\sqrt{\frac{d}{Tn\lambda_k}}
$ and
$\sqrt{\frac{d}{Tn^2\lambda_k^2}},
$
which exchange dominance at \(n\lambda_k\asymp 1\). Thus, tasks that add
signal in a weak direction can substantially improve recovery, whereas tasks
that add signal only in already well-represented directions need not improve
the relevant bottleneck.

	Extending the theory to approximately shared or nonlinear representations,
	unequal or dependent tasks, and unknown rank remains an important direction
	for future work. With unequal sample sizes, the relevant geometry may involve
	an information-weighted task covariance together with an optimal weighting of
	the taskwise summaries. Approximately shared representations would require
	separating statistical error from approximation bias and determining when
	pooling yields positive rather than negative transfer, particularly when the
	misspecification occurs along weakly represented directions. Dependence across
	tasks may likewise reduce the effective number of tasks or alter the rate more
	fundamentally.
	
	Unknown rank introduces a model-selection problem in addition to subspace
	estimation: a data-driven procedure must distinguish weak but genuine
	directions from the noise floor while retaining sharp dependence on
	\(\lambda_k\). For nonlinear classification,
	Appendix~\ref{sec:nonlinear-general-design} accommodates a common
	non-isotropic Gaussian covariance through whitening, including an unknown
	covariance estimated by sample splitting. Task-dependent covariances and
	non-Gaussian designs remain further challenges, potentially requiring
	score-function identities, higher-order moments, or model-specific
	preconditioning.
	
	\bibliographystyle{abbrvnat}
	\bibliography{references}

	\appendix

	{
\section{A Gaussian Mean-Vector Benchmark}
\label{app:gaussian-means}

This appendix isolates the phase transition in our three multitask
applications through a simpler Gaussian experiment.  The model should be
viewed as an experiment for the taskwise summaries themselves, rather than as
another regression model.  For each \(t\in[T]\), observe
\begin{equation}\label{eq:gaussian-means-model}
Z_t=\beta_t+\sigma\varepsilon_t,
\qquad
\varepsilon_t\stackrel{\mathrm{i.i.d.}}{\sim}N(0,I_d),
\end{equation}
where
\(\mathcal V=\operatorname{span}(\beta_1,\ldots,\beta_T)\) has dimension
\(k\).  Thus \(\sigma^2\) is the variance of a generic taskwise estimate.  In
the applications below, estimating a task parameter from \(n\) observations
typically makes this effective variance of order \(1/n\).

Let \(B,Z\in\R^{T\times d}\) have \(t\)th rows \(\beta_t^\top\) and
\(Z_t^\top\), respectively.  Then
\[
Z=B+\sigma G,
\]
where \(G\) has independent standard Gaussian entries, and
\[
\Sigma_{\bm\beta}=\frac1T B^\top B,
\qquad
\sigma_k^2(B)=T\lambda_k(\Sigma_{\bm\beta}).
\]
Let \(\widehat{\mathcal V}_{\rm gm}\) be the top-\(k\) right singular
subspace of \(Z\).  The following result records both the performance of this
spectral estimator and the corresponding minimax benchmark.

\begin{theorem}[Gaussian mean-vector benchmark]
\label{thm:gaussian-means-benchmark}
In model~\eqref{eq:gaussian-means-model}, let
\(\lambda_k=\lambda_k(\Sigma_{\bm\beta})>0\).  For every \(\tau\ge1\), with
probability at least \(1-e^{-\tau}\),
\begin{equation}\label{eq:gaussian-means-upper}
\sinth{\widehat{\mathcal V}_{\rm gm}}{\mathcal V}
\le
C\left(
\left[
\sigma\sqrt{\frac{d+\tau}{T\lambda_k}}
+
\sigma^2\sqrt{\frac{d+\tau}{T\lambda_k^2}}
\right]\wedge1
\right).
\end{equation}

For \(\lambda_k>0\), define
\[
\mathcal B_k(\lambda_k)
:=
\left\{
B\in\R^{T\times d}:
\operatorname{rank}(B)=k,
\lambda_k(B^\top B/T)\ge\lambda_k
\right\}.
\]
For \(B\in\mathcal B_k(\lambda_k)\), write
\(\mathcal V_B:=\operatorname{span}(B^\top)\).
If \(1\le k\le (T/16)\wedge(d/2)\), then
\begin{equation}\label{eq:gaussian-means-minimax}
\inf_{\widehat{\mathcal V}}\ 
\sup_{B\in\mathcal B_k(\lambda_k)}
\E_B\!\left[
\sinth{\widehat{\mathcal V}}{\mathcal V_B}^2
\right]
\asymp
\left[
\frac{d\sigma^2}{T\lambda_k}
+
\frac{d\sigma^4}{T\lambda_k^2}
\right]\wedge1.
\end{equation}
The constants in the upper bound are universal, and the constants in the
minimax comparison are numerical.
\end{theorem}

\begin{proof}
For the high-probability statement, use the summaries
\(\widehat\beta_t=Z_t\) in Theorem~\ref{thm:mtl-main}.  Their surrogate means
are \(\widetilde\beta_t=\beta_t\), their errors are independent centered
Gaussian vectors with covariance \(\sigma^2I_d\), and hence
\[
\widetilde{\mathcal V}=\mathcal V,\qquad
\widetilde\lambda_k=\lambda_k,\qquad
\rho=\lambda_k+\sigma^2,\qquad
\nu=\sigma^2,\qquad
\delta_1=0.
\]
Theorem~\ref{thm:mtl-main}, followed by
\(\sqrt{\lambda_k+\sigma^2}\le\sqrt{\lambda_k}+\sigma\), gives the two
displayed terms in~\eqref{eq:gaussian-means-upper}, together with
\(\sigma^2(d+\tau)/(T\lambda_k)\).  The latter is the square of the first
displayed term.  It is absorbed by that term whenever the bound is
nontrivial, and otherwise by the truncation at one.

For the minimax statement, divide the observation by \(\sigma\) and apply the
Gaussian matrix-denoising upper and lower bounds of
\citet[Theorems~3 and~4]{cai2018rate} with
\[
p_1=T,\qquad p_2=d,\qquad r=k,\qquad
t^2=\frac{T\lambda_k}{\sigma^2}.
\]
Indeed, \(\sigma_k(B)^2\ge T\lambda_k\) on
\(\mathcal B_k(\lambda_k)\), and the target is the right singular subspace.
Their squared operator sine-theta rate becomes
\[
\left[
\frac{d(t^2+T)}{t^4}
\right]\wedge1
=
\left[
\frac{d\sigma^2}{T\lambda_k}
+
\frac{d\sigma^4}{T\lambda_k^2}
\right]\wedge1.
\]
The rank restrictions in the statement are precisely the source lower
bound's conditions \(r\le p_1/16\) and \(r\le p_2/2\).
\end{proof}

\subsection{Where the two terms come from}

The centered empirical second moment has the exact decomposition
\begin{align}
\frac1T Z^\top Z-\sigma^2I_d
&=
\Sigma_{\bm\beta}
+
\frac{\sigma}{T}(B^\top G+G^\top B)
+
\sigma^2\left(\frac1T G^\top G-I_d\right).
\label{eq:gaussian-means-gram-decomposition}
\end{align}
The first fluctuation is linear in the noise and represents
signal--noise interaction.  The refined subspace perturbation calculation
turns it into the contribution
$
\sigma\sqrt{\frac{d}{T\lambda_k}}.
$
The second fluctuation is quadratic in the noise.  After division by the
population eigengap \(\lambda_k\), it contributes
$
\sigma^2\sqrt{\frac{d}{T\lambda_k^2}}.
$
The ratio of the quadratic contribution to the linear contribution is
$
\frac{\sigma}{\sqrt{\lambda_k}}.
$
Consequently, the transition occurs at
\begin{equation}\label{eq:gaussian-means-transition}
\lambda_k\asymp\sigma^2.
\end{equation}
When \(\lambda_k\gg\sigma^2\), the signal--noise term dominates and the
weakest direction behaves as a standard strong signal.  When
\(\lambda_k\ll\sigma^2\), the centered noise--noise fluctuation dominates.
The lower bound in~\eqref{eq:gaussian-means-minimax} shows that neither
contribution is merely an artifact of the spectral estimator or of our
analysis.

The rare-direction example from the introduction makes this threshold
especially transparent.  If one direction appears in only one of \(T\)
otherwise bounded tasks, then \(\lambda_k\asymp1/T\), and
\eqref{eq:gaussian-means-transition} becomes
\[
T\asymp\sigma^{-2}.
\]
For taskwise estimates based on \(n\) observations,
\(\sigma^2\asymp1/n\), so this is exactly the transition \(T\asymp n\).

\subsection{Connection to the multitask models}

Table~\ref{tab:gaussian-means-correspondence} summarizes how the benchmark
appears in the three applications.  Here ``effective variance'' refers to
the scale of a taskwise summary after accounting for fixed model constants;
it does not assert that the summary is itself Gaussian or has independent
coordinates.

\begin{table}[ht]
\centering
\small
\begin{tabular}{p{0.18\textwidth}p{0.285\textwidth}p{0.195\textwidth}p{0.23\textwidth}}
\toprule
Model & Taskwise summary and population mean
& Effective variance & Relation to the benchmark\\
\midrule
Linear regression
& Least squares; conditionally,
\(\E[\widehat\theta_t\mid X_t]=\beta_t\)
& \(\sigma_\xi^2/n\)
& Exact when \(X_t^\top X_t=nI_d\); whitening gives the heterogeneous
analogue for well-conditioned designs.\\
\addlinespace
Classification
& \(\widehat\beta_t=n^{-1}\sum_iY_{t,i}X_{t,i}\), with
\(\E\widehat\beta_t=a_t\beta_t\)
& \(1/n\), up to link constants
& Gaussian symmetry makes the population cross-block vanish, and
\(a_t\) preserves the task covariance up to fixed factors.\\
\addlinespace
Additive-noise regression
& The same first-moment summary, after truncation for the upper bound
& \(1/n\), inflated by the truncation level
& The resulting logarithmic inflation multiplies the two benchmark terms.\\
\bottomrule
\end{tabular}
\caption{The Gaussian mean-vector benchmark in the three multitask
applications.}
\label{tab:gaussian-means-correspondence}
\end{table}

For linear regression with Gaussian response noise and
\(X_t^\top X_t=nI_d\), the correspondence is literal:
\[
\widehat\theta_t\mid X_t
\sim
N\left(\beta_t,\frac{\sigma_\xi^2}{n}I_d\right).
\]
Thus \(\sigma^2=\sigma_\xi^2/n\) in the benchmark.  With general
well-conditioned designs, the least-squares errors are heterogeneous, but
the whitening step used in Section~\ref{sec:linear} makes their averaged
covariance isotropic.  Our multitask lemma is what allows the Gaussian
benchmark rate to persist without identical row distributions.

For classification, Stein's identity gives
\(\E\widehat\beta_t=a_t\beta_t\), where \(a_t\) is bounded away from zero
under the alignment condition.  Gaussian rotational symmetry gives
\(\delta_1=0\), while averaging \(n\) first-moment observations yields
taskwise error scale \(n^{-1/2}\).  Hence the surrogate signal strength is
comparable to \(\lambda_k\), the effective variance is of order \(1/n\), and
Theorem~\ref{thm:gsi} reproduces the two Gaussian benchmark terms.

The additive-noise model has the same population alignment and \(n^{-1/2}\)
taskwise scale.  Its responses are unbounded, so the upper-bound argument
truncates the first-moment summaries.  The truncation level inflates the
effective noise scale by logarithmic factors, which become the logarithmic
multipliers in Theorem~\ref{thm:gsi-subg}; the lower bound retains the
uninflated Gaussian benchmark.

The last two experiments are therefore not Gaussian mean experiments.
Rather, the multitask lemma shows that their possibly non-Gaussian,
coordinate-dependent errors have the same two relevant fluctuation
mechanisms as~\eqref{eq:gaussian-means-gram-decomposition}.  This is why all
three applications inherit the same transition after the substitution
\(\sigma^2\asymp1/n\).
}

	\section{Proof for Section~\ref{sec:setup}: Downstream Error Decomposition}
	
	\subsection{Proof of Proposition~\ref{prop:downstream-risk}}
	
	\begin{proof}[Proof of Proposition~\ref{prop:downstream-risk}]
		{
			Let \(P=VV^\top\) and \(\widehat P=\widehat V\widehat V^\top\).
			
			Since \(f^\star_{\widehat{\mathcal V}}(\widehat V^\top X)=\E[Y\mid \widehat V^\top X]\), the orthogonality principle for conditional expectation yields
			\[
			\E\!\left[
			\left(\widehat f(\widehat V^\top X)-f^\star_{\widehat{\mathcal V}}(\widehat V^\top X)\right)
			\left(f^\star_{\widehat{\mathcal V}}(\widehat V^\top X)-Y\right)
			\right]
			=0.
			\]
			Therefore, if we write
			\[
			\mathcal R(\widehat f;\widehat{\mathcal V}):=\E\left[\left(Y-\widehat f(\widehat V^\top X)\right)^2\right],
			\]
			then
			\begin{equation}\label{eq:prop1-exact-decomp}
				\mathcal R(\widehat f;\widehat{\mathcal V})
				=
				\E\!\left[\left(\widehat f(\widehat V^\top X)-f^\star_{\widehat{\mathcal V}}(\widehat V^\top X)\right)^2\right]
				+
				\mathcal R(f^\star_{\widehat{\mathcal V}};\widehat{\mathcal V}).
			\end{equation}
			
			Next we compare \(\mathcal R(f^\star_{\widehat{\mathcal V}};\widehat{\mathcal V})\) to
			\[
			\E\left[\left(Y-g(X^\top\beta)\right)^2\right].
			\]
			Since \(\E[Y\mid X]=g(X^\top\beta)\), we have
			\[
			f^\star_{\widehat{\mathcal V}}(\widehat V^\top X)-Y
			=
			\left(f^\star_{\widehat{\mathcal V}}(\widehat V^\top X)-g(X^\top\beta)\right)
			+
			\left(g(X^\top\beta)-Y\right).
			\]
			The first term is measurable with respect to \(X\), while the second has conditional mean zero given \(X\). Hence the cross term vanishes:
			\[
			\E\!\left[
			\left(f^\star_{\widehat{\mathcal V}}(\widehat V^\top X)-g(X^\top\beta)\right)
			\left(g(X^\top\beta)-Y\right)
			\right]
			=0.
			\]
			Expanding the square, we obtain
			\begin{equation}\label{eq:prop1-oracle-gap}
				\mathcal R(f^\star_{\widehat{\mathcal V}};\widehat{\mathcal V})-\E\left[\left(Y-g(X^\top\beta)\right)^2\right]
				=
				\E\!\left[\left(f^\star_{\widehat{\mathcal V}}(\widehat V^\top X)-g(X^\top\beta)\right)^2\right].
			\end{equation}
			
			Now define the candidate predictor
			\[
			\bar f(z):=g(z^\top \widehat V^\top\beta),
			\]
			so that \(\bar f(\widehat V^\top X)=g(X^\top \widehat P\beta)\). Since \(f^\star_{\widehat{\mathcal V}}(\widehat V^\top X)\) is the \(L^2\)-projection of \(Y\) onto the \(\sigma\)-field generated by \(\widehat V^\top X\),
			\[
			\mathcal R(f^\star_{\widehat{\mathcal V}};\widehat{\mathcal V})
			\le
			\mathcal R(\bar f;\widehat{\mathcal V}).
			\]
			Subtracting \(\E[(Y-g(X^\top\beta))^2]\) from both sides and using \eqref{eq:prop1-oracle-gap} gives
			\[
			\E\!\left[\left(f^\star_{\widehat{\mathcal V}}(\widehat V^\top X)-g(X^\top\beta)\right)^2\right]
			\le
			\E\!\left[\left(g(X^\top \widehat P\beta)-g(X^\top \beta)\right)^2\right].
			\]
			The \(L\)-Lipschitz property of \(g\) now gives
			\[
			|g(X^\top \widehat P\beta)-g(X^\top \beta)|
			\le
			L\,|X^\top(\widehat P-P)\beta|,
			\]
			so
			\[
			\mathcal R(f^\star_{\widehat{\mathcal V}};\widehat{\mathcal V})-\E\left[\left(Y-g(X^\top\beta)\right)^2\right]
			\le
			L^2\,\E\!\left[\left(X^\top(\widehat P-P)\beta\right)^2\right].
			\]
			Since \(X\) is mean zero and has covariance matrix \(\Sigma_X\),
			\[
			\E\!\left[\left(X^\top(\widehat P-P)\beta\right)^2\right]
			=
			\beta^\top(\widehat P-P)\Sigma_X(\widehat P-P)\beta
			\le
			\|\Sigma_X\|_{\mathrm{op}}\,\|(\widehat P-P)\beta\|_2^2.
			\]
			Finally, because \(\beta\in\mathcal V=\mathrm{span}(V)\), we have \(P\beta=\beta\), hence
			\[
			\|(\widehat P-P)\beta\|_2
			=
			\|(\widehat P-P)P\beta\|_2
			\le
			\|(\widehat P-P)P\|_{\mathrm{op}}\,\|\beta\|_2
			=
			\sinth{\widehat{\mathcal V}}{\mathcal V}\,\|\beta\|_2.
			\]
			Combining the last three displays yields
			\[
			\mathcal R(f^\star_{\widehat{\mathcal V}};\widehat{\mathcal V})-\E\left[\left(Y-g(X^\top\beta)\right)^2\right]
			\le
			L^2\|\Sigma_X\|_{\mathrm{op}}\|\beta\|_2^2 \cdot \sinth{\widehat{\mathcal V}}{\mathcal V}^2.
			\]
			Together with \eqref{eq:prop1-exact-decomp}, this gives the sharper bound
			\[
			\mathcal E(\widehat f;\widehat{\mathcal V})
			\le
			\E\!\left[\left(\widehat f(\widehat V^\top X)-f^\star_{\widehat{\mathcal V}}(\widehat V^\top X)\right)^2\right]
			+
			L^2\|\Sigma_X\|_{\mathrm{op}}\|\beta\|_2^2\sinth{\widehat{\mathcal V}}{\mathcal V}^2.
			\]
			This is the stated inequality.
		}
	\end{proof}
	
	\section{Proofs for Section~\ref{sec:general}: Multitask Lemma}
	
	\begin{lemma}[Stability under a well-conditioned map]\label{lem:subspace-map-stability}
		Let \(H:\R^d\to\R^d\) be invertible and satisfy
		\[
		m_H\|x\|_2\le \|Hx\|_2\le M_H\|x\|_2
		\qquad\text{for all }x\in\R^d.
		\]
		Then, for any two \(k\)-dimensional subspaces
		\(\mathcal U,\mathcal W\subseteq\R^d\),
		\[
		\sinth{H\mathcal U}{H\mathcal W}
		\le
		\frac{M_H}{m_H}\,
		\sinth{\mathcal U}{\mathcal W}.
		\]
	\end{lemma}
	
	\begin{proof}
		Let \(z\in H\mathcal U\) with \(\|z\|_2=1\), and write
		\(z=Hu/\|Hu\|_2\) for some unit vector \(u\in\mathcal U\). If
		\(w=\Proj_{\mathcal W}u\), then \(Hw\in H\mathcal W\), so
		\[
		\operatorname{dist}(z,H\mathcal W)
		\le
		\frac{\|H(u-w)\|_2}{\|Hu\|_2}
		\le
		\frac{M_H}{m_H}\operatorname{dist}(u,\mathcal W).
		\]
		Taking the supremum over unit \(z\in H\mathcal U\), and using the
		directed-distance representation of \(\sin_\Theta\) for equal-dimensional
		subspaces, gives the result.
	\end{proof}
	
	\subsection{Proof of Theorem~\ref{thm:mtl-main}}
	
	\paragraph{Normalization of the failure probability.}
	The concentration inequalities used below initially give failure probability
	at most \(Ce^{-\tau}\). Replacing \(\tau\) by \(\tau+\log C\) and adjusting
	the constants in the bounds yields the stated failure probability
	\(e^{-\tau}\).
	
	\begin{lemma}[Auxiliary concentration facts]\label{lem:mtl-aux}
		There exist constants \(C,c>0\), depending only on the sub-Gaussian norm constant in Condition~\ref{cond:surrogate-recovery}(1), such that the following hold.
		\begin{enumerate}
			\item If \(\xi_1,\dots,\xi_T\) are independent mean-zero random variables with
			\[
			\|\xi_s\|_{\psi_2}\le K_s
			\qquad\text{for all }s\in[T],
			\]
			then for every \(r\ge 0\),
			\[
			\Prob\!\left(\left|\sum_{s=1}^T \xi_s\right|>r\right)
			\le
			2\exp\!\left(
			-c\,\frac{r^2}{\sum_{s=1}^T K_s^2}
			\right).
			\]
			\item If \(\zeta\) is mean-zero and sub-Gaussian with \(\|\zeta\|_{\psi_2}\le K\), then
			\[
			\|\zeta^2-\E[\zeta^2]\|_{\psi_1}\le CK^2.
			\]
			Consequently, if \(\eta_1,\dots,\eta_T\) are independent mean-zero sub-exponential random variables with \(\|\eta_s\|_{\psi_1}\le L\), then for every \(r\ge 0\),
			\[
			\Prob\!\left(\left|\sum_{s=1}^T \eta_s\right|>r\right)
			\le
			2\exp\!\left(
			-c\min\!\left\{
			\frac{r^2}{TL^2},\frac{r}{L}
			\right\}
			\right).
			\]
			\item If \(\zeta,\zeta'\) are sub-Gaussian with \(\|\zeta\|_{\psi_2}\le K\) and \(\|\zeta'\|_{\psi_2}\le K'\), then
			\[
			\|\zeta\zeta'-\E[\zeta\zeta']\|_{\psi_1}\le CKK'.
			\]
		\end{enumerate}
	\end{lemma}
	
	\begin{proof}
		These are standard consequences of sub-Gaussian and sub-exponential calculus; see, for example, Chapters~2 and~4 of \cite{vershynin2018hdp}.
	\end{proof}
	
	We will also use the standard \(1/4\)-net bounds
	\begin{equation}\label{eq:mtl-net-symm}
		\|H\|_{\mathrm{op}}
		\le
		2\sup_{x\in\mathcal N_m}|x^\top Hx|
	\end{equation}
	for symmetric \(H\in\R^{m\times m}\), and
	\begin{equation}\label{eq:mtl-net-rect}
		\|R\|_{\mathrm{op}}
		\le
		2\sup_{x\in\mathcal N_m,\ y\in\mathcal N_n}|x^\top Ry|
	\end{equation}
	for rectangular \(R\in\R^{m\times n}\), where \(\mathcal N_m\subset S^{m-1}\) and
	\(\mathcal N_n\subset S^{n-1}\) are \(1/4\)-nets with cardinalities at most
	\(9^m\) and \(9^n\), respectively; see Lemma~4.4.1 of \cite{vershynin2018hdp}.
	
	\begin{lemma}[Population block structure]\label{lem:mtl-population}
		Let \(\widetilde A:=\E[\widehat A]\), let \(Z:=\widehat A-\widetilde A\), and define
		\[
		\Sigma_{\mathrm{sig}}
		:=
		\widetilde V^\top
		\left[
		\frac1T\widetilde A^\top\widetilde A+\Sigma_{\bm\varepsilon}
		\right]
		\widetilde V.
		\]
		Then
		\[
		\Sigma_{\mathrm{sig}}=\frac1T\E[Y^\top Y],
		\qquad
		\lambda_{\min}(\Sigma_{\mathrm{sig}})=\rho,
		\]
		and
		\[
		\frac1T\E[Y_\perp^\top Y_\perp]
		=
		\widetilde V_\perp^\top \Sigma_{\bm\varepsilon}\widetilde V_\perp.
		\]
		Moreover,
		\[
		\widetilde V^\top \E[Z^\top Z]\widetilde V_\perp
		=
		T\widetilde V^\top \Sigma_{\bm\varepsilon}\widetilde V_\perp
		\]
		{Consequently,
			\[
			\left\|
			\widetilde V^\top \E[Z^\top Z]\widetilde V_\perp
			\right\|_{\mathrm{op}}
			=
			T\delta_1\widetilde\lambda_k.
			\]}
	\end{lemma}
	
	\begin{proof}
		Since the rows of \(Z\) are mean zero and \(\widetilde A\widetilde V_\perp=0\),
		\[
		Y=\widetilde A\widetilde V+Z\widetilde V,
		\qquad
		Y_\perp=Z\widetilde V_\perp.
		\]
		Therefore
		\[
		\E[Y^\top Y]
		=
		\widetilde V^\top \widetilde A^\top\widetilde A\widetilde V
		+
		\widetilde V^\top \E[Z^\top Z]\widetilde V,
		\]
		which gives the formula for \(\Sigma_{\mathrm{sig}}\). The identity
		\(\lambda_{\min}(\Sigma_{\mathrm{sig}})=\rho\) is exactly the definition of
		\(\rho\) in Condition~\ref{cond:surrogate-recovery}(2). The formula for
		\(\E[Y_\perp^\top Y_\perp]\) follows from \(Y_\perp=Z\widetilde V_\perp\), and
		{the mixed-block identity follows from
			\(\E[Z^\top Z]=T\Sigma_{\bm\varepsilon}\). Its operator norm is
			\(T\delta_1\widetilde\lambda_k\) by~{\eqref{eq:mixing}} and
			invariance under the
			orthonormal coordinate maps \(\widetilde V\) and \(\widetilde V_\perp\).}
	\end{proof}
	
	\begin{lemma}[Signal-block concentration]\label{lem:mtl-signal}
		Define
		\[
		N:=\Sigma_{\mathrm{sig}}^{-1/2},
		\qquad
		G:=N^\top\left(\frac1T Y^\top Y\right)N.
		\]
		Then for every \(\tau\ge 1\), with probability at least \(1-e^{-\tau}\),
		\begin{equation}\label{eq:mtl-signal-conc}
			\|G-I_k\|_{\mathrm{op}}
			\le
			\delta_{\tau},
		\end{equation}
		where
		\begin{equation}\label{eq:mtl-delta}
			\delta_{\tau}
			\le
			C\left[
			\sigma\sqrt{\frac{d+\tau}{T\rho}}
			+
			\sigma^2\sqrt{\frac{d+\tau}{T\rho^2}}
			+
			\frac{\sigma^2(d+\tau)}{T\rho}
			\right].
		\end{equation}
		Consequently,
		\begin{equation}\label{eq:mtl-signal-lb}
			\sigma_k^2(Y)\ge (1-\delta_{\tau})T\rho.
		\end{equation}
	\end{lemma}
	
	\begin{proof}
		By Lemma~\ref{lem:mtl-population}, \(\E[G]=I_k\). Fix \(a\in S^{k-1}\), and set
		\[
		\alpha:=\widetilde A\widetilde VNa\in\R^T,
		\qquad
		v:=\widetilde VNa\in\R^d.
		\]
		Then
		\[
		a^\top(G-I_k)a
		=
		\frac{2}{T}\alpha^\top Zv
		+
		\frac{1}{T}\Bigl(v^\top Z^\top Zv-\E[v^\top Z^\top Zv]\Bigr).
		\]
		The normalization gives
		\[
		\|v\|_2^2=a^\top N^2a\le \rho^{-1}
		\]
		and
		\[
		\frac1T\|\alpha\|_2^2
		=
		a^\top N\widetilde V^\top
		\left(\frac1T\widetilde A^\top\widetilde A\right)\widetilde VNa
		\le
		a^\top N\Sigma_{\mathrm{sig}}Na
		=1.
		\]
		For the linear term, the summands
		\[
		\xi_s:=\alpha_s\langle Z_s,v\rangle
		\]
		are independent, mean zero, and satisfy
		\[
		\|\xi_s\|_{\psi_2}\le C|\alpha_s|\sigma\|v\|_2.
		\]
		Lemma~\ref{lem:mtl-aux}(1) therefore yields, for every fixed \(a\),
		\[
		\Prob\!\left(
		\left|\frac{2}{T}\alpha^\top Zv\right|>r
		\right)
		\le
		2\exp\!\left(-c\,\frac{Tr^2\rho}{\sigma^2}\right).
		\]
		Hence, with probability at least \(1-e^{-\tau}\),
		\[
		\left|\frac{2}{T}\alpha^\top Zv\right|
		\le
		C\sigma\sqrt{\frac{d+\tau}{T\rho}}.
		\]
		
		For the quadratic term, the variables
		\[
		\eta_s:=\langle Z_s,v\rangle^2-\E[\langle Z_s,v\rangle^2]
		\]
		are independent, mean zero, and satisfy
		\[
		\|\eta_s\|_{\psi_1}\le C\sigma^2\|v\|_2^2\le C\sigma^2/\rho
		\]
		by Lemma~\ref{lem:mtl-aux}(2). Bernstein's inequality therefore gives, for every fixed \(a\),
		\[
		\left|
		\frac{1}{T}\Bigl(v^\top Z^\top Zv-\E[v^\top Z^\top Zv]\Bigr)
		\right|
		\le
		C\left[
		\sigma^2\sqrt{\frac{d+\tau}{T\rho^2}}
		+
		\frac{\sigma^2(d+\tau)}{T\rho}
		\right]
		\]
		with probability at least \(1-e^{-\tau}\).
		
		Combining the last two displays, taking a union bound over a \(1/4\)-net of
		\(S^{k-1}\), and then applying \eqref{eq:mtl-net-symm} proves
		\eqref{eq:mtl-signal-conc}. Finally,
		\[
		\frac1T Y^\top Y=N^{-1}GN^{-1},
		\]
		so
		\[
		\lambda_{\min}\!\left(\frac1T Y^\top Y\right)
		\ge
		\lambda_{\min}(G)\lambda_{\min}(\Sigma_{\mathrm{sig}})
		\ge
		(1-\delta_{\tau})\rho,
		\]
		which is exactly \eqref{eq:mtl-signal-lb}.
	\end{proof}
	
	\begin{lemma}[Noise-floor bound]\label{lem:mtl-noise}
		For every \(\tau\ge 1\), with probability at least \(1-e^{-\tau}\),
		\begin{equation}\label{eq:mtl-noise-floor}
			\sigma_{k+1}^2(\widehat A)
			\le
			T\nu+C\sigma^2\bigl(\sqrt{T(d+\tau)}+(d+\tau)\bigr).
		\end{equation}
	\end{lemma}
	
	\begin{proof}
		Because \(\widetilde A\widetilde V_\perp=0\),
		\[
		\sigma_{k+1}(\widehat A)\le \|Z\widetilde V_\perp\|_{\mathrm{op}}.
		\]
		By Lemma~\ref{lem:mtl-population},
		\[
		\E[\widetilde V_\perp^\top Z^\top Z\widetilde V_\perp]
		=
		T\widetilde V_\perp^\top \Sigma_{\bm\varepsilon}\widetilde V_\perp.
		\]
		Fix \(b\in S^{d-k-1}\) and set \(z:=\widetilde V_\perp b\). Then
		\[
		b^\top \widetilde V_\perp^\top
		\bigl(Z^\top Z-\E[Z^\top Z]\bigr)\widetilde V_\perp b
		=
		\sum_{s=1}^T \Bigl(\langle Z_s,z\rangle^2-\E[\langle Z_s,z\rangle^2]\Bigr).
		\]
		Each summand is centered sub-exponential with \(\psi_1\)-norm at most
		\(C\sigma^2\) by Lemma~\ref{lem:mtl-aux}(2). Bernstein's inequality and a
		\(1/4\)-net argument therefore give
		\[
		\left\|
		\widetilde V_\perp^\top
		\bigl(Z^\top Z-\E[Z^\top Z]\bigr)\widetilde V_\perp
		\right\|_{\mathrm{op}}
		\le
		C\sigma^2\bigl(\sqrt{T(d+\tau)}+(d+\tau)\bigr)
		\]
		with probability at least \(1-e^{-\tau}\). Since
		\[
		\lambda_{\max}(\widetilde V_\perp^\top\Sigma_{\bm\varepsilon}\widetilde V_\perp)=\nu,
		\]
		this proves \eqref{eq:mtl-noise-floor}.
	\end{proof}
	
	\begin{lemma}[Cross-block bound]\label{lem:mtl-cross}
		For every \(\tau\ge 1\) such that \(\delta_{\tau}\le 1/2\), with probability at least
		\(1-e^{-\tau}\),
		\begin{equation}\label{eq:mtl-cross}
			\|\Proj_YY_\perp\|_{\mathrm{op}}
			\le
			C
			\left[
			\sigma\sqrt{d+\tau}
			+
			\frac{\sigma^2\sqrt{d+\tau}}{\sqrt{\rho}}
			+
			\frac{\sigma^2(d+\tau)}{\sqrt{T\rho}}
			+
			{\frac{\sqrt{T}\,\delta_1\widetilde\lambda_k}{\sqrt{\rho}}}
			\right],
		\end{equation}
		where \(\delta_{\tau}\) is the quantity from Lemma~\ref{lem:mtl-signal}.
	\end{lemma}
	
	\begin{proof}
		Since \(\Proj_Y=\Proj_{YN}\),
		\[
		\|\Proj_YY_\perp\|_{\mathrm{op}}
		\le
		T^{-1/2}\|G^{-1/2}\|_{\mathrm{op}}\,
		\|N^\top Y^\top Y_\perp\|_{\mathrm{op}}.
		\]
		Moreover,
		\begin{align*}
			N^\top Y^\top Y_\perp
			&=
			N^\top \widetilde V^\top \widetilde A^\top Z\widetilde V_\perp
			+
			N^\top \widetilde V^\top Z^\top Z\widetilde V_\perp \\
			&=
			N^\top \widetilde V^\top \widetilde A^\top Z\widetilde V_\perp
			+
			N^\top \widetilde V^\top
			\bigl(Z^\top Z-\E[Z^\top Z]\bigr)\widetilde V_\perp\\
			&\quad+
			{N^\top\widetilde V^\top
				\E[Z^\top Z]\widetilde V_\perp},
		\end{align*}
		where the second identity separates the centered quadratic fluctuation from
		its population cross-block.
		
		For the linear term, fix \(a\in S^{k-1}\), \(b\in S^{d-k-1}\), and set
		\[
		\alpha:=\widetilde A\widetilde VNa\in\R^T.
		\]
		Then
		\[
		a^\top N^\top \widetilde V^\top \widetilde A^\top Z\widetilde V_\perp b
		=
		\sum_{s=1}^T \alpha_s\langle Z_s,\widetilde V_\perp b\rangle.
		\]
		Since \(\|\alpha\|_2^2\le T\), Lemma~\ref{lem:mtl-aux}(1), followed by a
		rectangular \(1/4\)-net argument, gives
		\[
		\left\|
		N^\top \widetilde V^\top \widetilde A^\top Z\widetilde V_\perp
		\right\|_{\mathrm{op}}
		\le
		C\sigma\sqrt{T(d+\tau)}
		\]
		with probability at least \(1-e^{-\tau}\).
		
		For the quadratic term, fix \(a\in S^{k-1}\), \(b\in S^{d-k-1}\), and define
		\[
		w:=\widetilde VNa,
		\qquad
		z:=\widetilde V_\perp b.
		\]
		Then
		\[
		a^\top N^\top \widetilde V^\top
		\bigl(Z^\top Z-\E[Z^\top Z]\bigr)\widetilde V_\perp b
		=
		\sum_{s=1}^T
		\Bigl(
		\langle Z_s,w\rangle\langle Z_s,z\rangle
		-\E[\langle Z_s,w\rangle\langle Z_s,z\rangle]
		\Bigr).
		\]
		Because \(\|w\|_2\le \rho^{-1/2}\), Lemma~\ref{lem:mtl-aux}(3) shows that each
		summand is centered sub-exponential with \(\psi_1\)-norm at most
		\(C\sigma^2/\sqrt{\rho}\). Bernstein's inequality and another rectangular
		\(1/4\)-net argument therefore yield
		\[
		\left\|
		N^\top \widetilde V^\top
		\bigl(Z^\top Z-\E[Z^\top Z]\bigr)\widetilde V_\perp
		\right\|_{\mathrm{op}}
		\le
		C\left[
		\frac{\sigma^2\sqrt{T(d+\tau)}}{\sqrt{\rho}}
		+
		\frac{\sigma^2(d+\tau)}{\sqrt{\rho}}
		\right]
		\]
		with probability at least \(1-e^{-\tau}\).
		
		{For the population term, Lemma~\ref{lem:mtl-population} and
			\(\|N\|_{\mathrm{op}}=\rho^{-1/2}\) give
			\[
			\left\|
			N^\top\widetilde V^\top
			\E[Z^\top Z]\widetilde V_\perp
			\right\|_{\mathrm{op}}
			\le
			\frac{T\delta_1\widetilde\lambda_k}{\sqrt{\rho}}.
			\]}
		
		Combining the previous displays and using
		\[
		\|G^{-1/2}\|_{\mathrm{op}}\le (1-\delta_{\tau})^{-1/2}\le \sqrt{2}
		\]
		on the event \(\|G-I_k\|_{\mathrm{op}}\le \delta_{\tau}\) proves
		\eqref{eq:mtl-cross}.
	\end{proof}
	
	\begin{proof}
		Let \(\widetilde A:=\E[\widehat A]\), so that the \(t\)-th row of \(\widetilde A\) is \(\tilde\beta_t^\top\), and write
		\[
		Z:=\widehat A-\widetilde A.
		\]
		Then the \(t\)-th row of \(Z\) is \((\hat\beta_t-\tilde\beta_t)^\top\), the rows of \(Z\) are independent, and
		\[
		\widehat A=\widetilde A+Z.
		\]
		Since \(\widetilde{\mathcal V}=\mathrm{span}(\tilde\beta_1,\dots,\tilde\beta_T)\), we have
		\[
		\widetilde A\widetilde V_\perp=0,
		\qquad
		Y=\widetilde A\widetilde V+Z\widetilde V,
		\qquad
		Y_\perp=Z\widetilde V_\perp.
		\]
		
		Define the population signal-block matrix
		\[
		\Sigma_{\mathrm{sig}}
		:=
		\widetilde V^\top
		\left[
		\frac1T\sum_{t=1}^T \tilde\beta_t\tilde\beta_t^\top+\Sigma_{\bm\varepsilon}
		\right]
		\widetilde V
		=
		\frac1T\E[Y^\top Y].
		\]
		By definition, \(\lambda_{\min}(\Sigma_{\mathrm{sig}})=\rho\). Let
		\[
		N:=\Sigma_{\mathrm{sig}}^{-1/2},
		\qquad
		G:=N^\top\left(\frac1T Y^\top Y\right)N.
		\]
		Then \(\E[G]=I_k\).
		
		For the proof, set
		\[
		\eta_{\tau}
		:=
		\sigma\sqrt{\frac{(d+\tau)\rho}{T}}
		+
		\sigma^2\sqrt{\frac{d+\tau}{T}}
		+
		\frac{\sigma^2(d+\tau)}{T}.
		\]
		If \(\eta_{\tau}>c\widetilde\lambda_k\) for a sufficiently small constant \(c>0\), then
		\[
		\sigma\sqrt{\frac{(d+\tau)\rho}{T\widetilde\lambda_k^2}}
		+
		\sigma^2\sqrt{\frac{d+\tau}{T\widetilde\lambda_k^2}}
		+
		\frac{\sigma^2(d+\tau)}{T\widetilde\lambda_k}
		+
		{\delta_1}
		\ge c,
		\]
		so the stated bound is trivial after truncation by \(1\). It therefore suffices to work on the complementary regime
		\[
		\eta_{\tau}\le c\widetilde\lambda_k.
		\]
		
		Let \(N\), \(G\), and \(\delta_{\tau}\) be as in Lemma~\ref{lem:mtl-signal}. By
		\eqref{eq:mtl-delta},
		\[
		\delta_{\tau}\rho\le C\eta_{\tau}.
		\]
		Because \(\rho\ge\rho-\nu\ge\gamma_0\widetilde\lambda_k\), the regime
		\(\eta_{\tau}\le c\widetilde\lambda_k\) implies \(\delta_{\tau}\le 1/2\) when \(c\) is
		sufficiently small. We may therefore invoke the cross-block lemma. By
		Lemmas~\ref{lem:mtl-signal},~\ref{lem:mtl-noise}, and~\ref{lem:mtl-cross},
		with probability at least \(1-e^{-\tau}\) we have
		\[
		\sigma_k^2(Y)\ge (1-\delta_{\tau})T\rho,
		\]
		\[
		\sigma_{k+1}^2(\widehat A)
		\le
		T\nu+C\sigma^2\bigl(\sqrt{T(d+\tau)}+(d+\tau)\bigr),
		\]
		and
		\[
		\|\Proj_YY_\perp\|_{\mathrm{op}}
		\le
		C
		\left[
		\sigma\sqrt{d+\tau}
		+
		\frac{\sigma^2\sqrt{d+\tau}}{\sqrt{\rho}}
		+
		\frac{\sigma^2(d+\tau)}{\sqrt{T\rho}}
		+
		{\frac{\sqrt{T}\,\delta_1\widetilde\lambda_k}{\sqrt{\rho}}}
		\right].
		\]
		
		We now work on the intersection of these events. Since
		\[
		\delta_{\tau} T\rho
		\le
		C\left[
		\sigma\sqrt{T(d+\tau)\rho}
		+
		\sigma^2\sqrt{T(d+\tau)}
		+
		\sigma^2(d+\tau)
		\right]
		=
		CT\eta_{\tau},
		\]
		the assumption \(\eta_{\tau}\le c\widetilde\lambda_k\) and the population gap
		\(\rho-\nu\ge\gamma_0\widetilde\lambda_k\) imply, for \(c>0\) small enough,
		\[
		\sigma_k^2(Y)-\sigma_{k+1}^2(\widehat A)
		\ge
		T(\rho-\nu)-CT\eta_{\tau}
		\ge
		\frac{\gamma_0}{2}T\widetilde\lambda_k.
		\]
		Now let
		\[
		L^2:=(1-\delta_{\tau})T\rho,
		\qquad
		U^2:=T\nu+C\sigma^2\bigl(\sqrt{T(d+\tau)}+(d+\tau)\bigr).
		\]
		Then \(\sigma_k^2(Y)\ge L^2\), \(\sigma_{k+1}^2(\widehat A)\le U^2\), and
		\[
		L^2-U^2\ge \frac{\gamma_0}{2}T\widetilde\lambda_k,
		\qquad
		L\le \sqrt{T\rho}.
		\]
		Since \(x\mapsto x/(x^2-U^2)\) is decreasing on \((U,\infty)\),
		\[
		\frac{\sigma_k(Y)}{\sigma_k^2(Y)-\sigma_{k+1}^2(\widehat A)}
		\le
		\frac{L}{L^2-U^2}
		\le
		C\frac{\sqrt{\rho/T}}{\widetilde\lambda_k}.
		\]
		Applying the deterministic reduction~\eqref{eq:mtl-deterministic-reduction}
		and the bound on \(\|\Proj_YY_\perp\|_{\mathrm{op}}\) yields
		\begin{align*}
			\sinth{\widehat{\mathcal V}^{(s)}}{\widetilde{\mathcal V}}
			&\le
			C\frac{\sqrt{\rho/T}}{\widetilde\lambda_k}
			\left[
			\sigma\sqrt{d+\tau}
			+
			\frac{\sigma^2\sqrt{d+\tau}}{\sqrt{\rho}}
			+
			\frac{\sigma^2(d+\tau)}{\sqrt{T\rho}}
			+
			{\frac{\sqrt{T}\,\delta_1\widetilde\lambda_k}{\sqrt{\rho}}}
			\right]\\
			&\le
			C\left[
			\sigma\sqrt{\frac{(d+\tau)\rho}{T\widetilde\lambda_k^2}}
			+
			\sigma^2\sqrt{\frac{d+\tau}{T\widetilde\lambda_k^2}}
			+
			\frac{\sigma^2(d+\tau)}{T\widetilde\lambda_k}
			+
			{\delta_1}
			\right].
		\end{align*}
		Combining this with the trivial regime \(\eta_{\tau}>c\widetilde\lambda_k\) proves the displayed bound with the final truncation by \(1\).
	\end{proof}
	
	\section{Proofs for Section~\ref{sec:instantiations}}
	
	\subsection{Proof of Theorem~\ref{thm:linear}}
	
	\begin{proof}
		Condition on \(X_1,\dots,X_T\); all expectations and covariances below are
		conditional on these design matrices. Write
		\[
		\hat\theta_t=\beta_t+\eta_t,
		\qquad
		\eta_t:=(X_t^\top X_t)^{-1}X_t^\top \xi_t.
		\]
		By definition of the whitened surrogate,
		\[
		\hat\beta_t=\widehat S^{-1/2}\hat\theta_t,
		\qquad
		\tilde\beta_t=\E[\hat\beta_t\mid X_1,\dots,X_T]=\widehat S^{-1/2}\beta_t.
		\]
		{Hence the surrogate means span
			\(\widetilde{\mathcal V}=\widehat S^{-1/2}\mathcal V\). The required
			model-specific transformation is the linear map
			\[
			H=\widehat S^{1/2},
			\qquad
			H\widetilde{\mathcal V}=\mathcal V,
			\qquad
			H\widehat{\mathcal V}^{(s)}=\widehat{\mathcal V}.
			\]
			Moreover,
			\[
			\frac1{\kappa_{\max}}I_d\preceq \widehat S\preceq \frac1{\kappa_{\min}}I_d,
			\]
			so Lemma~\ref{lem:subspace-map-stability} gives
			\[
			\sinth{\widehat{\mathcal V}}{\mathcal V}
			\le
			\sqrt{\frac{\kappa_{\max}}{\kappa_{\min}}}\,
			\sinth{\widehat{\mathcal V}^{(s)}}{\widetilde{\mathcal V}}.
			\]
			Thus it remains only to apply the multitask lemma in the surrogate
			coordinates and compare the surrogate and target signal strengths.}
		
		Next, the residual rows are
		\[
		\hat\beta_t-\tilde\beta_t=\widehat S^{-1/2}\eta_t.
		\]
		For any \(a\in\R^d\),
		\[
		\langle \hat\beta_t-\tilde\beta_t,a\rangle
		=
		\xi_t^\top X_t (X_t^\top X_t)^{-1}\widehat S^{-1/2}a.
		\]
		Because the coordinates of \(\xi_t\) are independent sub-Gaussian and
		\[
		\|X_t (X_t^\top X_t)^{-1}\widehat S^{-1/2}a\|_2^2
		=
		a^\top \widehat S^{-1/2}(X_t^\top X_t)^{-1}\widehat S^{-1/2}a
		\le
		\frac{C}{n}\|a\|_2^2,
		\]
		Condition~\ref{cond:surrogate-recovery}(1) holds with \(\sigma^2\le C\sigma_\xi^2/n\).
		
		Moreover,
		\[
		\E[\eta_t\eta_t^\top\mid X_t]=\sigma_\xi^2(X_t^\top X_t)^{-1},
		\]
		and therefore
		\begin{align*}
			\Sigma_{\bm\varepsilon}
			&=
			\frac1T\sum_{t=1}^T \widehat S^{-1/2}\E[\eta_t\eta_t^\top\mid X_t]\widehat S^{-1/2}\\
			&=
			\frac{\sigma_\xi^2}{T}\widehat S^{-1/2}\left(\sum_{t=1}^T(X_t^\top X_t)^{-1}\right)\widehat S^{-1/2}\\
			&=
			\frac{\sigma_\xi^2}{n}I_d.
		\end{align*}
		{Thus the residual cross-block vanishes, while the orthogonal-block
			eigenvalue is
			\[
			\delta_1=0,
			\qquad
			\nu=\frac{\sigma_\xi^2}{n}.
			\]}
		Since the signal term is supported on \(\widetilde{\mathcal V}\),
		\[
		\rho=
		\widetilde\lambda_k+\frac{\sigma_\xi^2}{n},
		\]
		so \(\rho-\nu=\widetilde\lambda_k\), verifying Condition~\ref{cond:surrogate-recovery}(2).
		
		{Applying Theorem~\ref{thm:mtl-main} to
			\(\widehat{\mathcal V}^{(s)}\), using the preceding stability inequality and
			\(\delta_1=0\), gives
			\[
			\sinth{\widehat{\mathcal V}}{\mathcal V}
			\le
			C\left(
			\left[
			\sigma\sqrt{\frac{(d+\tau)\rho}{T\widetilde\lambda_k^2}}
			+
			\sigma^2\sqrt{\frac{d+\tau}{T\widetilde\lambda_k^2}}
			+
			\frac{\sigma^2(d+\tau)}{T\widetilde\lambda_k}
			\right]\wedge 1
			\right).
			\]
			The design bounds also imply
			\[
			\widetilde\lambda_k
			=
			\lambda_k\!\left(\widehat S^{-1/2}\Sigma_{\bm\beta}\widehat S^{-1/2}\right)
			\asymp \lambda_k,
			\]
			with constants depending only on \(\kappa_{\min},\kappa_{\max}\). Absorbing
			these fixed conditioning constants yields
			\[
			\sinth{\widehat{\mathcal V}}{\mathcal V}
			\le
			C\left(
			\left[
			\sigma\sqrt{\frac{(d+\tau)\rho}{T\lambda_k^2}}
			+
			\sigma^2\sqrt{\frac{d+\tau}{T\lambda_k^2}}
			+
			\frac{\sigma^2(d+\tau)}{T\lambda_k}
			\right]\wedge 1
			\right).
			\]}
		Substituting \(\sigma^2\le C\sigma_\xi^2/n\) and \(\rho=\widetilde\lambda_k+\sigma_\xi^2/n\), and using
		\[
		\widetilde\lambda_k
		=
		\lambda_k\!\left(\widehat S^{-1/2}\left[\frac1T\sum_{t=1}^T\beta_t\beta_t^\top\right]\widehat S^{-1/2}\right)
		\asymp
		\lambda_k
		\]
		with constants depending only on \(\kappa_{\min},\kappa_{\max}\), gives
		\[
		\sigma\sqrt{\frac{(d+\tau)\rho}{T\lambda_k^2}}
		\le
		C\left[
		\sigma_\xi\sqrt{\frac{d+\tau}{Tn\lambda_k}}
		+
		\sigma_\xi^2\sqrt{\frac{d+\tau}{Tn^2\lambda_k^2}}
		\right].
		\]
		Finally,
		\[
		\frac{\sigma^2(d+\tau)}{T\lambda_k}
		=
		\left(\sigma\sqrt{\frac{d+\tau}{T\lambda_k}}\right)^2.
		\]
		Thus this third term is bounded by the leading term when the latter is at most
		one, while otherwise the outer truncation makes the claimed bound immediate.
		This yields the stated bound.
	\end{proof}
	
	\subsection{Proof of Theorem~\ref{thm:linear-lb}}
	
	\begin{proof}
		\medskip\noindent\textbf{Existing result.}
		We use Theorem~5.1 and its proof in Section~9.2 of
		\citet[arXiv:2409.04919v3, revised November 22, 2025]{niu2024learning}.
		In the notation of that paper, there are \(M\) clients, client \(i\) has
		\(n_i\) observations, and \(N=\sum_{i=1}^M n_i\). The shared representation is
		an orthonormal matrix \(B^\star\in\R^{d\times k}\), and the client parameters
		satisfy
		\[
		\Gamma_i\theta_i^\star=B^\star\alpha_i^\star,
		\qquad
		\|\alpha_i^\star\|_2\le C_\alpha,
		\]
		where \(C_\alpha\) is a fixed numerical constant. For
		\[
		D_{\rm Niu}
		:=
		\frac1N\sum_{i=1}^M
		n_i\alpha_i^\star(\alpha_i^\star)^\top,
		\]
		their parameter class imposes, up to fixed numerical constants,
		\[
		c_\alpha\mu_k I_k
		\preceq D_{\rm Niu}\preceq
		C_\alpha'\mu_1 I_k.
		\]
		Here \(\mu_1\) and \(\mu_k\) denote the nominal largest and smallest diversity
		eigenvalues in the source theorem. Under standard Gaussian covariates and
		standard Gaussian noise, Theorem~5.1 gives
		\[
		\inf_{\widehat B}
		\sup_{B^\star,\{n_i\},\{\alpha_i^\star\}}
		\E\!\left[\|\sin\Theta(\widehat B,B^\star)\|_{\rm op}\right]
		\ge
		c\left(
		\left[
		\sqrt{\frac{d}{N\mu_k}}
		+
		\sqrt{\frac{Md}{N^2\mu_k^2}}
		\right]\wedge1
		\right)
		\]
		when \(k\ge c_0\log M\), \(d\ge(1+\rho_1)k\), and
		\(M\ge(1+\rho_2)k\). The constants in this display depend only on the fixed
		constants in the source parameter class and on \(\rho_1,\rho_2\).
		
		\medskip\noindent\textbf{Equal-sample specialization.}
		The statement of Theorem~5.1 permits heterogeneous sample allocations, so its
		display alone does not imply a lower bound for a prescribed equal allocation.
		The proof constructions in Section~9.2, however, use the equal allocation
		\[
		n_i=N/M\qquad\text{for every }i
		\]
		for both terms: the proof of the first term makes this choice explicitly, and
		the construction for the second term fixes the same allocation in
		Lemma~9.2. We therefore specialize those equal-allocation constructions, while
		retaining the packing and divergence arguments proved there.
		
		\medskip\noindent\textbf{Notation and model correspondence.}
		The correspondence between the source notation and ours is
		\[
		\begin{array}{c|c}
			\text{\citet{niu2024learning}} & \text{present paper}\\
			\hline
			M & T\\
			n_i & n\\
			N=\sum_i n_i & Tn\\
			B^\star & V\\
			\theta_i^\star & \beta_t\\
			\alpha_i^\star & \alpha_t\\
			\Gamma_i & I_d\\
			D_{\rm Niu} & T^{-1}\sum_{t=1}^T\alpha_t\alpha_t^\top.
		\end{array}
		\]
		Indeed, our Gaussian covariates give \(\Gamma_i=I_d\), so the source structural
		condition becomes
		\[
		\beta_t=V\alpha_t.
		\]
		Moreover,
		\[
		\Sigma_{\bm\beta}
		=
		\frac1T\sum_{t=1}^T\beta_t\beta_t^\top
		=
		V\left(\frac1T\sum_{t=1}^T\alpha_t\alpha_t^\top\right)V^\top.
		\]
		Consequently, the nonzero eigenvalues of \(\Sigma_{\bm\beta}\) are exactly the
		eigenvalues of \(D_{\rm Niu}\) under the equal allocation.
		{The source loss
			\(\|\sin\Theta(\widehat B,B^\star)\|_{\rm op}\) is likewise the operator
			sine-theta distance
			\(\sinth{\widehat{\mathcal V}}{\mathcal V}\) used here.}
		
		\medskip\noindent\textbf{Inclusion under the unit-norm constraint.}
		We now verify directly that the two equal-allocation constructions used in the
		source proof can be chosen inside our parameter class.  Nonemptiness and
		\(\|\beta_t\|_2\le1\) imply
		\[
		k\lambda_k
		\le \operatorname{tr}(\Sigma_{\bm\beta})
		\le \frac1T\sum_{t=1}^T\|\beta_t\|_2^2
		\le1.
		\]
		
		For the first term, use an equal-norm Parseval frame.  Namely, there exists
		\(Q\in\R^{T\times k}\) such that
		\[
		Q^\top Q=I_k,
		\qquad
		\|Q^\top e_t\|_2^2=\frac{k}{T}
		\quad\text{for every }t\in[T].
		\]
		For completeness, this follows from the finite-dimensional Schur--Horn
		theorem: the constant vector \((k/T,\ldots,k/T)\) is majorized by the spectrum
		\((1,\ldots,1,0,\ldots,0)\), so there is a rank-\(k\) orthogonal projection
		with constant diagonal \(k/T\); factoring that projection gives \(Q\).
		Set
		\[
		\alpha_t:=\sqrt{T\lambda_k}\,Q^\top e_t,
		\qquad
		\beta_t:=V\alpha_t.
		\]
		Then
		\[
		\|\beta_t\|_2^2=\|\alpha_t\|_2^2=k\lambda_k\le1,
		\qquad
		\frac1T\sum_{t=1}^T\alpha_t\alpha_t^\top=\lambda_k I_k.
		\]
		The proof of the first term in Section~9.2 of the source varies the shared
		subspace while using the coefficient vectors only through the identity
		\(\sum_t n\alpha_t\alpha_t^\top=N\lambda_k I_k\).  The same packing and
		divergence calculation therefore applies to the frame above and gives
		\begin{equation}\label{eq:linear-lb-first}
			\inf_{\widehat{\mathcal V}}
			\sup_{(\mathcal V,\beta_1,\ldots,\beta_T)\in
				\mathcal P(\lambda_k)}
			\E\big[\sinth{\widehat{\mathcal V}}{\mathcal V}\big]
			\ge
			c\left[\sqrt{\frac{d}{N\lambda_k}}\wedge1\right].
		\end{equation}
		
		For the second term, first consider the regime
		\(k\lambda_k\le a_0\), where \(a_0>0\) is a sufficiently small constant
		depending only on \(c_0\) and \(\rho_2\).  Let
		\[
		\alpha_t=\sqrt{L\lambda_k}\,g_t,
		\qquad
		g_t\stackrel{\mathrm{i.i.d.}}{\sim}N(0,I_k),
		\]
		where \(L\ge1\) is a sufficiently large fixed constant depending only on
		\(\rho_2\).  The standard extreme-singular-value bounds for the
		\(k\times T\) Gaussian matrix \([g_1,\ldots,g_T]\), together with
		\(T\ge(1+\rho_2)k\), show, after increasing \(L\) if necessary, that with
		probability at least \(7/8\),
		\[
		\lambda_k I_k
		\preceq
		\frac1T\sum_{t=1}^T\alpha_t\alpha_t^\top
		\preceq C\lambda_k I_k.
		\]
		Moreover, Gaussian norm concentration and a union bound give, with probability
		at least \(7/8\),
		\[
		\max_{t\le T}\|\alpha_t\|_2
		\le
		\sqrt{L\lambda_k}\bigl(\sqrt{k}+C\sqrt{\log T}\bigr)
		\le1
		\]
		provided \(a_0\) is chosen sufficiently small, because
		\(k\ge c_0\log T\).  Thus the conditioning event used in the source proof of
		the second term can be replaced by their intersection, which has probability
		at least \(3/4\) and on which the resulting coefficients belong to
		\(\mathcal P(\lambda_k)\).  Running the
		same Gaussian-mixture mutual-information calculation with variance
		\(L\lambda_k\) changes the result only by the fixed factor \(L\), and hence
		\begin{equation}\label{eq:linear-lb-second}
			\inf_{\widehat{\mathcal V}}
			\sup_{(\mathcal V,\beta_1,\ldots,\beta_T)\in
				\mathcal P(\lambda_k)}
			\E\big[\sinth{\widehat{\mathcal V}}{\mathcal V}\big]
			\ge
			c\left[\sqrt{\frac{Td}{N^2\lambda_k^2}}\wedge1\right]
			\qquad (k\lambda_k\le a_0).
		\end{equation}
		Combining \eqref{eq:linear-lb-first} and
		\eqref{eq:linear-lb-second}, and using that the maximum of the two truncated
		terms controls their truncated sum up to a factor of two, proves the claimed
		bound when \(k\lambda_k\le a_0\).
		
		It remains to consider \(k\lambda_k>a_0\).  Since \(n\ge d\ge(1+\rho_1)k\),
		we have \(n\lambda_k\ge(1+\rho_1)a_0\), and therefore
		\[
		\sqrt{\frac{Td}{N^2\lambda_k^2}}
		=
		\frac{1}{\sqrt{n\lambda_k}}
		\sqrt{\frac{d}{N\lambda_k}}
		\lesssim
		\sqrt{\frac{d}{N\lambda_k}}.
		\]
		Thus \eqref{eq:linear-lb-first} alone controls the claimed two-term rate in
		this regime.  Finally, substituting \(N=Tn\) yields
		\[
		c
		\left(
		\left[
		\sqrt{\frac{d}{Tn\lambda_k}}
		+
		\sqrt{\frac{d}{Tn^2\lambda_k^2}}
		\right]\wedge1
		\right),
		\]
		as claimed.
	\end{proof}
	
	\subsection{Proof of Corollary~\ref{cor:overview-linear-minimax}}
	
	\begin{proof}
		We first prove the upper bound, uniformly over fixed designs and sub-Gaussian
		noise laws. Fix \(X_{1:T}\in\mathfrak X(\kappa_{\min},\kappa_{\max})\),
		\(\xi\in\mathfrak E_{\rm sg}\), and
		\((\mathcal V,\beta_1,\ldots,\beta_T)\in\mathcal P(\lambda_k)\).
		Treating the sub-Gaussian noise scale as a fixed model constant, we absorb its
		dependence into the constants below. Write
		\[
		\delta:=\sinth{\widehat{\mathcal V}}{\mathcal V},
		\qquad
		\lambda_{\min}^{\beta}
		:=
		\lambda_k\!\left(\frac1T\sum_{t=1}^T\beta_t\beta_t^\top\right).
		\]
		Thus \(\lambda_{\min}^{\beta}=\lambda_k(\Sigma_{\bm\beta})\).
		By Theorem~\ref{thm:linear}, there are constants \(K_1,K_2>0\), depending only
		on the fixed noise and design-conditioning constants, such that for all \(\tau\ge1\),
		\[
		\Pr\!\left(
		\delta
		>
		K_1\left[
		\sqrt{\frac{d+\tau}{Tn\lambda_{\min}^{\beta}}}
		+
		\sqrt{\frac{d+\tau}{Tn^2(\lambda_{\min}^{\beta})^2}}
		\right]\wedge1
		\,\middle|\, X_1,\ldots,X_T
		\right)
		\le K_2e^{-\tau}.
		\]
		Since \(\lambda_{\min}^{\beta}\ge \lambda_k\) on
		\(\mathcal P(\lambda_k)\), we may use the simpler parameters
		\[
		a:=\frac{1}{\sqrt{Tn\lambda_k}}
		\qquad\text{and}\qquad
		b:=\frac{1}{\sqrt{Tn^2\lambda_k^2}}.
		\]
		Define
		\[
		r_{\rm lin}:=\bigl[(a+b)\sqrt d\bigr]\wedge1
		=\left[\sqrt{\frac{d}{Tn\lambda_k}}
		+\sqrt{\frac{d}{Tn^2\lambda_k^2}}\right]\wedge1.
		\]
		Enlarging \(K_1\) if necessary, the preceding display implies
		\[
		\Pr\!\left(
		\delta>K_1(a+b)\sqrt{d+\tau}\wedge1
		\,\middle|\, X_1,\ldots,X_T
		\right)
		\le K_2e^{-\tau}
		\qquad\text{for all }\tau\ge1.
		\]
		We claim that
		\[
		\E\!\left[\delta\,\middle|\, X_1,\ldots,X_T\right]
		\le
		C\bigl[(a+b)\sqrt d\bigr]\wedge1.
		\]
		If \(2K_1(a+b)\sqrt d\ge1\), this is immediate from \(0\le \delta\le1\). It remains
		to consider the case \(2K_1(a+b)\sqrt d<1\). By the layer-cake formula,
		\[
		\E[\delta\mid X_1,\ldots,X_T]
		=
		\int_0^1 \Pr(\delta>s\mid X_1,\ldots,X_T)\,ds.
		\]
		Split the integral at \(s_0:=2K_1(a+b)\sqrt d\). The first part is bounded by
		\(s_0\). For \(s\ge s_0\), define
		\[
		\tau_s:=\left(\frac{s}{K_1(a+b)}\right)^2-d.
		\]
		Then \(\tau_s\ge0\) and
		\[
		K_1(a+b)\sqrt{d+\tau_s}=s.
		\]
		Moreover, since \(s\ge2K_1(a+b)\sqrt d\),
		\[
		\tau_s
		=
		\left(\frac{s}{K_1(a+b)}\right)^2-d
		\ge
		\frac{s^2}{2K_1^2(a+b)^2}.
		\]
		Therefore the high-probability bound gives
		\begin{align*}
			\int_{s_0}^1 \Pr(\delta>s\mid X_1,\ldots,X_T)\,ds
			&\le
			K_2\int_{s_0}^1
			\exp\left\{-\frac{s^2}{2K_1^2(a+b)^2}\right\}\,ds\\
			&\le
			K_2\int_0^\infty
			\exp\left\{-\frac{s^2}{2K_1^2(a+b)^2}\right\}\,ds\\
			&\le
			C(a+b)
			\le
			C(a+b)\sqrt d,
		\end{align*}
		where the last inequality uses \(d\ge1\). Hence
		\[
		\E[\delta\mid X_1,\ldots,X_T]
		\le
		C(a+b)\sqrt d
		\]
		in the nontrivial regime, proving the claimed conditional expectation bound
		after truncating by \(1\).
		
		Since the preceding bound is uniform over
		\(X_{1:T}\in\mathfrak X(\kappa_{\min},\kappa_{\max})\), \(\xi\in\mathfrak E_{\rm sg}\), and
		\((\mathcal V,\beta_1,\ldots,\beta_T)\in\mathcal P(\lambda_k)\),
		we obtain
		\[
		\sup_{\substack{
				X_{1:T}\in\mathfrak X(\kappa_{\min},\kappa_{\max})\\
				\xi\in\mathfrak E_{\rm sg}\\
				(\mathcal V,\beta_1,\dots,\beta_T)\in
				\mathcal P(\lambda_k)}}
		\E\bigl[\sinth{\widehat{\mathcal V}}{\mathcal V}\bigr]
		\le
		C r_{\rm lin}.
		\]
		
		It remains to justify the lower bound for this larger fixed-design,
		sub-Gaussian-noise class. Fix \(s>0\) such that
		\(\kappa_{\min}<s^2<\kappa_{\max}\), and let
		\(X_{t,i}\stackrel{\mathrm{i.i.d.}}{\sim}N(0,s^2I_d)\). Define the
		good-design event
		\[
		\mathcal E_X:=
		\left\{
		\kappa_{\min}I_d\preceq \frac1nX_t^\top X_t\preceq \kappa_{\max}I_d
		\ \text{for all }t\in[T]
		\right\}.
		\]
		Because \(s^2\) lies strictly between the two fixed conditioning constants,
		standard Gaussian sample-covariance concentration gives
		\[
		\Pr(\mathcal E_X^c)\le CTe^{-c_{\rm cov}n}
		\]
		whenever \(n\ge C_{\rm des}d\), where \(C_{\rm des}\) may depend only on
		\(\kappa_{\min},\kappa_{\max}\). The lower-bound argument of
		Theorem~\ref{thm:linear-lb} applies unchanged to this scaled Gaussian design:
		writing \(X_{t,i}=sZ_{t,i}\) rescales the effective task vectors by \(s\),
		their covariance by \(s^2\), and the relevant Kullback--Leibler divergences
		by fixed factors determined by \(s\). Thus it yields the same rate up to
		constants depending only on
		\(\kappa_{\min},\kappa_{\max}\). Gaussian errors are contained in
		\(\mathfrak E_{\rm sg}\). For a fixed estimator, define the conditional
		worst-case Gaussian-noise risk
		\[
		R_{\widehat{\mathcal V}}(X)
		:=
		\sup_{(\mathcal V,\beta_1,\dots,\beta_T)\in
			\mathcal P(\lambda_k)}
		\E_{\xi}\bigl[\sinth{\widehat{\mathcal V}}{\mathcal V}\mid X\bigr],
		\]
		where \(\xi_{t,i}\stackrel{\mathrm{i.i.d.}}{\sim}N(0,1)\). Theorem~\ref{thm:linear-lb}
		gives, for every estimator,
		\[
		c r_{\rm lin}
		\le
		\sup_{(\mathcal V,\beta_1,\dots,\beta_T)\in
			\mathcal P(\lambda_k)}
		\E_{X,\xi}\bigl[\sinth{\widehat{\mathcal V}}{\mathcal V}\bigr]
		\le
		\E_X R_{\widehat{\mathcal V}}(X).
		\]
		Since the loss is bounded by \(1\), \(0\le R_{\widehat{\mathcal V}}(X)\le1\).
		Thus
		\[
		\E_X\left[
		\mathbf 1_{\mathcal E_X}
		R_{\widehat{\mathcal V}}(X)
		\right]
		\ge c r_{\rm lin}-\Pr(\mathcal E_X^c).
		\]
		It remains only to absorb the bad-design probability. Since
		\(\|\beta_t\|_2\le1\), nonemptiness of
		\(\mathcal P(\lambda_k)\) implies
		\[
		k\lambda_k
		\le
		\operatorname{tr}\!\left(\frac1T\sum_{t=1}^T\beta_t\beta_t^\top\right)
		\le 1,
		\]
		and hence \(\lambda_k\lesssim 1/k\). Therefore
		\[
		r_{\rm lin}
		\ge
		c\left[\sqrt{\frac{dk}{Tn}}\right]\wedge1
		\ge
		\frac{c}{\sqrt{Tn}}.
		\]
		Choosing \(c_1>0\) sufficiently small relative to \(c_{\rm cov}\), the condition
		\(T\le e^{c_1 n}\) gives \(CTe^{-c_{\rm cov}n}\lesssim r_{\rm lin}\) after
		adjusting constants: its left-hand side is at most
		\(Ce^{-(c_{\rm cov}-c_1)n}\), while
		\(r_{\rm lin}\gtrsim e^{-c_1n/2}/\sqrt n\). Hence, for every estimator, there exists a
		deterministic design realization in \(\mathfrak X(\kappa_{\min},\kappa_{\max})\)
		for which the Gaussian-noise risk is at least \(c r_{\rm lin}\). Taking the
		supremum over \(\mathfrak X(\kappa_{\min},\kappa_{\max})\) and
		\(\mathfrak E_{\rm sg}\), and then the infimum over estimators, gives the
		desired lower bound. Combining the upper and lower bounds proves the corollary.
	\end{proof}
	
	\subsection{Proof of Theorem~\ref{thm:gsi}}
	\label{sec:proof-of-classification-upper-bound}
	\begin{proof}[Proof of Theorem~\ref{thm:gsi}]
		We first specify the threshold \(n_{\mathrm{cls}}\) appearing in the theorem. Let
		\(Z\sim N(0,1)\) and define
		\begin{equation}\label{eq:bg-ng-definition}
			b_g:=\sup_{0\le r\le1}\Bigl(-\E[g''(rZ)]\Bigr)_+,
			\qquad
			n_{\mathrm{cls}}:=\max\left\{2,\,1+\frac{2b_g}{a_g^2}\right\}.
		\end{equation}
		Since \(b_g\le\|g''\|_\infty<\infty\) and \(a_g>0\), this threshold
		is finite and depends only on \(g\).

		For each task \(t\), define
		\[
		\tilde\beta_t:=\E[\hat\beta_t]
		=
		\E[Y_{t,1}X_{t,1}].
		\]
		Let \(r_t:=\|\beta_t\|_2\). By the model assumption \(\E[Y_t\mid X_t]=g(X_t^\top\beta_t)\) and Stein's identity,
		\[
		\tilde\beta_t
		=
		\E[g(X_{t,1}^\top\beta_t)X_{t,1}]
		=
		\E[g'(r_t Z)]\beta_t
		=:a_t\beta_t.
		\]
		{By the definition of \(a_g\), \(|a_t|\ge a_g\) for all \(t\in[T]\).
			Therefore
			\[
			\Sigma_{\bm{\tilde\beta}}
			=
			\frac1T\sum_{t=1}^T a_t^2\beta_t\beta_t^\top
			\succeq
			a_g^2\Sigma_{\bm\beta},
			\]
			and its range is \(\mathcal V\). Hence
			\(\widetilde{\mathcal V}=\mathcal V\), so the spectral estimator already
			targets the desired subspace without any transformation. In addition,
			\[
			\widetilde\lambda_k\ge a_g^2\lambda_k.
			\]}
		Also, by the definition of \(b_g\),
		\[
		\E[g''(r_tZ)]\ge -b_g
		\qquad\text{for all } t\in[T].
		\]
		
		Next, write \(m_t:=\tilde\beta_t\). Since
		\[
		\hat\beta_t-m_t
		=
		\frac1n\sum_{i=1}^n \bigl(Y_{t,i}X_{t,i}-m_t\bigr),
		\]
		the task residuals are independent and mean zero across \(t\). Since \(Y_{t,i}\in\{0,1\}\), each coordinate \(\langle Y_{t,i}X_{t,i},v\rangle\) is a bounded multiple of a Gaussian. Hence
		\[
		\|\langle Y_{t,i}X_{t,i}-m_t,v\rangle\|_{\psi_2}\le C\|v\|_2
		\qquad\text{for all }v\in\R^d,
		\]
		and standard averaging gives Condition~\ref{cond:surrogate-recovery}(1) with \(\sigma^2\le C/n\).
		
		We now verify the block structure of the residual covariance. If \(r_t>0\), write \(u_t:=\beta_t/r_t\), set \(S_t:=X_{t,1}^\top u_t\sim N(0,1)\), and decompose
		\[
		X_{t,1}=S_tu_t+W_t,
		\]
		where \(W_t\sim N(0,I_d-u_tu_t^\top)\) is independent of \(S_t\). Since \(Y_{t,1}\mid X_{t,1}\stackrel{d}{=}Y_{t,1}\mid X_{t,1}^\top\beta_t\), the conditional law of \(Y_{t,1}\) depends only on \(S_t\). Therefore \(Y_{t,1}\) is conditionally independent of \(W_t\) given \(S_t\), and
		\[
		\E[Y_{t,1}^2X_{t,1}X_{t,1}^\top]
		=
		\E[Y_{t,1}^2]I_d
		+
		\Bigl(\E[Y_{t,1}^2S_t^2]-\E[Y_{t,1}^2]\Bigr)u_tu_t^\top.
		\]
		Subtracting \(m_tm_t^\top\), we obtain
		\[
		\mathrm{Cov}(Y_{t,1}X_{t,1})
		=
		\E[Y_{t,1}^2]I_d
		+
		\Bigl(\E[Y_{t,1}^2S_t^2]-\E[Y_{t,1}^2]-\|m_t\|_2^2\Bigr)u_tu_t^\top.
		\]
		Since \(Y_{t,1}\in\{0,1\}\), we have \(Y_{t,1}^2=Y_{t,1}\), and Gaussian integration by parts gives
		\[
		\E[Y_{t,1}^2S_t^2]-\E[Y_{t,1}^2]
		=
		\E[g(r_tS_t)(S_t^2-1)]
		=
		r_t^2\E[g''(r_tS_t)].
		\]
		Also,
		\[
		\|m_t\|_2^2=a_t^2r_t^2.
		\]
		Therefore
		\[
		\mathrm{Cov}(Y_{t,1}X_{t,1})
		=
		\E[Y_{t,1}]I_d+r_t^2\bigl(\E[g''(r_tS_t)]-a_t^2\bigr)u_tu_t^\top.
		\]
		If \(r_t=0\), then \(\beta_t=0\), the conditional mean \(g(X_{t,1}^\top\beta_t)=g(0)\) is constant, and the single-index assumption implies \(Y_{t,1}\) is independent of \(X_{t,1}\). Consequently \(m_t=0\) and
		\[
		\mathrm{Cov}(Y_{t,1}X_{t,1})=\E[Y_{t,1}^2]I_d=\E[Y_{t,1}]I_d,
		\]
		which is consistent with the displayed formula above because the rank-one correction vanishes when \(r_t=0\).
		Thus each task covariance is isotropic on \(\mathcal V^\perp\) and differs from a scalar multiple of the identity only in the one-dimensional direction \(\mathrm{span}(\beta_t)\subset\mathcal V\). Averaging over tasks, we may write
		\[
		\Sigma_{\bm\varepsilon}
		=
		\frac1T\sum_{t=1}^T \E[(\hat\beta_t-\tilde\beta_t)(\hat\beta_t-\tilde\beta_t)^\top]
		=
		\omega^2 I_d + V R_\star V^\top
		\]
		for some \(\omega^2\ge 0\) and symmetric \(R_\star\).
		{Consequently, \(\delta_1=0\).} Moreover, because
		\[
		\nu=\lambda_{\max}(V_\perp^\top \Sigma_{\bm\varepsilon}V_\perp)=\omega^2,
		\]
		the signal-block contribution for task \(t\) equals
		\[
		\tilde\beta_t\tilde\beta_t^\top
		+
		\frac1n r_t^2\bigl(\E[g''(r_tS_t)]-a_t^2\bigr)u_tu_t^\top
		=
		r_t^2\left[\left(1-\frac1n\right)a_t^2+\frac1n\E[g''(r_tS_t)]\right]u_tu_t^\top.
		\]
		Using \(a_t^2\ge a_g^2\), \(\E[g''(r_tS_t)]\ge -b_g\), and \(n\ge n_{\mathrm{cls}}\), we obtain
		\[
		\left(1-\frac1n\right)a_t^2+\frac1n\E[g''(r_tS_t)]
		\ge
		\left(1-\frac1n\right)a_t^2-\frac{b_g}{n}
		\ge
		\left(1-\frac1n\right)a_t^2-\frac{n-1}{2n}a_g^2
		\ge
		\frac{1}{2}\left(1-\frac1n\right)a_t^2.
		\]
		Therefore the signal-block contribution is bounded below by
		\[
		\frac12\left(1-\frac1n\right)a_t^2r_t^2u_tu_t^\top
		=
		\frac12\left(1-\frac1n\right)\tilde\beta_t\tilde\beta_t^\top.
		\]
		Hence it is in particular bounded below by
		\[
		\gamma_\star\,\tilde\beta_t\tilde\beta_t^\top
		\]
		with
		\[
		\gamma_\star:=\frac12\left(1-\frac1n\right)>0.
		\]
		After averaging over \(t\), we obtain
		\[
		V^\top\left[\frac1T\sum_{t=1}^T \tilde\beta_t\tilde\beta_t^\top+\Sigma_{\bm\varepsilon}\right]V
		\succeq
		\omega^2 I_k+\gamma_\star V^\top\left[\frac1T\sum_{t=1}^T \tilde\beta_t\tilde\beta_t^\top\right]V.
		\]
		Therefore
		\[
		\rho\ge \omega^2+\gamma_\star\widetilde\lambda_k,
		\qquad
		\nu=\omega^2,
		\]
		so
		\[
		\rho-\nu\ge \gamma_\star\widetilde\lambda_k.
		\]
		Since \(n\ge n_{\mathrm{cls}}\ge2\), we have \(\gamma_\star\ge1/4\). Thus
		Condition~\ref{cond:surrogate-recovery}(2) holds with the fixed constant \(\gamma_0=1/4\).
		
		{Because \(\widetilde{\mathcal V}=\mathcal V\), no transformation is
			needed. Applying Theorem~\ref{thm:mtl-main} directly and using
			\(\delta_1=0\) gives, for every \(\tau\ge1\),
			\[
			\sinth{\widehat{\mathcal V}}{\mathcal V}
			\le
			C\left(
			\left[
			\sigma\sqrt{\frac{(d+\tau)\rho}{T\widetilde\lambda_k^2}}
			+
			\sigma^2\sqrt{\frac{d+\tau}{T\widetilde\lambda_k^2}}
			+
			\frac{\sigma^2(d+\tau)}{T\widetilde\lambda_k}
			\right]\wedge 1
			\right).
			\]
			Since \(\widetilde\lambda_k\ge a_g^2\lambda_k\), this implies
			\[
			\sinth{\widehat{\mathcal V}}{\mathcal V}
			\le
			C\left(
			\left[
			a_g^{-2}\left\{
			\sigma\sqrt{\frac{(d+\tau)\rho}{T\lambda_k^2}}
			+
			\sigma^2\sqrt{\frac{d+\tau}{T\lambda_k^2}}
			+
			\frac{\sigma^2(d+\tau)}{T\lambda_k}
			\right\}
			\right]\wedge 1
			\right).
			\]}
		Since \(\sigma^2\le C/n\) and every directional variance of \(\hat\beta_t-\tilde\beta_t\) is at most \(C/n\), we also have
		\[
		\lambda_{\max}(\Sigma_{\bm\varepsilon})\le \frac{C}{n}.
		\]
		Let
		\[
		A:=
		V^\top\left(\frac1T\sum_{t=1}^T \tilde\beta_t\tilde\beta_t^\top\right)V,
		\qquad
		B:=V^\top \Sigma_{\bm\varepsilon} V.
		\]
		Then \(\rho=\lambda_{\min}(A+B)\), so
		\[
		\rho
		\le
		\lambda_{\min}(A)+\lambda_{\max}(B)
		=
		\widetilde\lambda_k+\lambda_{\max}(B)
		\le
		\widetilde\lambda_k+\lambda_{\max}(\Sigma_{\bm\varepsilon}).
		\]
		Because \(\widetilde\lambda_k\le \lambda_k\max_t a_t^2\), this gives
		\[
		\rho\le C\left(\lambda_k+\frac{1}{n}\right),
		\]
		where \(C\) depends only on the parameter class \(\Beta\) and the link \(g\). Therefore
		\[
		a_g^{-2}\sigma\sqrt{\frac{(d+\tau)\rho}{T\lambda_k^2}}
		\le
		C\,a_g^{-2}\left[
		\sqrt{\frac{d+\tau}{Tn\lambda_k}}
		+
		\sqrt{\frac{d+\tau}{Tn^2\lambda_k^2}}
		\right].
		\]
		The second concentration term is bounded by the second term in the last
		display. For the remaining term, set
		\[
		r_1:=\sqrt{\frac{d+\tau}{Tn\lambda_k}},
		\qquad
		r_2:=\sqrt{\frac{d+\tau}{Tn^2\lambda_k^2}}.
		\]
		Then \(\sigma^2(d+\tau)/(T\lambda_k)\le Cr_1^2\). If
		\(r_1+r_2<1\), then \(r_1^2\le r_1\); if \(r_1+r_2\ge1\), the final
		truncation by one makes the desired bound trivial. Finally, choosing
		\(\|\beta\|_2=1\) in the definition of \(a_g\), Stein's identity and
		Cauchy--Schwarz give
		\[
		a_g\le \bigl|\E[g'(Z)]\bigr|
		=\bigl|\E[Zg(Z)]\bigr|\le1.
		\]
		Thus \(a_g^{-2}\ge1\), so the factor may be placed outside the truncation.
		Combining these observations yields the displayed bound.
	\end{proof}
	
	\subsection{Proof of Theorem~\ref{thm:gsi-subg}}
	\label{sec:proof-of-additive-upper-bound}
	
	For this proof, write \(a_0:=a_g\). The construction below produces the
	finite threshold denoted by \(n_{\mathrm{add}}\) in the theorem.
	Fix \(\tau\ge1\).
	
	For the proof, define the truncated response
	\[
	\phi_L(y):=y\,\mathbf 1\{|y|\le L\},
	\]
	and the truncated first-moment estimator
	\[
	\hat\beta_t^{(L)}
	:=
	\frac1n\sum_{i=1}^n \phi_L(Y_{t,i})X_{t,i}.
	\]
	Let
	\[
	\widehat A^{(L)}\in\R^{T\times d}
	\quad\text{be the stacked matrix whose \(t\)-th row is }
	(\hat\beta_t^{(L)})^\top,
	\]
	\[
	\widehat V_k^{(L)}
	\quad\text{be an orthonormal basis matrix for the top-\(k\) right singular
		subspace of }\widehat A^{(L)},
	\]
	\[
	{\widehat{\mathcal V}^{(L)}
		:=\operatorname{span}(\widehat V_k^{(L)}),}
	\]
	\[
	\tilde\beta_t^{(L)}:=\E[\hat\beta_t^{(L)}],
	\qquad
	\widetilde{\mathcal V}^{(L)}
	:=
	\mathrm{span}(\tilde\beta_1^{(L)},\dots,\tilde\beta_T^{(L)}),
	\]
	and let \(\widetilde V_L\) be an orthonormal basis of
	\(\widetilde{\mathcal V}^{(L)}\). Write
	\[
	\Sigma_{\bm\varepsilon}^{(L)}
	:=
	\frac1T\sum_{t=1}^T
	\E\!\left[
	(\hat\beta_t^{(L)}-\tilde\beta_t^{(L)})
	(\hat\beta_t^{(L)}-\tilde\beta_t^{(L)})^\top
	\right],
	\]
	\[
	\widetilde\lambda_k^{(L)}
	:=
	\lambda_k\!\left(
	\frac1T\sum_{t=1}^T
	\tilde\beta_t^{(L)}(\tilde\beta_t^{(L)})^\top
	\right),
	\]
	and
	\[
	\rho_L
	:=
	\lambda_{\min}\!\left(
	\widetilde V_L^\top
	\left[
	\frac1T\sum_{t=1}^T \tilde\beta_t^{(L)}(\tilde\beta_t^{(L)})^\top
	+
	\Sigma_{\bm\varepsilon}^{(L)}
	\right]
	\widetilde V_L
	\right).
	\]
	For \(r\in[0,1]\), let \(S\sim N(0,1)\), let \(\xi\) be an independent
	copy of the additive noise, and define
	\[
	Y_r:=g(rS)+\xi.
	\]
	Set
	\[
	A(r):=\E[Y_rS],
	\qquad
	Q(r):=\E[Y_r^2(S^2-1)],
	\qquad
	R(r):=Q(r)-A(r)^2.
	\]
	Also define
	\[
	\alpha(r):=\E[g'(rS)],
	\qquad
	\chi(r):=\E[(g^2)''(rS)].
	\]
	We assume throughout this appendix subsection that \(r\mapsto \alpha(r)\) and
	\(r\mapsto \chi(r)\) are well defined on \([0,1]\); this is a mild
	regularity condition on \(g\) ensuring that the Stein identities below are
	justified. Under the derivative bounds in~\eqref{eq:bounded-derivatives},
	the quantity \(\chi(r)\) is automatically finite on \([0,1]\), because
	\[
	|(g^2)''(x)|=|2(g'(x))^2+2g(x)g''(x)|
	\le
	2L_1^2+2L_2\bigl(|g(0)|+L_1 |x|\bigr),
	\]
	so \((g^2)''(rS)\) has finite expectation for every \(r\in[0,1]\). Because
	\(\Beta=\mathcal V\cap B_2(1)\), every task parameter satisfies
	\(\|\beta_t\|_2\in[0,1]\), so uniform control over \(r\in[0,1]\) is exactly
	what the proof needs. Also, since \(|g(rS)|\le |g(0)|+\|g'\|_\infty |S|\) for
	\(r\in[0,1]\), the signal term \(g(rS)\) is uniformly sub-Gaussian on
	\([0,1]\). Together with the uniform bound \(\|\xi\|_{\psi_2}\le K_\xi\), this
	implies
	\[
	\sup_{r\in[0,1]}\|Y_r\|_{\psi_2}\le K_Y
	\]
	after enlarging the constant in the definition of \(K_Y\) if necessary.
	Moreover, \(|\alpha(r)|\ge a_0\) for all \(r\in[0,1]\)
	by the definition of \(a_0\). By independence of \(\xi\) and \(S\), together
	with \(\E[\xi]=0\),
	\[
	A(r)=\E[g(rS)S]+\E[\xi]\E[S]=\E[g(rS)S].
	\]
	Stein's identity gives
	\[
	A(r)=r\,\alpha(r).
	\]
	Next,
	\begin{align*}
		Q(r)
		&=
		\E[g(rS)^2(S^2-1)]
		+
		2\E[g(rS)\xi(S^2-1)]
		+
		\E[\xi^2(S^2-1)].
	\end{align*}
	The cross term vanishes because \(\xi\perp S\) and \(\E[\xi]=0\), while the
	pure-noise term vanishes because \(\xi\perp S\) and \(\E[S^2-1]=0\). Hence
	\[
	Q(r)=\E[g(rS)^2(S^2-1)]=r^2\chi(r),
	\]
	again by Stein's identity, and therefore
	\[
	R(r)=r^2\bigl(\chi(r)-\alpha(r)^2\bigr).
	\]
	Define
	\begin{equation}\label{eq:n-add-definition}
		\kappa(r):=
		\begin{cases}
			\dfrac{(\alpha(r)^2-\chi(r))_+}{\alpha(r)^2}, & \alpha(r)\neq 0,\\[0.9em]
			0, & \alpha(r)=0 \text{ and } \chi(r)\ge 0,\\[0.4em]
			\infty, & \alpha(r)=0 \text{ and } \chi(r)<0,
		\end{cases}
		\qquad
		n_0:=1+\sup_{r\in[0,1]}\kappa(r),
		\qquad
		n_{\mathrm{add}}:=\lceil n_0\rceil.
	\end{equation}
	Since \(|\alpha(r)|\ge a_0\) on \([0,1]\), only the first branch can
	occur. Furthermore,
	\[
	\kappa(r)
	\le
	1+\frac{|\chi(r)|}{\alpha(r)^2}
	\le
	1+\frac{\sup_{s\in[0,1]}|\chi(s)|}{a_0^2},
	\]
	so \(n_0<\infty\). Then
	\[
	R(r)\ge -(n_0-1)A(r)^2
	\qquad\text{for all }r\in[0,1].
	\]
	Set
	\[
	L:=CK_Y\sqrt{\Gamma_{n,T,\lambda_k}+\tau},
	\]
	where \(C>0\) is chosen sufficiently large. For \(r\in[0,1]\), define
	\[
	A_L(r):=\E[\phi_L(Y_r)S],
	\qquad
	Q_L(r):=\E[\phi_L(Y_r)^2(S^2-1)],
	\qquad
	R_L(r):=Q_L(r)-A_L(r)^2.
	\]
	\begin{lemma}[Uniform truncation bounds]
		\label{lem:gsi-subg-convergence}
		\[
		\sup_{r\in[0,1]}
		|A_L(r)-A(r)|
		+
		\sup_{r\in[0,1]}
		|Q_L(r)-Q(r)|
		\le
		Ce^{-cL^2/K_Y^2}.
		\]
		Consequently,
		\[
		\sup_{r\in[0,1]}
		|R_L(r)-R(r)|
		\le
		Ce^{-cL^2/K_Y^2}.
		\]
	\end{lemma}
	
	\begin{proof}
		Fix \(r\in[0,1]\). Since
		\[
		A_L(r)-A(r)=\E\!\left[(\phi_L(Y_r)-Y_r)S\right],
		\]
		we have
		\[
		|A_L(r)-A(r)|
		\le
		\E\!\left[|Y_rS|\,\mathbf 1\{|Y_r|>L\}\right].
		\]
		By Cauchy--Schwarz,
		\[
		\E\!\left[|Y_rS|\,\mathbf 1\{|Y_r|>L\}\right]
		\le
		\bigl(\E[Y_r^2S^2]\bigr)^{1/2}\Prob(|Y_r|>L)^{1/2}.
		\]
		The Gaussian moment \(\E[S^4]\) is finite, and the uniform sub-Gaussian bound
		\(\|Y_r\|_{\psi_2}\le K_Y\) implies a uniform bound on \(\E[Y_r^4]\) together
		with
		\[
		\sup_{r\in[0,1]}\Prob(|Y_r|>L)\le Ce^{-cL^2/K_Y^2}.
		\]
		Therefore
		\[
		\sup_{r\in[0,1]}|A_L(r)-A(r)|
		\le
		Ce^{-cL^2/K_Y^2}.
		\]
		Next,
		\[
		Q_L(r)-Q(r)=\E\!\left[(\phi_L(Y_r)^2-Y_r^2)(S^2-1)\right],
		\]
		so
		\[
		|Q_L(r)-Q(r)|
		\le
		\E\!\left[Y_r^2|S^2-1|\,\mathbf 1\{|Y_r|>L\}\right].
		\]
		By Cauchy--Schwarz,
		\[
		\E\!\left[Y_r^2|S^2-1|\,\mathbf 1\{|Y_r|>L\}\right]
		\le
		\bigl(\E[Y_r^4(S^2-1)^2]\bigr)^{1/2}\Prob(|Y_r|>L)^{1/2}.
		\]
		Since \(\sup_r\E[Y_r^8]\le C K_Y^8\) and \(\E[(S^2-1)^4]<\infty\), the same tail
		estimate gives
		\[
		\sup_{r\in[0,1]}|Q_L(r)-Q(r)|
		\le
		Ce^{-cL^2/K_Y^2},
		\]
		and hence
		\[
		\sup_{r\in[0,1]}|R_L(r)-R(r)|
		\le
		\sup_{r\in[0,1]}|Q_L(r)-Q(r)|
		+
		\sup_{r\in[0,1]}|A_L(r)^2-A(r)^2|.
		\]
		Finally, \(\sup_r|A(r)|\le C\) and \(\sup_r|A_L(r)|\le C\), so
		\[
		\sup_{r\in[0,1]}|A_L(r)^2-A(r)^2|
		\le
		C\sup_{r\in[0,1]}|A_L(r)-A(r)|.
		\]
		This proves the claim.
	\end{proof}
	
	Let
	\[
	b_L:=\sup_{r\in[0,1]}|A_L(r)-A(r)|,
	\qquad
	c_L:=\sup_{r\in[0,1]}|R_L(r)-R(r)|.
	\]
	By Lemma~\ref{lem:gsi-subg-convergence}, both \(b_L\) and \(c_L\) tend to \(0\)
	as \(L\to\infty\). Fix
	\[
	\eta_L:=\max\{b_L^{1/2},\,L^{-1}\},
	\qquad
	\varepsilon_L:=(n_0-1)\eta_L,
	\qquad
	\delta_L:=c_L+(n_0-1)(1+\eta_L^{-1})b_L^2.
	\]
	Then \(\varepsilon_L\to 0\) and \(\delta_L\to 0\). Moreover, for every
	\(r\in[0,1]\),
	\[
	A(r)^2
	\le
	(1+\eta_L)A_L(r)^2+(1+\eta_L^{-1})b_L^2,
	\]
	by the inequality \((u-v)^2\le (1+\eta)u^2+(1+\eta^{-1})v^2\). Since
	\(R(r)\ge -(n_0-1)A(r)^2\), it follows that
	\[
	R_L(r)\ge R(r)-c_L\ge -(n_0-1+\varepsilon_L)A_L(r)^2-\delta_L.
	\]
	We also record the simpler upper bound
	\[
	A_L(r)^2\le 2A(r)^2+2b_L^2.
	\]
	After increasing the constant \(C\) in the definition of \(L\) if necessary, we
	may assume
	\[
	\varepsilon_L\le \frac14,
	\qquad
	\delta_L\le \frac18 a_0^2\lambda_k,
	\qquad
	b_L\le \frac14 a_0\sqrt{\lambda_k}.
	\]
	Under this choice of \(L\), the truncation bias is negligible on the scale of
	the population signal.
	
	\begin{lemma}[Gap transfer to the truncated surrogate]
		\label{lem:gsi-subg-gap-transfer}
		For every task \(t\),
		\[
		\tilde\beta_t^{(L)}(\tilde\beta_t^{(L)})^\top
		+
		\frac1n\|\beta_t\|_2^2 q_{t,L}u_tu_t^\top
		\succeq
		\left(1-\frac{n_0-1+\varepsilon_L}{n}\right)
		\tilde\beta_t^{(L)}(\tilde\beta_t^{(L)})^\top
		-
		\frac{\delta_L}{n}u_tu_t^\top.
		\]
	\end{lemma}
	
	\begin{proof}
		By definition,
		\[
		\|\beta_t\|_2^2 q_{t,L}=Q_{t,L}-A_{t,L}^2,
		\qquad
		\tilde\beta_t^{(L)}=A_{t,L}u_t.
		\]
		Hence the left-hand side equals
		\[
		\left(A_{t,L}^2+\frac1n(Q_{t,L}-A_{t,L}^2)\right)u_tu_t^\top.
		\]
		The lower bound on \(R_L(r)=Q_L(r)-A_L(r)^2\) applied at
		\(r=\|\beta_t\|_2\) gives
		\[
		Q_{t,L}-A_{t,L}^2
		\ge
		-(n_0-1+\varepsilon_L)A_{t,L}^2-\delta_L.
		\]
		Substituting this inequality proves the claim.
	\end{proof}
	
	\begin{proof}
		Write \(r_t:=\|\beta_t\|_2\), and when \(r_t>0\), set
		\[
		u_t:=\beta_t/r_t,
		\qquad
		S_{t,i}:=X_{t,i}^\top u_t,
		\qquad
		X_{t,i}=S_{t,i}u_t+W_{t,i},
		\]
		where \(S_{t,i}\sim N(0,1)\) and \(W_{t,i}\perp S_{t,i}\). Under the additive
		model,
		\[
		Y_{t,i}=g(r_tS_{t,i})+\xi_{t,i},
		\qquad
		\xi_{t,i}\perp X_{t,i},
		\]
		so \(Y_{t,i}\) depends on \(X_{t,i}\) only through \(S_{t,i}\) and an
		independent noise term. Since \(W_{t,i}\perp S_{t,i}\) and
		\(W_{t,i}\perp \xi_{t,i}\), we have \(Y_{t,i}\perp W_{t,i}\).
		
		For each task \(t\), define
		\[
		A_{t,L}:=\E[\phi_L(Y_{t,1})S_{t,1}],
		\qquad
		Q_{t,L}:=\E[\phi_L(Y_{t,1})^2(S_{t,1}^2-1)].
		\]
		When \(r_t>0\), these satisfy
		\[
		a_{t,L}=\frac{A_{t,L}}{r_t},
		\qquad
		q_{t,L}=\frac{Q_{t,L}-A_{t,L}^2}{r_t^2},
		\]
		so that
		\[
		\tilde\beta_t^{(L)}=A_{t,L}u_t.
		\]
		If \(r_t=0\), then \(X_{t,i}^\top\beta_t=0\) almost surely, hence
		\(Y_{t,i}\perp X_{t,i}\), \(A_{t,L}=0\), and \(\tilde\beta_t^{(L)}=0\).
		Therefore, for each task \(t\), there exists \(a_{t,L}\) such that
		\[
		\tilde\beta_t^{(L)}=a_{t,L}\beta_t,
		\qquad
		a_{t,L}=\frac{1}{\|\beta_t\|_2}\E[\phi_L(Y_{t,1})S_{t,1}]
		\quad (\beta_t\neq 0),
		\]
		and
		\[
		\mathrm{Cov}(\phi_L(Y_{t,1})X_{t,1})
		=
		\sigma_{t,L}^2I_d+\|\beta_t\|_2^2q_{t,L}u_tu_t^\top,
		\qquad
		\sigma_{t,L}^2:=\E[\phi_L(Y_{t,1})^2].
		\]
		Averaging over \(t\) and dividing by \(n\), we obtain
		\[
		\Sigma_{\bm\varepsilon}^{(L)}=\omega_L^2I_d+VM_LV^\top
		\]
		for some \(\omega_L^2\ge 0\) and symmetric \(M_L\).
		{Thus \(\delta_1=0\) for the truncated surrogate.}
		
		Since \(|\phi_L(Y_{t,i})|\le L\) almost surely and
		\(\langle X_{t,i},u\rangle\sim N(0,\|u\|_2^2)\), we have
		\[
		\|\langle \hat\beta_t^{(L)}-\tilde\beta_t^{(L)},u\rangle\|_{\psi_2}
		\le
		C\frac{L}{\sqrt n}\|u\|_2
		\qquad\text{for all }u\in\R^d.
		\]
		Thus Condition~\ref{cond:surrogate-recovery}(1) holds with \(\sigma_L\asymp L/\sqrt n\).
		
		Applying Lemma~\ref{lem:gsi-subg-gap-transfer} and averaging over \(t\) yields
		\[
		\rho_L-\nu_L
		\ge
		\left(1-\frac{n_0-1+\varepsilon_L}{n}\right)\widetilde\lambda_k^{(L)}
		-\frac{\delta_L}{n}.
		\]
		By the lower bound on \(\widetilde\lambda_k^{(L)}\) proved below, namely
		\(\widetilde\lambda_k^{(L)}\ge a_0^2\lambda_k/2\), and the choice
		\(\delta_L\le a_0^2\lambda_k/8\), we have
		\[
		\delta_L\le \frac14 \widetilde\lambda_k^{(L)}.
		\]
		Since \(n\ge n_0\) and \(\varepsilon_L\le 1/4\),
		\[
		1-\frac{n_0-1+\varepsilon_L}{n}\ge \frac{3}{4n_0}.
		\]
		Therefore
		\[
		\rho_L-\nu_L
		\ge
		\left(\frac{3}{4n_0}-\frac{1}{4n}\right)\widetilde\lambda_k^{(L)}
		\ge
		\frac{1}{2n_0}\widetilde\lambda_k^{(L)}.
		\]
		Thus the population gap condition in Condition~\ref{cond:surrogate-recovery}(2) holds for the
		truncated surrogate with \(\gamma_0=(2n_0)^{-1}\).
		
		To compare \(\widetilde\lambda_k^{(L)}\) and \(\lambda_k\), Stein's identity gives
		\[
		\tilde\beta_t=\E[g'(\|\beta_t\|_2 Z)]\,\beta_t=:a_t\beta_t,
		\]
		so
		\[
		\frac1T\sum_{t=1}^T \tilde\beta_t\tilde\beta_t^\top
		=
		\frac1T\sum_{t=1}^T a_t^2\beta_t\beta_t^\top
		\succeq
		a_0^2\left(\frac1T\sum_{t=1}^T \beta_t\beta_t^\top\right),
		\]
		hence
		\[
		\widetilde\lambda_k\ge a_0^2\lambda_k.
		\]
		Moreover,
		\[
		\|\tilde\beta_t\|_2=|a_t|\,\|\beta_t\|_2\le \|g'\|_\infty
		\qquad\text{for all }t\in[T].
		\]
		Also,
		\[
		\|\tilde\beta_t^{(L)}-\tilde\beta_t\|_2
		\le
		CK_Ye^{-cL^2/K_Y^2}
		\]
		by duality and the sub-Gaussian tail of \(Y_{t,1}\). Therefore, if
		\[
		b_L:=CK_Ye^{-cL^2/K_Y^2},
		\]
		then
		\[
		\left\|
		\frac1T\sum_{t=1}^T \tilde\beta_t^{(L)}(\tilde\beta_t^{(L)})^\top
		-
		\frac1T\sum_{t=1}^T \tilde\beta_t\tilde\beta_t^\top
		\right\|_{\mathrm{op}}
		\le
		2\|g'\|_\infty b_L+b_L^2.
		\]
		Because \(L=CK_Y\sqrt{\Gamma_{n,T,\lambda_k}+\tau}\), the constant \(C\) may be chosen large
		enough that
		\[
		2\|g'\|_\infty b_L+b_L^2\le \frac12\widetilde\lambda_k.
		\]
		Thus
		\[
		\widetilde\lambda_k^{(L)}\ge \frac12\widetilde\lambda_k\ge \frac12 a_0^2\lambda_k.
		\]
		Since each \(\tilde\beta_t^{(L)}\) is a scalar multiple of \(\beta_t\), we have
		\(\widetilde{\mathcal V}^{(L)}\subseteq \mathcal V\). The bound
		\(\widetilde\lambda_k^{(L)}>0\) implies that \(\widetilde{\mathcal V}^{(L)}\) has
		dimension \(k\). Since \(\mathcal V\) also has dimension \(k\), it follows
		that
		\[
		\widetilde{\mathcal V}^{(L)}=\mathcal V.
		\]
		
		Applying Theorem~\ref{thm:mtl-main} to the truncated surrogate with
		\(\widetilde V_L=V\) and \(\sigma_L\asymp L/\sqrt n\), we obtain, with
		probability at least \(1-e^{-\tau}\),
		\[
		{\sinth{\widehat{\mathcal V}^{(L)}}{\mathcal V}}
		\le
		C\left(
		\left[
		\frac{L}{\sqrt n}\sqrt{\frac{(d+\tau)\rho_L}{T(\widetilde\lambda_k^{(L)})^2}}
		+
		\frac{L^2}{n}\sqrt{\frac{d+\tau}{T(\widetilde\lambda_k^{(L)})^2}}
		+
		\frac{L^2(d+\tau)}{nT\widetilde\lambda_k^{(L)}}
		\right]\wedge 1
		\right).
		\]
		Let
		\[
		A_L
		:=
		V^\top\left(\frac1T\sum_{t=1}^T
		\tilde\beta_t^{(L)}(\tilde\beta_t^{(L)})^\top\right)V,
		\qquad
		B_L:=V^\top\Sigma_{\bm\varepsilon}^{(L)}V.
		\]
		Then \(\rho_L=\lambda_{\min}(A_L+B_L)\), so
		\[
		\rho_L
		\le
		\lambda_{\min}(A_L)+\lambda_{\max}(B_L)
		=
		\widetilde\lambda_k^{(L)}+\lambda_{\max}(B_L)
		\le
		\widetilde\lambda_k^{(L)}+\lambda_{\max}(\Sigma_{\bm\varepsilon}^{(L)}),
		\qquad
		\lambda_{\max}(\Sigma_{\bm\varepsilon}^{(L)})\le C\frac{L^2}{n},
		\]
		so
		\[
		\rho_L\le C\left(\widetilde\lambda_k^{(L)}+\frac{L^2}{n}\right).
		\]
		Substituting \(\widetilde\lambda_k^{(L)}\ge a_0^2\lambda_k/2\) yields
		\[
		{\sinth{\widehat{\mathcal V}^{(L)}}{\mathcal V}}
		\le
		C\left(
		\left[
		\frac{L}{a_0\sqrt n}\sqrt{\frac{d+\tau}{T\lambda_k}}
		+
		\frac{L^2}{a_0^2n}\sqrt{\frac{d+\tau}{T\lambda_k^2}}
		+
		\frac{L^2(d+\tau)}{a_0^2nT\lambda_k}
		\right]\wedge 1
		\right).
		\]
		Set
		\[
		s_1:=\frac{L}{a_0\sqrt n}\sqrt{\frac{d+\tau}{T\lambda_k}},
		\qquad
		s_2:=\frac{L^2}{a_0^2n}\sqrt{\frac{d+\tau}{T\lambda_k^2}}.
		\]
		The third term is exactly \(s_1^2\). If \(s_1+s_2<1\), then
		\(s_1^2\le s_1\); if \(s_1+s_2\ge1\), truncation by one makes the desired
		bound immediate. Substituting
		\(L=CK_Y\sqrt{\Gamma_{n,T,\lambda_k}+\tau}\) into \(s_1+s_2\) gives precisely the
		two terms displayed in the theorem. Finally, if
		\[
		\Omega_L:=\left\{\max_{t\le T,\;i\le n}|Y_{t,i}|\le L\right\},
		\]
		then
		\[
		\Prob(\Omega_L^c)\le CnTe^{-cL^2/K_Y^2}\le Ce^{-\tau}
		\]
		for the same choice \(L=CK_Y\sqrt{\Gamma_{n,T,\lambda_k}+\tau}\). On \(\Omega_L\),
		\(\hat\beta_t^{(L)}=\hat\beta_t\) for all \(t\), so
		\(\widehat V_k^{(L)}=\widehat V_k\). This yields the stated bound.
	\end{proof}
	
	\subsection{Proof of Corollary~\ref{cor:overview-cls-minimax}}
	
	\begin{proof}
		The lower bound is Theorem~\ref{thm:gsi-lower}. We prove the matching
		expectation upper bound for the first-moment spectral estimator
		\(\widehat V_k\). Set
		\[
		r_{\rm cls}:=
		\left[
		\sqrt{\frac{d}{Tn\lambda_k}}
		+
		\sqrt{\frac{d}{Tn^2\lambda_k^2}}
		\right]\wedge1.
		\]
		
		Fix \((\mathcal V,\beta_1,\ldots,\beta_T)\in
		\mathcal P(\lambda_k)\), let \(V\) be an orthonormal basis
		of \(\mathcal V\), and write
		\[
		L:={\sinth{\widehat{\mathcal V}}{\mathcal V}}.
		\]
		By Theorem~\ref{thm:gsi}, there exist constants \(K_1,K_2>0\), depending only
		on the fixed link constants, such that for all \(\tau\ge1\),
		\[
		\Pr\!\left(
		L>
		K_1\left[
		\sqrt{\frac{d+\tau}{Tn\lambda_k}}
		+
		\sqrt{\frac{d+\tau}{Tn^2\lambda_k^2}}
		\right]\wedge1
		\right)
		\le K_2e^{-\tau}.
		\]
		Set
		\[
		a:=\frac{1}{\sqrt{Tn\lambda_k}},
		\qquad
		b:=\frac{1}{\sqrt{Tn^2\lambda_k^2}}.
		\]
		We claim that
		\[
		\E L\le C\bigl[(a+b)\sqrt d\bigr]\wedge1.
		\]
		If \(2K_1(a+b)\sqrt d\ge1\), this follows from \(0\le L\le1\). Otherwise,
		by the layer-cake formula,
		\[
		\E L=\int_0^1\Pr(L>s)\,ds.
		\]
		Let \(s_0:=2K_1(a+b)\sqrt d\). The contribution from \([0,s_0]\) is at most
		\(s_0\). For \(s\ge s_0\), set
		\[
		\tau_s:=\left(\frac{s}{K_1(a+b)}\right)^2-d.
		\]
		Then \(\tau_s\ge0\),
		\[
		K_1(a+b)\sqrt{d+\tau_s}=s,
		\]
		and, since \(s\ge2K_1(a+b)\sqrt d\),
		\[
		\tau_s\ge \frac{s^2}{2K_1^2(a+b)^2}.
		\]
		Therefore
		\begin{align*}
			\int_{s_0}^1\Pr(L>s)\,ds
			&\le
			K_2\int_{s_0}^1
			\exp\left\{-\frac{s^2}{2K_1^2(a+b)^2}\right\}\,ds\\
			&\le C(a+b)
			\le C(a+b)\sqrt d,
		\end{align*}
		where \(d\ge1\). Hence
		\[
		\sup_{(\mathcal V,\beta_1,\ldots,\beta_T)\in
			\mathcal P(\lambda_k)}
		\E\!\left[{\sinth{\widehat{\mathcal V}}{\mathcal V}}\right]
		\le Cr_{\rm cls}.
		\]
		Combining this upper bound with Theorem~\ref{thm:gsi-lower} proves the
		corollary.
	\end{proof}
	
	\subsection{Proof of Corollary~\ref{cor:overview-add-minimax}}
	
	\begin{proof}
		The lower bound is Theorem~\ref{thm:gsi-additive-lower}. We establish the
		expectation upper bound for the first-moment spectral estimator. Write
		\[
		\Gamma:=\Gamma_{n,T,\lambda_k},
		\qquad
		A:=\frac{K_Y}{a_g\sqrt{Tn\lambda_k}},
		\qquad
		B:=\frac{K_Y^2}{a_g^2\sqrt{Tn^2\lambda_k^2}},
		\]
		and set
		\[
		q_0:=A\sqrt{\Gamma d}+B\Gamma\sqrt d.
		\]
		The standing nonemptiness assumption gives \(T\ge k\) and
		\(\lambda_k\le1/k\), so \(\Gamma=\log(e nT/\lambda_k)\ge1\); also \(d\ge1\).
		Therefore, for every \(\tau\ge1\),
		\[
		A\sqrt{(\Gamma+\tau)(d+\tau)}
		+B(\Gamma+\tau)\sqrt{d+\tau}
		\le q_0(1+\tau)^{3/2}.
		\]
		Theorem~\ref{thm:gsi-subg} consequently gives constants \(C_1,C_2>0\) such
		that, uniformly over \(\mathcal P(\lambda_k)\),
		\[
		\Prob\!\left(
		{\sinth{\widehat{\mathcal V}}{\mathcal V}}>
		\bigl[C_1q_0(1+\tau)^{3/2}\bigr]\wedge1
		\right)
		\le C_2e^{-\tau}.
		\]
		Set \(s_0:=2^{3/2}C_1q_0\). If \(s_0\ge1\), boundedness of the
		loss gives the desired expectation bound immediately. Otherwise, define
		\[
		\tau_s:=\left(\frac{s}{C_1q_0}\right)^{2/3}-1
		\qquad (s\ge s_0).
		\]
		Then \(\tau_s\ge1\), so the tail bound above applies. The layer-cake
		formula gives
		\begin{align*}
			\E\!\left[{\sinth{\widehat{\mathcal V}}{\mathcal V}}\right]
			&\le s_0
			+C_2\int_{s_0}^1
			\exp\!\left\{1-\left(\frac{s}{C_1q_0}\right)^{2/3}\right\}\,ds\\
			&\le Cq_0.
		\end{align*}
		For brevity, write
		\[
		r_1:=\sqrt{\frac{d}{Tn\lambda_k}},
		\qquad
		r_2:=\sqrt{\frac{d}{Tn^2\lambda_k^2}}.
		\]
		Thus the minimax risk \(R^\star\) satisfies
		\[
		c\left(\bigl[r_1+r_2\bigr]\wedge1\right)
		\le R^\star
		\le
		C\left(\bigl[\sqrt\Gamma\,r_1+\Gamma r_2\bigr]\wedge1\right),
		\]
		where the fixed factors involving \(K_Y\) and \(a_g\) are absorbed into
		\(C\). {This is exactly the pair of inequalities stated in the corollary.}
	\end{proof}

	\section{Upper Bounds for Task-Dependent Nonlinear Links}
	\label{sec:task-dependent-nonlinear-links}
	
	This section extends the nonlinear upper-bound analysis to settings in which
	the links may differ across tasks. The first-moment spectral estimator does
	not require knowledge of the links, and the multitask lemma already allows
	heterogeneous task statistics. The extension therefore amounts to checking
	the alignment and population-gap conditions uniformly over the task links;
	the concentration and perturbation arguments from the common-link case can
	then be reused. We treat binary classification and additive-noise regression.
	
	Replace the common-link model~\eqref{eq:gsi-model} by
	\begin{equation}\label{eq:task-dependent-nonlinear-model}
		X_{t,i}\sim N(0,I_d),
		\qquad
		\E[Y_{t,i}\mid X_{t,i}]=g_t(X_{t,i}^\top\beta_t),
		\qquad
		Y_{t,i}\mid X_{t,i}\stackrel d=
		Y_{t,i}\mid X_{t,i}^\top\beta_t.
	\end{equation}
	The pairs are independent and identically distributed within each task and
	independent across tasks. Each task has \(n\) samples, the links
	\(g_t:\R\to\R\) are fixed and may be unknown, and
	\(\|\beta_t\|_2\le1\). As before,
	\(\mathcal V=\operatorname{span}\{\beta_t:t\in[T]\}\) has known dimension
	\(1\le k<d\), and
	\(\lambda_k:=\lambda_k(\Sigma_{\bm\beta})>0\).
	We use the same estimator~\eqref{eq:gsi-first-moment-estimator}--\eqref{eq:gsi-subspace-estimator}.
	
	Throughout this section, assume that the links satisfy the uniform
	smoothness and alignment conditions
	\begin{equation}\label{eq:task-dependent-link-conditions}
		g_t\in C^2(\R),
		\qquad
		\sup_{t\in[T]}\|g_t'\|_\infty\le L_1,
		\qquad
		\sup_{t\in[T]}\|g_t''\|_\infty\le L_2,
		\qquad
		a_*:=\inf_{\substack{t\in[T]\\0\le r\le1}}
		\bigl|\E[g_t'(rZ)]\bigr|>0,
	\end{equation}
	where \(Z\sim N(0,1)\) and \(L_1,L_2<\infty\). For growing task
	collections, these derivative bounds and a positive lower bound on \(a_*\)
	are required uniformly in \(T\). The averaged slopes need not have the same
	sign across tasks.
	
	For classification, specialize~\eqref{eq:task-dependent-nonlinear-model} to
	\begin{equation}\label{eq:task-dependent-classification-model}
		Y_{t,i}\mid X_{t,i}\sim
		\operatorname{Ber}\!\left(g_t(X_{t,i}^\top\beta_t)\right),
		\qquad g_t:\R\to[0,1].
	\end{equation}
	The following result extends Theorem~\ref{thm:gsi} with the common-link
	alignment constant \(a_g\) replaced by its uniform counterpart \(a_*\).
	
	\begin{theorem}[Classification with task-dependent links]
		\label{thm:task-dependent-classification-upper}
		Under~\eqref{eq:task-dependent-classification-model} and
		\eqref{eq:task-dependent-link-conditions}, assume \(n\ge n_{\mathrm{cls}}^{\mathrm{td}}\),
		where \(n_{\mathrm{cls}}^{\mathrm{td}}\ge2\) is a finite threshold bounded in terms of
		\(a_*\) and \(L_2\) alone. Then, for every \(\tau\ge1\), with probability
		at least \(1-e^{-\tau}\),
		\begin{equation}\label{eq:task-dependent-classification-rate}
			\sinth{\widehat{\mathcal V}}{\mathcal V}
			\le Ca_*^{-2}\left(
			\left[
			\sqrt{\frac{d+\tau}{Tn\lambda_k}}
			+
			\sqrt{\frac{d+\tau}{Tn^2\lambda_k^2}}
			\right]\wedge1\right).
		\end{equation}
		Here \(C\) depends only on the fixed uniform derivative bounds, and not
		on \(n,T,d,\lambda_k\) or the individual links beyond those bounds.
	\end{theorem}
	
	\begin{remark}[Symmetric task links]
		If every link satisfies \(g_t(-u)=1-g_t(u)\), the threshold
		\(n_{\mathrm{cls}}^{\mathrm{td}}\) can be chosen as \(2\), even when the links differ across
		tasks. Thus the upper bound has the same sample-size requirement as in the
		common symmetric-link case.
	\end{remark}
	
	For additive-noise regression, specialize~\eqref{eq:task-dependent-nonlinear-model} to
	\begin{equation}\label{eq:task-dependent-additive-model}
		Y_{t,i}=g_t(X_{t,i}^\top\beta_t)+\xi_{t,i},
		\qquad \xi_{t,i}\perp X_{t,i},
		\qquad \E[\xi_{t,i}]=0,
		\qquad \|\xi_{t,i}\|_{\psi_2}\le K_\xi.
	\end{equation}
	The noise distributions may differ across tasks. In addition to
	\eqref{eq:task-dependent-link-conditions}, assume
	\begin{equation}\label{eq:task-dependent-link-offsets}
		\sup_{t\in[T]}|g_t(0)|\le B<\infty,
		\qquad
		K_Y:=C_0(B+L_1+K_\xi),
	\end{equation}
	where \(C_0\) is a sufficiently large universal constant and \(B,K_\xi\)
	are fixed uniformly in \(T\). This gives the uniform response-tail bound
	needed by the truncation argument, while the estimator remains the
	untruncated first-moment spectral estimator.

	\begin{theorem}[Additive-noise regression with task-dependent links]
	\label{thm:task-dependent-additive-upper}
		Under~\eqref{eq:task-dependent-additive-model},
		\eqref{eq:task-dependent-link-conditions}, and
		\eqref{eq:task-dependent-link-offsets}, assume \(n\ge n_{\mathrm{add}}^{\mathrm{td}}\),
		where \(n_{\mathrm{add}}^{\mathrm{td}}\ge1\) is a finite threshold bounded in terms of
		\(B,L_1,L_2,a_*\) alone and specified in
		\eqref{eq:task-dependent-additive-threshold} below. Set
		\(\Gamma_{n,T,\lambda_k}:=\log(e nT/\lambda_k)\).
		Then, for every \(\tau\ge1\), with probability at least \(1-e^{-\tau}\),
		\begin{equation}\label{eq:task-dependent-additive-rate}
			\sinth{\widehat{\mathcal V}}{\mathcal V}
			\le C\left(\left[
			\frac{K_Y\sqrt{\Gamma_{n,T,\lambda_k}+\tau}}{a_*}
			\sqrt{\frac{d+\tau}{Tn\lambda_k}}
			+
			\frac{K_Y^2(\Gamma_{n,T,\lambda_k}+\tau)}{a_*^2}
			\sqrt{\frac{d+\tau}{Tn^2\lambda_k^2}}
			\right]\wedge1\right).
		\end{equation}
		Here \(C\) depends only on the fixed uniform model constants, not on
		\(n,T,d,\lambda_k\) or the individual links and noise laws beyond those
		constants.
	\end{theorem}

	\begin{remark}[Reduction to a common link]
		If \(g_t=g\) for every task, take \(B=|g(0)|\) and \(a_*=a_g\).
		The threshold~\eqref{eq:task-dependent-additive-threshold} then reduces
		exactly to~\eqref{eq:n-add-definition}, and the assumptions and rate
		recover Theorem~\ref{thm:gsi-subg}.
	\end{remark}

	\subsection{Proof of the Classification Upper Bound}
	\label{subsec:proof-task-dependent-classification-upper}
	
	\begin{proof}[Proof of Theorem~\ref{thm:task-dependent-classification-upper}]
		We recall only the taskwise ingredients from
		Appendix~\ref{sec:proof-of-classification-upper-bound} that change when
		\(g\) is replaced by \(g_t\). Write
		\[
		r_t:=\|\beta_t\|_2,
		\qquad a_t:=\E[g_t'(r_tZ)],
		\qquad c_t:=\E[g_t''(r_tZ)],
		\qquad p_t:=\E[g_t(r_tZ)].
		\]
		
		\paragraph{Alignment.}
		The Stein calculation giving~\eqref{eq:gsi-population-alignment} is
		taskwise and therefore yields
		\begin{equation}\label{eq:task-dependent-population-alignment}
			\widetilde\beta_t=a_t\beta_t,
			\qquad
			\Sigma_{\bm{\tilde\beta}}
			=\frac1T\sum_{t=1}^Ta_t^2\beta_t\beta_t^\top
			\succeq a_*^2\Sigma_{\bm\beta}.
		\end{equation}
		Hence \(\widetilde{\mathcal V}=\mathcal V\) and
		\(\widetilde\lambda_k\ge a_*^2\lambda_k\), exactly as in the common-link
		proof.
		
		\paragraph{Error control and covariance.}
		The sub-Gaussian error argument in
		Appendix~\ref{sec:proof-of-classification-upper-bound} uses only
		\(Y_{t,i}\in\{0,1\}\) and the Gaussian design, so it gives the same
		uniform control with \(\sigma^2\le C/n\). The rotational covariance
		calculation there likewise applies separately to each task and gives
		\begin{equation}\label{eq:task-dependent-classification-covariance}
			\operatorname{Cov}(\widehat\beta_t)
			=\frac1n\left[p_tI_d+(c_t-a_t^2)\beta_t\beta_t^\top\right].
		\end{equation}
		Thus the averaged error covariance retains the block
		structure~\eqref{eq:gsi-covariance-block-structure}, and \(\delta_1=0\).
		Different response prevalences \(p_t\) change only its isotropic term;
		in the notation of the multitask lemma,
		\(\nu=(Tn)^{-1}\sum_{t=1}^Tp_t\).
		
		\paragraph{Uniform population gap.}
		Replace the proof-only constants in~\eqref{eq:bg-ng-definition} by
		\begin{equation}\label{eq:task-dependent-classification-threshold}
			b_*:=\sup_{\substack{t\in[T]\\0\le r\le1}}
			\Bigl(-\E[g_t''(rZ)]\Bigr)_+,
			\qquad
			n_{\mathrm{cls}}^{\mathrm{td}}:=\max\left\{2,\,1+\frac{2b_*}{a_*^2}\right\}.
		\end{equation}
		Since \(b_*\le L_2\), this threshold is finite and uniformly bounded
		in terms of \(a_*\) and \(L_2\). The coefficient calculation from the
		common-link proof now reads
		\begin{equation}\label{eq:task-dependent-classification-gap}
			\left(1-\frac1n\right)a_t^2+\frac{c_t}{n}
			\ge\left(1-\frac1n\right)a_t^2-\frac{b_*}{n}
			\ge\frac12\left(1-\frac1n\right)a_t^2.
		\end{equation}
		Averaging gives
		\(\rho-\nu\ge\frac12(1-n^{-1})\widetilde\lambda_k
		\ge\widetilde\lambda_k/4\). This verifies
		Condition~\ref{cond:surrogate-recovery} with the same fixed population-gap
		constant. For symmetric task links, each \(c_t=0\), so \(b_*=0\) and
		\(n_{\mathrm{cls}}^{\mathrm{td}}=2\), proving the remark.
		
		\paragraph{Application of the multitask lemma.}
		Apply Theorem~\ref{thm:mtl-main}. In the final rate conversion from
		Appendix~\ref{sec:proof-of-classification-upper-bound}, the uniform bound
		\(\lvert a_t\rvert\le L_1\) replaces the common-link upper bound on the
		averaged slopes. The same spectral argument gives
		\(\rho\le L_1^2\lambda_k+C/n\). Together with
		\eqref{eq:task-dependent-population-alignment}, \(\sigma^2\le C/n\),
		and \(\delta_1=0\), the remaining rate simplification is identical to
		that proof, with \(a_g\) replaced by \(a_*\). Its observation that
		\(a_g\le1\) also applies taskwise and gives \(a_*\le1\), so the same
		placement of the alignment factor outside the truncation is valid.
		This proves~\eqref{eq:task-dependent-classification-rate}.
	\end{proof}
	
	\subsection{Proof of the Additive-Noise Upper Bound}
	\label{subsec:proof-task-dependent-additive-upper}

	\begin{proof}[Proof of Theorem~\ref{thm:task-dependent-additive-upper}]
		We adapt Appendix~\ref{sec:proof-of-additive-upper-bound}, retaining its
		truncation construction and replacing common-link scalar quantities by
		their taskwise counterparts. For \(Z\sim N(0,1)\), \(r\in[0,1]\), and
		an independent noise \(\xi_t\) with the task's law, put
		\[
		Y_{t,r}:=g_t(rZ)+\xi_t,
		\qquad \alpha_t(r):=\E[g_t'(rZ)],
		\qquad \chi_t(r):=\E[(g_t^2)''(rZ)].
		\]
		The same Stein identities give
		\begin{equation}\label{eq:task-dependent-additive-moments}
			A_t(r):=\E[Y_{t,r}Z]=r\alpha_t(r),
			\qquad
			Q_t(r):=\E[Y_{t,r}^2(Z^2-1)]=r^2\chi_t(r),
		\end{equation}
		and \(R_t(r):=Q_t(r)-A_t(r)^2\). The noise contributions to \(Q_t\)
		vanish by independence and zero mean, regardless of its taskwise
		variance. Thus the relevant curvature is that of \(g_t^2\). Moreover,
		\[
		\sup_{t,r}|\chi_t(r)|
		\le 2L_1^2+2L_2\bigl(B+L_1\E|Z|\bigr)<\infty.
		\]
		The generalization of~\eqref{eq:n-add-definition} is
		\begin{equation}\label{eq:task-dependent-additive-threshold}
			\kappa_t(r):=
			\frac{(\alpha_t(r)^2-\chi_t(r))_+}{\alpha_t(r)^2},
			\qquad
			n_{0}^{\mathrm{td}}:=1+\sup_{\substack{t\in[T]\\0\le r\le1}}\kappa_t(r),
			\qquad n_{\mathrm{add}}^{\mathrm{td}}:=\lceil n_{0}^{\mathrm{td}}\rceil.
		\end{equation}
		Because \(|\alpha_t(r)|\ge a_*\), the denominator is nonzero and
		\(n_{0}^{\mathrm{td}}\le2+a_*^{-2}\sup_{t,r}|\chi_t(r)|<\infty\).
		In particular,
		\begin{equation}\label{eq:task-dependent-additive-relative-curvature}
			R_t(r)\ge-(n_{0}^{\mathrm{td}}-1)A_t(r)^2
			\qquad\text{for every }t\in[T],\ r\in[0,1].
		\end{equation}

		\paragraph{Uniform truncation and alignment.}
		From~\eqref{eq:task-dependent-link-offsets},
		\(\sup_{t,r}\|Y_{t,r}\|_{\psi_2}\le K_Y\). Define
		\(\phi_L(y):=y\mathbf 1\{|y|\le L\}\) and
		\(\widehat\beta_t^{(L)}:=n^{-1}\sum_i\phi_L(Y_{t,i})X_{t,i}\), with
		\(L=C_{\rm tr}K_Y\sqrt{\Gamma_{n,T,\lambda_k}+\tau}\).
		The constant \(C_{\rm tr}\) is chosen sufficiently large in terms of
		the fixed uniform model constants. Lemma~\ref{lem:gsi-subg-convergence}
		uses only a uniform response-tail bound, so its estimates now hold
		with the supremum over \(r\) replaced by the supremum over \((t,r)\).
		Here \(A_{t,L},Q_{t,L},R_{t,L}\) are the moments above with \(Y_{t,r}\)
		replaced by \(\phi_L(Y_{t,r})\).
		Proposition~\ref{prop:response-transformation-alignment} applies
		taskwise, so every truncated population vector remains in
		\(\operatorname{span}(\beta_t)\). The alignment identity
		\eqref{eq:task-dependent-population-alignment} and the same vector-bias
		and Weyl argument from the common-link proof therefore give
		\begin{equation}\label{eq:task-dependent-additive-truncated-signal}
			\widetilde\lambda_k^{(L)}\ge\frac12a_*^2\lambda_k,
			\qquad \widetilde{\mathcal V}^{(L)}=\mathcal V.
		\end{equation}

		\paragraph{Covariance structure and population gap.}
		The rotational covariance calculation in the common-link proof is
		also taskwise; it retains~\eqref{eq:gsi-covariance-block-structure}
		for the truncated errors and hence \(\delta_1=0\). Their directional
		sub-Gaussian scale remains \(\sigma_L\asymp L/\sqrt n\).
		Use the quantities \(b_L,c_L,\varepsilon_L,\delta_L\) from
		Appendix~\ref{sec:proof-of-additive-upper-bound}, now taking the
		truncation-error suprema over \((t,r)\), and replacing \(n_0,a_0\) by
		\(n_{0}^{\mathrm{td}},a_*\). The uniform convergence bounds allow the same choices
		\(\varepsilon_L\le1/4\) and \(\delta_L\le a_*^2\lambda_k/8\).
		Equation~\eqref{eq:task-dependent-additive-relative-curvature} is
		exactly the hypothesis used in Lemma~\ref{lem:gsi-subg-gap-transfer}.
		Applying that calculation at each realized task norm and averaging
		therefore gives
		\begin{equation}\label{eq:task-dependent-additive-gap}
			\rho_L-\nu_L
			\ge\left(1-\frac{n_{0}^{\mathrm{td}}-1+\varepsilon_L}{n}\right)
			\widetilde\lambda_k^{(L)}-\frac{\delta_L}{n}
			\ge\frac1{2n_{0}^{\mathrm{td}}}\widetilde\lambda_k^{(L)}.
		\end{equation}
		The last step uses \(n\ge\lceil n_{0}^{\mathrm{td}}\rceil\) and
		\eqref{eq:task-dependent-additive-truncated-signal}, just as in the
		common-link proof. Thus all conditions of Theorem~\ref{thm:mtl-main}
		hold with a gap constant controlled by the uniform model constants.

		\paragraph{Rate conversion and transfer.}
		Apply the multitask lemma to the truncated estimator. Its final
		rate conversion in Appendix~\ref{sec:proof-of-additive-upper-bound}
		uses only \(\rho_L\le\widetilde\lambda_k^{(L)}+CL^2/n\),
		\eqref{eq:task-dependent-additive-truncated-signal}, and boundedness
		of the subspace loss. It therefore applies unchanged with \(a_0=a_*\).
		The same uniform tail bound gives
		\(\Prob\{\max_{t,i}|Y_{t,i}|>L\}\le CnT e^{-cL^2/K_Y^2}\).
		Apply the multitask lemma with confidence parameter \(\tau+\log2\)
		and enlarge \(C_{\rm tr}\) so that this agreement event and the
		multitask event each fail with probability at most \(e^{-\tau}/2\).
		On their intersection, all truncated task statistics equal their
		untruncated counterparts. The resulting two-term bound is
		\eqref{eq:task-dependent-additive-rate}, with total failure probability
		at most \(e^{-\tau}\).
	\end{proof}
	
	\section{Nonlinear Models with General Gaussian Design}
	\label{sec:nonlinear-general-design}
	
	This section extends the binary classification analysis from isotropic
	Gaussian covariates to Gaussian covariates with a common positive definite
	covariance matrix \(\Sigma_X\). The extension illustrates how the multitask
	lemma guides the construction of taskwise estimators: even when their
	population means are aligned with the task parameters, their errors can
	mix the shared subspace with its orthogonal complement. Whitening removes
	this interaction and permits recovery of the original subspace after a
	change of coordinates.
	
	Section~\ref{subsec:nonlinear-known-sigma} treats known \(\Sigma_X\),
	identifies the required whitening transformation, and establishes the
	same two-term recovery rate as in the isotropic setting, up to constants
	depending on the conditioning of \(\Sigma_X\) and the link.
	Section~\ref{subsec:nonlinear-unknown-sigma} replaces \(\Sigma_X\) by an
	independently learned estimate, bounds the effect of covariance-estimation
	error, and shows that pooling covariates through sample splitting preserves
	this rate. Throughout, we give only the changes to the existing
	classification proof in Appendix~\ref{sec:proof-of-classification-upper-bound}.
	The same whitening transformation should also extend the additive-noise
	regression analysis, after verifying its truncation and population-gap
	conditions.
	
	\subsection{Known Common Covariance}
	\label{subsec:nonlinear-known-sigma}
	
	Replace the design in~\eqref{eq:gsi-model} by
	\begin{equation}\label{eq:common-sigma-classification-model}
		X_{t,i}\sim N(0,\Sigma_X),
		\qquad
		Y_{t,i}\mid X_{t,i}\sim
		\operatorname{Ber}\!\left(g(X_{t,i}^\top\beta_t)\right),
		\qquad
		\kappa_{\min}I_d\preceq\Sigma_X\preceq\kappa_{\max}I_d,
	\end{equation}
	with the same known \(\Sigma_X\) for every task, fixed constants
	\(0<\kappa_{\min}<\kappa_{\max}\), and independent observations
	across tasks and samples. Retain
	\(\mathcal V=\operatorname{span}\{\beta_t:t\in[T]\}\) with
	\(\dim(\mathcal V)=k\), \(\|\beta_t\|_2\le1\), and the smoothness assumptions on \(g\) in
	Theorem~\ref{thm:gsi}. The transformed task norms satisfy
	\[
		r_t:=(\beta_t^\top\Sigma_X\beta_t)^{1/2}
		\le\sqrt{\kappa_{\max}}.
	\]
	In this subsection, use the following versions of the link constants
	and sample-size threshold from~\eqref{eq:ag-def}
	and~\eqref{eq:bg-ng-definition}, with the norm range adjusted accordingly:
	\begin{equation}\label{eq:common-sigma-link-constants}
		a_g:=\inf_{0\le r\le\sqrt{\kappa_{\max}}}|\E[g'(rZ)]|>0,
		\qquad
		b_g:=\sup_{0\le r\le\sqrt{\kappa_{\max}}}
		\bigl(-\E[g''(rZ)]\bigr)_+,
		\qquad Z\sim N(0,1).
	\end{equation}
	Define the corresponding sample-size threshold by
	\begin{equation}\label{eq:common-sigma-classification-threshold}
		n_{\mathrm{cls}}:=\max\{2,1+2b_g/a_g^2\}.
	\end{equation}
	
	\paragraph{Why whitening is needed.}
	A natural taskwise estimator is the covariance-corrected first moment
	\(\widehat\beta_t^{(\mathrm{raw})}:=
	\Sigma_X^{-1}n^{-1}\sum_iY_{t,i}X_{t,i}\).
	Gaussian Stein identities give, with \(a_t:=\E[g'(r_tZ)]\),
	\begin{equation}\label{eq:common-sigma-raw-moments}
		\begin{aligned}
		\E[\widehat\beta_t^{(\mathrm{raw})}]&=a_t\beta_t,\\
		\operatorname{Cov}(\widehat\beta_t^{(\mathrm{raw})})
		&=\frac1n\left[\E[Y_{t,1}]\Sigma_X^{-1}+
		\bigl(\E[g''(r_tZ)]-a_t^2\bigr)
		\beta_t\beta_t^\top\right].
		\end{aligned}
	\end{equation}
	Thus its population span is exactly the target space \(\mathcal V\).
	However, as in the linear regression setting, population alignment alone
	does not ensure favorable error geometry for spectral aggregation.
	The averaged error covariance has cross block
	\[
		\Proj_{\mathcal V}\left[\frac1T\sum_t
		\operatorname{Cov}(\widehat\beta_t^{(\mathrm{raw})})\right]
		\Proj_{\mathcal V^\perp}
		=\left(\frac1{nT}\sum_t\E[Y_{t,1}]\right)
		\Proj_{\mathcal V}\Sigma_X^{-1}
		\Proj_{\mathcal V^\perp}.
	\]
	For a general \(\Sigma_X\), this cross block need not vanish.
	Consequently, directly aggregating these taskwise estimators can introduce
	error-covariance mixing, despite their exact population alignment. This
	motivates whitening before applying the spectral construction.
	
	\paragraph{Estimator.}
	We therefore whiten the taskwise estimators by defining
	\begin{equation}\label{eq:common-sigma-whitened-statistic}
		\widehat\beta_t:=\Sigma_X^{1/2}\widehat\beta_t^{(\mathrm{raw})}
		=\Sigma_X^{-1/2}\frac1n\sum_{i=1}^nY_{t,i}X_{t,i}.
	\end{equation}
	Apply the spectral construction~\eqref{eq:surrogate-subspace-estimator}
	to obtain \(\widehat{\mathcal V}^{(s)}=\operatorname{span}(\widehat V_k^{(s)})\),
	and define the target-subspace estimator by
	\begin{equation}\label{eq:common-sigma-map-back}
		\widehat{\mathcal V}:=
		\operatorname{span}\!\left(\Sigma_X^{-1/2}\widehat V_k^{(s)}\right).
	\end{equation}
	
	\begin{theorem}[Classification with known common covariance]
	\label{thm:classification-known-common-covariance}
		Assume the model and conditioning bounds in
		\eqref{eq:common-sigma-classification-model}, the task-parameter and
		link smoothness conditions above, and \(a_g>0\) as defined in
		\eqref{eq:common-sigma-link-constants}. Let
		\(\lambda_k:=\lambda_k(\Sigma_{\bm\beta})>0\) and assume
		\(n\ge n_{\mathrm{cls}}\), with \(n_{\mathrm{cls}}\) defined in
		\eqref{eq:common-sigma-classification-threshold}.
		Let \(\widehat{\mathcal V}\) be the estimator
		in~\eqref{eq:common-sigma-whitened-statistic}--\eqref{eq:common-sigma-map-back}.
		Then, for every \(\tau\ge1\), with probability at least \(1-e^{-\tau}\),
		\begin{equation}\label{eq:common-sigma-classification-rate}
			\sinth{\widehat{\mathcal V}}{\mathcal V}
			\le C\left(\left[
			\sqrt{\frac{d+\tau}{Tn\lambda_k}}
			+\sqrt{\frac{d+\tau}{Tn^2\lambda_k^2}}
			\right]\wedge1\right).
		\end{equation}
		The constant \(C\) depends only on the fixed link constants, including
		\(a_g^{-1}\), and the design-conditioning constants
		\(\kappa_{\min},\kappa_{\max}\).
	\end{theorem}
	
	\begin{proof}
	We record only the changes to the proof of Theorem~\ref{thm:gsi}
	in Appendix~\ref{sec:proof-of-classification-upper-bound}.
	\paragraph{Transformed coordinates and surrogate geometry.}
	Set \(G_{t,i}:=\Sigma_X^{-1/2}X_{t,i}\) and
	\(\theta_t:=\Sigma_X^{1/2}\beta_t\).
	Now \(G_{t,i}\sim N(0,I_d)\) and
	\(X_{t,i}^\top\beta_t=G_{t,i}^\top\theta_t\), so the proof of Theorem~\ref{thm:gsi} in
	Appendix~\ref{sec:proof-of-classification-upper-bound} applies with \(\beta_t\) replaced by \(\theta_t\)
	and the norm range \([0,1]\) replaced by
	\([0,\sqrt{\kappa_{\max}}]\). In particular, the population surrogates are
	\begin{equation}\label{eq:common-sigma-corrected-geometry}
		\widetilde\beta_t:=\E[\widehat\beta_t]=a_t\theta_t,
		\qquad
		\widetilde{\mathcal V}=\Sigma_X^{1/2}\mathcal V.
	\end{equation}
	With \(\varepsilon_t:=\widehat\beta_t-\widetilde\beta_t\), the averaged
	error covariance retains the form~\eqref{eq:gsi-covariance-block-structure},
	now relative to the surrogate space:
	\begin{equation}\label{eq:common-sigma-error-covariance}
		\Sigma_{\bm\varepsilon}
		=\omega^2I_d+\widetilde V R_\star\widetilde V^\top,
		\qquad \omega^2:=\frac1{nT}\sum_t\E[Y_{t,1}],
	\end{equation}
	where \(\widetilde V\) is an orthonormal basis of
	\(\widetilde{\mathcal V}\) and \(R_\star\) is symmetric. Hence the cross
	block relative to this surrogate space is zero and \(\delta_1=0\).
	The directional sub-Gaussian scale remains \(\sigma^2\le C/n\).
	The surrogate covariance is
	\[
		\Sigma_{\bm{\tilde\beta}}
		=\Sigma_X^{1/2}\left(\frac1T\sum_t a_t^2\beta_t\beta_t^\top\right)
		\Sigma_X^{1/2},
	\]
	so its smallest nonzero eigenvalue satisfies
	\begin{equation}\label{eq:common-sigma-signal-comparison}
		a_g^2\kappa_{\min}\lambda_k\le\widetilde\lambda_k
		\le\|g'\|_\infty^2\kappa_{\max}\lambda_k,
	\end{equation}
	where \(\lambda_k=\lambda_k(\Sigma_{\bm\beta})\).
	The population-gap calculation in the proof of Theorem~\ref{thm:gsi}, using
	\eqref{eq:common-sigma-link-constants}, gives
	\[
		\rho-\nu\ge\frac12(1-n^{-1})\widetilde\lambda_k
		\ge\frac14\widetilde\lambda_k,
		\qquad
		\rho\le\widetilde\lambda_k+C/n.
	\]
	These are the only changes needed to verify the multitask conditions;
	the concentration and rate simplification in the proof of
	Theorem~\ref{thm:gsi} apply
	using~\eqref{eq:common-sigma-signal-comparison}.
	
	\paragraph{Recovery of the original space.}
	For the target-subspace estimator in~\eqref{eq:common-sigma-map-back},
	the identity \(\Sigma_X^{-1/2}\widetilde{\mathcal V}=\mathcal V\) and the usual
	subspace comparison under an invertible linear map give
	\begin{equation}\label{eq:common-sigma-subspace-transfer}
		\sinth{\widehat{\mathcal V}}{\mathcal V}
		\le\sqrt{\frac{\kappa_{\max}}{\kappa_{\min}}}\,
		\sinth{\widehat{\mathcal V}^{(s)}}{\widetilde{\mathcal V}}.
	\end{equation}
	Consequently, Theorem~\ref{thm:mtl-main}, the signal comparison
	\eqref{eq:common-sigma-signal-comparison}, and the rate simplification
	from the proof of Theorem~\ref{thm:gsi} give~\eqref{eq:common-sigma-classification-rate}.
	\end{proof}
	
	\subsection{Unknown Common Covariance}
	\label{subsec:nonlinear-unknown-sigma}
	Retain the model, conditioning bounds, and link constants of
	Section~\ref{subsec:nonlinear-known-sigma}. Let \(\widehat\Sigma_X\) be a
	positive definite covariance estimate independent of the labeled pairs
	used to form the task statistics. It may be learned from additional
	unlabeled covariates or from a disjoint part of each task's sample.
	Measure its quality by the relative error
	\begin{equation}\label{eq:unknown-sigma-relative-error}
		\eta:=\left\|\Sigma_X^{-1/2}
		(\widehat\Sigma_X-\Sigma_X)\Sigma_X^{-1/2}\right\|_{\mathrm{op}}.
	\end{equation}
	Replace \(\Sigma_X\) by \(\widehat\Sigma_X\) in the preceding construction:
	\begin{equation}\label{eq:unknown-sigma-estimator}
		\widehat\beta_t:=\widehat\Sigma_X^{-1/2}\frac1n\sum_iY_{t,i}X_{t,i}.
	\end{equation}
	Apply~\eqref{eq:surrogate-subspace-estimator} to these task statistics and set
	\begin{equation}\label{eq:unknown-sigma-map-back}
		\widehat{\mathcal V}:=
		\operatorname{span}\!\left(\widehat\Sigma_X^{-1/2}\widehat V_k^{(s)}\right).
	\end{equation}
	
	\begin{theorem}[Classification with an estimated common covariance]
	\label{thm:classification-estimated-common-covariance}
		Assume the model, task-parameter, and link conditions of
		Theorem~\ref{thm:classification-known-common-covariance}, with
		\(\lambda_k:=\lambda_k(\Sigma_{\bm\beta})>0\) and
		\(n\ge n_{\mathrm{cls}}\) as defined in
		\eqref{eq:common-sigma-classification-threshold}.
		Let \(\widehat\Sigma_X\) be positive definite and independent of the
		labeled pairs used in~\eqref{eq:unknown-sigma-estimator}, and let
		\(\eta\) be its relative error from~\eqref{eq:unknown-sigma-relative-error}.
		Use \(\kappa_{\min}\) from~\eqref{eq:common-sigma-classification-model}
		and \(a_g\) from~\eqref{eq:common-sigma-link-constants}.
		For every fixed realization of \(\widehat\Sigma_X\) satisfying
		\begin{equation}\label{eq:unknown-sigma-small-error}
			\eta\le\min\left\{\frac14,
			\frac{n\kappa_{\min}a_g^2\lambda_k}{128}\right\},
		\end{equation}
		the estimator~\eqref{eq:unknown-sigma-estimator}--\eqref{eq:unknown-sigma-map-back}
		satisfies, for every \(\tau\ge1\), with conditional probability at least
		\(1-e^{-\tau}\) given \(\widehat\Sigma_X\),
		\begin{equation}\label{eq:unknown-sigma-conditional-rate}
			\sinth{\widehat{\mathcal V}}{\mathcal V}
			\le C\left(\left[
			\sqrt{\frac{d+\tau}{Tn\lambda_k}}
			+\sqrt{\frac{d+\tau}{Tn^2\lambda_k^2}}
			+\frac{\eta}{n\lambda_k}+\eta
			\right]\wedge1\right).
		\end{equation}
		The constant \(C\) has the same dependence as in
		Theorem~\ref{thm:classification-known-common-covariance} and is uniform
		over the realizations satisfying~\eqref{eq:unknown-sigma-small-error}.
	\end{theorem}
	
	\begin{proof}
	Condition throughout on \(\widehat\Sigma_X\). Independence of the covariance
	sample leaves the labeled pairs independent with their original law, so
	the multitask lemma applies conditionally. We give only the changes to
	the proof of Theorem~\ref{thm:classification-known-common-covariance}.
	
	\paragraph{Approximate whitening and the conditional surrogates.}
	Write \(H:=\widehat\Sigma_X^{-1/2}\Sigma_X^{1/2}\) and retain
	\(\theta_t=\Sigma_X^{1/2}\beta_t\) from the proof of
	Theorem~\ref{thm:classification-known-common-covariance}.
	The estimated-whitening statistics are \(H\) times the exactly whitened
	statistics. Since
	\(H^\top H=(\Sigma_X^{-1/2}\widehat\Sigma_X\Sigma_X^{-1/2})^{-1}\),
	\[
		\frac{I_d}{1+\eta}\preceq HH^\top\preceq\frac{I_d}{1-\eta},
		\qquad
		\|HH^\top-I_d\|_{\mathrm{op}}\le\frac{\eta}{1-\eta}.
	\]
	Consequently, the conditional surrogate quantities become
	\begin{equation}\label{eq:unknown-sigma-surrogate-geometry}
		\widetilde\beta_t:=\E[\widehat\beta_t\mid\widehat\Sigma_X]
		=a_tH\theta_t,
		\qquad
		\widetilde{\mathcal V}=\widehat\Sigma_X^{-1/2}\Sigma_X\mathcal V.
	\end{equation}
	The signal comparison~\eqref{eq:common-sigma-signal-comparison} and the
	sub-Gaussian error bound therefore change only to
	\[
		\frac{a_g^2\kappa_{\min}}{1+\eta}\lambda_k
		\le\widetilde\lambda_k
		\le\frac{\|g'\|_\infty^2\kappa_{\max}}{1-\eta}\lambda_k,
		\qquad \sigma^2\le\frac{C}{n(1-\eta)}.
	\]
	Here \(\varepsilon_t:=\widehat\beta_t-\widetilde\beta_t\) and
	\(\Sigma_{\bm\varepsilon}\) are defined conditionally on \(\widehat\Sigma_X\).
	
	\paragraph{Residual mixing and population gap.}
	Transforming the error covariance~\eqref{eq:common-sigma-error-covariance}
	by \(H\) gives
	\[
		\Sigma_{\bm\varepsilon}
		=\omega^2HH^\top+\widetilde V R_\star\widetilde V^\top,
		\qquad \omega^2=\frac1{nT}\sum_t\E[Y_{t,1}],
	\]
	for an orthonormal basis \(\widetilde V\) of the conditional surrogate space
	and a symmetric \(R_\star\). Only \(HH^\top-I_d\) contributes to its cross block:
	\begin{equation}\label{eq:unknown-sigma-mixing}
		\delta_1=
		\frac{\omega^2}{\widetilde\lambda_k}
		\left\|\Proj_{\widetilde{\mathcal V}}(HH^\top-I_d)
		\Proj_{\widetilde{\mathcal V}^\perp}\right\|_{\mathrm{op}}
		\le\frac{2\eta}{n\kappa_{\min}a_g^2\lambda_k}.
	\end{equation}
	The signal-supported part of the population Gram matrix still dominates
	\(\Sigma_{\bm{\tilde\beta}}/4\), by the same coefficient calculation and
	threshold~\eqref{eq:common-sigma-classification-threshold} used in the
	proof of Theorem~\ref{thm:classification-known-common-covariance}.
	Replacing its isotropic term \(\omega^2I_d\) by \(\omega^2HH^\top\) gives
	\[
		\rho-\nu\ge\frac14\widetilde\lambda_k
		-\frac{2\omega^2\eta}{1-\eta^2}
		\ge\frac14\widetilde\lambda_k-\frac{3\eta}{n}
		\ge\frac18\widetilde\lambda_k,
		\qquad
		\rho\le\widetilde\lambda_k+\frac1{n(1-\eta)},
	\]
	where the last gap bound follows from~\eqref{eq:unknown-sigma-small-error}.
	Theorem~\ref{thm:mtl-main} and the rate simplification in the proof of
	Theorem~\ref{thm:gsi} now give the
	two concentration terms in~\eqref{eq:unknown-sigma-conditional-rate},
	plus \(C\eta/(n\lambda_k)\), for recovery of
	\(\widetilde{\mathcal V}\).
	
	\paragraph{Mapping back and the shifted target.}
	Unlike~\eqref{eq:common-sigma-map-back}, the estimated map returns the
	conditional population space to
	\(\widehat\Sigma_X^{-1}\Sigma_X\mathcal V\). Its distance from the target obeys
	\[
		\sinth{\widehat\Sigma_X^{-1}\Sigma_X\mathcal V}{\mathcal V}
		\le\|\Sigma_X^{-1}(\widehat\Sigma_X-\Sigma_X)\|_{\mathrm{op}}
		\le\sqrt{\frac{\kappa_{\max}}{\kappa_{\min}}}\,\eta.
	\]
	Indeed, if \(u\in\widehat\Sigma_X^{-1}\Sigma_X\mathcal V\) is a unit vector,
	then \(\Sigma_X^{-1}\widehat\Sigma_Xu\in\mathcal V\), so its distance to
	\(\mathcal V\) is at most \(\|\Sigma_X^{-1}(\widehat\Sigma_X-\Sigma_X)u\|_2\).
	The condition number of \(\widehat\Sigma_X^{-1/2}\) is at most
	\(\sqrt{\kappa_{\max}(1+\eta)/(\kappa_{\min}(1-\eta))}\).
	The subspace-transfer argument used for
	\eqref{eq:common-sigma-subspace-transfer} in the proof of
	Theorem~\ref{thm:classification-known-common-covariance}, followed by
	the triangle inequality, therefore adds \(C\eta\) and proves the claim.
	\end{proof}
	
	\begin{corollary}[Pooled covariance estimation by sample splitting]
	\label{cor:classification-split-common-covariance}
		Assume the model, task-parameter, and link conditions of
		Theorem~\ref{thm:classification-known-common-covariance}, and let
		\(\lambda_k:=\lambda_k(\Sigma_{\bm\beta})>0\).
		Split each task's \(n\) observations into two groups of sizes
		\(\lfloor n/2\rfloor\) and \(\lceil n/2\rceil\), with
		\(\lceil n/2\rceil\ge n_{\mathrm{cls}}\), where \(n_{\mathrm{cls}}\) is
		defined in~\eqref{eq:common-sigma-classification-threshold}.
		Let \(\widehat\Sigma_X\) be the
		sample covariance matrix of the pooled covariates from the first groups,
		using the known zero mean. Apply
		\eqref{eq:unknown-sigma-estimator}--\eqref{eq:unknown-sigma-map-back}
		using only the labeled observations from the second groups.
		If \(\widehat\Sigma_X\) is singular, return any fixed \(k\)-dimensional space.
		Then the bound~\eqref{eq:common-sigma-classification-rate}, with \(n\)
		denoting the original sample size per task, holds with probability at least
		\(1-2e^{-\tau}\) for every \(\tau\ge1\), with the same constant
		dependence as in Theorem~\ref{thm:classification-known-common-covariance}.
	\end{corollary}
	
	\begin{proof}
	Let \(\eta\) be the relative covariance error in
	\eqref{eq:unknown-sigma-relative-error}. The two groups are independent,
	and the pooled covariance sample contains
	order \(nT\) observations. Gaussian sample-covariance concentration gives,
	with probability at least \(1-e^{-\tau}\),
	\begin{equation}\label{eq:unknown-sigma-pooled-concentration}
		\eta\le C\left[\sqrt{\frac{d+\tau}{nT}}+\frac{d+\tau}{nT}\right].
	\end{equation}
	When the sum of the two terms in~\eqref{eq:common-sigma-classification-rate}
	is sufficiently small, this event implies
	\begin{equation}\label{eq:unknown-sigma-absorption}
		\eta\le C\sqrt{\frac{d+\tau}{nT}}
		\le C\sqrt{\frac{d+\tau}{Tn\lambda_k}},
		\qquad
		\frac{\eta}{n\lambda_k}
		\le\frac{C}{\sqrt n}\sqrt{\frac{d+\tau}{Tn^2\lambda_k^2}}.
	\end{equation}
	Here \(\lambda_k\le1\) follows from \(\|\beta_t\|_2\le1\).
	These inequalities also ensure~\eqref{eq:unknown-sigma-small-error}
	for the retained labeled sample size: both \(\eta\) and
	\(\eta/(n\lambda_k)\) are sufficiently small, and retaining half the
	observations changes only fixed factors. Apply
	Theorem~\ref{thm:classification-estimated-common-covariance} conditionally
	on the first groups. Equation~\eqref{eq:unknown-sigma-absorption} absorbs
	both covariance-estimation terms into the two-term rate.
	Integrating the conditional failure probability and adding the covariance-event
	failure probability gives at most \(2e^{-\tau}\).
	If the two-term rate is not sufficiently small, the loss bound by one
	proves the result after enlarging \(C\), including when the covariance
	estimate is singular.
	\end{proof}
	
	Thus estimating the common covariance by sample splitting changes only
	the constants in~\eqref{eq:common-sigma-classification-rate}, leaving the
	representation-recovery rate unchanged. An independent unlabeled sample
	of size comparable to \(nT\) gives the same guarantee while retaining all
	labeled observations.
	
	\section{Lower-Bound Proofs}\label{sec:lower-bound-proofs}

This appendix collects the proof steps that are common to the lower bounds for
Gaussian single-index classification and additive Gaussian responses. The
model-specific appendices verify the one-sample Kullback--Leibler or
centered-reference \(\chi^2\) bounds, and the results below convert those
model-specific estimates into the final minimax lower bounds.

\paragraph{Roadmap.}
Section~\ref{sec:lower-bound-proofs} is organized as follows. In
Section~\ref{subsec:shared-lb-construction} we collect the packing, Fano, and
revealed-sign reductions that are shared by both lower bounds. We then prove
the classification lower bound from Theorem~\ref{thm:gsi-lower} in
Section~\ref{subsec:classification-lb}, and the additive Gaussian lower bound
from Theorem~\ref{thm:gsi-additive-lower} in
Section~\ref{subsec:additive-lb}. Auxiliary technical results used across these
proofs are collected in Section~\ref{subsec:shared-lb-technical-tools}.

\subsection{Shared Construction}\label{subsec:shared-lb-construction}

This subsection isolates the proof architecture shared by both lower bounds.
Section~\ref{subsubsec:shared-lb-geometry-fano} constructs the local packing
and the Fano reduction, Section~\ref{subsubsec:shared-lb-first-reduction}
packages the generic first-term argument, and
Section~\ref{subsubsec:shared-lb-second-reduction} collects the revealed-sign,
centered-reference, \(\chi^2\), and \(T\)-task KL reductions used for the
second term.

\subsubsection{Geometry and Fano}\label{subsubsec:shared-lb-geometry-fano}

\begin{lemma}[Local one-column packing]\label{lem:gsi-lb-packing}
Assume \(1\le k\le d/2\) and let \(\varepsilon\in(0,1/2)\). Then there exist
absolute constants \(c_{\mathrm{pack}},c_\rho,C_\rho>0\) and matrices
\(B^{(1)},\dots,B^{(M)}\in O(d\times k)\) such that
\[
\log M\ge c_{\mathrm{pack}}(d-k),
\]
the first \(k-1\) columns of all \(B^{(\ell)}\) are equal to
\(e_1,\dots,e_{k-1}\), and if
\[
u_\ell:=B^{(\ell)}e_k,
\qquad
\rho_{\ell m}:=\langle u_\ell,u_m\rangle,
\]
then for all \(\ell\neq m\),
\[
c_\rho\varepsilon\le 1-\rho_{\ell m}^2\le C_\rho\varepsilon.
\]
\end{lemma}

\begin{proof}
This is the usual local Grassmann packing obtained by varying only the last
column. More explicitly, one may take
\[
u_\ell=\sqrt{1-\varepsilon}\,e_k+\sqrt{\varepsilon}
\begin{bmatrix}
0_k\\ b_\ell
\end{bmatrix},
\]
where \(\{b_\ell\}_{\ell=1}^M\subset \mathbb S^{d-k-1}\) is an exponential
packing with pairwise angular separation bounded away from zero. Then
\(\log M\asymp d-k\) and \(1-\rho_{\ell m}^2\asymp \varepsilon\) uniformly in
\(\ell\neq m\).
\end{proof}

\begin{proposition}[Fano reduction]\label{prop:gsi-lb-fano}
There exist sufficiently small absolute constants \(\kappa,c_0>0\) such that
the following holds. Consider hypotheses \(P_1,\dots,P_M\) indexed by the
packing in Lemma~\ref{lem:gsi-lb-packing}, chosen as the two-point packing
below when \(d-k\) is below a fixed absolute threshold. If
\[
\max_{\ell\neq m}\mathrm{KL}(P_\ell\|P_m)\le \kappa(d-k),
\]
then
\[
\inf_{\widehat V}\frac1M\sum_{\ell=1}^M
\Prob_{P_\ell}\!\left(
\sinth{\widehat V}{\mathrm{span}(B^{(\ell)})}
\ge c_0\sqrt{\varepsilon}
\right)
\ge \frac38.
\]
Consequently,
\[
\inf_{\widehat V}\frac1M\sum_{\ell=1}^M
\E_{P_\ell}\!\left[\sinth{\widehat V}{\mathrm{span}(B^{(\ell)})}\right]
\ge \frac{3c_0}{8}\sqrt{\varepsilon}.
\]
\end{proposition}

\begin{proof}
Write \(q:=d-k\). By Lemma~\ref{lem:gsi-lb-packing}, the hypotheses
are separated in operator sine-theta loss at scale \(\sqrt{\varepsilon}\).
Nearest-subspace decoding therefore turns any subspace estimator into a test
whose error probability is bounded above by the probability of subspace loss
at least \(c_0\sqrt{\varepsilon}\), for a sufficiently small absolute
constant \(c_0>0\).

Let \(c_{\mathrm{exp}}>0\) be an absolute constant such that the exponential
packing construction gives \(\log M\ge c_{\mathrm{exp}}q\). First suppose
\(q\ge q_0\), where \(q_0\) is a fixed absolute threshold large enough that
\(c_{\mathrm{exp}}q_0\ge4\log2\). Fano's inequality gives
\[
p_{\mathrm{err}}
\ge 1-\frac{\kappa q+\log2}{\log M}
\ge 1-\frac{\kappa}{c_{\mathrm{exp}}}
       -\frac{\log2}{c_{\mathrm{exp}}q}
\ge \frac12,
\]
provided \(\kappa\le c_{\mathrm{exp}}/4\).

For \(1\le q<q_0\), choose the two-point packing given by
\(b_1=v\) and \(b_2=-v\), with any \(v\in\mathbb S^{q-1}\), in the
construction of Lemma~\ref{lem:gsi-lb-packing}. Here
\[
1-\rho_{12}^2=4\varepsilon(1-\varepsilon)\asymp\varepsilon,
\]
and \(\log M=\log2\) is comparable to \(q\) up to absolute constants,
so this is an admissible packing in that lemma (after adjusting its absolute
constants if necessary). The equal-prior two-point testing bound and
Pinsker's inequality give
\[
p_{\mathrm{err}}
\ge \frac{1-\operatorname{TV}(P_1,P_2)}2
\ge \frac{1-\sqrt{\mathrm{KL}(P_1\|P_2)/2}}2
\ge \frac38,
\]
provided \(\kappa q_0\le1/8\). Choosing \(\kappa\) to satisfy both
restrictions proves the uniform probability bound.
The expectation bound follows from
\[
\E[L]\ge c_0\sqrt{\varepsilon}\,\Prob(L\ge c_0\sqrt{\varepsilon})
\]
with \(L=\sinth{\widehat V}{\mathrm{span}(B^{(\ell)})}\), followed by
averaging over \(\ell\).
\end{proof}

\subsubsection{Shared First-Term Reduction}
\label{subsubsec:shared-lb-first-reduction}

\begin{proposition}[Generic first-term reduction]
\label{prop:shared-lb-first-reduction}
Fix a parameter class \(\mathcal P(\lambda)\) consisting of tuples
\((V,\beta_1,\dots,\beta_T)\) with
\[
V\in O(d\times k),
\qquad
\beta_t\in \mathrm{span}(V)\cap B_2(1)
\quad (t\in[T]),
\]
and
\[
\lambda_k\!\left(\frac1T\sum_{t=1}^T \beta_t\beta_t^\top\right)\ge \lambda.
\]
Assume also that
\[
1\le k\le d/2.
\]
Assume there exists \(C_{\mathrm{pair}}>0\) such that, for every local packing
from Lemma~\ref{lem:gsi-lb-packing}, one can construct hypotheses
\(P_1,\dots,P_M\) supported on \(\mathcal P(\lambda)\) and satisfying
\[
\mathrm{KL}(P_\ell\|P_m)\le C_{\mathrm{pair}}\,Tn\lambda(1-\rho_{\ell m}^2)
\qquad (\ell\neq m).
\]
Then
\[
\inf_{\widehat V}
\sup_{(V,\beta_1,\dots,\beta_T)\in \mathcal P(\lambda)}
\E\!\left[\sinth{\widehat V}{V}\right]
\ge
c\left(\left[\sqrt{\frac{d}{Tn\lambda}}\right]\wedge 1\right),
\]
where \(c>0\) depends only on \(C_{\mathrm{pair}}\).
\end{proposition}

\begin{proof}
Fix
\[
\varepsilon:=c_\ast\min\!\left\{\frac12,\frac{d-k}{Tn\lambda}\right\},
\]
where \(c_\ast>0\) is a sufficiently small constant depending only on
\(C_{\mathrm{pair}}\). By assumption and Lemma~\ref{lem:gsi-lb-packing},
\[
\max_{\ell\neq m}\mathrm{KL}(P_\ell\|P_m)
\le
C_{\mathrm{pair}}\,Tn\lambda\,\varepsilon
\le
C_{\mathrm{pair}}c_\ast(d-k).
\]
Choosing \(c_\ast\) sufficiently small gives
\[
\max_{\ell\neq m}\mathrm{KL}(P_\ell\|P_m)\le \kappa(d-k),
\]
where \(\kappa\) is the constant from Proposition~\ref{prop:gsi-lb-fano}.
Hence Proposition~\ref{prop:gsi-lb-fano} yields
\[
\inf_{\widehat V}
\sup_{(V,\beta_1,\dots,\beta_T)\in \mathcal P(\lambda)}
\E\!\left[\sinth{\widehat V}{V}\right]
\ge
c\,\sqrt{\varepsilon}.
\]
Since \(k\le d/2\), we have \(d-k\asymp d\), which gives the stated bound.
\end{proof}

\subsubsection{Shared Second-Term Reduction}
\label{subsubsec:shared-lb-second-reduction}

\begin{proposition}[Shared revealed-sign setup]
\label{prop:shared-lb-revealed-setup}
Fix a local packing \(B^{(1)},\dots,B^{(M)}\) from
Lemma~\ref{lem:gsi-lb-packing}, write
\[
\mu:=2\lambda,
\qquad
\omega_t=(\xi_{t,1},\dots,\xi_{t,k-1},\zeta_t)\in\{\pm1\}^k,
\qquad
\alpha_t:=\sqrt{\mu}\,\omega_t,
\]
and define
\[
\beta_t^{(\ell)}:=B^{(\ell)}\alpha_t
\qquad (\ell\in[M],\,t\in[T]).
\]
For each \(\ell\in[M]\), let \(P_{\ell,\omega}\) be a conditional full-sample
law based on the parameter tuple
\((B^{(\ell)},\beta_1^{(\ell)},\dots,\beta_T^{(\ell)})\), let \(\Pi_\ell\) be
the joint law of \((\omega,\text{data})\), let \(P_\ell\) be the data marginal,
and let \(\widetilde P_\ell\) be the law of
\((\xi_1,\dots,\xi_T,\text{data})\), where
\[
\xi_t:=(\xi_{t,1},\dots,\xi_{t,k-1})\in\{\pm1\}^{k-1}.
\]
If
\[
n\ge C_0k,
\qquad
n\lambda\le c_0,
\qquad
c_0\le \frac{C_0}{2},
\]
then:
\begin{enumerate}
    \item For every \(\ell\in[M]\) and \(t\in[T]\),
    \[
    \|\beta_t^{(\ell)}\|_2^2
    =
    2k\lambda
    \le
    \frac{2c_0k}{n}
    \le 1.
    \]
    \item On the event
    \[
    A:=
    \left\{
    \lambda_{\min}\!\left(
    \frac1T\sum_{t=1}^T \omega_t\omega_t^\top
    \right)\ge \frac12
    \right\},
    \]
    every conditional tuple
    \((B^{(\ell)},\beta_1^{(\ell)},\dots,\beta_T^{(\ell)})\) belongs to
    \(\mathcal P(\lambda)\).
    \item For every \(\ell,m\in[M]\),
    \[
    \chi^2(P_\ell\|P_m)\le \chi^2(\widetilde P_\ell\|\widetilde P_m).
    \]
\end{enumerate}
\end{proposition}

\begin{proof}
The norm bound follows from
\[
\|\beta_t^{(\ell)}\|_2^2
=
\|\alpha_t\|_2^2
=
\mu\|\omega_t\|_2^2
=
2k\lambda.
\]
On the event \(A\),
\[
\frac1T\sum_{t=1}^T \beta_t^{(\ell)}(\beta_t^{(\ell)})^\top
=
\mu B^{(\ell)}
\left(\frac1T\sum_{t=1}^T \omega_t\omega_t^\top\right)
(B^{(\ell)})^\top
\succeq
\frac{\mu}{2}B^{(\ell)}(B^{(\ell)})^\top
=
\lambda B^{(\ell)}(B^{(\ell)})^\top.
\]
Thus the covariance constraint holds on \(A\), while part (1) gives
\(\beta_t^{(\ell)}\in B_2(1)\). The final claim follows from the
data-processing inequality by dropping the revealed signs:
\[
(\xi_1,\dots,\xi_T,\text{data})\mapsto\text{data}.
\]
\end{proof}

\begin{proposition}[Generic centered-reference \(L^2\) reduction]
\label{prop:shared-lb-centered-l2}
Let \(R\) be a probability measure, let \(q_v,q_w\in L^2(R)\) be densities
with respect to \(R\), and let \(\rho\in[-1,1]\). Suppose there exists an even
function \(B:[-1,1]\to\R\) such that
\[
\E_R[q_vq_w]=B(\rho),
\qquad
\E_R[q_v^2]=\E_R[q_w^2]=B(1),
\]
and
\[
\sup_{s\in[-1,1]}|B''(s)|\le \Gamma.
\]
Then
\[
\|q_v-q_w\|_{L^2(R)}^2\le \Gamma(1-\rho^2).
\]
\end{proposition}

\begin{proof}
Since
\[
\|q_v-q_w\|_{L^2(R)}^2
=
\E_R[q_v^2]+\E_R[q_w^2]-2\E_R[q_vq_w]
=
2(B(1)-B(\rho)),
\]
it suffices to bound \(B(1)-B(\rho)\). Because \(B\) is even, \(B'(0)=0\).
Hence for \(s\in[0,1]\),
\[
|B'(s)|
\le
\sup_{u\in[-1,1]}|B''(u)|\,s
\le
\Gamma s.
\]
Therefore
\[
B(1)-B(\rho)
=
\int_{|\rho|}^1 B'(s)\,ds
\le
\Gamma\int_{|\rho|}^1 s\,ds
=
\frac{\Gamma}{2}(1-\rho^2).
\]
Multiplying by \(2\) yields the claim.
\end{proof}

\begin{proposition}[Generic one-task \(\chi^2\) reduction]
\label{prop:shared-lb-onetask-chi2}
Let \(Q_v,Q_w\ll R\) be probability measures with densities
\[
q_v:=\frac{dQ_v}{dR},
\qquad
q_w:=\frac{dQ_w}{dR}.
\]
Assume that
\[
\|q_v-q_w\|_{L^2(R)}^2\le D,
\qquad
\E_R[(q_v-q_w)^4]\le M_4,
\qquad
\E_R[q_w^{-2}\mathbf 1_{\{q_w<1/2\}}]\le \tau.
\]
Then
\[
\chi^2(Q_v\|Q_w)\le 2D+\sqrt{M_4\tau}.
\]
\end{proposition}

\begin{proof}
Write
\[
\chi^2(Q_v\|Q_w)
=
\int \frac{(q_v-q_w)^2}{q_w}\,dR.
\]
Split over the good event \(\{q_w\ge 1/2\}\) and its complement. On
\(\{q_w\ge 1/2\}\),
\[
\int_{\{q_w\ge 1/2\}} \frac{(q_v-q_w)^2}{q_w}\,dR
\le
2\|q_v-q_w\|_{L^2(R)}^2
\le 2D.
\]
On \(\{q_w<1/2\}\), Cauchy--Schwarz gives
\[
\int_{\{q_w<1/2\}} \frac{(q_v-q_w)^2}{q_w}\,dR
\le
\left(\E_R[(q_v-q_w)^4]\right)^{1/2}
\left(\E_R[q_w^{-2}\mathbf 1_{\{q_w<1/2\}}]\right)^{1/2}
\le \sqrt{M_4\tau}.
\]
Combining the two pieces proves the result.
\end{proof}

\begin{proposition}[Generic \(T\)-task KL reduction]
\label{prop:shared-lb-second-kl-reduction}
Suppose that for two hypotheses \(P_\ell,P_m\) there exist auxiliary laws
\(\widetilde P_\ell,\widetilde P_m,\widetilde Q_\ell,\widetilde Q_m\) such that
\[
\widetilde P_\ell=\widetilde Q_\ell^{\otimes T},
\qquad
\widetilde P_m=\widetilde Q_m^{\otimes T},
\]
and
\[
\chi^2(P_\ell\|P_m)\le \chi^2(\widetilde P_\ell\|\widetilde P_m).
\]
If
\[
\chi^2(\widetilde Q_\ell\|\widetilde Q_m)\le \Delta_{\ell m},
\]
then
\[
\mathrm{KL}(P_\ell\|P_m)\le T\Delta_{\ell m}.
\]
\end{proposition}

\begin{proof}
By tensorization,
\[
1+\chi^2(\widetilde P_\ell\|\widetilde P_m)
=
\bigl(1+\chi^2(\widetilde Q_\ell\|\widetilde Q_m)\bigr)^T
\le
e^{T\Delta_{\ell m}}.
\]
By data processing,
\[
\chi^2(P_\ell\|P_m)\le \chi^2(\widetilde P_\ell\|\widetilde P_m).
\]
Finally,
\[
\mathrm{KL}(P_\ell\|P_m)
\le
\log\!\bigl(1+\chi^2(P_\ell\|P_m)\bigr)
\le
T\Delta_{\ell m}.
\]
\end{proof}

\begin{proposition}[Generic second-term reduction]
\label{prop:shared-lb-second-reduction}
Fix a parameter class \(\mathcal P(\lambda)\) consisting of tuples
\((V,\beta_1,\dots,\beta_T)\) with
\[
V\in O(d\times k),
\qquad
\beta_t\in \mathrm{span}(V)\cap B_2(1)
\quad (t\in[T]),
\]
and
\[
\lambda_k\!\left(\frac1T\sum_{t=1}^T \beta_t\beta_t^\top\right)\ge \lambda.
\]
Assume also that
\[
1\le k\le d/2.
\]
Assume:
\begin{enumerate}
    \item There exists \(c_{\mathrm{first}}>0\) such that
    \[
    \inf_{\widehat V}
    \sup_{(V,\beta_1,\dots,\beta_T)\in \mathcal P(\lambda)}
    \E\!\left[\sinth{\widehat V}{V}\right]
    \ge
    c_{\mathrm{first}}
    \left(\left[\sqrt{\frac{d}{Tn\lambda}}\right]\wedge 1\right).
    \]
    \item There are hypotheses \(P_1,\dots,P_M\) indexed by the packing in
    Lemma~\ref{lem:gsi-lb-packing}, joint extensions \(\Pi_\ell\), and an event
    \(A\) such that on \(A\) the conditional tuple belongs to
    \(\mathcal P(\lambda)\).
    \item There exist constants \(C_0,c_0,C_{\mathrm{pair}}>0\) such that
    \(n\ge C_0k\), \(n\lambda\le c_0\), and
    \[
    \mathrm{KL}(P_\ell\|P_m)
    \le
    C_{\mathrm{pair}}\,Tn^2\lambda^2(1-\rho_{\ell m}^2)+R
    \qquad (\ell\neq m),
    \]
    where
    \[
    R\le c_0(d-k).
    \]
    \item Whenever \(T\ge C_{\mathrm{cov}}k\), Lemma~\ref{lem:gsi-lb-rademacher-cov}
    applies to the event \(A\), so \(\Prob(A)\ge 3/4\).
\end{enumerate}
Then
\[
\inf_{\widehat V}
\sup_{(V,\beta_1,\dots,\beta_T)\in \mathcal P(\lambda)}
\E\!\left[\sinth{\widehat V}{V}\right]
\ge
c\left(\left[\sqrt{\frac{d}{Tn^2\lambda^2}}\right]\wedge 1\right),
\]
where \(c>0\) depends only on \(c_{\mathrm{first}},C_0,c_0,C_{\mathrm{pair}}\).
\end{proposition}

\begin{proof}
If \(T<C_{\mathrm{cov}}k\), then under \(n\lambda\le c_0\),
\[
\sqrt{\frac{d}{Tn^2\lambda^2}}
\ge
\sqrt{\frac{d}{Tc_0^2}}
\ge
\frac{1}{\sqrt{C_{\mathrm{cov}}c_0^2}}\sqrt{\frac{d}{k}}
\ge
\frac{1}{\sqrt{2C_{\mathrm{cov}}}\,c_0},
\]
since \(k\le d/2\). Thus, after shrinking \(c_0\) if necessary, the target
\[
\left[\sqrt{\frac{d}{Tn^2\lambda^2}}\right]\wedge 1
\]
is bounded below by an absolute constant.

On the other hand, the same assumptions give
\[
\sqrt{\frac{d}{Tn\lambda}}
\ge
\sqrt{\frac{d}{Tc_0}}
\ge
\frac{1}{\sqrt{C_{\mathrm{cov}}c_0}}\sqrt{\frac{d}{k}}
\ge
\frac{1}{\sqrt{2C_{\mathrm{cov}}c_0}},
\]
again because \(k\le d/2\). Therefore assumption (1) yields an absolute
constant lower bound on
\[
\inf_{\widehat V}
\sup_{(V,\beta_1,\dots,\beta_T)\in \mathcal P(\lambda)}
\E\!\left[\sinth{\widehat V}{V}\right].
\]
After adjusting constants, this proves the claimed lower bound in the regime
\(T<C_{\mathrm{cov}}k\). We may therefore assume from now on that
\[
T\ge C_{\mathrm{cov}}k,
\]
so that \(\Prob(A)\ge 3/4\).

Fix
\[
\varepsilon:=c_\ast\min\!\left\{\frac12,\frac{d-k}{Tn^2\lambda^2}\right\},
\]
where \(c_\ast>0\) is sufficiently small depending only on
\(C_{\mathrm{pair}}\). By assumption (3) and Lemma~\ref{lem:gsi-lb-packing},
\[
\max_{\ell\neq m}\mathrm{KL}(P_\ell\|P_m)
\le
C_{\mathrm{pair}}\,Tn^2\lambda^2\,\varepsilon + R
\le
C_{\mathrm{pair}}c_\ast(d-k)+c_0(d-k).
\]
Choosing \(c_\ast\) and then \(c_0\) sufficiently small gives
\[
\max_{\ell\neq m}\mathrm{KL}(P_\ell\|P_m)\le \kappa(d-k),
\]
where \(\kappa\) is the constant from Proposition~\ref{prop:gsi-lb-fano}.
Hence, for every estimator \(\widehat V\),
\[
\frac1M\sum_{\ell=1}^M
\Prob_{P_\ell}\!\left(
\sinth{\widehat V}{\mathrm{span}(B^{(\ell)})}
\ge c_3\sqrt{\varepsilon}
\right)
\ge \frac38
\]
for an absolute constant \(c_3>0\).

Fix an arbitrary estimator \(\widehat V\), and let
\[
L_\ell:=
\sinth{\widehat V}{\mathrm{span}(B^{(\ell)})},
\qquad
E_\ell:=\{L_\ell\ge c_3\sqrt{\varepsilon}\}.
\]
Because \(L_\ell\le 1\), while on the event \(A\) the conditional tuples lie in
\(\mathcal P(\lambda)\), we have
\[
\frac1M\sum_{\ell=1}^M \E_{\Pi_\ell}[L_\ell\mathbf 1_A]
\le
\Prob(A)\,
\sup_{(V,\beta_1,\dots,\beta_T)\in \mathcal P(\lambda)}
\E\!\left[\sinth{\widehat V}{V}\right].
\]
Also,
\[
\frac1M\sum_{\ell=1}^M \E_{\Pi_\ell}[L_\ell\mathbf 1_A]
\ge
c_3\sqrt{\varepsilon}\,
\frac1M\sum_{\ell=1}^M \Prob_{\Pi_\ell}(E_\ell\cap A).
\]
By the union bound,
\[
\Prob_{\Pi_\ell}(E_\ell\cap A)\ge \Prob_{P_\ell}(E_\ell)+\Prob(A)-1,
\]
and therefore
\[
\frac1M\sum_{\ell=1}^M \Prob_{\Pi_\ell}(E_\ell\cap A)
\ge
\frac38+\Prob(A)-1
\ge
\frac18.
\]
Combining the previous displays gives
\[
\inf_{\widehat V}
\sup_{(V,\beta_1,\dots,\beta_T)\in \mathcal P(\lambda)}
\E\!\left[\sinth{\widehat V}{V}\right]
\ge
c\,\sqrt{\varepsilon}.
\]
Since \(k\le d/2\), we have \(d-k\asymp d\), which gives the stated bound.
\end{proof}

\begin{lemma}[Regime comparison]
\label{lem:shared-lb-regime-comparison}
Fix \(c_0>0\). If \(n\lambda\le c_0\), then
\[
\sqrt{\frac{d}{Tn\lambda}}
\le
c_0^{1/2}\sqrt{\frac{d}{Tn^2\lambda^2}}.
\]
If \(n\lambda>c_0\), then
\[
\sqrt{\frac{d}{Tn^2\lambda^2}}
\le
c_0^{-1/2}\sqrt{\frac{d}{Tn\lambda}}.
\]
\end{lemma}

\begin{proof}
Both inequalities are immediate after multiplying through by
\(\sqrt{T}n\lambda/\sqrt{d}\).
\end{proof}

	\subsection{Classification Lower Bound}\label{subsec:classification-lb}

This subsection proves Theorem~\ref{thm:gsi-lower}. Section
~\ref{subsubsec:gsi-lb-setup} records the standing assumptions and the proof
reduction, Section~\ref{subsubsec:gsi-lb-first-term} verifies the
classification-specific first-term KL bound, Section
~\ref{subsubsec:gsi-lb-second-term} proves the second-term lower bound via the
shared revealed-sign scaffold and Bernoulli-specific overlap estimates, and
Section~\ref{subsubsec:gsi-lb-completion} combines the two regimes to finish
the theorem.

\subsubsection{Setup and Reduction}\label{subsubsec:gsi-lb-setup}

Write
\[
\lambda:=\lambda_k,
\qquad
h:=\mathrm{logit}\circ g.
\]
Throughout the appendix, we identify an orthonormal basis
\(V\in O(d\times k)\) with its column space \(\operatorname{span}(V)\) when
writing membership in \(\mathcal P(\lambda)\).
The theorem assumes that \(g\in C^2(\R)\), \(g:\R\to(0,1)\), and
\[
\sup_{t\in\R}\left|\frac{g'(t)}{g(t)(1-g(t))}\right|\le L_1,
\qquad
\sup_{t\in\R}\left|\frac{g''(t)}{g(t)(1-g(t))}\right|\le L_2.
\]
We also work throughout under the theorem-level dimensional assumption
\[
1\le k\le d/2.
\]
The theorem-level assumption that \(\mathcal P(\lambda)\) is
nonempty is also in force.

\begin{lemma}[Logit derivative identities]\label{lem:gsi-lb-logit-derivatives}
Let \(h:=\mathrm{logit}\circ g\). Then
\[
h'(t)=\frac{g'(t)}{g(t)(1-g(t))}
\]
and
\[
h''(t)=\frac{g''(t)}{g(t)(1-g(t))}-(1-2g(t))(h'(t))^2.
\]
Consequently,
\[
\|h'\|_\infty\le L_1,
\qquad
\|h''\|_\infty\le L_2+L_1^2.
\]
Conversely, if \(\|h'\|_\infty\le L_1\) and \(\|h''\|_\infty\le \widetilde L_2\),
\[
\sup_{t\in\R}\left|\frac{g'(t)}{g(t)(1-g(t))}\right|\le L_1,
\qquad
\sup_{t\in\R}\left|\frac{g''(t)}{g(t)(1-g(t))}\right|
\le \widetilde L_2+L_1^2.
\]
In particular, bounded \(h'\) and \(h''\) is equivalent to bounded
\(g'(t)/[g(t)(1-g(t))]\) and \(g''(t)/[g(t)(1-g(t))]\).
\end{lemma}

\begin{proof}
Differentiate \(h(t)=\log(g(t))-\log(1-g(t))\) to obtain the formula for
\(h'\). Differentiating once more gives the displayed identity for \(h''\).
The bounds then follow from \(|1-2g(t)|\le 1\).
\end{proof}

By Lemma~\ref{lem:gsi-lb-logit-derivatives}, throughout the proof we may use
the standing bounds
\[
\|h'\|_\infty\le L_1,
\qquad
\|h''\|_\infty\le L_2+L_1^2,
\qquad
n\ge C_0 k,
\qquad
T e^{-c_0 n}\le c_0 d.
\]
We shall prove the two auxiliary bounds
\[
\inf_{\widehat V}
\sup_{(V,\beta_1,\dots,\beta_T)\in \mathcal P(\lambda)}
\E\!\left[\sinth{\widehat V}{V}\right]
\ge
c\left(\left[\sqrt{\frac{d}{Tn\lambda}}\right]\wedge 1\right),
\]
using only bounded \(g'(t)/[g(t)(1-g(t))]\), and
\[
\inf_{\widehat V}
\sup_{(V,\beta_1,\dots,\beta_T)\in \mathcal P(\lambda)}
\E\!\left[\sinth{\widehat V}{V}\right]
\ge
c\left(\left[\sqrt{\frac{d}{Tn^2\lambda^2}}\right]\wedge 1\right),
\]
using in addition bounded \(g''(t)/[g(t)(1-g(t))]\) and \(n\ge C_0 k\). The
theorem then follows by a regime split in \(n\lambda\).

\paragraph{Proof roadmap.}
The shared appendix supplies the local packing, the operator sine-theta
Fano reduction, the Rademacher covariance event, and the generic first-term and
second-term reductions. The present appendix therefore only verifies the
Bernoulli/logit-specific KL and centered-reference \(\chi^2\) estimates needed
to invoke those shared reductions.

\subsubsection{First-Term Lower Bound}\label{subsubsec:gsi-lb-first-term}

\begin{lemma}[Bernoulli KL in natural parameters]\label{lem:gsi-lb-ber-kl}
For any \(p,q\in(0,1)\),
\[
\mathrm{KL}(\mathrm{Ber}(p)\|\mathrm{Ber}(q))
\le
\frac18\left(\log\frac{p}{1-p}-\log\frac{q}{1-q}\right)^2.
\]
\end{lemma}

\begin{proof}
Let
\[
\eta:=\log\frac{p}{1-p},
\qquad
\eta':=\log\frac{q}{1-q},
\qquad
\phi(t):=\log(1+e^t).
\]
Then
\[
\mathrm{KL}(\mathrm{Ber}(p)\|\mathrm{Ber}(q))
=
\phi(\eta')-\phi(\eta)-\phi'(\eta)(\eta'-\eta).
\]
By Taylor's theorem,
\[
\mathrm{KL}(\mathrm{Ber}(p)\|\mathrm{Ber}(q))
=
\frac12\phi''(\tilde\eta)(\eta-\eta')^2
\]
for some \(\tilde\eta\) between \(\eta\) and \(\eta'\). Since
\[
\phi''(t)=\frac{e^t}{(1+e^t)^2}\le \frac14,
\]
the result follows.
\end{proof}

\begin{proposition}[First-term pairwise KL bound]\label{prop:gsi-lb-first-kl}
Assume
\[
\sup_{t\in\R}\left|\frac{g'(t)}{g(t)(1-g(t))}\right|\le L_1.
\]
Then there exists \(C=C(L_1)\) such that for every \(\ell\neq m\),
\[
\mathrm{KL}(P_\ell\|P_m)\le C\,Tn\lambda(1-\rho_{\ell m}^2),
\]
for a family of hypotheses \(P_\ell\) supported on
\(\mathcal P(\lambda)\).
\end{proposition}

\begin{proof}
By the theorem-level nonemptiness assumption and
Lemma~\ref{lem:shared-lb-frame}, there exists a deterministic family
\(a_1,\dots,a_T\in\R^k\) satisfying
\[
\|a_t\|_2^2=k\quad (t\in[T]),
\qquad
\frac1T\sum_{t=1}^T a_ta_t^\top=I_k,
\]
and \(\lambda\le 1/k\). Under hypothesis \(\ell\), define
\[
\beta_t^{(\ell)}:=\sqrt{\lambda}\,B^{(\ell)}a_t
\qquad (t\in[T]).
\]
Then \(\beta_t^{(\ell)}\in \mathrm{span}(B^{(\ell)})\) for every \(t\), and
\[
\|\beta_t^{(\ell)}\|_2^2
=
\lambda\|a_t\|_2^2
=
\lambda k
\le 1.
\]
Therefore \(\beta_t^{(\ell)}\in \mathrm{span}(B^{(\ell)})\cap B_2(1)\), and
\[
\frac1T\sum_{t=1}^T \beta_t^{(\ell)}(\beta_t^{(\ell)})^\top
=
\lambda B^{(\ell)}\left(\frac1T\sum_{t=1}^T a_ta_t^\top\right)(B^{(\ell)})^\top
=
\lambda B^{(\ell)}(B^{(\ell)})^\top,
\]
so the hypotheses lie in \(\mathcal P(\lambda)\).

By independence across tasks and samples,
\[
\mathrm{KL}(P_\ell\|P_m)
=
n\sum_{t=1}^T
\E_X\!\left[
\mathrm{KL}\!\left(
\mathrm{Ber}(g(X^\top\beta_t^{(\ell)}))
\;\middle\|\;
\mathrm{Ber}(g(X^\top\beta_t^{(m)}))
\right)
\right],
\]
where \(X\sim N(0,I_d)\). By Lemma~\ref{lem:gsi-lb-ber-kl},
\[
\mathrm{KL}\!\left(
\mathrm{Ber}(g(X^\top\beta_t^{(\ell)}))
\;\middle\|\;
\mathrm{Ber}(g(X^\top\beta_t^{(m)}))
\right)
\le
\frac18\bigl(h(X^\top\beta_t^{(\ell)})-h(X^\top\beta_t^{(m)})\bigr)^2.
\]
Using \(\|h'\|_\infty\le L_1\),
\[
\bigl|h(X^\top\beta_t^{(\ell)})-h(X^\top\beta_t^{(m)})\bigr|
\le
L_1\bigl|X^\top(\beta_t^{(\ell)}-\beta_t^{(m)})\bigr|,
\]
and therefore
\[
\mathrm{KL}(P_\ell\|P_m)
\le
\frac{L_1^2}{8}\,n\sum_{t=1}^T
\E\bigl[(X^\top(\beta_t^{(\ell)}-\beta_t^{(m)}))^2\bigr].
\]
Since \(X\sim N(0,I_d)\),
\[
\E\bigl[(X^\top v)^2\bigr]=\|v\|_2^2
\qquad (v\in\R^d),
\]
so
\[
\mathrm{KL}(P_\ell\|P_m)
\le
\frac{L_1^2}{8}\,n\sum_{t=1}^T \|\beta_t^{(\ell)}-\beta_t^{(m)}\|_2^2.
\]
Because the first \(k-1\) columns of \(B^{(\ell)}\) and \(B^{(m)}\) agree,
\[
\beta_t^{(\ell)}-\beta_t^{(m)}
=
\sqrt{\lambda}\,a_{t,k}(u_\ell-u_m),
\]
hence
\[
\sum_{t=1}^T \|\beta_t^{(\ell)}-\beta_t^{(m)}\|_2^2
=
\lambda \|u_\ell-u_m\|_2^2\sum_{t=1}^T a_{t,k}^2.
\]
The frame identity gives \(T^{-1}\sum_{t=1}^T a_{t,k}^2=1\), while
\[
\|u_\ell-u_m\|_2^2=2(1-\rho_{\ell m})\le 2(1-\rho_{\ell m}^2).
\]
Combining the preceding displays yields the claim.
\end{proof}

\begin{proposition}[First-term lower bound]\label{prop:gsi-lb-first}
Assume
\[
1\le k\le d/2,
\qquad
\sup_{t\in\R}\left|\frac{g'(t)}{g(t)(1-g(t))}\right|\le L_1.
\]
Then
\[
\inf_{\widehat V}
\sup_{(V,\beta_1,\dots,\beta_T)\in \mathcal P(\lambda)}
\E\!\left[\sinth{\widehat V}{V}\right]
\ge
c\left(\left[\sqrt{\frac{d}{Tn\lambda}}\right]\wedge 1\right),
\]
where \(c>0\) depends only on \(L_1\).
\end{proposition}

\begin{proof}
Proposition~\ref{prop:gsi-lb-first-kl} verifies the pairwise
\(\mathrm{KL}\)-bound hypothesis in
Proposition~\ref{prop:shared-lb-first-reduction} for the class
\(\mathcal P(\lambda)\). Applying the shared reduction gives
the claim.
\end{proof}

\subsubsection{Second-Term Lower Bound}\label{subsubsec:gsi-lb-second-term}

We first instantiate the shared revealed-sign construction from
Section~\ref{subsec:shared-lb-construction}. It remains to verify the
Bernoulli/logit overlap, fourth-moment, and lower-tail ingredients for the
present model. The final one-task \(\chi^2\) and \(T\)-task KL reductions are
then supplied by the shared lemmas from that subsection.

For the second-term construction, let
\[
\mu:=2\lambda,
\]
\[
\omega_t=(\xi_{t,1},\dots,\xi_{t,k-1},\zeta_t)\in\{\pm1\}^k
\]
have i.i.d.\ Rademacher coordinates and set
\[
\alpha_t:=\sqrt{\mu}\,\omega_t,
\qquad
\beta_t^{(\ell)}:=B^{(\ell)}\alpha_t.
\]
For a sign realization \(\omega:=(\omega_1,\dots,\omega_T)\), let
\(P_{\ell,\omega}\) be the conditional law of the full sample under
hypothesis \(\ell\), and let
\[
P_\ell:=\E_\omega[P_{\ell,\omega}]
\]
be the corresponding mixture law after integrating out the latent signs. This
induces a joint law \(\Pi_\ell\) of \((\omega,\text{data})\), whose data
marginal is \(P_\ell\). This
is the same product-prior construction used throughout these lower-bound
appendices. By Proposition~\ref{prop:shared-lb-revealed-setup}, in the
weak-signal regime \(n\lambda\le c_0\) with \(n\ge C_0k\), the vectors
\(\beta_t^{(\ell)}\) lie in \(B_2(1)\), on the event
\[
A:=
\left\{
\lambda_{\min}\!\left(
\frac1T\sum_{t=1}^T \omega_t\omega_t^\top
\right)\ge \frac12
\right\}
\]
the conditional tuples belong to \(\mathcal P(\lambda)\), and
for every \(\ell,m\in[M]\),
\[
\chi^2(P_\ell\|P_m)\le \chi^2(\widetilde P_\ell\|\widetilde P_m).
\]

For each task \(t\), define the revealed vector
\[
\xi_t:=(\xi_{t,1},\dots,\xi_{t,k-1})\in\{\pm1\}^{k-1},
\]
and let \(\widetilde P_\ell\) be the joint law of
\((\xi_1,\dots,\xi_T,\text{data})\) under hypothesis \(\ell\).

Fix \(\ell,m\in[M]\) and let
\[
\rho_{\ell m}:=\langle u_\ell,u_m\rangle.
\]
For a revealed sign vector \(\xi\in\{\pm1\}^{k-1}\), define
\[
\beta_0(\xi):=\sqrt{\mu}\sum_{j=1}^{k-1}\xi_j e_j,
\]
\[
Q_{\ell,\xi}
:=
\frac12 \mathbb P_{\beta_0(\xi)+\sqrt{\mu}u_\ell}^{\otimes n}
+
\frac12 \mathbb P_{\beta_0(\xi)-\sqrt{\mu}u_\ell}^{\otimes n},
\]
\[
Q_{m,\xi}
:=
\frac12 \mathbb P_{\beta_0(\xi)+\sqrt{\mu}u_m}^{\otimes n}
+
\frac12 \mathbb P_{\beta_0(\xi)-\sqrt{\mu}u_m}^{\otimes n},
\qquad
R_\xi:=\mathbb P_{\beta_0(\xi)}^{\otimes n},
\]
where \(\mathbb P_\beta\) denotes the one-sample Bernoulli single-index law
with parameter \(\beta\). Let \(\pi\) be the uniform law on
\(\{\pm1\}^{k-1}\), and define
\[
\widetilde Q_\ell(d\xi,dz):=\pi(d\xi)\,Q_{\ell,\xi}(dz),
\qquad
\widetilde Q_m(d\xi,dz):=\pi(d\xi)\,Q_{m,\xi}(dz),
\]
\[
\widetilde R(d\xi,dz):=\pi(d\xi)\,R_\xi(dz).
\]
Then
\[
\widetilde P_\ell=\widetilde Q_\ell^{\otimes T},
\qquad
\widetilde P_m=\widetilde Q_m^{\otimes T}.
\]

\subsubsection*{Centered-reference kernel}

Fix a baseline \(\beta_0\in\R^d\) and perturbations \(\delta_1,\delta_2\in\R^d\).
Write
\[
u_0:=X^\top\beta_0,
\qquad
a:=X^\top\delta_1,
\qquad
b:=X^\top\delta_2,
\]
\[
r_a:=h(u_0+a)-h(u_0),
\qquad
r_b:=h(u_0+b)-h(u_0),
\qquad
\phi(t):=\log(1+e^t),
\]
and define
\[
F_h(u_0;a,b)
:=
\phi(h(u_0)+r_a+r_b)-\phi(h(u_0)+r_a)-\phi(h(u_0)+r_b)+\phi(h(u_0)),
\]
\[
K_h(u_0;a,b):=\exp(F_h(u_0;a,b)).
\]

\begin{lemma}[Centered-reference overlap identity]\label{lem:gsi-lb-overlap}
For one observation, let
\[
\ell_\delta(X,Y):=\frac{d\mathbb P_{\beta_0+\delta}}{d\mathbb P_{\beta_0}}(X,Y).
\]
Then
\[
\E_{\beta_0}[\ell_{\delta_1}\ell_{\delta_2}\mid X]=K_h(u_0;a,b).
\]
\end{lemma}

\begin{proof}
The Bernoulli likelihood in natural-parameter form gives
\[
\ell_\delta(X,Y)
=
\exp\!\Big(
Y[h(X^\top(\beta_0+\delta))-h(X^\top\beta_0)]
-[\phi(h(X^\top(\beta_0+\delta)))-\phi(h(X^\top\beta_0))]
\Big).
\]
Multiplying the two likelihood ratios and taking conditional expectation under
\(\mathbb P_{\beta_0}\) yields the displayed formula.
\end{proof}

\begin{lemma}[Derivative identities]\label{lem:gsi-lb-kernel-derivatives}
\[
\partial_aF_h
=
h'(u_0+a)\Big[\phi'(h(u_0)+r_a+r_b)-\phi'(h(u_0)+r_a)\Big],
\]
\[
\partial_bF_h
=
h'(u_0+b)\Big[\phi'(h(u_0)+r_a+r_b)-\phi'(h(u_0)+r_b)\Big],
\]
and
\[
\partial_{ab}F_h
=
h'(u_0+a)h'(u_0+b)\phi''(h(u_0)+r_a+r_b).
\]
\end{lemma}

\begin{proof}
Differentiate the definition of \(F_h\) directly.
\end{proof}

\begin{proposition}[Uniform log-kernel bounds]\label{prop:gsi-lb-F}
There exists \(C=C(L_1,L_2)\) such that for all \(u_0,a,b\in\R\),
\[
|\partial_aF_h|\le C|b|,
\qquad
|\partial_bF_h|\le C|a|,
\qquad
|\partial_{ab}F_h|\le C,
\qquad
|F_h|\le C|ab|,
\]
and moreover
\[
|\partial_{aab}F_h|+|\partial_{abb}F_h|+|\partial_{abab}F_h|\le C.
\]
\end{proposition}

\begin{proof}
By Lemma~\ref{lem:gsi-lb-kernel-derivatives} and
\(\sup_t \phi''(t)\le 1/4\),
\[
|\partial_aF_h|
\le
\|h'\|_\infty \sup|\phi''|\,|r_b|
\le
C|b|,
\]
since \(|r_b|\le \|h'\|_\infty |b|\). The same argument gives
\(|\partial_bF_h|\le C|a|\), and
\[
|\partial_{ab}F_h|
\le
\|h'\|_\infty^2\sup|\phi''|
\le
C.
\]
Because \(F_h(u_0;0,b)=0\),
\[
|F_h(u_0;a,b)|
\le
\int_0^{|a|} |\partial_aF_h(u_0;s,b)|\,ds
\le
C|ab|.
\]
Differentiating the explicit formula for \(\partial_{ab}F_h\) once in \(a\)
and once in \(b\) shows that every higher mixed derivative is a finite sum of
terms built from bounded factors among \(h',h'',\phi'',\phi''',\phi''''\),
which yields the displayed bounds.
\end{proof}

\begin{proposition}[Kernel derivative bounds]\label{prop:gsi-lb-K}
There exists \(C=C(L_1,L_2)\) such that
\[
|\partial_{ab}K_h(u_0;a,b)|
\le
C\,e^{C(a^2+b^2)}(1+a^2+b^2),
\]
\[
|\partial_{abab}K_h(u_0;a,b)|
\le
C\,e^{C(a^2+b^2)}(1+a^4+b^4).
\]
\end{proposition}

\begin{proof}
Since \(K_h=e^{F_h}\),
\[
\partial_{ab}K_h
=
K_h\big(\partial_{ab}F_h+(\partial_aF_h)(\partial_bF_h)\big).
\]
Using Proposition~\ref{prop:gsi-lb-F},
\[
|\partial_{ab}F_h+(\partial_aF_h)(\partial_bF_h)|
\le
C(1+|ab|)
\le
C(1+a^2+b^2),
\]
and the bound
\[
K_h=e^{F_h}\le e^{C|ab|}\le e^{C(a^2+b^2)}.
\]
This proves the first display.

For the fourth mixed derivative, set
\[
G:=\partial_{ab}F_h+(\partial_aF_h)(\partial_bF_h),
\]
so that
\[
\partial_{ab}K_h=K_hG.
\]
Differentiating first in \(a\) and then in \(b\) gives
\[
\partial_{abab}K_h
=
\partial_{ab}(K_hG)
=
(\partial_{ab}K_h)G+(\partial_aK_h)(\partial_bG)+(\partial_bK_h)(\partial_aG)+K_h\partial_{ab}G.
\]
Since
\[
\partial_aK_h=K_h\partial_aF_h,
\qquad
\partial_bK_h=K_h\partial_bF_h,
\qquad
\partial_{ab}K_h=K_hG,
\]
we may rewrite this as
\[
\partial_{abab}K_h
=
K_h\Big(G^2+(\partial_aF_h)(\partial_bG)+(\partial_bF_h)(\partial_aG)+\partial_{ab}G\Big).
\]
We now bound each factor.

First, Proposition~\ref{prop:gsi-lb-F} gives
\[
|G|
\le
|\partial_{ab}F_h|+|\partial_aF_h||\partial_bF_h|
\le
C(1+|ab|)
\le
C(1+a^2+b^2).
\]
Hence
\[
|G|^2\le C(1+a^4+b^4).
\]

Next,
\[
\partial_aG
=
\partial_{aab}F_h+(\partial_{aa}F_h)(\partial_bF_h)+(\partial_aF_h)(\partial_{ab}F_h),
\]
\[
\partial_bG
=
\partial_{abb}F_h+(\partial_{ab}F_h)(\partial_bF_h)+(\partial_aF_h)(\partial_{bb}F_h).
\]
By Proposition~\ref{prop:gsi-lb-F},
\[
|\partial_{aab}F_h|+|\partial_{abb}F_h|+|\partial_{ab}F_h|\le C,
\qquad
|\partial_aF_h|\le C|b|,
\qquad
|\partial_bF_h|\le C|a|.
\]
Moreover, because \(F_h(u_0;a,0)=0\) for every \(a\), we have
\(\partial_{aa}F_h(u_0;a,0)=0\). The bound
\(|\partial_{aab}F_h|\le C\), followed by integration in \(b\), yields
\[
|\partial_{aa}F_h(u_0;a,b)|
\le C|b|.
\]
Similarly, using \(\partial_{bb}F_h(u_0;0,b)=0\) and
\(|\partial_{abb}F_h|\le C\),
\[
|\partial_{bb}F_h(u_0;a,b)|
\le C|a|.
\]
Substituting these bounds gives
\[
|\partial_aG|+|\partial_bG|
\le
C(1+a^2+b^2).
\]
Therefore
\[
|(\partial_aF_h)(\partial_bG)|+|(\partial_bF_h)(\partial_aG)|
\le
C(|a|+|b|)(1+a^2+b^2)
\le
C(1+a^4+b^4).
\]

Finally,
\begin{align*}
\partial_{ab}G
&=
\partial_{abab}F_h
+(\partial_{aab}F_h)(\partial_bF_h)
+(\partial_{aa}F_h)(\partial_{bb}F_h) \\
&\qquad
+(\partial_{ab}F_h)^2
+(\partial_aF_h)(\partial_{abb}F_h).
\end{align*}
Using again Proposition~\ref{prop:gsi-lb-F} together with
\[
|\partial_{aa}F_h|\le C|b|,
\qquad
|\partial_{bb}F_h|\le C|a|,
\qquad
|\partial_aF_h|\le C|b|,
\qquad
|\partial_bF_h|\le C|a|,
\]
we obtain
\[
|\partial_{ab}G|
\le
C(1+|ab|)
\le
C(1+a^2+b^2)
\le
C(1+a^4+b^4).
\]

Combining the preceding bounds, every term inside the large parentheses is
bounded by \(C(1+a^4+b^4)\). Multiplying by
\[
K_h=e^{F_h}\le e^{C(a^2+b^2)}
\]
yields
\[
\bigl|\partial_{abab}K_h(u_0;a,b)\bigr|
\le
C\,e^{C(a^2+b^2)}(1+a^4+b^4),
\]
as claimed.
\end{proof}

\begin{lemma}[Uniform Gaussian exponential moments]\label{lem:gsi-lb-gauss-exp}
For every fixed \(t>0\) there exist constants \(c_t,C_t>0\) such that if
\((A,B)\sim N(0,\Sigma(c))\) with \(|c|\le \mu\) and \(0<\mu\le c_t\),
then
\[
\E[e^{t(A^2+B^2)}]\le C_t.
\]
\end{lemma}

\begin{proof}
For \(Z=(A,B)^\top\sim N(0,\Sigma(c))\),
\[
\E[e^{tZ^\top Z}]=\det(I-2t\Sigma(c))^{-1/2},
\]
whenever \(I-2t\Sigma(c)\) is positive definite. Since
\(\lambda_{\max}(\Sigma(c))\le 2\mu\) for \(|c|\le \mu\), we get
\[
\det(I-2t\Sigma(c))\ge (1-4t\mu)^2,
\]
as soon as \(4t\mu<1\), which proves the claim.
\end{proof}

Now assume \(v,w\in\mathbb S^{d-1}\) satisfy
\[
\langle \beta_0,v\rangle=\langle \beta_0,w\rangle=0,
\]
define
\[
\delta_v:=\sqrt{\mu}\,v,
\qquad
\delta_w:=\sqrt{\mu}\,w,
\]
and let
\[
U_0:=X^\top\beta_0,
\qquad
A:=X^\top\delta_v,
\qquad
B:=X^\top\delta_w.
\]
Then \(U_0\perp(A,B)\), and
\[
(A,B)\sim N\!\left(
0,
\begin{pmatrix}
\mu & c\\
c & \mu
\end{pmatrix}
\right)
\]
for some \(|c|\le \mu\). Define
\[
m(c):=\E[K_h(U_0;A,B)].
\]

\begin{proposition}[Overlap-function bounds]\label{prop:gsi-lb-m}
There exist constants \(\lambda_0,C>0\), depending only on \(L_1,L_2\), such
that for all \(0<\mu\le \lambda_0\) and all \(|c|\le \mu\),
\[
|m(c)-1|\le C\mu,
\qquad
|m'(c)|\le C,
\qquad
|m''(c)|\le C.
\]
The constants are uniform in the baseline \(\beta_0\).
\end{proposition}

\begin{proof}
By Proposition~\ref{prop:gsi-lb-F},
\[
|K_h-1|\le |F_h|e^{|F_h|}\le C|AB|e^{C|AB|}.
\]
Since \(|AB|\le \tfrac12(A^2+B^2)\), Lemma~\ref{lem:gsi-lb-gauss-exp} yields
\[
|m(c)-1|
\le
C\,\E[|AB|e^{C|AB|}]
\le
C\mu
\]
for sufficiently small \(\mu\).

Condition on \(U_0\). By Lemma~\ref{lem:shared-lb-gaussian-cov} applied to
\[
f_{u_0}(a,b):=K_h(u_0;a,b),
\]
and then averaged over \(U_0\),
\[
m'(c)=\E[\partial_{ab}K_h(U_0;A,B)],
\qquad
m''(c)=\E[\partial_{aabb}K_h(U_0;A,B)]
=\E[\partial_{abab}K_h(U_0;A,B)].
\]
Proposition~\ref{prop:gsi-lb-K} and
Lemma~\ref{lem:gsi-lb-gauss-exp} give integrable envelopes for both
derivatives, uniformly over \(|c|\le \mu\), so
\(|m'(c)|\le C\) and \(|m''(c)|\le C\) for \(|c|<\mu\). The same bounds at
\(c=\pm\mu\) follow by continuity.
\end{proof}

\begin{proposition}[Centered-reference \(L^2\) bound]\label{prop:gsi-lb-centered}
There exist constants \(c_1,c_2,C>0\), depending only on \(L_1,L_2\), such
that the following holds. Fix \(n\ge 1\), a baseline \(\beta_0\in\R^d\), and
unit vectors \(v,w\in\mathbb S^{d-1}\) with
\[
\langle \beta_0,v\rangle=\langle \beta_0,w\rangle=0.
\]
Let
\[
\rho:=\langle v,w\rangle,
\qquad
R:=\mathbb P_{\beta_0}^{\otimes n},
\]
\[
Q_v:=\frac12 \mathbb P_{\beta_0+\sqrt{\mu}v}^{\otimes n}
+
\frac12 \mathbb P_{\beta_0-\sqrt{\mu}v}^{\otimes n},
\qquad
Q_w:=\frac12 \mathbb P_{\beta_0+\sqrt{\mu}w}^{\otimes n}
+
\frac12 \mathbb P_{\beta_0-\sqrt{\mu}w}^{\otimes n}.
\]
If
\[
0<\mu\le c_1,
\qquad
n\mu\le c_2,
\]
then
\[
\left\|
\frac{dQ_v}{dR}-\frac{dQ_w}{dR}
\right\|_{L^2(R)}^2
\le
C n^2\mu^2(1-\rho^2).
\]
\end{proposition}

\begin{proof}
For one observation, Lemma~\ref{lem:gsi-lb-overlap} yields
\[
\E_R[L_{\delta_1}L_{\delta_2}]
=
\bigl(\E[K_h(U_0;A,B)]\bigr)^n.
\]
Taking \(\delta_1=s\sqrt{\mu}v\) and \(\delta_2=s'\sqrt{\mu}w\) with
\(s,s'\in\{\pm1\}\), we obtain
\[
\E_R[L_{s\sqrt{\mu}v}L_{s'\sqrt{\mu}w}]=m(ss'\mu\rho)^n.
\]
Hence, with
\[
B(\rho):=
\frac12\bigl(m(\mu\rho)^n+m(-\mu\rho)^n\bigr),
\]
we have
\[
\E_R\!\left[\frac{dQ_v}{dR}\frac{dQ_w}{dR}\right]=B(\rho),
\qquad
\E_R\!\left[\left(\frac{dQ_v}{dR}\right)^2\right]
=
\E_R\!\left[\left(\frac{dQ_w}{dR}\right)^2\right]
=
B(1).
\]
The function \(B\) is even. By Proposition~\ref{prop:gsi-lb-m}, if
\[
g(\rho):=m(\mu\rho),
\]
then
\[
|g'(\rho)|\le C\mu,
\qquad
|g''(\rho)|\le C\mu^2,
\qquad
g(\rho)\le 1+C\mu.
\]
If \(n\mu\le c_2\) with \(c_2\) sufficiently small, then
\[
g(\rho)^n\le (1+C\mu)^n\le e^{Cn\mu}\le C,
\]
and differentiating \(B(\rho)=\frac12(g(\rho)^n+g(-\rho)^n)\) twice gives
\[
\sup_{\rho\in[-1,1]}|B''(\rho)|\le C n^2\mu^2.
\]
Applying Proposition~\ref{prop:shared-lb-centered-l2} proves the claim.
\end{proof}

\begin{lemma}[Averaged \(L^2\) bound]\label{lem:gsi-lb-revealed-l2}
There exist constants \(c_1,c_2,C>0\) such that, whenever
\[
0<\mu\le c_1,
\qquad
n\mu\le c_2,
\]
one has
\[
\left\|
\frac{d\widetilde Q_\ell}{d\widetilde R}
-
\frac{d\widetilde Q_m}{d\widetilde R}
\right\|_{L^2(\widetilde R)}^2
\le
C n^2\mu^2(1-\rho_{\ell m}^2).
\]
\end{lemma}

\begin{proof}
For fixed \(\xi\), write
\[
q_{\ell,\xi}:=\frac{dQ_{\ell,\xi}}{dR_\xi},
\qquad
q_{m,\xi}:=\frac{dQ_{m,\xi}}{dR_\xi}.
\]
By construction,
\[
\beta_0(\xi)\in \mathrm{span}(e_1,\dots,e_{k-1}),
\qquad
u_\ell,u_m\in \mathrm{span}(e_1,\dots,e_{k-1})^\perp,
\]
so
\[
\langle \beta_0(\xi),u_\ell\rangle
=
\langle \beta_0(\xi),u_m\rangle
=
0.
\]
Thus Proposition~\ref{prop:gsi-lb-centered} applies with
\[
\beta_0=\beta_0(\xi),
\qquad
v=u_\ell,
\qquad
w=u_m,
\]
and yields
\[
\|q_{\ell,\xi}-q_{m,\xi}\|_{L^2(R_\xi)}^2
\le
C n^2\mu^2(1-\rho_{\ell m}^2)
\]
uniformly in \(\xi\). Averaging over \(\xi\sim \pi\) gives the result.
\end{proof}

\begin{lemma}[Fourth moment]\label{lem:gsi-lb-fourth}
There exist constants \(c_1,c_2,C>0\) such that, whenever
\[
0<\mu\le c_1,
\qquad
n\mu\le c_2,
\]
one has uniformly in \(\xi\),
\[
\E_{R_\xi}[q_{\ell,\xi}^4]+\E_{R_\xi}[q_{m,\xi}^4]\le C e^{Cn\mu},
\]
and consequently
\[
\E_{R_\xi}[(q_{\ell,\xi}-q_{m,\xi})^4]\le C e^{Cn\mu}.
\]
\end{lemma}

\begin{proof}
We treat \(q_{m,\xi}\) only. Set
\[
\delta:=\sqrt{\mu}\,u_m,
\qquad
L_\delta:=\frac{d\mathbb P_{\beta_0(\xi)+\delta}^{\otimes n}}{dR_\xi},
\qquad
L_{-\delta}:=\frac{d\mathbb P_{\beta_0(\xi)-\delta}^{\otimes n}}{dR_\xi}.
\]
Then
\[
q_{m,\xi}=\frac12(L_\delta+L_{-\delta}).
\]
For one sample, write
\[
u_0:=X^\top\beta_0(\xi),
\qquad
a:=X^\top\delta,
\qquad
t:=h(u_0),
\qquad
r:=h(u_0+a)-h(u_0).
\]
The one-sample likelihood ratio is
\[
\ell_\delta(X,Y)
=
\exp\!\Big(
Yr-[\phi(t+r)-\phi(t)]
\Big).
\]
Hence
\[
\E_{R_\xi}[\ell_\delta^4\mid X]
=
\exp\!\Big(\phi(t+4r)-4\phi(t+r)+3\phi(t)\Big).
\]
Since \(f(s):=\phi(t+s)\) has \(0\le f''\le 1/4\), a second-order Taylor
bound gives
\[
f(4r)-4f(r)+3f(0)\le Cr^2.
\]
As \(|r|\le \|h'\|_\infty |a|\le L_1|a|\),
\[
\E_{R_\xi}[\ell_\delta^4\mid X]\le e^{Ca^2}.
\]
Because \(a\sim N(0,\mu)\),
\[
\E[e^{Ca^2}]=(1-2C\mu)^{-1/2}\le e^{C'\mu}
\]
for \(\mu\) small. Independence over the \(n\) samples gives
\[
\E_{R_\xi}[L_\delta^4]\le e^{Cn\mu}.
\]
The same holds for \(L_{-\delta}\), so
\[
\E_{R_\xi}[q_{m,\xi}^4]\le C e^{Cn\mu}.
\]
The last claim follows from \((a-b)^4\le 8(a^4+b^4)\).
\end{proof}

\begin{lemma}[Lower tail]\label{lem:gsi-lb-lower-tail}
There exist constants \(c_1,c_2,c,C>0\) such that, whenever
\[
0<\mu\le c_1,
\qquad
n\mu\le c_2,
\]
one has uniformly in \(\xi\),
\[
\E_{R_\xi}\!\left[
q_{m,\xi}^{-2}\mathbf 1_{\{q_{m,\xi}<1/2\}}
\right]
\le
e^{-c/\mu}.
\]
\end{lemma}

\begin{proof}
Fix \(\xi\) and let \(\delta:=\sqrt{\mu}\,u_m\). Write
\[
q_{m,\xi}=\frac12(L_\delta+L_{-\delta}).
\]
For one sample, define
\[
u_0:=X^\top\beta_0(\xi),
\qquad
a:=X^\top\delta,
\qquad
t:=h(u_0),
\]
\[
r_+:=h(u_0+a)-h(u_0),
\qquad
r_-:=h(u_0-a)-h(u_0).
\]
Then
\[
\ell_\delta\ell_{-\delta}
=
\exp\!\Big(
Y(r_++r_-)-H(t;r_+,r_-)
\Big),
\]
where
\[
H(t;r_+,r_-):=\phi(t+r_+)+\phi(t+r_-)-2\phi(t).
\]
By the second-order finite-difference bound for \(h\),
\[
|r_++r_-|
=
|h(u_0+a)+h(u_0-a)-2h(u_0)|
\le
\|h''\|_\infty a^2.
\]
Also,
\begin{align*}
H(t;r_+,r_-)
&=
\Big(\phi(t+r_+)-\phi(t)-\phi'(t)r_+\Big)
+
\Big(\phi(t+r_-)-\phi(t)-\phi'(t)r_-\Big) \\
&\qquad +\phi'(t)(r_++r_-).
\end{align*}
Since \(0\le \phi'(t)\le 1\), \(\sup\phi''\le 1/4\), and
\(|r_\pm|\le L_1|a|\), we obtain
\[
H(t;r_+,r_-)\le Ca^2.
\]
Therefore
\[
Y(r_++r_-)-H(t;r_+,r_-)\ge -Ca^2,
\]
uniformly over \(Y\in\{0,1\}\), and so
\[
\sqrt{\ell_\delta\ell_{-\delta}}\ge e^{-Ca^2}.
\]
By the arithmetic--geometric mean inequality,
\[
q_{m,\xi}\ge \sqrt{L_\delta L_{-\delta}}
\ge
\exp\!\Big(-C\sum_{i=1}^n A_i^2\Big),
\qquad
A_i:=X_i^\top\delta\sim N(0,\mu).
\]
Hence
\[
\{q_{m,\xi}<1/2\}
\subseteq
\left\{\sum_{i=1}^n A_i^2>\frac{\log 2}{C}\right\},
\]
and on this event
\[
q_{m,\xi}^{-2}\le \exp\!\Big(2C\sum_{i=1}^n A_i^2\Big).
\]
If
\[
S_n:=\sum_{i=1}^n A_i^2=\mu\chi_n^2,
\]
then
\[
\E[q_{m,\xi}^{-2}\mathbf 1_{\{q_{m,\xi}<1/2\}}]
\le
\E[e^{\eta S_n}\mathbf 1_{\{S_n>\kappa\}}],
\]
for suitable constants \(\eta,\kappa>0\). Choose \(\mu\) so small that
\(\eta\mu<1/4\), and let
\[
s:=\frac{1}{4\mu}-\eta>0.
\]
Then Markov's inequality gives
\[
\E[e^{\eta S_n}\mathbf 1_{\{S_n>\kappa\}}]
\le
e^{-s\kappa}\E[e^{(\eta+s)S_n}]
=
e^{-s\kappa}\E[e^{\chi_n^2/4}]
=
e^{-s\kappa}2^{n/2}.
\]
Since \(n\mu\le c_2\), we have \(n\le c_2/\mu\), hence
\[
2^{n/2}\le \exp\!\Big(\frac{c_2\log 2}{2\mu}\Big).
\]
Choosing \(c_2\) small enough yields the desired bound.
\end{proof}

\begin{proposition}[One-task \(\chi^2\) bound]\label{prop:gsi-lb-one-task}
There exist constants \(c_1,c_2,c,C>0\) such that, whenever
\[
0<\mu\le c_1,
\qquad
n\mu\le c_2,
\]
we have
\[
\chi^2(\widetilde Q_\ell\|\widetilde Q_m)
\le
C n^2\mu^2(1-\rho_{\ell m}^2)+e^{-c/\mu}.
\]
\end{proposition}

\begin{proof}
Fix \(\xi\), and write
\[
q_{\ell,\xi}:=\frac{dQ_{\ell,\xi}}{dR_\xi},
\qquad
q_{m,\xi}:=\frac{dQ_{m,\xi}}{dR_\xi}.
\]
By Proposition~\ref{prop:gsi-lb-centered}, applied conditionally on the fixed
value of \(\xi\),
\[
\|q_{\ell,\xi}-q_{m,\xi}\|_{L^2(R_\xi)}^2
\le
C n^2\mu^2(1-\rho_{\ell m}^2).
\]
By Lemma~\ref{lem:gsi-lb-fourth},
\[
\E_{R_\xi}[(q_{\ell,\xi}-q_{m,\xi})^4]\le C e^{Cn\mu},
\]
and by Lemma~\ref{lem:gsi-lb-lower-tail},
\[
\E_{R_\xi}\!\left[q_{m,\xi}^{-2}\mathbf 1_{\{q_{m,\xi}<1/2\}}\right]
\le
e^{-c/\mu}.
\]
Since \(n\mu\le c_2\), after reducing \(c_2\) if necessary we have
\[
\sqrt{C e^{Cn\mu}e^{-c/\mu}}\le e^{-c'/\mu}
\]
for some \(c'>0\). Proposition~\ref{prop:shared-lb-onetask-chi2} therefore
gives
\[
\chi^2(Q_{\ell,\xi}\|Q_{m,\xi})
\le
C n^2\mu^2(1-\rho_{\ell m}^2)+e^{-c'/\mu}
\]
uniformly in \(\xi\). Averaging over \(\xi\sim\pi\) proves the claim.
\end{proof}

\begin{proposition}[\(T\)-task KL bound]\label{prop:gsi-lb-second-kl}
There exist constants \(c_1,c_2,c,C>0\) such that, whenever
\[
0<\mu\le c_1,
\qquad
n\mu\le c_2,
\]
we have
\[
\mathrm{KL}(P_\ell\|P_m)
\le
C T n^2\mu^2(1-\rho_{\ell m}^2)+T e^{-c/\mu}.
\]
\end{proposition}

\begin{proof}
Proposition~\ref{prop:gsi-lb-one-task} gives
\[
\chi^2(\widetilde Q_\ell\|\widetilde Q_m)
\le
C n^2\mu^2(1-\rho_{\ell m}^2)+e^{-c/\mu}.
\]
Applying Proposition~\ref{prop:shared-lb-second-kl-reduction} with
\[
\Delta_{\ell m}:=
C n^2\mu^2(1-\rho_{\ell m}^2)+e^{-c/\mu}
\]
proves the claim.
\end{proof}

\begin{proposition}[Second-term lower bound]\label{prop:gsi-lb-second}
Assume
\[
1\le k\le d/2,
\qquad
\sup_{t\in\R}\left|\frac{g'(t)}{g(t)(1-g(t))}\right|\le L_1,
\qquad
\sup_{t\in\R}\left|\frac{g''(t)}{g(t)(1-g(t))}\right|\le L_2.
\]
Then there exist constants \(c,c_0,C_0>0\), depending only on \(L_1,L_2\),
such that
\[
\inf_{\widehat V}
\sup_{(V,\beta_1,\dots,\beta_T)\in \mathcal P(\lambda)}
\E\!\left[\sinth{\widehat V}{V}\right]
\ge
c\left(\left[\sqrt{\frac{d}{Tn^2\lambda^2}}\right]\wedge 1\right),
\]
whenever
\[
n\ge C_0 k,
\qquad
n\lambda\le c_0,
\]
and
\[
T e^{-c_0 n}\le c_0 d.
\]
\end{proposition}

\begin{proof}
If \(n\lambda\le c_0\), then with \(\mu=2\lambda\) we have \(n\mu\le 2c_0\), so
after reducing \(c_0\) if necessary the assumptions of
Proposition~\ref{prop:gsi-lb-second-kl} hold. Also, after reducing \(c_0\)
again if necessary,
\[
e^{-c/\mu}\le e^{-c'n}
\]
for some constant \(c'>0\). Therefore the theorem-level condition
\[
T e^{-c_0 n}\le c_0 d
\]
implies, after adjusting constants,
\[
T e^{-c/\mu}\le c_0(d-k).
\]
The support event \(A\) from the revealed-sign construction satisfies the
hypotheses of Lemma~\ref{lem:gsi-lb-rademacher-cov}. Thus
Proposition~\ref{prop:shared-lb-second-reduction} applies to the class
\(\mathcal P(\lambda)\), using
Proposition~\ref{prop:gsi-lb-first} for the first-term input and
Proposition~\ref{prop:gsi-lb-second-kl} for the pairwise
\(\mathrm{KL}\)-bound. This yields the claim.
\end{proof}

\subsubsection{Completion of the Proof}\label{subsubsec:gsi-lb-completion}

\begin{proof}[Proof of Theorem~\ref{thm:gsi-lower}]
By Proposition~\ref{prop:gsi-lb-first}, there exists \(c_1>0\), depending only
on \(L_1\), such that
\[
\inf_{\widehat V}
\sup_{(V,\beta_1,\dots,\beta_T)\in \mathcal P(\lambda)}
\E\!\left[\sinth{\widehat V}{V}\right]
\ge
c_1\left(\left[\sqrt{\frac{d}{Tn\lambda}}\right]\wedge 1\right).
\]
By Proposition~\ref{prop:gsi-lb-second}, there exists \(c_2>0\), depending only
on \(L_1,L_2\), such that, whenever \(n\ge C_0 k\) and \(n\lambda\le c_2\),
\[
\inf_{\widehat V}
\sup_{(V,\beta_1,\dots,\beta_T)\in \mathcal P(\lambda)}
\E\!\left[\sinth{\widehat V}{V}\right]
\ge
c_2\left(\left[\sqrt{\frac{d}{Tn^2\lambda^2}}\right]\wedge 1\right),
\]
after reducing constants if necessary.

We now split into two regimes.

\medskip
\noindent\emph{Case 1: \(n\lambda\le c_2\).}
Since the standing assumptions include \(n\ge C_0 k\),
Proposition~\ref{prop:gsi-lb-second} applies in this regime. By
Lemma~\ref{lem:shared-lb-regime-comparison},
\[
\sqrt{\frac{d}{Tn\lambda}}
\le c_2^{1/2}\sqrt{\frac{d}{Tn^2\lambda^2}},
\]
so the second term dominates the first up to an absolute constant. Hence
Proposition~\ref{prop:gsi-lb-second} yields
\[
\inf_{\widehat V}
\sup_{(V,\beta_1,\dots,\beta_T)\in \mathcal P(\lambda)}
\E\!\left[\sinth{\widehat V}{V}\right]
\ge
c\left(
\left[
\sqrt{\frac{d}{Tn\lambda}}
+
\sqrt{\frac{d}{Tn^2\lambda^2}}
\right]\wedge 1
\right).
\]

\medskip
\noindent\emph{Case 2: \(n\lambda>c_2\).}
By Lemma~\ref{lem:shared-lb-regime-comparison},
\[
\sqrt{\frac{d}{Tn^2\lambda^2}}
\le c_2^{-1/2}\sqrt{\frac{d}{Tn\lambda}},
\]
so the first term dominates the second up to an absolute constant. Therefore
Proposition~\ref{prop:gsi-lb-first} gives
\[
\inf_{\widehat V}
\sup_{(V,\beta_1,\dots,\beta_T)\in \mathcal P(\lambda)}
\E\!\left[\sinth{\widehat V}{V}\right]
\ge
c\left(
\left[
\sqrt{\frac{d}{Tn\lambda}}
+
\sqrt{\frac{d}{Tn^2\lambda^2}}
\right]\wedge 1
\right).
\]
This proves the theorem.
\end{proof}

	\subsection{Additive Gaussian Lower Bound}\label{subsec:additive-lb}

This subsection proves Theorem~\ref{thm:gsi-additive-lower}. Section
~\ref{subsubsec:agsi-lb-setup} records the standing assumptions and the proof
reduction, Section~\ref{subsubsec:agsi-lb-first-term} verifies the
additive-Gaussian first-term KL bound, Section
~\ref{subsubsec:agsi-lb-second-term} proves the second-term lower bound via
the shared revealed-sign scaffold and Gaussian overlap calculations, and
Section~\ref{subsubsec:agsi-lb-completion} combines the two regimes to finish
the theorem.

\subsubsection{Setup and Reduction}\label{subsubsec:agsi-lb-setup}

Write
\[
\lambda:=\lambda_k.
\]
Throughout this appendix, we identify an orthonormal basis
\(V\in O(d\times k)\) with its column space \(\operatorname{span}(V)\) when
writing membership in \(\mathcal P(\lambda)\).
The theorem assumes that \(g\in C^2(\R)\) satisfies
\eqref{eq:bounded-derivatives}. We also assume
\[
n\ge C_0 k,
\qquad
T e^{-c_0\sqrt n}\le c_0 d.
\]
We also work throughout under the theorem-level dimensional assumption
\[
1\le k\le d/2.
\]

The geometric part of the proof is identical to the classification lower bound.
We therefore reuse the local packing in Lemma~\ref{lem:gsi-lb-packing}, the
Fano reduction in Proposition~\ref{prop:gsi-lb-fano}, and the Rademacher
covariance event from Lemma~\ref{lem:gsi-lb-rademacher-cov}.

We shall prove two auxiliary bounds:
\[
\inf_{\widehat V}
\sup_{(V,\beta_1,\dots,\beta_T)\in \mathcal P(\lambda)}
\E\!\left[\sinth{\widehat V}{V}\right]
\ge
c\left(\left[\sqrt{\frac{d}{Tn\lambda}}\right]\wedge 1\right),
\]
using only \(\|g'\|_\infty\le L_1\), and
\[
\inf_{\widehat V}
\sup_{(V,\beta_1,\dots,\beta_T)\in \mathcal P(\lambda)}
\E\!\left[\sinth{\widehat V}{V}\right]
\ge
c\left(\left[\sqrt{\frac{d}{Tn^2\lambda^2}}\right]\wedge 1\right),
\]
in the weak-signal regime \(n\lambda\le c_1\), using in addition
\(\|g''\|_\infty\le L_2\). The theorem then follows by a regime split in
\(n\lambda\), exactly as in the classification lower-bound appendix.

\paragraph{Proof roadmap.}
The shared appendix supplies the packing/Fano geometry, the frame reduction, the
Rademacher covariance event, the generic first-term reduction, and the generic
second-term reduction. The present appendix therefore only verifies the
Gaussian-KL and centered-reference \(\chi^2\) calculations specific to the
additive-response model.

\subsubsection{First-Term Lower Bound}\label{subsubsec:agsi-lb-first-term}

\begin{lemma}[One-sample Gaussian KL]\label{lem:agsi-lb-gaussian-kl}
For every \(\beta,\beta'\in\R^d\), under the additive Gaussian model
\[
Y=g(X^\top\beta)+\xi,
\qquad
X\sim N(0,I_d),
\qquad
\xi\sim N(0,\sigma^2),
\qquad
\xi\perp X,
\]
we have
\[
\mathrm{KL}(P_\beta\|P_{\beta'})
=
\frac{1}{2\sigma^2}
\E\!\left[
\bigl(g(X^\top\beta)-g(X^\top\beta')\bigr)^2
\right].
\]
\end{lemma}

\begin{proof}
Conditional on \(X\), the two response laws are Gaussian with common variance
\(\sigma^2\) and means \(g(X^\top\beta)\) and \(g(X^\top\beta')\). The usual
Gaussian location-family formula gives
\[
\mathrm{KL}(P_\beta(\,\cdot\,|X)\|P_{\beta'}(\,\cdot\,|X))
=
\frac{1}{2\sigma^2}
\bigl(g(X^\top\beta)-g(X^\top\beta')\bigr)^2.
\]
Taking expectation over \(X\) proves the claim.
\end{proof}

\begin{proposition}[First-term pairwise KL bound]\label{prop:agsi-lb-first-kl}
There exists \(C=C(L_1,\sigma)\) such that for every \(\ell\neq m\),
\[
\mathrm{KL}(P_\ell\|P_m)\le C\,Tn\lambda(1-\rho_{\ell m}^2),
\]
for a family of hypotheses \(P_\ell\) supported on
\(\mathcal P(\lambda)\).
\end{proposition}

\begin{proof}
By the standing nonemptiness assumption and
Lemma~\ref{lem:shared-lb-frame}, there exist
vectors \(a_1,\dots,a_T\in\R^k\) such that
\[
\|a_t\|_2^2=k\quad (t\in[T]),
\qquad
\frac1T\sum_{t=1}^T a_ta_t^\top=I_k,
\qquad
\lambda\le \frac1k.
\]
Under hypothesis \(\ell\), define
\[
\beta_t^{(\ell)}:=\sqrt{\lambda}\,B^{(\ell)}a_t
\qquad (t\in[T]).
\]
Then \(\beta_t^{(\ell)}\in \operatorname{span}(B^{(\ell)})\) for all \(t\), and
\[
\|\beta_t^{(\ell)}\|_2^2
=
\lambda\|a_t\|_2^2
=
\lambda k
\le 1,
\]
so \(\beta_t^{(\ell)}\in \operatorname{span}(B^{(\ell)})\cap B_2(1)\). Also,
\[
\frac1T\sum_{t=1}^T \beta_t^{(\ell)}(\beta_t^{(\ell)})^\top
=
\lambda B^{(\ell)}
\left(\frac1T\sum_{t=1}^T a_ta_t^\top\right)
(B^{(\ell)})^\top
=
\lambda B^{(\ell)}(B^{(\ell)})^\top,
\]
so the hypotheses lie in \(\mathcal P(\lambda)\).

By independence across tasks and samples and
Lemma~\ref{lem:agsi-lb-gaussian-kl},
\[
\mathrm{KL}(P_\ell\|P_m)
=
\frac{n}{2\sigma^2}\sum_{t=1}^T
\E\!\left[
\bigl(
g(X^\top\beta_t^{(\ell)})-g(X^\top\beta_t^{(m)})
\bigr)^2
\right].
\]
Using \(\|g'\|_\infty\le L_1\),
\[
\bigl|
g(X^\top\beta_t^{(\ell)})-g(X^\top\beta_t^{(m)})
\bigr|
\le
L_1|X^\top(\beta_t^{(\ell)}-\beta_t^{(m)})|,
\]
and therefore
\[
\mathrm{KL}(P_\ell\|P_m)
\le
\frac{L_1^2 n}{2\sigma^2}
\sum_{t=1}^T \|\beta_t^{(\ell)}-\beta_t^{(m)}\|_2^2.
\]
Because the first \(k-1\) columns of \(B^{(\ell)}\) and \(B^{(m)}\) agree,
\[
\beta_t^{(\ell)}-\beta_t^{(m)}
=
\sqrt{\lambda}\,a_{t,k}(u_\ell-u_m),
\]
hence
\[
\sum_{t=1}^T \|\beta_t^{(\ell)}-\beta_t^{(m)}\|_2^2
=
\lambda\|u_\ell-u_m\|_2^2\sum_{t=1}^T a_{t,k}^2.
\]
The frame identity gives \(T^{-1}\sum_{t=1}^T a_{t,k}^2=1\), while
\[
\|u_\ell-u_m\|_2^2
=
2(1-\rho_{\ell m})
\le
2(1-\rho_{\ell m}^2).
\]
Combining the preceding displays proves the claim.
\end{proof}

\begin{proposition}[First-term lower bound]\label{prop:agsi-lb-first}
Assume
\[
1\le k\le d/2.
\]
Then there exists \(c>0\), depending only on \(L_1\) and \(\sigma\), such that
\[
\inf_{\widehat V}
\sup_{(V,\beta_1,\dots,\beta_T)\in \mathcal P(\lambda)}
\E\!\left[\sinth{\widehat V}{V}\right]
\ge
c\left(\left[\sqrt{\frac{d}{Tn\lambda}}\right]\wedge 1\right).
\]
\end{proposition}

\begin{proof}
Proposition~\ref{prop:agsi-lb-first-kl} verifies the pairwise
\(\mathrm{KL}\)-bound hypothesis in
Proposition~\ref{prop:shared-lb-first-reduction} for the class
\(\mathcal P(\lambda)\). Applying the shared reduction gives
the claim.
\end{proof}

\subsubsection{Second-Term Lower Bound}\label{subsubsec:agsi-lb-second-term}

We first instantiate the shared revealed-sign construction from
Section~\ref{subsec:shared-lb-construction}. It remains to verify the
Gaussian overlap, fourth-moment, and lower-tail ingredients for the present
model. The final one-task \(\chi^2\) and \(T\)-task KL reductions are then
supplied by the shared lemmas from that subsection.

For the weak-signal second-term construction, let
\[
\mu:=2\lambda,
\qquad
\omega_t=(\xi_{t,1},\dots,\xi_{t,k-1},\zeta_t)\in\{\pm1\}^k
\]
have i.i.d.\ Rademacher coordinates and set
\[
\alpha_t:=\sqrt{\mu}\,\omega_t,
\qquad
\beta_t^{(\ell)}:=B^{(\ell)}\alpha_t.
\]
For a sign realization \(\omega:=(\omega_1,\dots,\omega_T)\), let
\(P_{\ell,\omega}\) be the conditional law of the full sample under hypothesis
\(\ell\), and let
\[
P_\ell:=\E_\omega[P_{\ell,\omega}]
\]
be the corresponding mixture law after integrating out the latent signs. This
induces a joint law \(\Pi_\ell\) of \((\omega,\text{data})\), whose data
marginal is \(P_\ell\). This
is the same product-prior construction used throughout these lower-bound
appendices. By Proposition~\ref{prop:shared-lb-revealed-setup}, in the
weak-signal regime \(n\lambda\le c_1\) with \(n\ge C_0k\), the vectors
\(\beta_t^{(\ell)}\) lie in \(B_2(1)\), on the event
\[
A:=
\left\{
\lambda_{\min}\!\left(
\frac1T\sum_{t=1}^T \omega_t\omega_t^\top
\right)\ge \frac12
\right\}
\]
the conditional tuples belong to \(\mathcal P(\lambda)\), and
for every \(\ell,m\in[M]\),
\[
\chi^2(P_\ell\|P_m)\le \chi^2(\widetilde P_\ell\|\widetilde P_m).
\]

For each task \(t\), define the revealed vector
\[
\xi_t:=(\xi_{t,1},\dots,\xi_{t,k-1})\in\{\pm1\}^{k-1},
\]
and let \(\widetilde P_\ell\) be the joint law of
\((\xi_1,\dots,\xi_T,\text{data})\) under hypothesis \(\ell\).

Fix \(\ell,m\in[M]\) and write
\[
\rho_{\ell m}:=\langle u_\ell,u_m\rangle.
\]
For a revealed sign vector \(\xi\in\{\pm1\}^{k-1}\), define
\begin{equation}\label{eq:agsi-lb-beta0-xi}
\beta_0(\xi):=\sqrt{\mu}\sum_{j=1}^{k-1}\xi_j e_j,
\end{equation}
\begin{equation}\label{eq:agsi-lb-centered-laws}
\begin{aligned}
Q_{\ell,\xi}
&:=
\frac12 P_{\beta_0(\xi)+\sqrt{\mu}u_\ell}^{\otimes n}
+
\frac12 P_{\beta_0(\xi)-\sqrt{\mu}u_\ell}^{\otimes n},\\
Q_{m,\xi}
&:=
\frac12 P_{\beta_0(\xi)+\sqrt{\mu}u_m}^{\otimes n}
+
\frac12 P_{\beta_0(\xi)-\sqrt{\mu}u_m}^{\otimes n},
\qquad
R_\xi:=P_{\beta_0(\xi)}^{\otimes n},
\end{aligned}
\end{equation}
and let \(\pi\) be the uniform law on \(\{\pm1\}^{k-1}\). Define
\[
\widetilde Q_\ell(d\xi,dz):=\pi(d\xi)\,Q_{\ell,\xi}(dz),
\qquad
\widetilde Q_m(d\xi,dz):=\pi(d\xi)\,Q_{m,\xi}(dz).
\]
Then
\[
\widetilde P_\ell=\widetilde Q_\ell^{\otimes T},
\qquad
\widetilde P_m=\widetilde Q_m^{\otimes T}.
\]

\subsubsection*{Centered-reference Gaussian comparison}

\begin{lemma}[Conditional Gaussian likelihood ratio]
\label{lem:agsi-lb-conditional-lr}
Fix a baseline \(\beta_0\in\R^d\) and a perturbation \(\delta\in\R^d\). Under
the additive Gaussian model, let \(R_{\beta_0}\) denote the one-sample law with
parameter \(\beta_0\), and \(P_{\beta_0+\delta}\) the law with parameter
\(\beta_0+\delta\). Then
\[
\ell_\delta(X,Y)
:=
\frac{dP_{\beta_0+\delta}}{dR_{\beta_0}}(X,Y)
=
\exp\!\left(
\frac{Y-g(X^\top\beta_0)}{\sigma^2}\,\Delta_{\beta_0,\delta}(X)
-\frac{\Delta_{\beta_0,\delta}(X)^2}{2\sigma^2}
\right),
\]
where
\[
\Delta_{\beta_0,\delta}(X)
:=
g(X^\top(\beta_0+\delta))-g(X^\top\beta_0).
\]
Moreover, for two perturbations \(\delta_1,\delta_2\),
\[
\E_{R_{\beta_0}}[\ell_{\delta_1}\ell_{\delta_2}\mid X]
=
\exp\!\left(
\frac{\Delta_{\beta_0,\delta_1}(X)\Delta_{\beta_0,\delta_2}(X)}{\sigma^2}
\right).
\]
\end{lemma}

\begin{proof}
Conditional on \(X\), under \(R_{\beta_0}\),
\[
Y\mid X\sim N(g(X^\top\beta_0),\sigma^2),
\]
while under \(P_{\beta_0+\delta}\),
\[
Y\mid X\sim N(g(X^\top(\beta_0+\delta)),\sigma^2).
\]
Writing
\[
m_0:=g(X^\top\beta_0),
\qquad
m_1:=g(X^\top(\beta_0+\delta))
=
m_0+\Delta_{\beta_0,\delta}(X),
\]
the conditional density ratio is
\[
\frac{\exp(-(Y-m_1)^2/(2\sigma^2))}{\exp(-(Y-m_0)^2/(2\sigma^2))}
=
\exp\!\left(
\frac{(Y-m_0)(m_1-m_0)}{\sigma^2}
-\frac{(m_1-m_0)^2}{2\sigma^2}
\right),
\]
which is the first identity.

For the second, write \(\Delta_j:=\Delta_{\beta_0,\delta_j}(X)\). Then
\[
\ell_{\delta_1}\ell_{\delta_2}
=
\exp\!\left(
\frac{Y-m_0}{\sigma^2}(\Delta_1+\Delta_2)
-\frac{\Delta_1^2+\Delta_2^2}{2\sigma^2}
\right).
\]
Since \(Y-m_0\mid X\sim N(0,\sigma^2)\), its moment generating function gives
\[
\E\!\left[
\exp\!\left(\frac{Y-m_0}{\sigma^2}(\Delta_1+\Delta_2)\right)\middle|X
\right]
=
\exp\!\left(\frac{(\Delta_1+\Delta_2)^2}{2\sigma^2}\right).
\]
Multiplying by the deterministic factor
\(\exp(-(\Delta_1^2+\Delta_2^2)/(2\sigma^2))\) yields
\[
\E_{R_{\beta_0}}[\ell_{\delta_1}\ell_{\delta_2}\mid X]
=
\exp\!\left(\frac{\Delta_1\Delta_2}{\sigma^2}\right),
\]
as claimed.
\end{proof}

\begin{corollary}[Conditional \(n\)-sample overlap]
\label{cor:agsi-lb-conditional-overlap}
With the notation from Lemma~\ref{lem:agsi-lb-conditional-lr}, define
\[
M_{\beta_0}(\delta_1,\delta_2)
:=
\E_{R_{\beta_0}}\!\left[
\exp\!\left(
\frac{\Delta_{\beta_0,\delta_1}(X)\Delta_{\beta_0,\delta_2}(X)}{\sigma^2}
\right)
\right].
\]
If
\[
L_\delta:=\frac{dP_{\beta_0+\delta}^{\otimes n}}{dR_{\beta_0}^{\otimes n}},
\]
then
\[
\E_{R_{\beta_0}^{\otimes n}}[L_{\delta_1}L_{\delta_2}]
=
\bigl(M_{\beta_0}(\delta_1,\delta_2)\bigr)^n.
\]
In the revealed-sign construction above, we apply this with baseline
\(\beta_0(\xi)\) from \eqref{eq:agsi-lb-beta0-xi} and centered laws
\(Q_{\ell,\xi},Q_{m,\xi},R_\xi\) from \eqref{eq:agsi-lb-centered-laws}.
Consequently, if
\[
q_{\ell,\xi}:=\frac{dQ_{\ell,\xi}}{dR_\xi},
\qquad
q_{m,\xi}:=\frac{dQ_{m,\xi}}{dR_\xi},
\]
then
\[
\E_{R_\xi}[q_{\ell,\xi}q_{m,\xi}]
=
\frac14\sum_{s,t\in\{\pm1\}}
\bigl(M_{\beta_0(\xi)}(s\delta_\ell,t\delta_m)\bigr)^n,
\]
where
\[
\delta_\ell:=\sqrt{\mu}\,u_\ell,
\qquad
\delta_m:=\sqrt{\mu}\,u_m.
\]
\end{corollary}

\begin{proof}
Conditional on \(\xi\), the \(n\) samples are i.i.d.\ under each pure
component, so
\[
L_\delta=\prod_{i=1}^n \ell_\delta(X_i,Y_i).
\]
Applying Lemma~\ref{lem:agsi-lb-conditional-lr} independently over the \(n\)
samples gives
\[
\E[L_{\delta_1}L_{\delta_2}]
=
\prod_{i=1}^n \E[\ell_{\delta_1}(X_i,Y_i)\ell_{\delta_2}(X_i,Y_i)]
=
M_{\beta_0}(\delta_1,\delta_2)^n.
\]
For the specialization, set \(\beta_0=\beta_0(\xi)\) from
\eqref{eq:agsi-lb-beta0-xi}, take \(\delta_\ell=\sqrt{\mu}\,u_\ell\) and
\(\delta_m=\sqrt{\mu}\,u_m\), and use the centered laws in
\eqref{eq:agsi-lb-centered-laws}. The mixture identity then follows by expanding
\[
q_{\ell,\xi}=\frac12(L_{\delta_\ell}+L_{-\delta_\ell}),
\qquad
q_{m,\xi}=\frac12(L_{\delta_m}+L_{-\delta_m}).
\]
\end{proof}

\begin{lemma}[Uniform derivative bounds for the Gaussian log-kernel]
\label{lem:agsi-lb-F-bounds}
Define
\[
F_{s,t}(u;a,b):=\frac{\Delta_s(u,a)\Delta_t(u,b)}{\sigma^2}, \qquad
\Delta_s(u,a):=g(u+sa)-g(u).
\]
There exists \(C=C(L_1,L_2,\sigma)>0\) such that for all
\(u,a,b\in\R\) and all signs \(s,t\in\{\pm1\}\),
\[
|\partial_a F_{s,t}(u;a,b)|\le C|b|,
\qquad
|\partial_b F_{s,t}(u;a,b)|\le C|a|,
\]
\[
|\partial_{ab}F_{s,t}(u;a,b)|\le C,
\qquad
|F_{s,t}(u;a,b)|\le C|ab|,
\]
and moreover
\[
|\partial_{aab}F_{s,t}(u;a,b)|
+
|\partial_{abb}F_{s,t}(u;a,b)|
+
|\partial_{aabb}F_{s,t}(u;a,b)|
\le C.
\]

\end{lemma}

\begin{proof}
By the mean-value theorem,
\[
|\Delta_s(u,a)|\le L_1|a|,
\qquad
|\Delta_t(u,b)|\le L_1|b|.
\]
Hence
\[
|F_{s,t}(u;a,b)|\le \frac{L_1^2}{\sigma^2}|ab|.
\]
Also,
\[
\partial_a\Delta_s(u,a)=s\,g'(u+sa),
\qquad
\partial_{aa}\Delta_s(u,a)=g''(u+sa),
\]
and similarly in \(b\). Therefore
\[
\partial_aF_{s,t}(u;a,b)
=
\frac{\partial_a\Delta_s(u,a)\,\Delta_t(u,b)}{\sigma^2},
\]
so
\[
|\partial_aF_{s,t}(u;a,b)|\le \frac{L_1^2}{\sigma^2}|b|.
\]
The same argument gives the bound for \(\partial_bF_{s,t}\), while
\[
\partial_{ab}F_{s,t}(u;a,b)
=
\frac{\partial_a\Delta_s(u,a)\,\partial_b\Delta_t(u,b)}{\sigma^2}
\]
is uniformly bounded. The higher mixed derivatives are controlled in the same
way using \(\|g''\|_\infty\le L_2\).
\end{proof}

\begin{lemma}[Uniform derivative bounds for the Gaussian overlap kernel]
\label{lem:agsi-lb-K-bounds}
Define  \(K_{s,t}:=e^{F_{s,t}}\) then there exists \(C=C(L_1,L_2,\sigma)>0\) such that for all
\(u,a,b\in\R\) and all signs \(s,t\in\{\pm1\}\),
\[
|\partial_{ab}K_{s,t}(u;a,b)|
\le
C e^{C(a^2+b^2)}(1+a^2+b^2),
\]
\[
|\partial_{aabb}K_{s,t}(u;a,b)|
\le
C e^{C(a^2+b^2)}(1+a^4+b^4),
\]
\end{lemma}

\begin{proof}
Since \(K_{s,t}=e^{F_{s,t}}\),
\[
\partial_{ab}K_{s,t}
=
K_{s,t}\Big(\partial_{ab}F_{s,t}+(\partial_aF_{s,t})(\partial_bF_{s,t})\Big).
\]
Lemma~\ref{lem:agsi-lb-F-bounds} gives
\[
|\partial_{ab}F_{s,t}+(\partial_aF_{s,t})(\partial_bF_{s,t})|
\le
C(1+a^2+b^2),
\]
while
\[
K_{s,t}=e^{F_{s,t}}\le e^{C|ab|}\le e^{C(a^2+b^2)}.
\]
This proves the first display. The second follows by differentiating twice more
and using Lemma~\ref{lem:agsi-lb-F-bounds} again.
\end{proof}

\begin{proposition}[Centered-reference Gaussian \(L^2\) bound]
\label{prop:agsi-lb-centered}
There exist constants \(c_0,c_1,C>0\), depending only on
\(L_1,L_2,\sigma\), such that whenever
\[
0<\mu\le c_1,
\qquad
n\mu\le c_0,
\]
the following holds. Let \(\beta_0\in\R^d\) and unit vectors
\(v,w\in\mathbb S^{d-1}\) satisfy
\[
\langle \beta_0,v\rangle=\langle \beta_0,w\rangle=0,
\qquad
\rho:=\langle v,w\rangle.
\]
Set
\[
\delta_v:=\sqrt{\mu}\,v,
\qquad
\delta_w:=\sqrt{\mu}\,w,
\]
\[
R:=P_{\beta_0}^{\otimes n},
\]
\[
Q_v:=\frac12 P_{\beta_0+\delta_v}^{\otimes n}
+\frac12 P_{\beta_0-\delta_v}^{\otimes n},
\qquad
Q_w:=\frac12 P_{\beta_0+\delta_w}^{\otimes n}
+\frac12 P_{\beta_0-\delta_w}^{\otimes n},
\]
\[
q_v:=\frac{dQ_v}{dR},
\qquad
q_w:=\frac{dQ_w}{dR}.
\]
Then
\[
\|q_v-q_w\|_{L^2(R)}^2
\le
C n^2\mu^2(1-\rho^2).
\]
\end{proposition}

\begin{proof}
We first identify the one-task overlap function, then express the
\(n\)-sample overlap in terms of it, and finally apply the shared
\(L^2\)-reduction from Proposition~\ref{prop:shared-lb-centered-l2}.

\paragraph{Step 1: one-task overlap.}
For \(s,t\in\{\pm1\}\), define
\[
M_{s,t}^{(\beta_0)}(\rho)
:=
\E\!\left[
K_{s,t}(U_0;A_\rho,B_\rho)
\right],
\]
where \(U_0\sim N(0,\|\beta_0\|_2^2)\) is independent of
\((A_\rho,B_\rho)\), and \((A_\rho,B_\rho)\) is centered Gaussian with
variance \(\mu\) and covariance \(\mu\rho\). By
Lemma~\ref{lem:shared-lb-gaussian-cov} with \(c=\mu\rho\), the chain rule, and
Lemma~\ref{lem:agsi-lb-K-bounds}, for \(|\rho|<1\),
\[
|M_{s,t}^{(\beta_0)}(\rho)-1|\le C\mu,
\qquad
\left|\frac{d}{d\rho}M_{s,t}^{(\beta_0)}(\rho)\right|\le C\mu,
\qquad
\left|\frac{d^2}{d\rho^2}M_{s,t}^{(\beta_0)}(\rho)\right|\le C\mu^2,
\]
uniformly over \(s,t,\beta_0\) and \(|\rho|<1\), provided \(\mu\) is
sufficiently small. The bounds at \(\rho=\pm1\) follow by continuity: the
kernel bounds in Lemma~\ref{lem:agsi-lb-K-bounds} provide a uniform Gaussian
integrable envelope as \(\rho\to\pm1\).

\paragraph{Step 2: \(n\)-sample overlap.}
Define
\[
B_{\beta_0}(\rho)
:=
\frac14\sum_{s,t\in\{\pm1\}}
\bigl(M_{s,t}^{(\beta_0)}(\rho)\bigr)^n.
\]
By Corollary~\ref{cor:agsi-lb-conditional-overlap},
\[
\E_R[q_vq_w]=B_{\beta_0}(\rho),
\qquad
\E_R[q_v^2]=\E_R[q_w^2]=B_{\beta_0}(1).
\]
The function \(B_{\beta_0}\) is even in \(\rho\). Using the product rule
together with the derivative bounds above, and using \(n\mu\le c_0\) to keep
\(M_{s,t}^{(\beta_0)}(\rho)\) uniformly bounded, we obtain
\[
\frac{d^2}{d\rho^2}\bigl(M_{s,t}^{(\beta_0)}(\rho)\bigr)^n
=
n\bigl(M_{s,t}^{(\beta_0)}(\rho)\bigr)^{n-1}
\frac{d^2}{d\rho^2}M_{s,t}^{(\beta_0)}(\rho)
\;+\;
n(n-1)\bigl(M_{s,t}^{(\beta_0)}(\rho)\bigr)^{n-2}
\left(\frac{d}{d\rho}M_{s,t}^{(\beta_0)}(\rho)\right)^2.
\]
Since \(|M_{s,t}^{(\beta_0)}(\rho)-1|\le C\mu\), the condition \(n\mu\le c_0\)
implies
\[
\sup_{\rho\in[-1,1]}\bigl|M_{s,t}^{(\beta_0)}(\rho)\bigr|^n\le C.
\]
Therefore
\[
\sup_{\rho\in[-1,1]}
\left|
\frac{d^2}{d\rho^2}\bigl(M_{s,t}^{(\beta_0)}(\rho)\bigr)^n
\right|
\le
Cn\mu^2+Cn^2\mu^2
\le
Cn^2\mu^2,
\]
and summing over \(s,t\in\{\pm1\}\) yields
\[
\sup_{\rho\in[-1,1]}|B_{\beta_0}''(\rho)|\le C n^2\mu^2.
\]

\paragraph{Step 3: apply the shared \(L^2\)-reduction.}
The identities
\[
\E_R[q_vq_w]=B_{\beta_0}(\rho),
\qquad
\E_R[q_v^2]=\E_R[q_w^2]=B_{\beta_0}(1),
\]
the evenness of \(B_{\beta_0}\), and the bound on \(B_{\beta_0}''\) verify
the hypotheses of Proposition~\ref{prop:shared-lb-centered-l2} with
\(\Gamma=Cn^2\mu^2\). Therefore
\[
\|q_v-q_w\|_{L^2(R)}^2
\le
Cn^2\mu^2(1-\rho^2),
\]
which is the claimed bound.
\end{proof}

\begin{lemma}[Fourth moments for centered Gaussian mixtures]
\label{lem:agsi-lb-fourth}
There exist constants \(c_0,C>0\), depending only on \(L_1\) and \(\sigma\),
such that the following holds. Let \(\beta_0\in\R^d\), let \(v\in\mathbb S^{d-1}\),
and set \(\delta:=\sqrt{\mu}\,v\). If
\[
0<\mu\le c_0,
\qquad
n\mu\le c_0,
\]
and
\[
R:=P_{\beta_0}^{\otimes n},
\qquad
Q:=\frac12 P_{\beta_0+\delta}^{\otimes n}
+\frac12 P_{\beta_0-\delta}^{\otimes n},
\qquad
q:=\frac{dQ}{dR},
\]
then
\[
\E_R[q^4]\le C.
\]
Consequently, if \(q_v\) and \(q_w\) are the centered-reference mixture
densities from Proposition~\ref{prop:agsi-lb-centered}, then
\[
\E_R[(q_v-q_w)^4]\le C.
\]
\end{lemma}

\begin{proof}
Write
\[
q=\frac12(L_++L_-),
\]
where \(L_\pm=dP_{\beta_0\pm\delta}^{\otimes n}/dR\). By Jensen,
\[
q^4\le \frac12(L_+^4+L_-^4).
\]
Thus it suffices to bound \(\E_R[L_+^4]\). For one sample, let
\[
\alpha(X):=
\frac{g(X^\top(\beta_0+\delta))-g(X^\top\beta_0)}{\sigma}.
\]
Under \(R\), the one-sample likelihood ratio is
\[
\ell_+(X,Y)
=
\exp\!\left(\alpha(X)G-\frac12\alpha(X)^2\right),
\]
where \(G=(Y-g(X^\top\beta_0))/\sigma\sim N(0,1)\) conditionally on \(X\). Thus
\[
\E_R[\ell_+^4\mid X]=\exp(6\alpha(X)^2).
\]
Using \(\|g'\|_\infty\le L_1\),
\[
|\alpha(X)|\le \frac{L_1}{\sigma}|X^\top\delta|.
\]
Hence
\[
\E_R[L_+^4]
\le
\E\!\left[
\exp\!\left(
C\sum_{i=1}^n (X_i^\top\delta)^2
\right)
\right].
\]
Since \(X_i^\top\delta\sim N(0,\mu)\) independently,
\[
\sum_{i=1}^n (X_i^\top\delta)^2=\mu\chi_n^2,
\]
so for \(\mu\) sufficiently small,
\[
\E_R[L_+^4]
\le
(1-C\mu)^{-n/2}
\le
e^{Cn\mu}
\le C
\]
under \(n\mu\le c_0\). The final claim follows from
\[
(q_v-q_w)^4\le 8(q_v^4+q_w^4).
\]
\end{proof}

\begin{lemma}[Lower tail with stretched-exponential remainder]
\label{lem:agsi-lb-lower-tail}
There exist constants \(c_0,c,C>0\), depending only on \(L_1,L_2,\sigma\),
such that the following holds. Let \(\beta_0\in\R^d\), let
\(\delta\in\R^d\) satisfy \(\|\delta\|_2^2=\mu\), and define
\[
R:=P_{\beta_0}^{\otimes n},
\qquad
Q:=\frac12 P_{\beta_0+\delta}^{\otimes n}
+\frac12 P_{\beta_0-\delta}^{\otimes n},
\qquad
q:=\frac{dQ}{dR}.
\]
If
\[
0<\mu\le c_0,
\qquad
n\mu\le c_0,
\]
then
\[
\E_R\!\left[q^{-2}\mathbf 1_{\{q<1/2\}}\right]
\le
C e^{-c/\sqrt{\mu}}.
\]
\end{lemma}

\begin{proof}
For \(i=1,\dots,n\), define
\[
\Delta_{i,+}:=g(X_i^\top(\beta_0+\delta))-g(X_i^\top\beta_0),
\qquad
\Delta_{i,-}:=g(X_i^\top(\beta_0-\delta))-g(X_i^\top\beta_0),
\]
\[
\alpha_i:=\frac{\Delta_{i,+}}{\sigma},
\qquad
\beta_i:=\frac{\Delta_{i,-}}{\sigma},
\qquad
A_i:=X_i^\top\delta.
\]
Also write
\[
U:=\sum_{i=1}^n \alpha_i G_i-\frac12\sum_{i=1}^n \alpha_i^2,
\qquad
V:=\sum_{i=1}^n \beta_i G_i-\frac12\sum_{i=1}^n \beta_i^2,
\]
where \(G_i=(Y_i-g(X_i^\top\beta_0))/\sigma\sim N(0,1)\) under \(R\). Then
\[
q=\frac12(e^U+e^V).
\]
Set
\[
S:=U+V.
\]
Since \(q=e^{S/2}\cosh((U-V)/2)\), we have
\[
q^{-2}\mathbf 1_{\{q<1/2\}}
\le
e^{-S}\mathbf 1_{\{S<-2\log 2\}}.
\]

Conditioning on \(X_1,\dots,X_n\), the variable \(S\) is Gaussian with mean
\[
m_X:=-\frac12\sum_{i=1}^n(\alpha_i^2+\beta_i^2)
\]
and variance
\[
v_X:=\sum_{i=1}^n(\alpha_i+\beta_i)^2.
\]
Using \(\|g'\|_\infty\le L_1\) and \(\|g''\|_\infty\le L_2\),
\[
|\alpha_i|+|\beta_i|\le C|A_i|,
\qquad
|\alpha_i+\beta_i|\le C\min\{|A_i|^2,|A_i|\}.
\]
Hence
\[
-m_X\le C\sum_{i=1}^n A_i^2,
\qquad
v_X\le C\sum_{i=1}^n \min\{A_i^4,A_i^2\}.
\]
Now define the good event
\[
G:=
\left\{\sum_{i=1}^n A_i^2\le K\right\}
\cap
\left\{\max_{1\le i\le n}|A_i|\le \mu^{1/4}\right\},
\]
where \(K>0\) is sufficiently small. On \(G\),
\[
v_X\le CK\sqrt{\mu},
\]
and therefore the Gaussian tail bound yields
\[
\E_R[q^{-2}\mathbf 1_{\{q<1/2\}}\mid X]\mathbf 1_G
\le
C e^{-c/\sqrt{\mu}}.
\]

On \(G^c\), we use the crude bound
\[
\E_R[q^{-2}\mathbf 1_{\{q<1/2\}}\mid X]
\le
\exp\!\left(C\sum_{i=1}^n A_i^2\right).
\]
Since \(\sum_i A_i^2=\mu\chi_n^2\), the exponential moment is bounded under
\(n\mu\le c_0\), after reducing \(c_0\) so that the same is true with \(2C\)
in place of \(C\). Also,
\[
\Prob\!\left(\sum_{i=1}^n A_i^2>K\right)\le e^{-c/\mu},
\]
and
\[
\Prob\!\left(\max_{1\le i\le n}|A_i|>\mu^{1/4}\right)
\le
n e^{-c/\sqrt{\mu}}
\le
e^{-c'/\sqrt{\mu}},
\]
since \(n\le c_0/\mu\). Therefore
\[
\Prob(G^c)\le C e^{-c/\sqrt{\mu}}.
\]
Consequently, Cauchy--Schwarz gives
\[
\E\!\left[
e^{C\sum_{i=1}^n A_i^2}\mathbf 1_{G^c}
\right]
\le
\left(\E e^{2C\sum_{i=1}^n A_i^2}\right)^{1/2}
\Prob(G^c)^{1/2}
\le
C e^{-c/\sqrt\mu}.
\]
Combining the good-event and bad-event contributions proves the claim.
\end{proof}

\begin{proposition}[One-task \(\chi^2\) bound]
\label{prop:agsi-lb-onetask}
There exist constants \(c_0,c,C>0\), depending only on \(L_1,L_2,\sigma\),
such that whenever
\[
0<\mu\le c_0,
\qquad
n\mu\le c_0,
\]
the centered-reference mixtures from Proposition~\ref{prop:agsi-lb-centered}
satisfy
\[
\chi^2(Q_v\|Q_w)
\le
C n^2\mu^2(1-\rho^2)+C e^{-c/\sqrt{\mu}}.
\]
\end{proposition}

\begin{proof}
By Proposition~\ref{prop:agsi-lb-centered},
\[
\|q_v-q_w\|_{L^2(R)}^2\le C n^2\mu^2(1-\rho^2).
\]
By Lemma~\ref{lem:agsi-lb-fourth},
\[
\E_R[(q_v-q_w)^4]\le C,
\]
after reducing \(c_0\) if necessary, and by
Lemma~\ref{lem:agsi-lb-lower-tail},
\[
\E_R\!\left[q_w^{-2}\mathbf 1_{\{q_w<1/2\}}\right]
\le
C e^{-c/\sqrt{\mu}}.
\]
Proposition~\ref{prop:shared-lb-onetask-chi2} therefore gives
\[
\chi^2(Q_v\|Q_w)
\le
C n^2\mu^2(1-\rho^2)+C e^{-c/\sqrt{\mu}}.
\]
This proves the claim.
\end{proof}

\begin{proposition}[\(T\)-task KL bound]\label{prop:agsi-lb-second-kl}
There exist constants \(c_1,c_2,c,C>0\) such that, whenever
\[
0<\mu\le c_1,
\qquad
n\mu\le c_2,
\]
we have
\[
\mathrm{KL}(P_\ell\|P_m)
\le
C T n^2\mu^2(1-\rho_{\ell m}^2)+T e^{-c/\sqrt{\mu}}.
\]
\end{proposition}

\begin{proof}
Fix \(\xi\). Proposition~\ref{prop:agsi-lb-onetask}, applied with baseline
\(\beta_0(\xi)\), gives uniformly in \(\xi\)
\[
\chi^2(Q_{\ell,\xi}\|Q_{m,\xi})
\le
C n^2\mu^2(1-\rho_{\ell m}^2)+e^{-c/\sqrt{\mu}}.
\]
Because \(\widetilde Q_\ell(d\xi,dz)=\pi(d\xi)Q_{\ell,\xi}(dz)\) and likewise
for \(m\),
\[
\chi^2(\widetilde Q_\ell\|\widetilde Q_m)
=
\E_{\xi\sim\pi}\!\left[
\chi^2(Q_{\ell,\xi}\|Q_{m,\xi})
\right]
\le
C n^2\mu^2(1-\rho_{\ell m}^2)+e^{-c/\sqrt{\mu}}.
\]
Applying Proposition~\ref{prop:shared-lb-second-kl-reduction} with
\[
\Delta_{\ell m}:=
C n^2\mu^2(1-\rho_{\ell m}^2)+e^{-c/\sqrt{\mu}}
\]
proves the claim.
\end{proof}

\begin{proposition}[Second-term lower bound]\label{prop:agsi-lb-second}
Assume
\[
1\le k\le d/2.
\]
Then there exist constants \(c,c_0,C_0>0\), depending only on
\(L_1,L_2,\sigma\), such that
\[
\inf_{\widehat V}
\sup_{(V,\beta_1,\dots,\beta_T)\in \mathcal P(\lambda)}
\E\!\left[\sinth{\widehat V}{V}\right]
\ge
c\left(\left[\sqrt{\frac{d}{Tn^2\lambda^2}}\right]\wedge 1\right),
\]
whenever
\[
n\ge C_0 k,
\qquad
n\lambda\le c_0,
\qquad
T e^{-c_0\sqrt n}\le c_0 d.
\]
\end{proposition}

\begin{proof}
If \(n\lambda\le c_0\), then with \(\mu=2\lambda\) we have \(n\mu\le 2c_0\), so
after reducing \(c_0\) if necessary the assumptions of
Proposition~\ref{prop:agsi-lb-second-kl} hold. Also, if \(n\lambda\le c_0\),
then
\[
e^{-c/\sqrt{\mu}}
\le
e^{-c'\sqrt n}
\]
for a suitably adjusted constant \(c'>0\). Therefore the theorem-level
condition
\[
T e^{-c_0\sqrt n}\le c_0 d
\]
implies, after adjusting constants,
\[
T e^{-c/\sqrt{\mu}}\le c_0(d-k).
\]
The support event \(A\) from the revealed-sign construction satisfies the
hypotheses of Lemma~\ref{lem:gsi-lb-rademacher-cov}. Thus
Proposition~\ref{prop:shared-lb-second-reduction} applies to the class
\(\mathcal P(\lambda)\), using
Proposition~\ref{prop:agsi-lb-first} for the first-term input and
Proposition~\ref{prop:agsi-lb-second-kl} for the pairwise
\(\mathrm{KL}\)-bound. This yields the claim.
\end{proof}

\subsubsection{Completion of the Proof}\label{subsubsec:agsi-lb-completion}

\begin{proof}[Proof of Theorem~\ref{thm:gsi-additive-lower}]
By Proposition~\ref{prop:agsi-lb-first}, there exists \(c_1>0\), depending
only on \(L_1\) and \(\sigma\), such that
\[
\inf_{\widehat V}
\sup_{(V,\beta_1,\dots,\beta_T)\in \mathcal P(\lambda)}
\E\!\left[\sinth{\widehat V}{V}\right]
\ge
c_1\left(\left[\sqrt{\frac{d}{Tn\lambda}}\right]\wedge 1\right).
\]
By Proposition~\ref{prop:agsi-lb-second}, there exists \(c_2>0\), depending
only on \(L_1,L_2,\sigma\), such that, whenever \(n\ge C_0k\),
\(n\lambda\le c_2\), and \(T e^{-c_2\sqrt n}\le c_2 d\),
\[
\inf_{\widehat V}
\sup_{(V,\beta_1,\dots,\beta_T)\in \mathcal P(\lambda)}
\E\!\left[\sinth{\widehat V}{V}\right]
\ge
c_2\left(\left[\sqrt{\frac{d}{Tn^2\lambda^2}}\right]\wedge 1\right).
\]

We now split into two regimes.

\medskip
\noindent\emph{Case 1: \(n\lambda\le c_2\).}
Since the standing assumptions include \(n\ge C_0k\) and
\(T e^{-c_0\sqrt n}\le c_0 d\), Proposition~\ref{prop:agsi-lb-second} applies
after adjusting constants. By Lemma~\ref{lem:shared-lb-regime-comparison},
\[
\sqrt{\frac{d}{Tn\lambda}}
\le c_2^{1/2}\sqrt{\frac{d}{Tn^2\lambda^2}},
\]
so the second term dominates the first up to an absolute constant. Hence
\[
\inf_{\widehat V}
\sup_{(V,\beta_1,\dots,\beta_T)\in \mathcal P(\lambda)}
\E\!\left[\sinth{\widehat V}{V}\right]
\ge
c\left(
\left[
\sqrt{\frac{d}{Tn\lambda}}
+
\sqrt{\frac{d}{Tn^2\lambda^2}}
\right]\wedge 1
\right).
\]

\medskip
\noindent\emph{Case 2: \(n\lambda>c_2\).}
By Lemma~\ref{lem:shared-lb-regime-comparison},
\[
\sqrt{\frac{d}{Tn^2\lambda^2}}
\le c_2^{-1/2}\sqrt{\frac{d}{Tn\lambda}},
\]
so the first term dominates the second up to an absolute constant. Therefore
Proposition~\ref{prop:agsi-lb-first} gives
\[
\inf_{\widehat V}
\sup_{(V,\beta_1,\dots,\beta_T)\in \mathcal P(\lambda)}
\E\!\left[\sinth{\widehat V}{V}\right]
\ge
c\left(
\left[
\sqrt{\frac{d}{Tn\lambda}}
+
\sqrt{\frac{d}{Tn^2\lambda^2}}
\right]\wedge 1
\right).
\]
This proves the theorem.
\end{proof}

	\subsection{Auxiliary Technical Results}
\label{subsec:shared-lb-technical-tools}

This subsection collects reusable technical lemmas that are invoked across the
lower-bound appendices. Section~\ref{subsubsec:shared-lb-frame} provides the
frame reduction for nonempty shared-subspace classes, Section
~\ref{subsubsec:shared-lb-rademacher-cov} gives the Rademacher covariance good
event used in the weak-signal construction, and
Section~\ref{subsubsec:shared-lb-gaussian-cov} states the Gaussian covariance
differentiation identity used in the overlap calculations.

\subsubsection{Frame Reduction for Shared-Subspace Classes}
\label{subsubsec:shared-lb-frame}

\begin{lemma}[Frame reduction for shared-subspace classes]
\label{lem:shared-lb-frame}
Assume there exists a tuple \((V,\beta_1,\dots,\beta_T)\) such that
\[
V\in O(d\times k),
\qquad
\beta_t\in \mathrm{span}(V)\cap B_2(1)
\quad (t\in[T]),
\]
and
\[
\lambda_k\!\left(\frac1T\sum_{t=1}^T \beta_t\beta_t^\top\right)\ge \lambda.
\]
Then
\[
T\ge k
\qquad\text{and}\qquad
\lambda\le \frac1k.
\]
Moreover, for every \(T\ge k\), there exist vectors \(a_1,\dots,a_T\in\R^k\)
such that
\[
\|a_t\|_2^2=k\quad (t\in[T]),
\qquad
\frac1T\sum_{t=1}^T a_ta_t^\top=I_k.
\]
\end{lemma}

\begin{proof}
Write
\[
\beta_t=V\alpha_t
\qquad (t\in[T])
\]
with \(\alpha_t\in\R^k\). Since \(V\) has orthonormal columns,
\[
\|\alpha_t\|_2=\|\beta_t\|_2\le 1
\qquad (t\in[T]).
\]
Set
\[
A:=
\frac1T\sum_{t=1}^T \alpha_t\alpha_t^\top.
\]
Then
\[
\frac1T\sum_{t=1}^T \beta_t\beta_t^\top
=
VAV^\top,
\]
so
\[
\lambda_k(A)=
\lambda_k\!\left(\frac1T\sum_{t=1}^T \beta_t\beta_t^\top\right)\ge \lambda.
\]
In particular, \(A\) has rank \(k\), which forces \(T\ge k\). Also,
\[
k\lambda
\le
\operatorname{tr}(A)
=
\frac1T\sum_{t=1}^T \|\alpha_t\|_2^2
\le 1,
\]
which yields \(\lambda\le 1/k\).

For the final claim, let \(\{u_t\}_{t=1}^T\subset \R^k\) be a finite unit norm
tight frame:
\[
\|u_t\|_2=1\quad (t\in[T]),
\qquad
\frac{k}{T}\sum_{t=1}^T u_tu_t^\top=I_k.
\]
Setting \(a_t:=\sqrt{k}\,u_t\) gives the required family.
\end{proof}

\subsubsection{Rademacher Covariance Good Event}
\label{subsubsec:shared-lb-rademacher-cov}

\begin{lemma}[Rademacher covariance good event]\label{lem:gsi-lb-rademacher-cov}
There exist absolute constants \(C_{\mathrm{cov}},c_{\mathrm{cov}}>0\) such
that the following holds. Let \(\omega_1,\dots,\omega_T\in\{\pm1\}^k\) be
independent vectors with i.i.d.\ Rademacher coordinates, and define
\[
\widehat \Sigma_\omega:=\frac1T\sum_{t=1}^T \omega_t\omega_t^\top.
\]
If \(T\ge C_{\mathrm{cov}}k\), then
\[
\Prob\!\left(
\left\|\widehat \Sigma_\omega-I_k\right\|_{\mathrm{op}}>\frac12
\right)
\le 2e^{-c_{\mathrm{cov}}T}.
\]
In particular, on the event
\[
A:=\left\{
\lambda_{\min}(\widehat \Sigma_\omega)\ge \frac12
\right\},
\]
we have \(\Prob(A)\ge 3/4\).
\end{lemma}

\begin{proof}
Each \(\omega_t\) is an isotropic sub-Gaussian vector in \(\R^k\) with
\(\|\omega_t\|_{\psi_2}\le C\) for an absolute constant \(C\), because its
coordinates are independent centered Rademacher variables. Hence the stated
operator-norm concentration is a standard sample-covariance bound for isotropic
sub-Gaussian rows. More precisely, Theorem~4.6.1 of
\cite{vershynin2018hdp} yields absolute constants \(C_*,c_*>0\) such that for
every \(u\ge 1\),
\[
\Prob\!\left(
\left\|\widehat \Sigma_\omega-I_k\right\|_{\mathrm{op}}
>
C_*\Bigl(\sqrt{\frac{k}{T}}+\frac{u}{\sqrt T}\Bigr)
+C_*^2\Bigl(\sqrt{\frac{k}{T}}+\frac{u}{\sqrt T}\Bigr)^2
\right)
\le 2e^{-c_*u^2}.
\]
Choose
\[
u=\sqrt{\frac{T}{16C_*^2}}
\]
and then increase \(C_{\mathrm{cov}}\) so that, whenever \(T\ge C_{\mathrm{cov}}k\),
\[
C_*\Bigl(\sqrt{\frac{k}{T}}+\frac{u}{\sqrt T}\Bigr)
+C_*^2\Bigl(\sqrt{\frac{k}{T}}+\frac{u}{\sqrt T}\Bigr)^2
\le \frac12.
\]
This gives
\[
\Prob\!\left(
\left\|\widehat \Sigma_\omega-I_k\right\|_{\mathrm{op}}>\frac12
\right)
\le
2e^{-c_*T/(16C_*^2)}
=
2e^{-c_{\mathrm{cov}}T},
\]
for \(c_{\mathrm{cov}}:=c_*/(16C_*^2)\). Increasing \(C_{\mathrm{cov}}\) once
more if necessary, we may arrange
\[
2e^{-c_{\mathrm{cov}}T}\le \frac14
\qquad\text{whenever }T\ge C_{\mathrm{cov}}k.
\]
The final claim follows because
\[
\left\|\widehat \Sigma_\omega-I_k\right\|_{\mathrm{op}}\le \frac12
\quad\Longrightarrow\quad
\lambda_{\min}(\widehat \Sigma_\omega)\ge \frac12.
\qedhere
\]
\end{proof}

\subsubsection{Gaussian Covariance Differentiation}
\label{subsubsec:shared-lb-gaussian-cov}

\begin{lemma}[Gaussian covariance differentiation]
\label{lem:shared-lb-gaussian-cov}
Let \((A_c,B_c)\) be centered Gaussian with covariance matrix
\[
\Sigma(c)=
\begin{pmatrix}
\mu & c\\
c & \mu
\end{pmatrix},
\qquad |c|<\mu,
\]
and let \(f:\R^2\to\R\) have continuous mixed derivatives
\(\partial_{ab}f\) and \(\partial_{aabb}f\). Assume that \(f\) and these two
derivatives have sufficient Gaussian integrability to justify differentiation
under the integral and integration by parts. Then
\[
\frac{d}{dc}\E[f(A_c,B_c)]=\E[\partial_{ab}f(A_c,B_c)],
\]
and
\[
\frac{d^2}{dc^2}\E[f(A_c,B_c)]
=
\E[\partial_{aabb}f(A_c,B_c)].
\]
\end{lemma}

\begin{proof}
Let \(p_c\) denote the density of \(N(0,\Sigma(c))\). Its Fourier transform is
\[
\widehat p_c(\xi,\eta)
=
\exp\!\Big(-\tfrac12(\mu\xi^2+2c\xi\eta+\mu\eta^2)\Big).
\]
Hence
\[
\partial_c\widehat p_c(\xi,\eta)
=
-\xi\eta\,\widehat p_c(\xi,\eta)
=
\widehat{\partial_{ab}p_c}(\xi,\eta),
\]
so \(\partial_c p_c=\partial_{ab}p_c\). Differentiation under the integral and
integration by parts give the first formula. Repeating the density
differentiation and integrating by parts once more yields
\[
\frac{d^2}{dc^2}\E[f(A_c,B_c)]=\E[\partial_{aabb}f(A_c,B_c)].
\]
\end{proof}

\end{document}